\documentclass[12pt]{article}

\usepackage[margin=1in]{geometry}
\usepackage{color}
\usepackage{graphicx}
\usepackage{epsfig}
\usepackage{epstopdf}
\usepackage{subfig}

\usepackage{amsmath}
\usepackage{amssymb}
\usepackage{amsthm}

\usepackage{rotating}
\usepackage{verbatim}
\usepackage{hyperref}
\usepackage{bm}
\usepackage[normalem]{ulem}
\usepackage[authoryear]{natbib}
\usepackage{bbm}
\usepackage{array}
\usepackage{enumerate}
\usepackage{float}
\usepackage{authblk}% delete this for AOS template

\usepackage[perpage]{footmisc}

\newtheorem{theorem}{Theorem}[section]
\newtheorem{lemma}[theorem]{Lemma}
\theoremstyle{plain}
\newtheorem{definition}{Definition}[section]

\usepackage{commands}
\usepackage{multirow}

\usepackage{amsmath}
\usepackage{array}
\newcommand\undermat[2]{%
  \makebox[0pt][l]{$\smash{\underbrace{\phantom{%
    \begin{matrix}#2\end{matrix}}}_{\text{$#1$}}}$}#2}

 \newcommand{\kwz}[1]{
  {\color{magenta} [KWZ: {#1}]}
 }

\usepackage{thmtools}
\numberwithin{equation}{section}

\title{Statistical Inference After Adaptive Sampling for Longitudinal Data}
\author[1]{Kelly W. Zhang}
\author[1,2]{Lucas Janson}
\author[1,2]{Susan A. Murphy}
\date{}
\affil[1]{Department of Computer Science, Harvard University}
\affil[2]{Department of Statistics, Harvard University}

\begin{document}
\maketitle

\begin{abstract}
\noindent Online reinforcement learning and other adaptive sampling algorithms are increasingly used in digital intervention experiments to optimize treatment delivery for users over time. In this work, we focus on longitudinal user data collected by a large class of adaptive sampling algorithms that are designed to optimize treatment decisions online using accruing data from multiple users. Combining or ``pooling'' data across users allows adaptive sampling algorithms to potentially learn faster. However, by pooling, these algorithms induce dependence between the sampled user data trajectories; we show that this can cause standard variance estimators for i.i.d. data to underestimate the true variance of common estimators on this data type. We develop novel methods to perform a variety of statistical analyses on such adaptively sampled data via Z-estimation. Specifically, we introduce the \textit{adaptive} sandwich variance estimator, a corrected sandwich estimator that leads to consistent variance estimates under adaptive sampling. Additionally, to prove our results we develop novel theoretical tools for empirical processes on non-i.i.d., adaptively sampled longitudinal data which may be of independent interest. This work is motivated by our efforts in designing experiments in which online reinforcement learning algorithms optimize treatment decisions, yet statistical inference is essential for conducting analyses after experiments conclude.
\end{abstract}

% 01_introduction.tex
%%%%%%%%%%%%%%%%%%%%%%%%%%%%%%%%%%%%%%%%%%%%
%%%%%%%%%%%%%%%%%%%%%%%%%%%%%%%%%%%%%%%%%%%%
\section{Introduction} %%%%%%%%%%%%%%%%%%%%%%%%%%%%
%%%%%%%%%%%%%%%%%%%%%%%%%%%%%%%%%%%%%%%%%%%%
\label{sec:intro}

Online reinforcement learning (RL) and other adaptive sampling algorithms are increasingly used in digital intervention experiments to optimize treatment delivery for users over time \cite{cai2021bandit,figueroa2021adaptive,forman2019can,liao2020personalized,piette2022artificial,trellaPCS,trella2022reward,yom2017encouraging}. For example, in mobile health, online RL algorithms have been used in longitudinal clinical trials developing Just-In-Time interventions for people with a variety of chronic health problems \cite{figueroa2021adaptive,liao2020personalized,piette2022artificial,yom2017encouraging}.
These trials are longitudinal because they involve making multiple treatment decisions for users over time. Online RL algorithms are used during the experiment to optimize treatment decisions; specifically, the RL algorithm uses the outcomes of previous treatment decisions to inform future treatment selection. A critical consideration when designing experiments that use such RL algorithms is ensuring that one can use the resulting data collected to perform valid statistical inference after the experiment is over. For example, one might want to construct confidence intervals for a treatment effect on a variety of different outcomes, including the reward. This work is motivated by experimental trials for digital health interventions in which RL algorithms are used to optimize treatments over time for multiple users. In these settings, valid post-trial statistical inference is critical to inform decisions about whether to roll out or how to improve a given digital intervention after the trial is over \cite{figueroa2021adaptive,liao2020personalized,trellaPCS}. A significant challenge is to develop valid statistical inference methods that are applicable to the data collected by the variety of RL algorithms that those designing digital interventions want to use.

Recently in the longitudinal digital intervention space, there has been great interest in online RL algorithms that combine or ``pool'' data across multiple users to inform future treatment decisions, because they can potentially learn faster how best to select treatments.  In fact, there is so much interest that several digital health intervention trials have already used such pooling RL algorithms \cite{figueroa2021adaptive,piette2022artificial,tomkins2021intelligentpooling,yom2017encouraging}.
However, it is unclear whether existing statistical inference methods for longitudinal data \cite{boruvka2018assessing,fitzmaurice2012applied,qian2019estimating,robins1997causal,zeger1986longitudinal}, which assume independent user data trajectories, should be used on data collected with adaptive sampling algorithms that pool data online. This is because, by pooling, adaptive sampling algorithms induce dependence between the collected user data trajectories. For example, if the algorithm uses the outcomes of one user to inform future treatment decisions for another user, the data trajectories collected from these two users will not be independent. 

There are existing approaches for statistical inference after adaptive sampling that account for the dependence induced by the algorithm. However, these approaches make a variety of restrictive assumptions on how users' outcomes can evolve over time and can be affected by treatments. For example, many works assume a classical contextual bandit environment in which user states are i.i.d. over time and the mean reward only depends on the most recent state and treatment \cite{bibaut2021post,bibaut2021risk,chen2020statistical,deshpande,athey,hu2006theory,zhan2021off,zhang2020inference,zhang2021mestimator}. 
These inference approaches are not applicable to classical longitudinal data settings in which (a) treatment decisions may affect users' future responsiveness to treatments or (b) user outcomes may be non-stationary. 

Moreover, we want our statistical inference approach to be robust to misspecification of the model used by the RL algorithm. Specifically, online RL algorithms make treatment decisions using approximate models for the users' outcomes (i.e., models of the environment) that they repeatedly fit using the data collected during the experiment. The models used in online RL algorithms are chosen to appropriately trade off bias and variance so the algorithm can quickly learn to select effective treatments. For example, even in environments in which treatment decisions may impact users' responsiveness to treatments multiple decision times into the future, in order to reduce variance, often simpler algorithms that do not model these delayed effects of treatment (like bandit algorithms) are preferred  \cite{figueroa2021adaptive,trellaPCS,yom2017encouraging}. After the digital intervention experiment is over, we argue that the validity of the statistical inference using the resulting adaptively sampled longitudinal data should not require that these approximate models used by the online RL algorithm are correctly specified. 

%%%%%%%%%%%%%%%%%%%%%%%%%%%%%%%%%%%%%%%%%%%%%%%
%%%%%%%%%%%%%%%%%%%%%%%%%%%%%%%%%%%%%%%%%%%%%%%
\subsection{Our Contribution}

In this work, we consider pooling adaptive sampling algorithms that, for each decision time $t \in [1 \colon T]$, form a policy $\pihat{t}$ that appropriately converges to a \textit{target} policy $\pistar{t}$ as the number of users $n$ grows; see Remark \ref{remark:targetPolicy} and Section \ref{sec:policyAssumptions} for further discussion of this assumption. We provide statistical theory for Z-estimators \cite[Chapter 5]{van2000asymptotic} on data collected by such pooling adaptive sampling algorithms. Z-estimators encompass most classical statistical estimators (e.g., least squares and maximum likelihood estimators) and are often used in estimating time-varying causal effects \cite{robins1997causal}. We derive the asymptotic distribution of these Z-estimators as the number of users $n\to\infty$ to facilitate the construction of asymptotically valid confidence regions. Specifically, we prove that the commonly used standard sandwich variance estimator \cite{huber1967under,zeileis2006object}, can underestimate the true variance of Z-estimators when data is adaptively sampled via algorithms that learn by pooling data across users. We develop the \textit{adaptive sandwich estimator}, a corrected sandwich estimator that leads to consistent variance estimates under adaptive sampling. Specifically, our contributions are as follows:
\begin{enumerate}
    \item \bo{Facilitating Statistical Inference after Using Pooling Adaptive Sampling Algorithms on Longitudinal Data:}
    Our approach for inference via Z-estimators is the first method that is applicable to longitudinal datasets collected by adaptive sampling algorithms that learn by pooling data across users. Moreover, the validity of our approach does not require the approximate outcome models learned by the adaptive sampling algorithm to be correct. This work enables digital intervention researchers to use pooling adaptive sampling algorithms in their experiments without sacrificing the statistical validity in performing a wide variety of after-study analyses.
    
    \smallskip %%%%%%%%%%%%%%%%%%%%%%%%%%%%%%%%%%%
    \item \bo{Novel use of Radon-Nikodym Derivatives to Facilitate Theory for Adaptively Sampled Data:}
    A significant technical challenge is that standard methods for empirical processes are insufficient for proving our asymptotic normality results since the adaptively sampled user data trajectories are not i.i.d. A key approach we use to facilitate theory for this non-i.i.d. data type is Radon-Nikodym derivative weighting. Specifically, we consider settings in which the estimator of the parameter of interest and the estimators used by the adaptive sampling policies $\pihat{t}$ are each a solution to some standard estimating function. Incorporating Radon-Nikodym derivative weights is integral to defining \textit{joint} estimating functions for the parameter of interest and the policy parameters. Note that the joint estimating functions (and the Radon-Nikodym derivative weights) are used \textit{solely} to analyze the asymptotic distribution of these estimators and \textit{not} needed to form the estimators themselves. We introduce these weights in Section \ref{sec:proofIdeas}.
    
    \smallskip %%%%%%%%%%%%%%%%%%%%%%%%%%%%%%%%%%%
    \item \bo{Empirical Process Theory for Adaptively Sampled Longitudinal Data:} To prove our results we develop novel theoretical tools for empirical processes on non-i.i.d., adaptively sampled longitudinal data, which may be of independent interest. These empirical processes are weighted by the Radon-Nikodym derivatives mentioned earlier. Specifically, we develop a Weighted Martingale Central Limit Theorem for functions of adaptively sampled data (Theorem \ref{thm:weightedCLT}), as well as a novel Weighted Martingale Bernstein Inequality (Lemma \ref{lemma:bernstein}). Using these two results, we prove a functional asymptotic normality result for Radon-Nikodym derivative weighted empirical processes under bracketing number conditions. See Section \ref{sec:functionalNormality} for more details.
    %Additionally we develop a novel Weighted Martingale Bernstein Inequality (Lemma \ref{lemma:bernstein}) that can be used to prove functional asymptotic normality results for adaptively sampled data; see Section \ref{sec:functionalNormality} for more details.
\end{enumerate}

% 01a_preliminaries.tex
%%%%%%%%%%%%%%%%%%%%%%%%%%%%%%%%%%%%%%%%%%%%
%%%%%%%%%%%%%%%%%%%%%%%%%%%%%%%%%%%%%%%%%%%%
\section{Preliminaries} %%%%%%%%%%%%%%%%%%%%%%%%%%%%
%%%%%%%%%%%%%%%%%%%%%%%%%%%%%%%%%%%%%%%%%%%%
%%%%%%%%%%%%%%%%%%%%%%%%%%%
%%%%%%%%%%%%%%%%%%%%%%%%%%%
\label{sec:setup}

We consider a batch dataset collected by an adaptive sampling algorithm that pools across users. The dataset is comprised of data on $n$ users over $T$ decision times. For each decision time $t \in [1 \colon T]$ and user $i \in [1 \colon n]$, the observations consist of a multi-dimensional vector of random variables which we call the state, $\state{i}{t}\in\real^{d_S}$; a scalar action (i.e., treatment), $\action{i}{t} \in \MC{A}$ (here $\MC{A}$ is a finite set, so $| \MC{A} | < \infty$); and lastly the multi-dimensional outcome vector of random variables, $\reward{i}{t} \in \real^{d_Y}$. Often adaptive sampling algorithms are designed to maximize a reward; in this case, the reward $R_t^{(i)} \in \real$ is some known function of the outcome vector $\reward{i}{t}$. We define $\reward{i}{t}$ because often we are interested in inference regarding quantities that are not the reward. For example, the reward in a physical activity digital health study could be the user's step count, but we may be interested in other outcomes like the user's heart rate.

We use potential outcomes \cite{imbens2015causal} to represent counter-factual outcomes. We consider a longitudinal data setting in which the potential outcomes for $\reward{i}{t}$ may depend on all actions taken on user $i$ up to decision time $t$, $\action{i}{1:t}$; we use the notation $\action{i}{1:t} \triangleq \big\{ \action{i}{t'} \big\}_{t'=1}^t$ to denote collections of random variables. This means $\reward{i}{t}$ has $|\MC{A}|^t$ different potential outcomes, $\big\{ \reward{i}{t} (a_{1:t}) : a_{1:t} \in \MC{A}^t \big\}$, where $\MC{A}^t$ denotes the $t$-fold Cartesian product of $\MC{A}$. Similarly, states have potential outcomes $\big\{ \state{i}{t} (a_{1:t-1}) : a_{1:t-1} \in \MC{A}^{t-1} \big\}$. The observed variables are $\reward{i}{t} \triangleq \reward{i}{t} \big( \action{i}{1:t} \big)$ and $\state{i}{t} \triangleq \state{i}{t} \big( \action{i}{1:t-1} \big)$. 

We consider the setting in which  the potential outcomes,  $i \in [1 \colon n]$, are i.i.d. according to an unknown $\MC{P}$, i.e.,
\begin{equation}
	\label{eqn:potentialOutcomes}
	D^{(i)} =\left\{ \state{i}{t}(a_{1:t-1}), \reward{i}{t} (a_{1:t}) : a_{1:t} \in \MC{A}^t \right\}_{t=1}^T 
	\iidsim \mathcal{P}, ~\TN{ i.i.d over users } i \in [1 \colon n].
\end{equation}
\noindent Note that the above allows for the trajectory of the observed user states and outcomes to be non-stationary and dependent over time. This setting encompasses both Markovian and non-Markovian user environments and is widely used in the longitudinal data analysis literature \cite{fitzmaurice2012applied,robins1986new,robins1997causal}.

We consider adaptively sampled data for which at decision time $t = 1$,  a pre-specified policy $\pi_1$, where $\PP \big( \action{i}{1}\big| \state{i}{1} \big) \triangleq \pi_1 \big( \action{i}{1}, \state{i}{1} \big)$, is used to select the treatment actions independently for all users. Then, for each $t \geq 2$, an adaptive sampling algorithm may use all the observed data so far across all users %$\history{1:n}{t-1}$, 
to form a policy $\pihat{t}$.
Specifically the policy $\pihat{t}$ can be formed using the history $\history{i}{t-1} \triangleq \big\{ \state{i}{t'}, \action{i}{t'}, \reward{i}{t'} \big\}_{t'=1}^{t-1}$ for all users $i \in [1 \colon n]$; for convenience we will use the notation $\history{1:n}{t-1} \triangleq \big\{ \history{i}{t-1} \big\}_{i=1}^n$ to represent the collective history for all users.

%$i^{\TN{th}}$ 
The policy $\pihat{t}$ takes as input the user's current state, $\state{i}{t}$, and outputs a sampling probability distribution over the action space $\MC{A}$. For $a \in \MC{A}$,
\begin{equation}
	\pihat{t} \big( a, \state{i}{t} \big)
	= \PP \left( \action{i}{t} = a \big| \state{i}{t}, \history{1:n}{t-1} \right).
\end{equation}
We consider data collected by adaptive sampling algorithms that, conditional on $\history{1:n}{t-1}$ and $\state{1:n}{t}$, select actions $\action{1}{t}, \action{2}{t}, \dots, \action{n}{t}$ independently using policy $\pihat{t}$. Note that the actions are not identically distributed conditional on $\history{1:n}{t-1}$ as the realized value of users' states at time $t$ may differ. 

\begin{remark}[Dependent User Data Trajectories]
    Note that even though the users' potential outcomes $D^{(i)}$ are i.i.d. (as seen in display \eqref{eqn:potentialOutcomes}), the observed user data trajectories, $\history{i}{t}$, are generally not independent over $i \in [1 \colon n]$ due to algorithm's use of the common history, $\history{1:n}{t-1}$ in sampling the actions $\action{1}{t}, \action{2}{t}, \dots, \action{n}{t}$.
\end{remark}

For our statistical analyses, we consider asymptotics as the number of users, $n$, goes to infinity and keep the total number of decision times, $T$, fixed. 
This decision is motivated by our work in digital intervention experiments. These experiments are primarily concerned with using inference methods to draw scientific conclusions about a population of individuals over a fixed period of time, e.g., a $90$-day physical activity mobile health intervention for individuals with stage-1 hypertension \cite{liao2020personalized}. 

We now informally provide several key assumptions that we make on the online pooling adaptive sampling algorithm used to collect the data (see Section \ref{sec:policyAssumptions} for more details). We assume that policies $\pihat{t}$ belong to a parametric class 
\begin{equation*}
    \left\{ \pi_t \big( \cdotspace; \beta_{t-1} \big) : \beta_{t-1} \in \real^{ d_{ t-1 } } \right\}.
\end{equation*}
In particular, $\pihat{t}(\cdotspace) \triangleq \pi_t \big( \cdotspace; \betahat{t-1} \big)$, where $\betahat{t-1}$ is a function of all users'  data prior to time $t$, $\history{1:n}{t-1}$. For a given action $a \in \MC{A}$ and state $s \in \real^{d_S}$, $\pi_t \big(a, s; \betahat{t-1} \big)$ is a probability of selecting action $a$ conditional on state $s$. 
We will assume conditions under which $\betahat{t-1}$ converges to a deterministic $\betastar{t-1}$ as the number of users $n \to \infty$. Hence, we call $\pi_t(\cdotspace; \betastar{t-1})$, which we abbreviate as $\pistar{t}$, the \textit{target policy} at time $t$. 

\begin{remark}[Target Policies]
    \label{remark:targetPolicy}
    Note that the assumption that there exists a target policy $\pistar{t}$ for each decision time $t$ is rather mild; this is because the asymptotic arguments derived here are as $n\to\infty$ with the total number of decision times $T$ fixed. Further, $\betastar{t-1}$ can be an arbitrary deterministic limit, e.g., $\betastar{t-1}$ does not have to be a parameter in a correctly specified model of the reward, and $\pistar{t}$ does not have to be optimal in any way. We also allow the target policy $\pistar{t}$ to change with $t \in [2 \colon T]$; this allows for non-stationarity in the users' outcomes that is not accounted for by the algorithm. In the special case that the environment is stationary and any models assumed by the algorithm are correctly specified, the target policy $\pistar{t}$ may be the same for all $t \in [2 \colon T]$. 
    See Section \ref{sec:policyAssumptions} for more on the assumptions made on the policies.
\end{remark}

%%%%%%%%%%%%%%%%%%%%%%%%%%%%%%%%%%%%%%%%%%%%%%
%%%%%%%%%%%%%%%%%%%%%%%%%%%%%%%%%%%%%%%%%%%%%%
\section{Problem Statement}

\subsection{Inference Objective} %%%%%%%%%%%%%%%%%%%%%%%
\label{sec:infObjective}

We consider estimands that are defined with respect to the distribution in which the target policies $\pistar{2:T} \triangleq \big\{ \pistar{t} \big\}_{t=2}^T$ are used to select actions. 
Specifically, we aim to conduct inference about a parameter $\thetastar{}$ that solves
\begin{equation}
	\label{eqn:ZestimandDef}
	\bs{0} = \Estar{2:T} \left[ \est{} \big( \history{i}{T} ; \theta \big) \right],
\end{equation}
where $\est{}$ is a measurable function of $\history{i}{t} $ indexed by a finite-dimensional $\theta \in \real^{d_\theta}$. 

The above expectation is indexed by the target policies $\pistar{2:T}$ to indicate that the expectation is over the distribution of $\history{i}{T}$ in which the actions are selected using the target policies and user potential outcomes are drawn from $\MC{P}$ as described in display \eqref{eqn:potentialOutcomes}; we will use $\Pstar$ to refer to this distribution. Note that when the target policies $\pistar{2:T}$ are used to select actions, the data is no longer ``adaptively sampled'' so the user trajectories $\history{i}{T}$ are i.i.d.; thus, $\thetastar{}$ is not indexed by $i$.

To estimate $\thetastar{}$ we use Z-estimation;  the estimator $\thetahat{}$ satisfies 
\begin{equation}
    \label{eqn:ZestimatorDef}
     o_P \big( 1/\sqrt{n} \big) = \frac{1}{n} \sum_{i=1}^n \est{} \big( \history{i}{T} ; \thetahat{} \big).
\end{equation}
This setup encompasses many types of standard estimators (e.g., least squares and maximum likelihood) and includes minimizers of differentiable loss functions. 
We are interested in constructing confidence regions for $\thetastar{}$. We do this by characterizing the asymptotic distribution of $\thetahat{}$ as the number of users $n \to \infty$ and using the asymptotic distribution to approximate the finite-sample distribution of $\thetahat{}$.

To enhance expositional clarity we illustrate the ideas using the running example of a least squares estimator in a binary action setting, $\MC{A} = \{ 0, 1 \}$, with the following $\est{}$:
\begin{equation}
\label{eqn:leastSquaresInferenceObjective}
	\est{} \big( \history{i}{T} ; \theta \big) = \frac{1}{T} \sum_{t=1}^T \left( \reward{i}{t} - \theta_{0}^\top \state{i}{t} - \action{i}{t} \theta_{1}^\top \state{i}{t} \right) \begin{bmatrix}
	    \state{i}{t} \\
	    \action{i}{t} \state{i}{t}
	\end{bmatrix}.
\end{equation}
Above, $\reward{i}{t} \in \real$, $\theta = [\theta_{0}, \theta_{1}]$, and the first entry of $\state{i}{t}$ is $1$ (intercept term) for all $t,i$.

\begin{remark}[Interpretation when the Model used by $\psi$ is  Misspecified]
    \label{remark:misspecification}
    Often in Z-estimation, $\est{}$ corresponds to the derivative of a likelihood function for a parameter in a particular (possibly semi- or non-parametric) model for the data; in this case we can think of $\est{}$ as ``correctly specified'' if that model holds in our data. In the least squares example, $\est{}$ is correctly specified if $\E \big[ \reward{i}{t} \big| \state{i}{t}, \action{i}{t}, \history{i}{t-1} \big] = \theta_{0}^{*,\top} \state{i}{t} + \action{i}{t} \theta_{1}^{*,\top} \state{i}{t}$ a.s. for all $t$. As is standard for Z-estimators, if $\est{}$ is not correctly specified, then $\thetastar{}$ is the best \emph{projected} solution; for example, in the least squares example from display \eqref{eqn:leastSquaresInferenceObjective}, $\thetastar{}$ corresponds to the best fitting linear model. The projection is with respect to the distribution $\Pstar$ in which target policies $\pistar{2:T}$ are used to select actions.  In this case $\thetastar{}$ is a function of the target policies $\pistar{2:T}$. However in the correctly specified model case, $\thetastar{}$ does not depend on the target policies. See Section \ref{sec:simplifying} for how correct model specification affects the adaptive sandwich variance.
\end{remark}

\subsubsection{Excursion Effects are a Key Use Case}
\label{sec:excursion}
Excursion effects, which are used for the primary analysis in micro-randomized trials \cite{boruvka2018assessing,qian2021micro,qian2019estimating}, are a key use case for our inference method. In these longitudinal trials, treatment actions for each individual are repeatedly randomized using stochastic policies. The primary analysis for these trials concerns treatment effect excursions from the experiment's target policies. An example excursion effect is the following excursion from the target policy at time $t$:
\begin{equation}
    \label{eqn:excursionExample}
    \Estar{2:t-1} \left[ \reward{i}{t} \big( \action{i}{1:t-1}, a_t=1 \big) - \reward{i}{t} \big( \action{i}{1:t-1}, a_t=0 \big) \right].
\end{equation}

In the simplified setting in which the outcome $\reward{i}{t}$ only depends on the most recent action $\action{i}{t}$, the excursion effect simplifies to the standard treatment effect
\begin{equation*}
    \E \big[ \reward{i}{t} ( a_t=1 ) - \reward{i}{t} ( a_t=0 ) \big].
\end{equation*}
Thus, the excursion effect from display \eqref{eqn:excursionExample} can be considered a generalization of the standard treatment effect to environments in which all actions taken so far, $\action{i}{1:t}$, can affect the distribution of the outcome $\reward{i}{t}$.

\subsection{Policies Formed by the Adaptive Sampling Algorithm} %%%%%%%%%%%%%%%%%%%%%%%%%%%
\label{sec:policyAssumptions}

As discussed in Section \ref{sec:setup} (Preliminaries), at each decision time $t \in [2 \colon T]$, the adaptive sampling algorithm uses all previously observed user data, $\history{1:n}{t-1}$, to form a policy $\pihat{t}$ and uses this policy to select actions. In particular, we assume that there are policy function classes
\begin{equation}
    \label{eqn:policyClasses}
	\left\{ \pi_t \big( \cdotspace ; \beta_{t-1} \big) : \beta_{t-1} \in \real^{d_{t-1}} \right\}
\end{equation}
for each $t \in [2 \colon T]$ and that $\pihat{t}(\cdotspace) \triangleq \pi_{t}(\cdotspace; \betahat{t-1})$ where $\betahat{t-1} \in \real^{d_{t-1}}$ is a statistic formed using $\history{1:n}{t-1}$. 

Recall that we assume that for each decision time $t \in [2 \colon T]$, the statistic $\betahat{t-1}$ formed by the algorithm converges in probability to a deterministic target policy parameter $\betastar{t}$ as the number of users $n \to \infty$. The parameter $\betastar{t}$ could parameterize a model of the expected reward used by the adaptive sampling algorithm. The target policy parameters $\big\{ \betastar{t} \big\}_{t=1}^{T-1}$, which parameterize the target policies $\big\{ \pistar{t} \big\}_{t=2}^{T} \triangleq \big\{ \pi_{t}(\cdotspace; \betastar{t-1}) \big\}_{t=2}^{T}$.
To allow for a large class of possible target policy parameters, we assume $\betastar{t}$ is the solution to some estimating equation; we formally define these parameters below.
%We now formally define the target policy parameters $\big\{ \betastar{t} \big\}_{t=1}^{T-1}$, which parameterize the target policies $\big\{ \pistar{t} \big\}_{t=2}^{T} \triangleq \big\{ \pi_{t}(\cdotspace; \betastar{t-1}) \big\}_{t=2}^{T}$. 

Recall that at the first decision time actions are selected using a pre-specified policy $\pi_1$.
For the second decision time, the target policy $\pistar{2}(\cdotspace) \triangleq \pi_{2}(\cdotspace; \betastar{1})$ where the target parameter $\betastar{1}$ solves
\begin{equation*}
    \bs{0} = \E \left[ \piest{1} \big( \history{i}{1} ; \betastar{1} \big) \right],
\end{equation*}
for some measurable function $\piest{1}$ of $\history{i}{1}$ indexed by a finite dimensional $\beta_1 \in \real^{d_1}$.

For the third decision time, the target policy $\pistar{3}(\cdotspace) \triangleq \pi_{3}(\cdotspace; \betastar{2})$ where the target parameter $\betastar{2}$ solves
\begin{equation}
    \label{eqn:betastarDef2}
    \bs{0} = \Estar{2} \left[ \piest{2} \big( \history{i}{2} ; \betastar{2} \big) \right],
\end{equation}
for some measurable function $\piest{2}$ of $\history{i}{2}$ indexed by a finite dimensional $\beta_2 \in \real^{d_2}$.
In display \eqref{eqn:betastarDef2} above, the expectation is indexed by target policies $\pistar{2}(\cdotspace)$. This means that the definition of the target policy for the third decision time, $\pistar{3}$, depends on the definition of target policy for the second decision time, $\pistar{2}$.

Continuing this pattern, for the $t^{\TN{th}}$ decision time, the target policy $\pistar{t}(\cdotspace) \triangleq \pi_{t}(\cdotspace; \betastar{t-1})$ where the target parameter $\betastar{t}$ solves
\begin{equation}
    \label{eqn:betastarDef}
    \bs{0} = \Estar{2:t-1} \left[ \piest{t-1} \big( \history{i}{t-1} ; \betastar{t-1} \big) \right], 
\end{equation}
for some measurable function $\piest{t-1}$ of $\history{i}{t-1}$ indexed by a finite dimensional $\beta_{t-1} \in \real^{d_{t-1}}$.
Again, the above expectation is indexed by the previous target policies $\pistar{2:t-1}$.
%For example, $\piest{t-1}$ might correspond to 

An example function $\piest{t-1}$ is the following, which corresponds to a least squares solution:
\begin{equation}
	\label{eqn:psiPolicyParameter}
	\piest{t-1} \big( \history{i}{t-1} ; \beta_{t-1} \big) =  \sum_{t'=1}^{t-1} \left( R_{t'}^{(i)} - \beta_{0,t-1}^\top \state{i}{t'} - \action{i}{t'} \beta_{1,t-1}^\top \state{i}{t'} \right)
	\begin{bmatrix}
	    \state{i}{t'} \\
	    \action{i}{t'} \state{i}{t'}
	\end{bmatrix}.
\end{equation}
In our simulations (Section \ref{sec:simulations}) we consider a Boltzmann (or Softmax) exploration adaptive sampling algorithm \cite{asadi2017alternative,cesa2017boltzmann,sutton2018reinforcement} that forms policies using estimators of the least squares solution $\betastar{t}$, defined with the estimating function from display \eqref{eqn:psiPolicyParameter}. 
%In our simulations (Section \ref{sec:simulations}) we consider a Boltzmann (or Softmax) exploration adaptive sampling algorithm \cite{asadi2017alternative,cesa2017boltzmann,sutton2018reinforcement} that forms policies with a least squares estimator $\betahat{t}$ as shown in display \eqref{eqn:psiPolicyParameter}. 

Similarly, we assume that $\betahat{t}$, the estimators of the target policy parameters are Z-estimators, i.e., solutions to the empirical estimating functions. % assume that the estimators $\betahat{t-1}$ formed by the algorithm are Z-estimators. 
Formally, this means that $\betahat{t-1}$ satisfies
\begin{equation}
    \label{eqn:betahatDef}
    o_P(1/\sqrt{n}) = \frac{1}{n} \sum_{i=1}^n \piest{t-1} \big( \history{i}{t-1} ; \betahat{t-1} \big) \in \real^{d_{t-1}}.
\end{equation}

\begin{remark}[Misspecification of the Model used by the Adaptive Algorithm]
    %In general, the adaptive sampling algorithm may use $\betahat{t}$ defined in display \eqref{eqn:betahatDef}
    In general, the adaptive sampling algorithms target parameter $\betastar{t}$ defined in \eqref{eqn:betastarDef} can parameterize a model for parts of the multivariate distribution of $D^{(i)}$, the user's underlying potential outcomes from display \eqref{eqn:potentialOutcomes}. We do not require that this model is correct. For example, even if the algorithm is developed assuming the environment corresponds to that of a stochastic contextual bandit, the validity of our statistical analysis will not be affected if this assumption is wrong.
\end{remark}

%To allow for a large class of possible statistics $\betahat{t-1}$, we consider Z-estimators.
%for some measurable function $\piest{t-1}$ of $\history{i}{t-1}$ indexed by a finite-dimensional $\beta_{t-1} \in \real^{d_{t-1}}$. 

\subsubsection{Key Assumptions on Policies}
\label{sec:policyKeyAssumptions}

Rather than assume the adaptive sampling algorithm's model is correctly specified, we instead will make assumptions on the estimators $\big\{ \betahat{t} \big\}_{t=1}^{T-1}$ of the policy parameters and the policy function classes $\left\{ \pi_t \big( \cdotspace ; \beta_{t-1} \big) : \beta_{t-1} \in \real^{d_{t-1}} \right\}$, from display \eqref{eqn:policyClasses}. We now introduce the three foremost assumptions we place on the adaptive sampling policies, Conditions \ref{cond:consistencyPolicy}-\ref{cond:lipschitzPolicy} below (we introduce the other assumptions we place on $\betahat{t}$ in Section \ref{sec:policyConditions}).

Condition \ref{cond:consistencyPolicy} is a consistency condition that ensures that the policy parameter estimator $\betahat{t}$ formed by the algorithm converges in probability to a target parameter value $\betastar{t}$ as the number of users $n \to \infty$. 
%We formally define the parameters $\betastar{t}$ below the statement of Condition \ref{cond:consistencyPolicy}.

\begin{condition}[Consistency of Policy Estimators] 
    \label{cond:consistencyPolicy}
    For each $t \in [1 \colon T-1]$,
    \begin{equation*}
        \betahat{t} \Pto \betastar{t}.
    \end{equation*}
\end{condition}

\begin{remark}[Sufficient Assumptions for Condition \ref{cond:consistencyPolicy}]
    In Theorem \ref{thm:consistencyBeta} of Appendix \ref{app:algConditions} we state simple sufficient conditions for Condition \ref{cond:consistencyPolicy} to hold.
\end{remark}

The next two key assumptions we place on the adaptive sampling algorithm, Conditions \ref{cond:exploration} and \ref{cond:lipschitzPolicy}, both concern the policy function classes $\left\{ \pi_t \big( \cdotspace ; \beta_{t-1} \big) : \beta_{t-1} \in \real^{d_{t-1}} \right\}$, from display \eqref{eqn:policyClasses}.

\begin{condition}[Minimum Exploration]
	\label{cond:exploration}
    Let $0 < \pi_{\min} < 1$ be a constant. 
	For all $t \in [2 \colon T]$, 
    \begin{equation*}
		\inf_{\beta_{t-1} \in \real^{d_{t-1}} }
		\pi_t \big( a, s ; \beta_{t-1} \big) \geq \pi_{\min}
	\end{equation*}
	for all $a \in \MC{A}$ and $s \in \MC{S}$. Also for $t = 1$, $\pi_1 ( a, s ) \geq \pi_{\min}$ for all $a \in \MC{A}$ and $s \in \MC{S}$.
\end{condition}

Condition \ref{cond:exploration} ensures that the policy class produces action selection probabilities that are strictly bounded above zero for all actions. 
Note this ensures that the policy is stochastic, as is necessary for micro-randomized trials (discussed earlier in Section \ref{sec:excursion}).
Note this condition excludes deterministic policies, which means target policies that maximize the expected reward in classical contextual bandit and Markov decision process environments are excluded. However, in general, the fewer structural assumptions that are placed on the environment, the more need there is for reward-maximizing algorithms to continually explore. For example, in non-stationary and adversarial sequential decision-making problem settings it is common both theoretically and in practice to prevent the RL algorithm's action selection probabilities to go to zero for any action \cite{bubeck2012towards,cesa2006prediction, chandak2020optimizing,lattimore2020bandit} in order to ensure the algorithm can detect changes in the reward distribution. Action selection probabilities are also commonly constrained away from $0$ and $1$ to facilitate causal inference and off-policy evaluation after the experiment is over \cite{athey,liao2016sample,thomas2016data,trellaPCS,yao2020power}.

\medskip
In Condition \ref{cond:lipschitzPolicy} below, for each $t \in [1 \colon T-1]$, we use $B_t \subset \real^{d_t}$ to denote some compact subset whose interior contains $\betastar{t}$.

\begin{condition}[Lipschitz Policy Functions]
	\label{cond:lipschitzPolicy}
	For all $t \in [2 \colon T]$, there is a non-negative, real-valued function $\dot{\pi}_t(\action{i}{t}, \state{i}{t})$ such that 
    (i) $\Estar{2:t} \big[ \big| \dot{\pi}_t(\action{i}{t}, \state{i}{t}) \big|^{2+\alpha} \big] < \infty$ for some $\alpha > 0$, and (ii) for any $\beta_{t-1}, \beta_{t-1}' \in B_{t-1}$,
    \begin{equation*}
        \big| \pi_t \big(\action{i}{t}, \state{i}{t}; \beta_{t-1} \big) - \pi_t \big( \action{i}{t}, \state{i}{t}; \beta_{t-1} \big) \big| 
        \leq \dot{\pi}_t(\action{i}{t}, \state{i}{t}) \big\| \beta_{t-1} - \beta_{t-1}' \big\|_2 ~~~ \TN{a.s.}
    \end{equation*}
\end{condition}

Condition \ref{cond:lipschitzPolicy} is a smoothness condition on the policy function classes, which excludes policies that are a discontinuous function of parameters $\beta_{t-1}$. It is well known in the inference after adaptive sampling literature that standard estimators, like the sample mean, can be asymptotically non-normal on data collected by adaptive algorithms that do not satisfy such smoothness conditions \citep{deshpande,athey,zhang2020inference}. Although this smoothness condition may appear rather mild, note that the reward-maximizing policy in a stochastic bandit problem is a discontinuous function of the margin because of the argmax operation; for example, in a two-armed bandit setting with $\beta^* \triangleq \E[ R_t(1)] - \E[ R_t(0) ]$, the optimal policy is $\PP( A_t = 1 ) = \II_{\beta^* > 0}$. Despite this, as mentioned after Condition \ref{cond:exploration}, there are standard reinforcement learning algorithms developed for more complex environments (e.g., non-stationary) which satisfy this smoothness condition.  

\begin{remark}[Example Algorithms that Satisfy Conditions \ref{cond:exploration} and \ref{cond:lipschitzPolicy}]
    In Appendix \ref{appEx:examples} we show that a Boltzmann (or Softmax) exploration algorithm \cite{asadi2017alternative,cesa2017boltzmann,sutton2018reinforcement} and a stochastic mirror descent algorithm (based on those from \cite[pg 361]{lattimore2020bandit} and \cite{bubeck2012towards}) both satisfy Conditions \ref{cond:exploration} and \ref{cond:lipschitzPolicy} above. 
\end{remark}

% 02_related_work.tex
%%%%%%%%%%%%%%%%%%%%%%%%%%%%%%%%%%%%%%%%%%%%%%%%%%%%%%
%%%%%%%%%%%%%%%%%%%%%%%%%%%%%%%%%%%%%%%%%%%%%%%%%%%%%%
%%%%%%%%%%%%%%%%%%%%%%%%%%%%%%%%%%%%%%%%%%%%%%%%%%%%%
%%%%%%%%%%%%%%%%%%%%%%%%%%%%%%%%%%%%%%%%%%%%
\section{Related Work} %%%%%%%%%%%%%%%%%%%%%%%%%%%
%%%%%%%%%%%%%%%%%%%%%%%%%%%%%%%%%%%%%%%%%%%%
\label{sec:relatedWork}

Recently, many inference methods have been developed for adaptively sampled data focused on multi-armed and contextual bandit environments. These include inference methods via asymptotic approximations \cite{bibaut2021post,chen2020statistical,deshpande,athey,zhan2021off,zhang2020inference,zhang2021mestimator} as well as approaches that use high probability bounds \cite{abbasi2011improved,brennan2020estimating,howard2018uniform,karampatziakis2021off}. These works for the most part consider asymptotics as $T \to \infty$. These methods are more restrictive than ours in that they assume an underlying contextual bandit environment that does not allow a user's potential outcomes to be dependent over time. Moreover, most of these approaches consider inference for particular estimands, e.g., the value of a policy or a specific treatment effect, rather than an all-purpose Z-estimand. However, these methods are more general than ours in that they put fewer restrictions on the adaptive sampling policies used to collect the batch data, e.g., many allow the action selection probabilities to go to zero at some rate for some actions and do not require their policy function classes to be smooth in its parameters. Additionally, most of these prior methods require that the reward model used by the adaptive sampling algorithm is correctly specified.

The adaptive clinical trial literature provides methods for inference after using policies that satisfy conditions akin to Conditions \ref{cond:exploration} (Minimum Exploration) and \ref{cond:lipschitzPolicy} (Lipschitz Policy Functions) above, e.g., see Theorems 3.1 and 9.1 of \cite{hu2006theory}. There are two ways in which our results differ from these classical results. The first is that we consider a setting in which the adaptive sampling algorithm repeatedly selects treatment actions for each of multiple individuals sequentially over time. Since the adaptive sampling algorithm selects actions probabilistically, each individual is sequentially randomized. In contrast, the adaptive clinical trial literature classically considers settings in which at each decision time a new individual is drawn independently from the population and the adaptive sampling algorithm makes one treatment action decision per individual. The second major difference is that these classical results assume that both the model used for inference and the model used by the adaptive algorithm are correctly specified. For our results, we do not assume either of these models is correctly specified; see Remark \ref{remark:misspecification} for more on model misspecification. Additionally, in Section \ref{sec:simplifying} we discuss how our asymptotic results simplify when the estimating function, $\psi$ uses a correctly specified model. 

Another area of related work is inference methods for longitudinal data. This literature assumes the same underlying potential outcomes model, display \eqref{eqn:potentialOutcomes}, that allows for non-stationarity and dependent outcomes over time within each user \cite{fitzmaurice2012applied,robins1997causal,zeger1986longitudinal}. However, this literature considers batch datasets in which user data trajectories are independent across users, which excludes datasets collected by adaptive sampling algorithms that learn across users. This literature also includes methods for inferring excursion effects \cite{boruvka2018assessing,qian2019estimating}, which were discussed in Section \ref{sec:excursion}.

Here we generalize techniques from the classical literature on empirical processes for i.i.d. data \cite{van2000asymptotic,van1996weak} to adaptively sampled data. In particular, we develop functional asymptotic normality and maximal inequality results for \lq\lq Radon-Nikodym derivative weighted empirical processes\rq\rq; see Section \ref{sec:functionalNormality} for more details. Note that \cite{bibaut2021risk} develops a maximal inequality for adaptively sampled data assuming a classical contextual bandit environment. Besides the differences in the underlying environment assumptions, our maximal inequality results also differ from theirs because they consider asymptotics as $T \to \infty$, while here $T$ is fixed and we consider asymptotics as $n \to \infty$.

% 03_asymptotic_results.tex
%%%%%%%%%%%%%%%%%%%%%%%%%%%%%%%%%%%%%%%%%%%%%%%%%%%%%%
%%%%%%%%%%%%%%%%%%%%%%%%%%%%%%%%%%%%%%%%%%%%%%%%%%%%%%
%%%%%%%%%%%%%%%%%%%%%%%%%%%%%%%%%%%%%%%%%%%%%%%%%%%%%
\section{Main Results} %%%%%%%%%%%%%%%%%%%
%%%%%%%%%%%%%%%%%%%%%%%%%%%%%%%%%%%%%%%%%%%%
\label{sec:asymptoticResults}

Note that if the batch data were collected using the fixed target policies $\pistar{2:T}$, rather than the data-dependent, adaptive policies $\pihat{2:T}$, then the data trajectories would be independent across users, i.e., $\history{i}{T}$ would be i.i.d. across $i \in [1 \colon n]$. In that i.i.d. setting, we could use standard asymptotic normality results for Z-estimators \cite[Theorem 5.21]{van2000asymptotic} to get that $\thetahat{}$ is asymptotically normal with the standard sandwich variance, i.e., 
\begin{equation}
    \label{eqn:normalityStandardSandwich}
    \sqrt{n} \left( \thetahat{} - \thetastar{} \right) \Dto \N \left( \bs{0}, ~ [\EstDotStar{}]^{-1} \Sigma [\EstDotStar{}]^{-1, \top} \right),
\end{equation} 
with ``bread'' $\EstDotStar{} \triangleq \fracpartial{}{\theta} \Estar{2:T} \big[ \est{} \big( \history{i}{T}; \theta \big) \big] \big|_{\theta = \thetastar{}}$ and ``meat'' $\Sigma \triangleq \Estar{2:T} \big[ \est{}( \history{i}{T}; \thetastar{} \big)^{\otimes 2} \big]$; we use the notation $x^{\otimes 2} \triangleq x x^\top$.

However, in the adaptively sampled data setting in which the random, data-dependent policies $\pihat{2:T}$ produced by the adaptive sampling algorithm are used to select actions, we show that the limiting variance is different, specifically, 
\begin{equation}
    \label{eqn:normalityresult}
	\sqrt{n} \left( \thetahat{} - \thetastar{} \right)
	\Dto \N \left( \bs{0}, ~ [\EstDotStar{}]^{-1} \Madapt{} [\EstDotStar{}]^{-1, \top} \right),
\end{equation}
where 
\begin{equation}
    \label{eqn:Madapt}
    \Madapt{} \triangleq \Estar{2:T} \bigg[ \bigg\{ \est{} \big(\history{i}{T}; \thetastar{} \big) + \EstDotStar{} \sum_{t=1}^{T-1} M_t ~ \piest{t} \big( \history{i}{t} ; \betastar{t} \big) \bigg\}^{\otimes 2} \bigg].
\end{equation}
Above $M_t \in \real^{d_\theta \by d_t}$. See display \eqref{eqn:Qmatrices} for the definition of the $M_t$ matrices.

We call the limiting variance in display \eqref{eqn:normalityresult}, the \textit{adaptive} sandwich variance. Comparing $\Sigma$ and $\Madapt{}$ from display \eqref{eqn:Madapt}, we can interpret the term $\EstDotStar{} \sum_{t=1}^{T-1} M_t ~ \piest{t} \big( \history{i}{t} ; \betastar{t} \big)$ as the ``cost'' or ``inflation'' in variance due to using the estimated $\pihat{2:T}$ to select actions rather than $\pistar{2:T}$. In special cases, under a property we call ``policy invariance'', $M_t = 0$ so the limiting sandwich and adaptive sandwich variances are equal; see Section \ref{sec:simplifying} for more details. We provide estimators of the adaptive sandwich variance (see Appendix \ref{app:simulationsEstimators}), which we use in Section \ref{sec:simulations} for our simulation results.

We now outline the remainder of this asymptotic results section. In order to provide a high-level understanding of how our proof techniques and results differ from the i.i.d. data case, in Section \ref{sec:proofIdeas} we discuss the ideas behind our asymptotic normality proof; specifically, we introduce the Radon-Nikodym derivative weighting we use in Section \ref{sec:introRNweights} and provide an overview of the functional asymptotic normality results for adaptively sampled data that we develop in Section \ref{sec:functionalNormality}.
Then in Section \ref{sec:formalResults}, we state our results formally; specifically in Section \ref{sec:policyConditions} we introduce additional assumptions on the policy parameters and in Section \ref{sec:theoremStatement} we have our formal theorem statements. In Section \ref{sec:proofSketch}, we provide a more detailed proof sketch of our main asymptotic normality result, display \eqref{eqn:normalityresult}. Finally, in Section \ref{sec:simplifying}, we discuss cases in which the limiting adaptive sandwich variance equals the standard sandwich variance.

%%%%%%%%%%%%%%%%%%%%%%%%%%%%%%%%%%%%%%%%%%
%%%%%%%%%%%%%%%%%%%%%%%%%%%%%%%%%%%%%%%%%%
%%%%%%%%%%%%%%%%%%%%%%%%%%%%%%%%%%%%%%%%%%
%%%%%%%%%%%%%%%%%%%%%%%%%%%%%%%%%%%%%%%%%%
%\subsection{Understanding how the Adaptively Sampled Data Case Differs from the i.i.d. Data Case}
%\label{sec:understanding}

\subsection{Ideas Behind the Proof of Asymptotic Normality}
\label{sec:proofIdeas}
%%%%%%%%%%%%%%%%%%%%%%%%%%%%%%%%%%%%%%%%%%
%%%%%%%%%%%%%%%%%%%%%%%%%%%%%%%%%%%%%%%%%%

In order to provide a high-level understanding of how proving results for adaptively sampled data differs from the i.i.d. data case, we now discuss the key ideas that we use in our proof of asymptotic normality. The foremost technical challenge in our proof of asymptotic normality of $\sqrt{n} \big( \thetahat{} - \thetastar{} \big)$ is to account for how the data is collected using estimated policies $\pihat{t}(\cdotspace) = \pi_t \big( \cdotspace; \betahat{t-1} \big)$. Specifically, the challenge is accounting for how the error of estimator $\thetahat{}$ is impacted by the error of the estimated policies $\pihat{t}(\cdotspace) = \pi_t \big( \cdotspace; \betahat{t-1} \big)$ used to collect the data.

A key insight of this work is that \textit{a Z-estimator $\thetahat{}$ formed on adaptively sampled data can be framed as a Z-estimator in which the estimated policy parameters used to collect the data, $\betahat{1:T-1}$, are plug-in estimates of nuisance parameters.}
Classically on i.i.d. data, one constructs Z-estimators with a plug-in estimate of nuisance parameters which are fitted on the same data as the one used to form the Z-estimator itself \cite{tsiatis2006semiparametric}.
In deriving the asymptotic results for the Z-estimator in these classical settings, one must account for the dependence between the Z-estimator of interest and the plug-in estimator because they are constructed using a shared dataset. 

However, how to frame the inference after adaptive sampling problem as a problem of inference via a Z-estimator with a plug-in nuisance parameter estimator does not follow straightforwardly from the classical literature. Recall from the definition of $\thetahat{}$ from display \eqref{eqn:ZestimatorDef} that $\thetahat{}$ is formed by the data analyzer without constructing any plug-in estimators for nuisance parameters. We are interested in using the methods from the literature on plug-in estimates to account for the error of the policy parameters $\betahat{1:T-1}$, which impacted how the data was \textit{collected}. On i.i.d. data though, plug-in estimators do not affect \textit{data collection} and are only used for the \textit{data analysis}.

The critical step that will allow us to treat the policy parameters as plug-in nuisance parameters is to write the estimating function for the Z-estimator $\thetahat{}$ and the policy parameters $\betahat{1:T-1}$ jointly. The key tool we will use to do this is Radon-Nikodym derivative weighting.

\subsubsection{Radon-Nikodym Derivative Weights}
\label{sec:introRNweights}
Note the following ratios:
\begin{equation}
    \label{eqn:radonNikodym}
    \frac{ \pi_t \big( \cdotspace, \state{i}{t} ; \beta_{t-1} \big) }{ \pihat{t} \big( \cdotspace, \state{i}{t} \big) } : \MC{A} \mapsto [0, \infty).
\end{equation}
Conditional on $\history{1:n}{t-1}$ and $\state{i}{t}$, the functions $\pihat{t} \big( \cdotspace, \state{i}{t} \big)$ and $\pi_t \big( \cdotspace, \state{i}{t} ; \beta_{t-1} \big)$ each define a probability distribution over the action space $\MC{A}$. The ratio of these two probability distributions, as seen in display \eqref{eqn:radonNikodym} above, is a Radon-Nikodym derivative; see Lemma \ref{lemma:radonNikodym} for a formal statement of this result. Note that in the proof of Lemma \ref{lemma:radonNikodym}, we use the minimum exploration Condition \ref{cond:exploration} (Minimum Exploration), to ensure that these Radon-Nikodym derivatives exist. 

For notational convenience, we define the following weighting functions for any $\beta_{t-1}, \beta_{t-1}' \\
\in \real^{d_{t-1}}$,
\begin{equation}
    \label{eqn:weightsDef}
    \WW{i}{t}( \beta_{t-1}, \beta_{t-1}') ~ \triangleq ~ \frac{ \pi_t \big( \action{i}{t}, \state{i}{t}; \beta_{t-1} \big) }{ \pi_t \big( \action{i}{t}, \state{i}{t}; \beta_{t-1}' \big) }.
\end{equation}
Additionally, for any $\beta_{1:t-1}, \beta_{1:t-1}' \in \real^{d_{1:t-1}}$ (where $d_{1:t-1} \triangleq \sum_{t'=1}^{t-1} d_{t'}$) we define,
\begin{equation*}
    \WW{i}{2:t}( \beta_{1:t-1}, \beta_{1:t-1}') ~ \triangleq ~ \prod_{t'=2}^t \WW{i}{t'}( \beta_{t'-1}, \beta_{t'-1}').
\end{equation*}

The Radon-Nikodym weights above allow us to define estimating functions such as 
\begin{equation}
    \label{eqn:estDefWeights}
    \Est \big( \beta_{1:T-1}, \theta \big) 
    \triangleq \E_{\pi(\beta_{1:T-1})} \left[ \est{} \big( \history{i}{T}; \theta \big) \right]
    = \E \left[ \WW{i}{2:T}(\beta_{1:T-1}, \betahat{1:T-1}) \est{} \big( \history{i}{T}; \theta \big) \right].
\end{equation}
Above we use $\E$ to denote expectations with respect to the distribution used to collect the data. Thus, in the expectations from displays \eqref{eqn:estDefWeights} and \eqref{eqn:diffPolicy} above, the Radon-Nikodym weights have the effect of changing the distribution the expectation is taken over. Specifically, it changes the policy with which the actions are selected. Above we use the notation $\E_{\pi(\beta_{1:T-1})}$ to denote the expectation with respect to the distribution in which (i) the policies $\big\{ \pi_t(\cdotspace; \beta_{t-1}) \big\}_{t=2}^T$ are used to select actions and (ii) user potential outcomes are drawn from $\MC{P}$, as described in display \eqref{eqn:potentialOutcomes}.

We also define an empirical version of the limiting estimating function $\Est \big( \beta_{1:T-1}, \theta \big)$ above:
\begin{equation}
    \label{proofSktech:EstHat}
    \EstHat \big( \beta_{1:T-1}, \theta \big) \triangleq \frac{1}{n} \sum_{i=1}^n \WW{i}{2:T}(\beta_{1:T-1}, \betahat{1:T-1}) \est{} \big( \history{i}{T}; \theta \big).
\end{equation}

Additionally, in our proofs we will use the following limiting and empirical estimating functions for the policy parameters:
\begin{equation}
    \label{eqn:diffPolicy}
    \piEst{1:T-1}(\beta_{1:T-1}) \triangleq \E \begin{bmatrix}
        \piest1(\history{i}{1}; \beta_1) \\
        \WW{i}{2}(\beta_1, \betahat{1}) \piest2(\history{i}{2}; \beta_2) \\
        \WW{i}{2:3}(\beta_{1:2}, \betahat{1:2})  \piest3(\history{i}{3}; \beta_3) \\
        \vdots \\
        \WW{i}{2:T-1}(\beta_{1:T-2}, \betahat{1:T-2}) \piest{T-1}(\history{i}{T-1}; \beta_{T-1}) 
    \end{bmatrix}
\end{equation}
\begin{equation}
    \label{proofSketch:piEstHat}
    \piEstHat{1:T-1}(\beta_{1:T-1}) \triangleq \frac{1}{n} \sum_{i=1}^n \begin{bmatrix}
        \piest1(\history{i}{1}; \beta_1) \\
        \WW{i}{2}(\beta_1, \betahat{1}) \piest2(\history{i}{2}; \beta_2) \\
        \WW{i}{2:3}(\beta_{1:2}, \betahat{1:2}) \piest3(\history{i}{3}; \beta_3) \\
        \vdots \\
        \WW{i}{2:T-1}(\beta_{1:T-2}, \betahat{1:T-2}) \piest{T-1}(\history{i}{T-1}; \beta_{T-1}) 
    \end{bmatrix}
\end{equation}
%Above $\EstHat \big( \beta_{1:T-1}, \theta \big)$ is the empirical version of the limiting estimating function \\
%$\Est \big( \beta_{1:T-1}, \theta \big)$. Similarly, $\piEstHat{1:T-1}(\beta_{1:T-1})$ above is the empirical version of the limiting estimating function $\piEst{1:T-1} ( \beta_{1:T-1} )$.
%Additionally, above we use $\E$ to denote expectations with respect to the distribution used to collect the data. Thus, in the expectations from displays \eqref{eqn:estDefWeights} and \eqref{eqn:diffPolicy} above, the Radon-Nikodym weights have the effect of changing the distribution the expectation is taken over. Specifically, it changes the policy with which the actions are selected.

%We use these estimating functions defined above to write joint estimating functions for both the policy parameters $\beta_{1:T-1}$ and the parameter of interest $\theta$. 
We use these joint estimating functions for both the policy parameters $\beta_{1:T-1}$ and the parameter of interest $\theta$ to derive the joint limiting distribution of $\betahat{1:T-1}$ and $\thetahat{}$. Specifically, we prove that $\betahat{1:T-1}$ and $\thetahat{}$ are jointly asymptotically normal, as seen in display \eqref{mainNorm:limit} below; see Section \ref{sec:proofSketch} for a proof sketch. Note that since our adaptively sampled data is non-i.i.d. the proof of this result relies heavily on novel functional asymptotic normality results for Radon-Nikodym weighted functions on adaptively sampled data, which we discuss in detail in Section \ref{sec:functionalNormality}.

\begin{equation}
    \label{mainNorm:limit}
    \sqrt{n} \begin{pmatrix}
        \betahat{1:T-1} - \betastar{1:T-1} \\
        \thetahat{} - \thetastar{}
    \end{pmatrix} 
    \Dto \N \left( 0, \begin{bmatrix} 
        \piEstDotStar{1:T-1} && \bs{0} \\
        \bs{V}_{T,1:T-1} && \EstDotStar{}
    \end{bmatrix}^{-1} \Sigma_{1:T} \begin{bmatrix} 
        \piEstDotStar{1:T-1} && \bs{0} \\
        \bs{V}_{T,1:T-1} && \EstDotStar{}
    \end{bmatrix}^{-1, \top} \right).
\end{equation}
Above,
\begin{equation}
    \label{eqnMain:SigmaStackedDef}
    \Sigma_{1:T} \triangleq 
    \Estar{2:T} \left[ \begin{pmatrix}
        \piest1(\history{i}{1}; \betastar1) \\
        \piest2(\history{i}{2}; \betastar2) \\
        \vdots \\
        \piest{T-1}(\history{i}{T-1}; \betastar{T-1}) \\
        \est{}(\history{i}{T}; \thetastar{}) 
    \end{pmatrix}^{\otimes2} \right].
\end{equation}
and
\begin{equation}
    \label{eqnMain:stackedBread}
    \begin{bmatrix}
    \piEstDotStar{1:T-1} && \bs{0} \\
        \bs{V}_{T,1:T-1} && \EstDotStar{}
    \end{bmatrix}
    \triangleq \begin{bmatrix}
        \fracpartial{}{\beta_{1:T-1}} \piEst{1:T-1}(\beta_{1:T-1}) && \fracpartial{}{\theta} \piEst{1:T-1}(\beta_{1:T-1}) \\
        \fracpartial{}{\beta_{1:T-1}} \Est{}(\beta_{1:T-1}, \theta) && \fracpartial{}{\theta} \Est{}(\beta_{1:T-1}, \theta)
    \end{bmatrix} \bigg|_{( \beta_{1:T-1}, \theta ) = ( \betastar{1:T-1}, \thetastar{} )}.
\end{equation}
Note above that $\piEst{1:T-1}(\betastar{1:T-1})$ is not a function of $\theta$, thus $\fracpartial{}{\theta} \piEst{1:T-1}(\betastar{1:T-1}) \big|_{ \theta = \thetastar{} } = \bs{0}$.

Display \eqref{mainNorm:limit} above is sufficient for showing that $\sqrt{n} \big( \thetahat{} - \thetastar{} \big)$ is asymptotically normal with the adaptive sandwich variance from display \eqref{eqn:normalityStandardSandwich} holds. Specifically, by Lemma \ref{lemma:normalityAdaptiveUnpack} (Equivalent Formulations for the Adaptive Sandwich Variance), the lower $d_\theta \by d_\theta$ block of the limiting variance matrix from display \eqref{mainNorm:limit} above is equivalent to the adaptive sandwich variance $[\EstDotStar{}]^{-1} \Madapt{} [\EstDotStar{}]^{-1, \top}$ from display \eqref{eqn:normalityresult}.

%%%%%%%%%%%%%%%%%%%%%%%%%%%%%%%%%%%%%%%%%%%%%%%%%%%%%%%%
\subsubsection{Overview of Functional Asymptotic Normality Result for Adaptively Sampled Data}
\label{sec:functionalNormality}

The proof of the asymptotic normality result from display \eqref{mainNorm:limit} relies heavily on a novel functional asymptotic normality results we develop for adaptively sampled data. Functional asymptotic normality results are classical results from the empirical process literature that are used in many Z-estimator asymptotic normality proofs. These classical results concern stochastic processes of the following form:
\begin{equation}
	\label{eqn:empiricalProcessClassical}
	\bigg\{ \frac{1}{ \sqrt{n} } \sum_{i=1}^n \left( f(\history{i}{T}) - \E \big[ f(\history{i}{T}) \big] \right) \TN{~~s.t.~~} f \in \F \bigg\},
\end{equation}
for a class of functions $\F$.
If user data trajectories $\history{i}{T}$ were i.i.d. over $i \in [1 \colon n]$ and the complexity of a class of real-valued functions $\F$ was properly controlled, then the stochastic process from display \eqref{eqn:empiricalProcessClassical} would converge in distribution to a Gaussian process; see Theorem 19.5 of \cite{van2000asymptotic}.

For our inference after adaptive sampling problem, we are interested in showing a functional asymptotic normality result for stochastic processes like the following: 
\begin{multline}
	\label{eqn:empiricalProcessFull}
	\left\{ \sqrt{n} \big[ \EstHat(\beta_{1:T-1}, \theta) - \Est(\beta_{1:T-1}, \theta) \big] \TN{~~s.t.~~} \beta_{1:T-1} \in B_{1:T-1}, \theta \in \Theta \right\} \\
	= \bigg\{ \frac{1}{ \sqrt{n} } \sum_{i=1}^n \left( \WW{i}{2:T}(\beta_{1:T-1}, \betahat{1:T-1}) \psi(\history{i}{T}; \theta) - \E \left[ \WW{i}{2:T}(\beta_{1:T-1}, \betahat{1:T-1}) \psi(\history{i}{T}; \theta) \right] \right) \\
	\TN{~~s.t.~~} \beta_{1:T-1} \in B_{1:T-1}, ~ \theta \in \Theta\bigg\}.
\end{multline}
Above $B_{1:T-1} \subset \real^{d_{1:T-1}}$ and $\Theta \subset \real^{d_\theta}$ are compact balls whose interiors contain $\betastar{1:T-1}$ and $\thetastar{}$ respectively.
Also, recall that above we use the expectation $\E$ (not indexed by any policies) to refer to the distribution which was used to generate the data. Note that this means that $\E \left[ \WW{i}{2:T}(\beta_{1:T-1}, \betahat{1:T-1}) \psi(\history{i}{T}; \theta) \right] = \E_{\pi(\beta_{1:T-1})} \left[ \psi(\history{i}{T}; \theta) \right]$.

Note that to show that the stochastic process from display \eqref{eqn:empiricalProcessFull} is functionally asymptotically normal, it is sufficient to show a functional asymptotic normality result for empirical processes of the form
\begin{equation}
	\label{eqn:empiricalProcess}
	\bigg\{ \frac{1}{ \sqrt{n} } \sum_{i=1}^n \left( \rhohat{i}{2:T}f(\history{i}{T}) - \E \left[ \rhohat{i}{2:T} f(\history{i}{T}) \right] \right) \TN{~~s.t.~~} f \in \F \bigg\},
\end{equation}
where $\hat{\pi}^{(i)}_{2:T} \triangleq \prod_{t=2}^T \hat{\pi}_t \big( \action{i}{t}, \state{i}{t} \big)$.
Specifically, the stochastic process from display \eqref{eqn:empiricalProcessFull} is equivalent to the stochastic process in display \eqref{eqn:empiricalProcess} for the following choice of $\F$:
\begin{equation*}
	\F = \bigg\{ \bigg[ \prod_{t=2}^t \pi_t(\cdotspace; \beta_{t-1}) \bigg] \est{}(\cdotspace; \theta) \TN{~~s.t.~~} \beta_{1:T-1} \in B_{1:T-1}, \theta \in \Theta \bigg\}.
\end{equation*}

By Theorem 18.14 of \cite{van2000asymptotic}, the two conditions needed to ensure a functional normality result for display \eqref{eqn:empiricalProcess} are (i) a joint asymptotic normality result for the stochastic process evaluated at any finite number of functions $f_1, f_2, \dots, f_k \in \F$, and (ii) a maximal inequality result over the function class $\F$. To show part (i) holds for adaptively sampled data, we prove a Weighted Martingale Triangular Array Central Limit Theorem (Theorem \ref{thm:weightedCLT}). Specifically this Theorem can be used to show asymptotic normality results like the following:
\begin{multline*}
    \frac{1}{\sqrt{n}} \sum_{i=1}^n \left( \rhohat{i}{2:T} f(\history{i}{T}) - \E \left[ \rhohat{i}{2:T} f(\history{i}{T}) \right] \right) \\
    \Dto \N \bigg( 0, 
    ~\Estar{2:T} \left[ \big\{ \pi_{2:T}^{*,(i)} \big\}^{-2} f(\history{i}{T})^2 \right] - \Estar{2:T} \left[ \big\{ \pi_{2:T}^{*,(i)} \big\}^{-1} f(\history{i}{T}) \right]^2 \bigg).
\end{multline*}
%\begin{multline*}
%    \frac{1}{\sqrt{n}} \sum_{i=1}^n \bigg\{ \WW{i}{2:t}(\betastar{1:T-1}, \betahat{1:T-1}) f(\history{i}{T}) - \E \left[ \WW{i}{2:t}(\betastar{1:T-1}, \betahat{1:T-1}) f(\history{i}{T}) \right] \bigg\} \\
%    \Dto \N \left( 0, 
%    ~\Estar{2:T} \big[ f(\history{i}{T})^2 \big] - \Estar{2:T} \big[ f(\history{i}{T}) \big]^2 \right).
%\end{multline*}
Above $\pi^{*,(i)}_{2:T} \triangleq \prod_{t=2}^T \pistar{t} \big( \action{i}{t}, \state{i}{t} \big)$. The proof of our asymptotic normality result heavily relies on Lipschitz policy function Condition \ref{cond:lipschitzPolicy} and builds on the martingale Central Limit Theorem from Theorem 2.2 of \cite{dvoretzky1972asymptotic}.

%Functional asymptotic normality results require showing (a) a central limit result, and (b) a maximal inequality. For adaptively sampled data, we cannot use the standard limit theorems and maximal inequalities for i.i.d. data. Instead, for part (b) we develop a maximal inequality for adaptively sampled data built on a novel Bernstein type inequality that we prove holds using Radon-Nikodym derivative weightings; see Section \ref{sec:functionalNormality} and Appendix \ref{app:maximalInequalities} for more details on this inequality. For part (a) we use a Lindeberg martingale central limit theorem; this is why we require our brackets be bounded in $L_{2+\alpha}$ norm for some $\alpha > 0$, rather than the $L_{2}$ norm, which is standard for i.i.d. data.  

For part (ii), we prove a maximal inequality for adaptively sampled data as a function of the bracketing integral of $\F$, Lemma \ref{lemma:maximalBracketing}. We prove our maximal inequality, Lemma \ref{lemma:maximalBracketing}, using a novel Weighted Martingale Bernstein Inequality, Lemma \ref{lemma:bernstein}. This inequality modifies the classical Bernstein inequality for i.i.d. data \cite[Lemma 19.32]{van2000asymptotic}. 
Specifically, our Bernstein inequality ensures that on our adaptively sampled data type, for any real-valued function $f$ of $\history{i}{T}$ with $\| f \|_\infty < \infty$, 
\begin{multline}
	\label{main:berstein}
	\PP\bigg( \bigg| \frac{1}{ \sqrt{n} } \sum_{i=1}^n \rhohat{i}{2:T} f(\history{i}{T}) - \E \left[ \rhohat{i}{2:T} f(\history{i}{T}) \right] \bigg| > x \bigg) \\
	\leq 2 \exp \bigg( -\frac{ \pi_{\min}^{T-1} }{4} \frac{ x^2}{ \Estar{2:T} \big[ \rhostar{i}{2:T} f(\history{i}{T})^2 \big] + x \| f \|_\infty / \sqrt{n} } \bigg),
\end{multline}
for any $x > 0$ and $n \geq 1$. 
	
We now discuss the key techniques used in the proof of Lemma \ref{lemma:bernstein}.
Our proof of Lemma \ref{lemma:bernstein}, similar to the classical Bernstein inequality proof, starts by using a Chernoff bound to get an upper tail bound. Specifically, for any $\lambda > 0$,
\begin{equation*}
	\PP \bigg( \frac{1}{ \sqrt{n} } \sum_{i=1}^n \left( \rhohat{i}{2:T} f(\history{i}{T}) - \E \left[ \rhohat{i}{2:T} f(\history{i}{T}) \right] \right) > x \bigg)
\end{equation*}
\begin{equation*}
		\leq e^{-\lambda x } \E \bigg[ \exp \bigg\{ \frac{\lambda}{ \sqrt{n} } \sum_{i=1}^n \bigg( \rhohat{i}{2:T} f(\history{i}{T}) - \E \left[ \rhohat{i}{2:T} f(\history{i}{T}) \right] \bigg) \bigg\} \bigg].
\end{equation*}
Changing the summation in exponent into a product,
\begin{equation*}
	= e^{- \lambda x } \E \bigg[ \prod_{i=1}^n \exp \bigg\{ \frac{\lambda}{ \sqrt{n} } \bigg(\rhohat{i}{2:T} f(\history{i}{T}) - \E \left[ \rhohat{i}{2:T} f(\history{i}{T}) \right] \bigg) \bigg\} \bigg].
\end{equation*}

In the original proof for i.i.d. data, the next step is to move the product over $i=1, 2, \dots n$ above outside of the expectation. If we omit the terms $\rhohat{i}{2:T}$ above and if user trajectories $\history{i}{T}$ were i.i.d. over $i \in [1 \colon n]$, moving the product outside the expectation would be trivial. However, this is not the case for our adaptively sampled data setting.

The key insight we use in our proof is Lemma \ref{lemma:tightnessLemma}, a result that allows us to move products out of expectations using the $\rhohat{i}{2:T}$ weighting. Specifically, this Lemma proves that for any constant $c$,
\begin{equation*}
	 \E \bigg[ \prod_{i=1}^n \left( \rhohat{i}{2:T} f(\history{i}{T}) + c \right) \bigg]
	 = \prod_{i=1}^n \Estar{2:T} \bigg[ \rhostar{i}{2:T} f(\history{i}{T}) + c \bigg].
\end{equation*}
The proof leverages the conditional independence of the action selection at each time step and the fact that the underlying potential outcomes are i.i.d.
See Appendix \ref{app:maximalInequalities} for more details on all our maximal inequality results.

%%%%%%%%%%%%%%%%%%%%%%%%%%%%%%%%%%%%%%%%%%
%%%%%%%%%%%%%%%%%%%%%%%%%%%%%%%%%%%%%%%%%%
\subsection{Formal Statement of Results}
\label{sec:formalResults}
We now formally state the additional conditions we use to show consistency and asymptotic normality of $\thetahat{}$. Below we first provide assumptions on the estimated policy parameters $\betahat{1:T-1}$ (Section \ref{sec:policyConditions}) and then provide the theorem for, and assumptions on, the estimator $\thetahat{}$ based on the resulting adaptively sampled data (Section \ref{sec:theoremStatement}). The rationale for this is first,  separating out the conditions on the inferential estimator $\thetahat{}$ will also make explicit the conditions placed on the $\thetahat{}$ Z-estimator due to the adaptive sampling. A second consideration is that designers of adaptive sampling algorithms will know what assumptions on the algorithm are sufficient so that the resulting data can be used in a wide variety of after-study data analyses. 
A third consideration is that a data analyst who is provided an adaptively sampled dataset (with a known algorithm) can devise tests on the data to challenge the assumptions made on the policy parameters $\betahat{1:T-1}$.

\subsubsection{Assumptions on the Policy Parameters}
\label{sec:policyConditions}

We now discuss formally the remaining assumptions we place on the policy parameter estimators $\betahat{t}$ and their estimating functions $\piest{t}$. (Recall the assumption that estimators $\betahat{t}$ are consistent for $\betastar{t}$ and the assumptions placed on the policy function classes $\left\{ \pi_t \big( \cdotspace ; \beta_{t-1} \big) : \beta_{t-1} \in \real^{d_{t-1}} \right\}$ were introduced earlier in Section \ref{sec:policyAssumptions}.) The first of these assumptions, Condition \ref{cond:differentiabilityPolicy} below, will use the notation $\E_{\pi(\beta_{1:T-1})}$ to denote the expectation with respect to the distribution in which (i) the policies $\big\{ \pi_t(\cdotspace; \beta_{t-1}) \big\}_{t=2}^T$ are used to select actions and (ii) user potential outcomes are drawn from $\MC{P}$, as described in display \eqref{eqn:potentialOutcomes}.

\begin{condition}[Differentiability of Policy Parameter Estimating Functions]
    \label{cond:differentiabilityPolicy}
    The following mapping is differentiable at $\beta_{1:T-1} = \betastar{1:T-1}$
    \begin{equation}
        \label{eqn:diffPolicyCond}
        \beta_{1:T-1} \mapsto \piEst{1:T-1}(\beta_{1:T-1}) 
        \triangleq \E_{\pi(\beta_{1:T-1})} \begin{bmatrix}
        \piest1(\history{i}{1}; \beta_1) \\
        \piest2(\history{i}{2}; \beta_2) \\
        \vdots \\
        \piest{T-1}(\history{i}{T-1}; \beta_{T-1})
        \end{bmatrix}.
    \end{equation}
    Above the function $\piEst{1:T-1}(\beta_{1:T-1})$ was first defined in display \eqref{eqn:diffPolicyCond}.
    We also assume that for each $t \in [1 \colon T-1]$, the derivative matrix $\piEstDotStar{t} \triangleq \fracpartial{}{\beta_{t}} \Estar{2:t} \big[ \piest{t}(\history{i}{t}; \beta_{t}) \big] \big|_{\beta_{t} = \betastar{t}}$ is invertible. 
    
    Note in the expectation above in display \eqref{eqn:diffPolicyCond} that $\history{i}{t}$ only depends on the policies used to select actions up to decision time $t$, i.e., policies $\big\{ \pi_{t'}(\cdotspace; \beta_{t'-1}) \big\}_{t'=2}^t$. Thus, $\E_{\pi(\beta_{1:T-1})} \big[ \piest{t}(\history{i}{t}; \beta_t) \big] = \E_{\pi(\beta_{1:t-1})} \big[ \piest{t}(\history{i}{t}; \beta_t) \big]$.
\end{condition}

Condition \ref{cond:differentiabilityPolicy} ensures that the estimating functions for the policy parameters $\beta_{1:T-1}$ are differentiable at $\beta_{1:T-1} = \betastar{1:T-1}$. This kind of differentiability condition is common for Z-estimators \cite[Theorem 5.21]{van2000asymptotic}.
What is notable about Condition \ref{cond:differentiabilityPolicy} is that in display \eqref{eqn:diffPolicyCond}, the policy parameters $\beta_{1:T-1}$ parameterize not only the estimating functions $\piest{t}(\history{i}{t}; \beta_{t})$, they also parameterize the distribution with which the expectation is taken, $\E_{\pi(\beta_{1:T-1})}$.

The final assumption we place on the policies, Condition \ref{cond:lipschitzPolicyEstimatingFunction}, is a Lipschitz condition on the policy parameter estimating functions $\piest{t}$. This condition has the effect of restricting the complexity of the function class $\big\{ \piest{t}(\cdotspace; \beta_t) : \beta_t \in B_t \big\}$ and has been used in other standard proofs for the asymptotic normality of Z-estimators, e.g., see Theorem 5.21 of \cite{van2000asymptotic}.
Below, for each $t \in [1 \colon T-1]$, we use $B_t \subset \real^{d_t}$ to denote some compact subset whose interior contains $\betastar{t}$. 

\begin{condition}[Lipschitz Policy Estimating Function]
    \label{cond:lipschitzPolicyEstimatingFunction}
    Let $\alpha > 0$ be a constant. For each $t \in [2 \colon T]$, there is a non-negative valued function $\dot{\phi}_t(\history{i}{t})$ such that  
    \begin{enumerate}[label=(\roman*)]
        \item For any $\beta_{t}, \beta_{t}' \in B_{t}$,
        \begin{equation}
            \label{eqn:piestLipschitz}
            \big\| \piest{t} \big(\history{i}{t}; \beta_{t} \big) - \piest{t} \big( \history{i}{t}; \beta_{t}' \big) \big\|_2 
            \leq \dot{\phi}_t(\history{i}{t}) \big\| \beta_{t} - \beta_{t}' \big\|_2 ~~~ \TN{a.s.} 
        \end{equation}
        %%%%%%%%%%%%
        \item $\Estar{2:t} \big[ \big| \dot{\phi}_t(\history{i}{t}) \big|^{2+\alpha} \big] < \infty$.
        \label{condLipschitz:pidotPsidot}
        %%%%%%%%%%%%
        \item  $\Estar{2:t} \big[ \big| \dot{\phi}_t(\history{i}{t}) \dot{\pi}_{t'}(\action{i}{t'}, \state{i}{t'}) \big|^{2+\alpha} \big] < \infty$ for all $t' \in [2 \colon t]$ (the function $\dot{\pi}_{t'}$ is from Condition \ref{cond:lipschitzPolicy}). \\
        %%%%%%%%%%%%
    
    \noindent Additionally, for each $t \in [2 \colon T]$, let 
    \item $\Estar{2:t} \left[ \big\| \piest{t}(\history{i}{t}; \betastar{t}) \big\|_2^{2+\alpha} \right] < \infty$ and $\Estar{2:t} \left[ \big\| \piest{t}(\history{i}{t}; \betastar{t}) \dot{\pi}_{t'}(\action{i}{t'}, \state{i}{t'}) \big\|_2^{2+\alpha} \right] < \infty$ for all $t' \in [2 \colon t]$. 
    \label{condLipschitz:pidotPsi}
    \end{enumerate}
\end{condition}

In Condition \ref{cond:lipschitzPolicyEstimatingFunction} parts \ref{condLipschitz:pidotPsidot} and \ref{condLipschitz:pidotPsi} above, we assume finite moment conditions that involve the functions $\dot{\pi}_{t'}$ from Condition \ref{cond:lipschitzPolicy}. This will allow us to control the complexity of the function classes $\big\{ \big[ \prod_{t'=2}^t \pi_{t'}(\cdotspace; \beta_{t'-1}) \big] \piest{t}(\cdotspace; \beta_t) \TN{~~s.t.~~} \beta_{1:t} \in B_{1:t} \big\}$ for $t \in [2 \colon T-1]$; see the Remark below Theorem \ref{thm:asymptoticEquicontinuityPolicy} in Appendix \ref{mainapp:equicontinuityPolicy} for more details. Note that these assumptions involving $\dot{\pi}_{t'}$ in Condition \ref{cond:lipschitzPolicyEstimatingFunction} are relatively mild and are satisfied if $\Estar{2:t} \big[ \big| \dot{\pi}_{t}(\action{i}{t}, \state{i}{t}) \big|^{4+2\alpha} \big] < \infty$ for all $t \in [2 \colon T]$, and $\Estar{2:t} \big[ \big| \dot{\phi}_t(\history{i}{t}) \big|^{4+2\alpha} \big] < \infty$ and $\Estar{2:t} \left[ \big\| \piest{t}(\history{i}{t}; \betastar{t}) \big\|_2^{4+2\alpha} \right] < \infty$ for all $t \in [1 \colon T-1]$.

\begin{remark}[More General Policy Estimating Functions]
    We use Condition \ref{cond:lipschitzPolicyEstimatingFunction} to help ensure a stochastic equicontinuity result holds for the policy parameters. Condition \ref{cond:lipschitzPolicyEstimatingFunction} can be replaced by more general conditions involving bracketing numbers for function classes $\big\{ \piest{t}(\cdotspace; \beta_t) \TN{~~s.t.~~} \beta_t \in B_t \big\}$ for $t \in [1 \colon T-1]$. See Appendix \ref{mainapp:equicontinuityPolicy} for the statement of the stochastic equicontinuity result and the more general sufficient conditions.
\end{remark}

%%%%%%%%%%%%%%%%%%%%%%%%%%%%%%%%%%%%%%%%%%
%%%%%%%%%%%%%%%%%%%%%%%%%%%%%%%%%%%%%%%%%%
\subsubsection{Theorem Statements}
\label{sec:theoremStatement}

We have two main theorems. The first, Theorem \ref{thm:consistency}, shows the consistency of $\thetahat{}$, i.e., that $\thetahat{} \Pto \thetastar{}$.
The second, Theorem \ref{thm:normality}, shows that $\sqrt{n} \big( \thetahat{} - \thetastar{} \big)$ is asymptotically normal with the adaptive sandwich limiting variance from display \eqref{eqn:normalityresult}. Conditions \ref{cond:consistencyPolicy}-\ref{cond:lipschitzPolicyEstimatingFunction} introduced earlier in Sections \ref{sec:policyKeyAssumptions} and \ref{sec:policyConditions} are all the assumptions we make on the adaptive sampling algorithm for these two theorems. The remaining assumptions will concern the Z-estimation function $\est{}$ (used to define the inferential target $\thetastar{}$ and estimator $\thetahat{}$). We will require that $\est{}$ ``plays nicely'' with the adaptive sampling algorithm. In other words, there may be choices of $\est{}$ that make it incompatible with the adaptive sampling algorithm used to collect the data. 

We now state our main theorems. In the conditions for these theorems, we use bracketing numbers to control the complexity of function classes. The bracketing number of a class of real-valued functions $\F$ is the number of brackets, i.e., pairs functions, of a certain ``size'' needed to cover $\F$. Following the notation used in Chapter 19 of \cite{van2000asymptotic}, for any function class $\F$ of real-valued, measurable functions of $\history{i}{T}$, we use $N_{[~]} \big( \epsilon, \F, L_p(\Pstar) \big)$ to denote the number of brackets of size $\epsilon$ in $L_p(\Pstar)$ norm needed to cover $\F$; see Appendix \ref{sec:bracketingNumberDef} for a formal definition of bracketing numbers.

%%%%%%%%%%%%%%%%%%%%%%%%%%%%%%%%%
%%%%%%%%%%%%%%%%%%%%%%%%%%%%%%%%%
\begin{theorem}[Consistency]
    \label{thm:consistency}
    We assume that Conditions \ref{cond:consistencyPolicy}-\ref{cond:lipschitzPolicy} hold for the adaptive sampling algorithm. 
    Then  
    \begin{equation*}
        \thetahat{} \Pto \thetastar{}
    \end{equation*}
    under the following assumptions on the estimator $\thetahat{}$ and its corresponding estimating function $\est{}$:
    \begin{enumerate}[label=(C\arabic*)] %[label=Consistency (\roman*)]
        \item \bo{Well-Separated Solution:} 
        \label{consistency:wellSeparated}
        For any $\epsilon > 0$, there exists some $\eta > 0$ such that
        \begin{equation*}
        		\label{eqn:consistencySeparatedTheta}
            \inf_{ \theta \in \real^{d_\theta} \TN{~s.t.~} \| \theta - \thetastar{} \|_1 > \epsilon} \left\| \Estar{2:T} \big[ \est{}(\history{i}{T}; \theta) \big]  \right\|_1 > \eta > 0.
        \end{equation*}
        %\smallskip %%%%%%%%%%%%%%%%%%
        \item \bo{Asymptotically Tight:} 
        \label{consistency:tight}
        For any $\epsilon > 0$, there exists some $k < \infty$ such that
        \begin{equation*}
        		\label{eqn:consistencyTightTheta}
            \limsup_{n \to \infty} \PP \big( \big\| \thetahat{} \big\|_1 > k \big) \leq \epsilon.
         \end{equation*}
        %%%%%%%%%%%%%%%%%%%%%%%%%%%%%%%%%%%
        \item \bo{Finite Bracketing Number:}
         \label{consistency:finiteBracketing}
        Let $\alpha > 0$ be a constant. For any compact subset $K_\theta \subset \real^{d_\theta}$, 
        \begin{enumerate}[label=(\roman*)]
            \item For any $\epsilon > 0$ and any vector $c \in \real^{d_\theta}$, the bracketing number 
            \begin{equation}
                \label{condconsistency:bracketing}
                N_{[~]} \left( \epsilon, ~ \big\{ c^\top \est{}(\cdotspace; \theta) \TN{~~s.t.~~} \theta \in K_\theta \big\}, ~ L_{1+\alpha}(\Pstar) \right) < \infty.
            \end{equation}
            %%%%%%%%%%%%%
            \item There exists a function $F_{\est{}}$ where for any $\theta \in K_\theta$, $\big\| \est{}(\history{i}{T}; \theta ) \big\|_1 \leq F_{\est{}}(\history{i}{T})$ a.s. and 
            \begin{equation}
        		\label{condconsistency:envelope}
                \Estar{2:T} \left[ \big| F_{ \est{}}(\history{i}{T}) \dot{\pi}_t(\action{i}{t}, \state{i}{t}) \big|^{1+\alpha}
                \right] < \infty
            \end{equation}
            for all $t \in [2 \colon T]$; the functions $\dot{\pi}_t$ above are from Condition \ref{cond:lipschitzPolicy}.
        \end{enumerate}
    \end{enumerate}
\end{theorem}

Assumption \ref{consistency:wellSeparated} is used to ensure that there is a well-separated solution for the inferential quantity of interest, $\thetastar{}$; this also ensures that $\thetastar{}$ is a unique root of the Z-estimation criterion from display \eqref{eqn:ZestimandDef}. This well-separated condition is commonly used in consistency proofs for Z-estimators \cite[Theorem 5.9]{van2000asymptotic}. 

Next, assumption \ref{consistency:tight} is used to ensure that the estimator $\thetahat{}$ is asymptotically tight, i.e., does not tend towards infinity. In general, this assumption holds when the $\theta$ parameter space is bounded or when the estimating function $\est{}$ is the derivative of a concave function, e.g., see Theorem 5.14 of \cite{van2000asymptotic}.

Finally, assumption \ref{consistency:finiteBracketing} restricts the complexity of the function classes \\
$\big\{ c^\top \est{}(\cdotspace; \theta) \TN{~~s.t.~~} \theta \in K_\theta \big\}$ via bracketing numbers. For i.i.d. data, finite bracketing number conditions akin to display \eqref{condconsistency:bracketing} of assumption \ref{consistency:finiteBracketing} are used to show uniform law of large number results \cite[Theorem 19.4]{van1996weak}. However, since the adaptively sampled data is not i.i.d., we use assumption \ref{consistency:finiteBracketing} to prove a martingale version of a uniform law of large numbers (Theorem \ref{thm:weightedUWLLN}). Specifically, our martingale uniform law of large numbers will concern functions weighted by Radon-Nikodym derivatives. To facilitate use of these Radon-Nikodym weights we  control the bracketing complexity of the function class
\begin{equation}
    \label{eqn:FpiestDef}
    \F_{ \Pi c^\top \est{}}\big( B_{1:T-1}, K_\theta) \triangleq \bigg\{ \bigg[ \prod_{t=2}^T \pi_{t} ( \cdotspace; \beta_{t-1} ) \bigg] c^\top \est{}(\cdotspace; \theta \big) \TN{~~s.t.~~} \beta_{1:T-1} \in B_{1:T-1}, \theta \in K_\theta \bigg\}.
\end{equation}
In particular, we use display \eqref{condconsistency:envelope} of assumption \ref{consistency:finiteBracketing} to help ensure that for any $\epsilon > 0$ and any $c \in \real^{d_\theta}$, $N_{[~]} \left( \epsilon, ~ \F_{\Pi c^\top \est{} }( B_{1:T-1}, K_\theta), ~ L_{1+\alpha}(\Pstar) \right) < \infty$; see Lemma \ref{lemma:lipschitzPolicyProduct}.

\begin{theorem}[Asymptotic Normality]
    \label{thm:normality}
    We assume that Conditions \ref{cond:consistencyPolicy}-\ref{cond:lipschitzPolicyEstimatingFunction} hold for the adaptive sampling algorithm. 
    Furthermore, we assume that $\thetahat{} \Pto \thetastar{}$ holds (result of Theorem \ref{thm:consistency}). Then, for $\EstDotStar{} \triangleq \fracpartial{}{\theta} \Estar{2:T} \big[ \est{} \big( \history{i}{T}; \theta \big) \big] \big|_{\theta = \thetastar{}}$ and for $\Madapt{}$ as defined in display \eqref{eqn:Madapt},
    \begin{equation*}
	   \sqrt{n} \left( \thetahat{} - \thetastar{} \right)
	   \Dto \N \left( \bs{0}, ~ [\EstDotStar{}]^{-1} \Madapt{} [\EstDotStar{}]^{-1, \top} \right),
    \end{equation*}
    under the following additional assumptions:
    \begin{enumerate}[label=(N\arabic*)] %[label=Normality (\roman*)]
        %%%%%%%%%%%%%%%%%%%%%%%%%%%%%%%%%%%
        \item \bo{Invertible ``Bread'':}
        \label{normality:bread}
        The mapping $\theta \mapsto \Estar{2:T} \big[ \est{}\big( \history{i}{T}; \theta \big) \big]$ is differentiable at $\theta = \thetastar{}$. Moreover, the derivative matrix $\EstDotStar{} \triangleq \fracpartial{}{\theta} \Estar{2:T} \big[ \est{} \big( \history{i}{T}; \theta \big) \big] \big|_{\theta = \thetastar{}}$ is invertible.
        \smallskip
        %%%%%%%%%%%%%%%%%%%%%%%%%%%%%%%%%%%
        \item \bo{Differentiable with Respect to Policy Parameters:} 
        \label{normality:differentiable}
        
        \noindent The mapping $\beta_{1:T-1} \mapsto \E_{\pi_{2:T}(\beta_{1:T-1})} \big[ \est{}\big( \history{i}{T}; \thetastar{} \big) \big]$ is  differentiable at $\beta_{1:T-1} = \betastar{1:T-1}$.
        \smallskip
        %%%%%%%%%%%%%%%%%%%%%%%%%%%%%%%
        \item \bo{Continuity Condition:}
        \label{normality:L2convergence}
        The following mapping is continuous at $\theta = \thetastar{}$:
        \begin{equation*}
            \theta \mapsto 
            \Estar{2:T} \left[ \big\| \est{}\big(\history{i}{T}; \theta \big)
            - \est{} \big(\history{i}{T}; \thetastar{} \big) \big\|_2^2 \right].
        \end{equation*}
        %%%%%%%%%%%%%%%%%%%%%%%%%%%%%%%
        \item \bo{Finite Bracketing Integral:} 
        \label{normality:bracketing}
        Let $\Theta \subset \real^{d_\theta}$ be a compact subset whose interior contains $\thetastar{}$ and let $\alpha > 0$ be a constant. 
        \begin{enumerate}[label=(\roman*)]
            \item For any vector $c \in \real^{d_\theta}$,
            \begin{equation}
                \label{normalityeqn:bracketingIntegral}
                \int_0^1 \sqrt{ \log N_{[~]} \left( \epsilon, ~ \big\{ c^\top \est{}(\cdotspace; \theta) \TN{~~s.t.~~} \theta \in \Theta \big\}, ~ L_{2+\alpha}(\Pstar) \right) } d \epsilon < \infty.
            \end{equation}
            %%%%%%%%%%%%%%%%
            \item There exists a function $F_{\est{}}$ such that for all $\theta \in \Theta$, $\big\| \est{}(\history{i}{T}; \theta ) \big\|_1 \leq F_{\est{}}(\history{i}{T})$ a.s. and 
            \begin{equation}
                \label{normalityeqn:envelope}
                \Estar{2:T} \left[ \big| F_{ \est{}}(\history{i}{T}) \dot{\pi}_t(\action{i}{t}, \state{i}{t}) \big|^{2+\alpha} \right] < \infty
            \end{equation}
            for all $t \in [2 \colon T]$; the functions $\dot{\pi}_t$ above are from Condition \ref{cond:lipschitzPolicy}.
        \end{enumerate}
    \end{enumerate}
\end{theorem}

We now discuss each of the four parts of the assumptions of Theorem \ref{thm:normality} (Asymptotic Normality). Assumption \ref{normality:bread}  is used to ensure the ``bread'' part of the adaptive sandwich variance, $\EstDotStar{}$, is invertible. Note that the standard sandwich variance for Z-estimators on i.i.d. data also requires this condition to hold; see the discussion above display \eqref{eqn:normalityStandardSandwich}.

Assumption \ref{normality:differentiable} ensures that the estimating function $\E_{\pi_{2:T}(\beta_{1:T-1})} \big[ \est{}\big( \history{i}{T}; \thetastar{} \big) \big]$ is differentiable with respect to the policy parameters $\beta_{1:T-1}$. This condition is specific to the adaptively sampled data setting. The derivative $\fracpartial{}{\beta_{1:T-1}} \E_{\pi_{2:T}(\beta_{1:T-1})} \big[ \est{}\big( \history{i}{T}; \thetastar{} \big) \big] \big|_{\beta_{1:T-1} = \betastar{1:T-1}}$ is a component of the terms $M_t$ from $\Madapt{}$ in the adaptive sandwich variance; see display \eqref{eqn:Madapt} for the definition of $\Madapt{}$ and see display \eqref{eqn:Qmatrices} for the definition of $M_t$. 

Assumption \ref{normality:L2convergence} is used to ensure that if $\thetahat{} \Pto \thetastar{}$, then the $L_2(\Pstar)$ distance between $\est{} \big(\cdotspace; \thetahat{} \big)$ and $\est{}  \big(\cdotspace; \thetastar{} \big)$ converges in probability to zero. This type of condition is classically used in the empirical processes literature to show stochastic equicontinuity results \cite[Lemma 19.24]{van2000asymptotic}.

Finally, display \eqref{normalityeqn:bracketingIntegral} of assumption \ref{normality:bracketing} is a finite bracketing integral condition on the function class $\big\{ c^\top \est{}(\cdotspace; \theta) \TN{~~s.t.~~} \theta \in \Theta \big\}$ for any $c \in \real^{d_\theta}$. Note that if this function class is Lipschitz, this is sufficient for the finite bracketing integral condition in display \eqref{normalityeqn:bracketingIntegral} to hold \cite[Example 19.7]{van2000asymptotic}. For i.i.d. data, this kind of bracketing integral condition is used to show functional asymptotic normality results \cite[Theorem 19.5]{van2000asymptotic}. Similarly, we show a functional asymptotic normality result for adaptively sampled data using display \eqref{normalityeqn:bracketingIntegral}; this was discussed in more detail earlier in Section \ref{sec:functionalNormality}. Our central limit theorem considers functions weighted by particular Radon-Nikodym derivatives. Due to our use of these Radon-Nikodym weights, we will need to control the bracketing integral of the function class $\F_{ \Pi c^\top \est{}}\big( B_{1:T-1}, \Theta)$ as defined earlier in display \eqref{eqn:FpiestDef}. In particular, we use display \eqref{normalityeqn:envelope} of assumption \ref{normality:bracketing} to help ensure that for any $c \in \real^{d_\theta}$, $\int_0^1 \sqrt{ \log N_{[~]} \left( \epsilon, ~ \F_{\Pi c^\top \est{} }( B_{1:T-1}, \Theta), ~ L_{2+\alpha}(\Pstar) \right) } d\epsilon < \infty$; see Lemma \ref{lemma:lipschitzPolicyProduct}.

% 03b_proof_approach.tex
%%%%%%%%%%%%%%%%%%%%%%%%%%%%%%%%%%%%%%%%%%%%%%%%%%%%%%
%%%%%%%%%%%%%%%%%%%%%%%%%%%%%%%%%%%%%%%%%%%%%%%%%%%%%%
%%%%%%%%%%%%%%%%%%%%%%%%%%%%%%%%%%%%%%%%%%%%%%%%%%%%%

%%%%%%%%%%%%%%%%%%%%%%%%%%%%%%%%%%%%%%%%%%%%%%%%%%%%%%%%
\subsection{Proof Sketch for Theorem \ref{thm:normality} (Asymptotic Normality)}
\label{sec:proofSketch}

As first mentioned below display \eqref{mainNorm:limit}, by Lemma \ref{lemma:normalityAdaptiveUnpack} (Equivalent Formulations for the Adaptive Sandwich Variance), the following result is sufficient for the Theorem and will be the main result we show in this proof:
\begin{equation}
    \label{mainNorm:limit2}
    \sqrt{n} \begin{pmatrix}
        \betahat{1:T-1} - \betastar{1:T-1} \\
        \thetahat{} - \thetastar{}
    \end{pmatrix} 
    \Dto \N \left( 0, \begin{bmatrix} 
        \piEstDotStar{1:T-1} && \bs{0} \\
        \bs{V}_{T,1:T-1} && \EstDotStar{}
    \end{bmatrix}^{-1} \Sigma_{1:T} \begin{bmatrix} 
        \piEstDotStar{1:T-1} && \bs{0} \\
        \bs{V}_{T,1:T-1} && \EstDotStar{}
    \end{bmatrix}^{-1, \top} \right).
\end{equation}
The derivative terms $\EstDotStar{}$, $\bs{V}_{T,1:T-1}$, and $\piEstDotStar{1:T-1}$  above exist by assumptions \ref{normality:bread},  \ref{normality:differentiable}, and Condition \ref{cond:differentiabilityPolicy} respectively. 

We now state several equalities and discuss why they hold below:
\begin{multline}
    \label{eqn:proofMain1}
    - \sqrt{n} \begin{bmatrix}
        \piEstHat{1:T-1} \big( \betahat{1:T-1} \big) - \piEst{1:T-1} \big(\betahat{1:T-1} \big) \\
        \undermat{= o_P(1/\sqrt{n})}{ \EstHat \big( \betahat{1:T-1}, \thetahat{} \big) } - \Est \big(\betahat{1:T-1}, \thetahat{} \big)
    \end{bmatrix} \\
    ~~ \\
    \underbrace{=}_{(a)} - \sqrt{n} \begin{bmatrix}
        \piEst{1:T-1} \big( \betastar{1:T-1} \big) - \piEst{1:T-1} \big(\betahat{1:T-1} \big) \\
        \undermat{= 0}{ \Est \big( \betastar{1:T-1}, \thetastar{} \big) } - \Est \big(\betahat{1:T-1}, \thetahat{} \big)
    \end{bmatrix} + o_P(1) \\
    ~~ \\
    \underbrace{=}_{(b)} \sqrt{n} \begin{bmatrix}
    \piEstDotStar{1:T-1} && \bs{0} \\
        \bs{V}_{T,1:T-1} && \EstDotStar{}
    \end{bmatrix} \begin{bmatrix}
        \betahat{1:T-1} - \betastar{1:T-1} \\
        \thetahat{} - \thetastar{}
    \end{bmatrix} 
    + \sqrt{n} o_P \left( \left\| \begin{matrix}
        \betahat{1:T-1} - \betastar{1:T-1} \\
        \thetahat{} - \thetastar{}
    \end{matrix} \right\|_2 \right)
    + o_P(1).
\end{multline}

Equality (a) above holds since $\EstHat \big( \betahat{1:T-1}, \thetahat{} \big) = o_P(1/\sqrt{n})$  and $\Est \big( \betastar{1:T-1}, \thetastar{} \big) = 0$ by the definitions of $\thetahat{}$ and $\thetastar{}$ from displays \eqref{eqn:ZestimatorDef} and \eqref{eqn:ZestimandDef} respectively; 
also since $\piEstHat{1:T-1} \big( \betahat{1:T-1} \big) = o_P(1/\sqrt{n})$ and $\piEst{1:T-1} \big( \betastar{1:T-1} \big) = 0$ by the definitions of $\betahat{t}$ and $\betastar{t}$ from displays \eqref{eqn:betahatDef} and \eqref{eqn:betastarDef}.

Equality (b) above holds by a Taylor series expansion. Specifically by assumptions \ref{normality:bread} and \ref{normality:differentiable}, the mapping $\big( \beta_{1:T-1}, \theta \big) \mapsto \Est(\beta_{1:T-1}, \theta )$ is differentiable at $\big( \beta_{1:T-1}, \theta \big) = \big( \betastar{1:T-1}, \thetastar{} \big)$. Also  the mapping $\beta_{1:T-1} \mapsto \piEst{1:T-1}(\beta_{1:T-1})$ is differentiable at $\beta_{1:T-1} = \betastar{1:T-1}$ by Condition \ref{cond:differentiabilityPolicy}. As mentioned below display \eqref{eqnMain:stackedBread}, since $\piEst{1:T-1}(\betastar{1:T-1})$ is not a function of $\theta$, $\fracpartial{}{\theta} \piEst{1:T-1}(\betastar{1:T-1}) \big|_{ \theta = \thetastar{} } = \bs{0}$.

\medskip
We now state the next set of results and discuss why they hold below:
\begin{multline}
    \label{eqn:proofMain2}
    -\sqrt{n} \begin{bmatrix}
        \piEstHat{1:T-1} \big( \betahat{1:T-1} \big) - \piEst{1:T-1} \big(\betahat{1:T-1} \big) \\
        \EstHat \big( \betahat{1:T-1}, \thetahat{} \big) - \Est \big(\betahat{1:T-1}, \thetahat{} \big)
    \end{bmatrix} \\
    \underbrace{=}_{(c)} -\sqrt{n} \begin{bmatrix}
        \piEstHat{1:T-1} \big( \betastar{1:T-1} \big) - \piEst{1:T-1} \big(\betastar{1:T-1} \big) \\
        \EstHat \big( \betastar{1:T-1}, \thetastar{} \big) - \Est \big(\betastar{1:T-1}, \thetastar{} \big)
    \end{bmatrix} + o_P(1)
    \underbrace{\Dto}_{(d)} \N \big( 0, \Sigma_{1:T} \big).
\end{multline}

Equality (c) above is an stochastic equicontinuity result that holds by applying Lemma \ref{lemma:stochasticEquicontinuity} (Stochastic Equicontinuity). Lemma \ref{lemma:stochasticEquicontinuity} uses the fact that $\thetahat{} \Pto \thetastar{}$ (by assumption of the Theorem) and $\betahat{1:T-1} \Pto \betastar{1:T-1}$ (by Condition \ref{cond:consistencyPolicy}). Ensuring that the other assumptions needed to apply Lemma \ref{lemma:stochasticEquicontinuity} are satisfied is more involved; see the proofs of Theorems \ref{thm:asymptoticEquicontinuityPolicy} and \ref{thm:normality} for more details. The proof of Lemma \ref{lemma:stochasticEquicontinuity} relies on a functional asymptotic normality result that we prove holds on adaptively sampled data for functions weighted by the Radon-Nikodym derivatives (see Section \ref{sec:functionalNormality}).

Asymptotic normality result (d) above holds by Theorem \ref{thm:weightedCLT} (Weighted Martingale Triangular Array Central Limit Theorem). This Theorem proves a martingale central limit theorem for functions weighted by the Radon-Nikodym derivatives on adaptively sampled data.

\medskip
By consolidating the results from displays \eqref{eqn:proofMain1} and \eqref{eqn:proofMain2} above, and applying Slutsky's theorem we get the following result:
\begin{equation*}
    \sqrt{n} \begin{bmatrix}
    \piEstDotStar{1:T-1} && \bs{0} \\
        \bs{V}_{T,1:T-1} && \EstDotStar{}
    \end{bmatrix} \begin{bmatrix}
        \betahat{1:T-1} - \betastar{1:T-1} \\
        \thetahat{} - \thetastar{}
    \end{bmatrix} 
    + \sqrt{n} o_P \left( \left\| \begin{matrix}
        \betahat{1:T-1} - \betastar{1:T-1} \\
        \thetahat{} - \thetastar{}
    \end{matrix} \right\|_2 \right)
    \Dto \N \big( 0, \Sigma_{1:T} \big).
\end{equation*}

Note that $\EstDotStar{}$ is invertible by assumption \ref{normality:bread} and $\piEstDotStar{1:T-1}$ is invertible by Condition \ref{cond:differentiabilityPolicy} and Lemma \ref{lemma:invertibilityPiDotStar}. By Proposition \ref{prop:blockInversion} (Blockwise Inversion of Matrices), this is sufficient for $\begin{bmatrix}
    \piEstDotStar{1:T-1} && \bs{0} \\
        \bs{V}_{T,1:T-1} && \EstDotStar{}
    \end{bmatrix}$ to be invertible.
Thus, by the continuous mapping theorem,
\begin{multline*}
    \sqrt{n} \begin{bmatrix}
        \betahat{1:T-1} - \betastar{1:T-1} \\
        \thetahat{} - \thetastar{}
    \end{bmatrix} 
    + \sqrt{n} O(1) o_P \left( \left\| \begin{matrix}
        \betahat{1:T-1} - \betastar{1:T-1} \\
        \thetahat{} - \thetastar{}
    \end{matrix} \right\|_2 \right) \\
    \Dto \N \left( 0, \begin{bmatrix}
    \piEstDotStar{1:T-1} && \bs{0} \\
        \bs{V}_{T,1:T-1} && \EstDotStar{}
    \end{bmatrix}^{-1} \Sigma_{1:T} \begin{bmatrix}
    \piEstDotStar{1:T-1} && \bs{0} \\
        \bs{V}_{T,1:T-1} && \EstDotStar{}
    \end{bmatrix}^{-1, \top} \right).
\end{multline*}
The above implies that $\sqrt{n} \begin{bmatrix}
        \betahat{1:T-1} - \betastar{1:T-1} \\
        \thetahat{} - \thetastar{}
    \end{bmatrix} = O_P(1)$. This means that \\
    $\sqrt{n} O(1) o_P \left( \left\| \begin{matrix}
        \betahat{1:T-1} - \betastar{1:T-1} \\
        \thetahat{} - \thetastar{}
    \end{matrix} \right\|_2 \right) = o_P(1)$.
Thus, display \eqref{mainNorm:limit2} holds by Slutky's Theorem.

% 03c_extensions.tex
%%%%%%%%%%%%%%%%%%%%%%%%%%%%%%%%%%%%%%%%%%%%%%%%%%%%%%
%%%%%%%%%%%%%%%%%%%%%%%%%%%%%%%%%%%%%%%%%%%%%%%%%%%%%%
%%%%%%%%%%%%%%%%%%%%%%%%%%%%%%%%%%%%%%%%%%%%%%%%%%%%%
%%%%%%%%%%%%%%%%%%%%%%%%%%%%%%%%%%%%%%%%%%

\subsection{Cases in which the Adaptive and Standard Sandwich Variances are Equal}
\label{sec:simplifying}

In this section, we now discuss formally the conditions under which the adaptive sandwich variance equals the standard sandwich variance. Specifically, we show that this equivalence holds under a property of estimands $\thetastar{}$, which we call \textit{policy invariance}.

\begin{definition}[Policy Invariance]
    \label{def:policyInvariance}
    We say that $\thetastar{}$ is policy invariant if 
	\begin{equation}
	    \label{eqn:policyInvariance}
	    0 = \bs{V}_{T,1:T-1} \triangleq \fracpartial{}{\beta_{1:T-1}} \E_{\pi(\beta_{1:T-1})} \left[ \est{} \big( \history{i}{T}; \theta \big) \right] \bigg|_{ \beta_{1:T-1} = \betastar{1:T-1} }.
    \end{equation}
    We use also use the notation $\big[ V_{T,1}, V_{T,2}, \dots, V_{T,T-1} \big] \triangleq \bs{V}_{T,1:T-1} \in \real^{d_\theta \by d_{1:T-1}}$.
\end{definition}

Using the Radon-Nikodym derivative weighting from \eqref{eqn:weightsDef} we can equivalently write $V_{T, t}$ as follows:
\begin{equation}
    \label{eqn:Vdef}
    V_{T, t} \triangleq \fracpartial{}{\beta_t} \Estar{2:T} \left[ \WW{i}{t+1}(\beta_{t}, \betastar{t}) \est{}(\history{i}{T}; \thetastar{}) \right] \bigg|_{\beta_t = \betastar{t}}
    \in \real^{d_\theta \by d_{t}}.
\end{equation}
We can interpret $V_{T,t}$ as how the estimating function for $\thetastar{}$ changes with small changes in the limiting policy parameter $\betastar{t}$. A particular case in which $\thetastar{}$ is policy invariant was first discussed informally in Remark \ref{remark:misspecification}; specifically when $\est{}$ is chosen to be a derivative of the likelihood function of a model for a particular outcome, if that model is correctly specified then the estimand $\thetastar{}$ will not be a projection and will not depend on the target policy parameters $\betastar{1:T-1}$.

Recall from display \eqref{eqn:Madapt} that the adaptive sandwich variance is $[ \EstDotStar{} ]^{-1} \Madapt{} [ \EstDotStar{} ]^{-1}$ where
\begin{equation}
    \label{eqn:MadaptSecond}
    \Madapt{} \triangleq \Estar{2:T} \bigg[ \bigg\{ \est{} \big(\history{i}{T}; \thetastar{} \big) + \EstDotStar{} \sum_{t=1}^{T-1} M_t ~ \piest{t} \big( \history{i}{t} ; \betastar{t} \big) \bigg\}^{\otimes 2} \bigg].
\end{equation}
We now discuss how the policy invariance property will ensure that the sandwich and adaptive sandwich variances are equivalent. 

Recall from the proof sketch from Section \ref{sec:proofSketch} that by Lemma \ref{lemma:normalityAdaptiveUnpack} (Equivalent Formulations for the Adaptive Sandwich Variance), $[ \EstDotStar{} ]^{-1} \Madapt{} [ \EstDotStar{} ]^{-1}$, the adaptive sandwich variance, equals the lower-right $d_\theta \by d_\theta$ block of limiting variance from display \eqref{mainNorm:limit}, \begin{equation}
    \label{simplifying:stackedVariance}
    \begin{bmatrix} 
    \piEstDotStar{1:T-1} && \bs{0} \\
    \bs{V}_{T,1:T-1} && \EstDotStar{}
    \end{bmatrix}^{-1} \Sigma_{1:T} \begin{bmatrix} 
        \piEstDotStar{1:T-1} && \bs{0} \\
        \bs{V}_{T,1:T-1} && \EstDotStar{}
    \end{bmatrix}^{-1, \top}.
\end{equation}
By Proposition \ref{prop:blockInversion} (Blockwise Inversion of Matrices),
\begin{equation}
    \label{eqn:piEstStarDotInvdef}
    \begin{bmatrix}
    \piEstDotStar{1:T-1} && \bs{0} \\
        \bs{V}_{T,1:T-1} && \EstDotStar{}
    \end{bmatrix}^{-1} = \begin{bmatrix}
    \big\{ \piEstDotStar{1:T-1} \big\}^{-1} && \bs{0} \\
        -\big\{ \EstDotStar{} \big\}^{-1} \bs{V}_{T,1:T-1} \big\{ \piEstDotStar{1:T-1} \big\}^{-1} && \big\{ \EstDotStar{} \big\}^{-1}
    \end{bmatrix}.
\end{equation}

The matrices $M_t \in \real^{d_\theta \by d_t}$ in $\Madapt{}$ above are defined as the lower left block of the inverse matrix above in display \eqref{eqn:piEstStarDotInvdef}:
\begin{equation}
    \label{eqn:Qmatrices}
    \big[ M_1, M_2, \dots, M_{T-1} \big] 
    \triangleq -\big\{ \EstDotStar{} \big\}^{-1} \bs{V}_{T,1:T-1} \big\{ \piEstDotStar{1:T-1} \big\}^{-1} \in \real^{d_\theta \by d_{1:T-1}}.
\end{equation}
Above we use $d_{1:T-1} \triangleq \sum_{t=1}^{T-1} d_t$. % and $\piEstDotStar{1:T-1} \triangleq \fracpartial{}{\beta_{1:T-1}} \piEst{1:T-1}(\beta_{1:T-1})$; recall that the function $\piEst{1:T-1} ( \beta_{1:T-1} )$ above was defined earlier in display \eqref{eqn:diffPolicy}. 
It is clear from the definition of $M_t$ from display \eqref{eqn:Qmatrices} above that if $\bs{V}_{T,1:T-1} = \bs{0}$ (i.e., the policy invariance property holds), then $\Madapt{} = \Estar{2:T} \big[ \est{} \big(\history{i}{T}; \thetastar{} \big)^{\otimes 2} \big]$, and the limiting sandwich and adaptive sandwich variances are equivalent. 

Note that the asymptotic normality result with the standard sandwich variance for adaptively sampled longitudinal data does not follow from existing results (see Section \ref{sec:relatedWork} for further discussion of related work). Additionally, when the limiting standard and adaptive sandwich variances are equal, their variance estimators in general will not be equal in small samples (see the formulas for both the sandwich and adaptive sandwich variance estimators in Appendix \ref{app:simulationsEstimators}). In general, we advocate for using the adaptive sandwich variance over the standard sandwich variance since it is rare in digital intervention experiments for inference models to be exactly correctly specified.

%, which do not consider the longitudinal data setting and do not allow for model-misspecification of the algorithm's model

%Consider the following anaology. For maximum likelihood inference on i.i.d. data, the standard sandwich variance estimator is used over the inverse of the empirical Fisher information matrix to preserve validity of inference under misspecification of the likelihood model, and correlated and heteroskedastic errors \citep{white1982maximum}; when the likelihood model happens to be correctly specified and there happens to be uncorrelated and homoskedastic errors, the limiting sandwich variance simplifies to the inverse Fisher information matrix. Analogously, for adaptively collected, the adaptive sandwich variance estimator can be used over the standard sandwich variance estimator to preserve validity of inference for the best projected solution under adaptive sampling when the model is misspecified; when the model happens to be correctly specified, the limiting adaptive sandwich variance simplifies to the standard sandwich variance. 

% 05_simulation_results.tex
%%%%%%%%%%%%%%%%%%%%%%%%%%%%%%%%%%%%%%%%%%%%%%%%%%%%%%
%%%%%%%%%%%%%%%%%%%%%%%%%%%%%%%%%%%%%%%%%%%%%%%%%%%%%%
\section{Simulation Results} %%%%%%%%%%%%%%%%%%%
%%%%%%%%%%%%%%%%%%%%%%%%%%%%%%%%%%%%%%%%%%%%
\label{sec:simulations}
%%%%%%%%%%%%%%%%%%%%%%%%%%%%%%%%%%%%%%%%%%%%%%%%%%%%%
%%%%%%%%%%%%%%%%%%%%%%%%%%%%%%%%%%%%%%%%%%
%%%%%%%%%%%%%%%%%%%%%%%%%%%%%%%%%%%%%%%%%%%%

\subsection{Data Generation}
We consider a binary action setting with $T=50$ decision times. We use a Boltzmann (or Softmax) exploration type adaptive sampling algorithm \cite{asadi2017alternative,cesa2017boltzmann,sutton2018reinforcement}. The state used by the algorithm is the previous time step's reward, i.e., $\state{i}{t} = \big[ 1, R^{(i)}_{t-1} \big]$. Specifically, the adaptive sampling algorithm forms action selection probabilities as follows:
\begin{equation}
    \label{eqn:piSimulations}
    \pi_t \big(1, \state{i}{t}; \beta_{t-1} \big) = \TN{Clip}_{0.1} \left[ \TN{expit} \big( \rho \cdot \beta_{t-1,1}^\top \state{i}{t} \big) \right],
\end{equation}
where $\TN{Clip}_{0.1}[x] \triangleq \min( \max(x, 0.1), 0.9 )$. We can interpret $\beta_{t-1,1}^\top \state{i}{t}$ above as the adaptive sampling algorithm's working model of a treatment effect. Above the parameter $\rho$ is a positive constant that controls the steepness of the Softmax function; larger values of $\rho$ make the Softmax function steeper. We vary the value of $\rho$ in our experiments.
The policy parameter estimators $\betahat{t} = \big[ \betahat{t,0}, \betahat{t,1} \big]$ are those from the least squares example defined earlier in display \eqref{eqn:psiPolicyParameter}. 

We generate the rewards as follows:
\begin{equation}
    \label{eqn:rewardGen}
    R^{(i)}_{t} = \kappa_0 + \kappa_1 \bigg[ \frac{1}{c_\gamma} \sum_{t'=1}^{t-1} \gamma^{t-1-t'} \action{i}{t'} \bigg] + \kappa_2 \action{i}{t} + \epsilon^{(i)}_t.
\end{equation}
Above, the errors $\epsilon^{(i)}_t$ are generated so that they are correlated over time within a user. We use $\gamma = 0.95$, so $\sum_{t'=1}^{t-1} \gamma^{t-1-t'} \action{i}{t'}$ is a discounted sum of the user's recent ``dosage'', i.e., the number of times action $\action{i}{t'} = 1$ was previously chosen for that user. We create this dosage variable because the impact of dosage on user receptivity is of great interest in mobile health trials \cite{liao2020personalized,trellaPCS}.
The dosage is normalized by $c_\gamma \triangleq 1/(1-\gamma)$ to ensure the variable is between $[0, 1]$. In our experiments, we vary the magnitude of the dosage coefficient $\kappa_1$. Note that the reward model used by the adaptive sampling algorithm, from display \eqref{eqn:psiPolicyParameter}, is incorrectly specified, since it does not take dosage into account.
See Appendix \ref{app:simDetails} for more details.

\subsection{Data Analysis}

For inference we use the following choice of $\est{}$, which corresponds to a least squares criterion:
\begin{equation}
    \label{eqn:esteqnSimulations}
    \est{} \big( \history{i}{T}; \theta \big) = \sum_{t=1}^T \big( R^{(i)}_{t} - \theta_0^\top 
    \state{i}{t} - \theta_1 \action{i}{t} \big) \begin{bmatrix}
        \state{i}{t} \\
        \action{i}{t}
    \end{bmatrix}.
\end{equation}
Above $\theta_1 \in \real$ parameterizes the marginal treatment effect and $\theta_0 \in \real^{2}$ parameterize the model of the reward under $\action{i}{t} = 0$. Note that the data analysis model is misspecified because it does not take into account the dosage variable, which was used to generate rewards as described in display \eqref{eqn:rewardGen}. Thus, the parameters $\thetastar{1}$ and $\thetastar{0}$ are projections. When we vary the coefficient for the dosage variable, $\kappa_1$, we can see how results are affected by the degree of model misspecification.

We construct $95\%$ confidence intervals for $\thetastar{1}$, the projection of the treatment effect. We compare the empirical coverage of confidence intervals constructed using both the standard sandwich and adaptive sandwich variance estimators. We include the formulas for both the sandwich and adaptive sandwich variance estimators in Appendix \ref{app:simulationsEstimators}.

\medskip
\begin{table}[H]
\caption{\bo{Empirical Coverage of Confidence $95\%$ Intervals for Projected Treatment Effect $\thetastar{1}$.}  %All standard errors $< 1.5$. 
2000 Monte Carlo repetitions; standard errors are in parentheses.}
\begin{center}
{\renewcommand{\arraystretch}{1.2}%
\begin{tabular}{|l|l|l|l|l|}
\hline
\bo{Dosage Coeff.} & \bo{Alg. Steepness} & \bo{Sample Size} & \bo{Sandwich} & \bo{Adaptive Sandwich}  \\ \hline
\multirow{9}{*}{$\kappa_1=1$}
& \multirow{3}{*}{$\rho=0.5$} & $n=50$ & 93.65\% (0.55) & 96.95\% (0.39)  \\ \cline{3-5}
                            & & $n=100$ & 93.45\% (0.55) & 97\% (0.38) \\ \cline{3-5}
                            & & $n=500$ & 93.3\% (0.56) & 95.5\% (0.46)  \\ \cline{2-5}
& \multirow{3}{*}{$\rho=1$} & $n=50$ & 92.3\% (0.6) & 96.7\% (0.4)  \\ \cline{3-5}
                            & & $n=100$ & 90.85\% (0.65) & 97.3\% (0.36) \\ \cline{3-5}
                            & & $n=500$ & 89.85\% (0.68) & 95.6\% (0.46)  \\ \cline{2-5}
& \multirow{3}{*}{$\rho=5$} & $n=50$ & 75.8\% (0.96) & 95.4\% (0.47)  \\ \cline{3-5}
                            & & $n=100$ & 77.6\% (0.93) & 96.5\% (0.41) \\ \cline{3-5}
                            & & $n=500$ & 73.05\% (0.99) & 95.7\% (0.45)  \\ \hline
%%%%%%%%%%%%%%%%%%%%%%%%%%%%%%%%%%%%%%%%%%%%%%%%%%%%%%%%%%%%%%%
\multirow{9}{*}{$\kappa_1=5$}
& \multirow{3}{*}{$\rho=0.5$} & $n=50$ & 86.7\% (0.76) & 95.15\% (0.48) \\ \cline{3-5}
                            & & $n=100$ & 87.6\% (0.74) & 95.85\% (0.45) \\ \cline{3-5}
                            & & $n=500$ & 85.8\% (0.78) & 94.65\% (0.5) \\ \cline{2-5}
& \multirow{3}{*}{$\rho=1$} & $n=50$ & 83.5\% (0.83) & 96.2\% (0.43)  \\ \cline{3-5}
                            & & $n=100$ & 83.65\% (0.83) & 95.65\% (0.46) \\ \cline{3-5}
                            & & $n=500$ & 80.0\% (0.89) & 95.35\% (0.47)  \\ \cline{2-5}
& \multirow{3}{*}{$\rho=5$} & $n=50$ & 54.85\% (1.1) & 90.9\% (0.64) \\ \cline{3-5}
                            & & $n=100$ & 52.85\% (1.1) & 94.35\% (0.52) \\ \cline{3-5}
                            & & $n=500$ & 45.9\% (1.1) & 95.25\% (0.48) \\ \hline
\end{tabular}
}
\label{table:mainText}
\end{center}
\end{table}

\subsection{Discussion of Results} As seen in Table \ref{table:mainText}, the adaptive sandwich estimator consistently outperforms the standard sandwich variance estimator across all sample sizes and all simulation variants. Moreover, the performance gap increases with the magnitude of the dosage coefficient $\kappa_1$. This pattern is expected because as we discussed in Section \ref{sec:simplifying}, the adaptive and sandwich variances are equivalent when the inference model $\est{}$ is correctly specified; increasing the magnitude of $\kappa_1$ in the generative model increases the degree of model misspecification for our inference model from display \eqref{eqn:esteqnSimulations} because it does not include dosage. 

Additionally, note that the performance gap between the sandwich and adaptive sandwich variances increases with the algorithm's Softmax steepness parameter $\rho$. Note that when $\rho = 0$, then there is no adaptive sampling because the action selection probabilities from display \eqref{eqn:piSimulations} always equal $0.5$, i.e., $\pi_t \big(1, \state{i}{t}; \beta_{t-1} \big) = 0.5$ a.s. As one increases the value of $\rho$, the steeper the Softmax curve becomes, and the more ``adaptive'' the algorithm is allowed to be (the algorithm is able to make greater changes in the action selection probabilities). Specifically, when we increase $\rho$, we expect the norm of the matrices $V_{T, t}$ from display \eqref{eqn:Vdef} to grow. Recall that
\begin{equation*}
    V_{T, t} \triangleq 
    \fracpartial{}{\beta_t} \Estar{2:T} \left[ \WW{i}{t+1}(\beta_{t}, \betastar{t}) \est{}(\history{i}{T}; \thetastar{}) \right] \bigg|_{\beta_t = \betastar{t}}
    \in \real^{d_\theta \by d_{t}}
\end{equation*}
captures how the expectation of the estimating function changes with small changes in the policy parameter $\beta_t$. Note that as discussed in Section \ref{sec:simplifying}, when $V_{T, t} = 0$ for all $t \in [1 \colon T-1]$, then the limiting sandwich and adaptive sandwich variances are equivalent.

% 06_discussion.tex
%%%%%%%%%%%%%%%%%%%%%%%%%%%%%%%%%%%%%%%%%%%%
%%%%%%%%%%%%%%%%%%%%%%%%%%%%%%%%%%%%%%%%%%%%
%%%%%%%%%%%%%%%%%%%%%%%%%%%%%%%%%%%%%%%%%%%%
\section{Discussion} %%%%%%%%%%%%%%%%%%%
%%%%%%%%%%%%%%%%%%%%%%%%%%%%%%%%%%%%%%%%%%%%
%%%%%%%%%%%%%%%%%%%%%%%%%%%%%%%%%%%%%%%%%%%%

On adaptively sampled data the error of $\pihat{t}$ in estimating the target policy $\pistar{t}$ impacts what data is collected at the $t^{\TN{th}}$ decision time, and thus the error of future estimated policies ($\pihat{t'}$ for $t' > t$) and the final Z-estimator $\thetahat{}$. A key conceptual contribution of this work is to provide an approach to represent how the errors in the estimated policies impact the final Z-estimator $\thetahat{}$. In particular, we show that $\betahat{t-1}$, which parameterizes the estimated policy $\pihat{t}$, can be treated like a plug-in estimator for $\betastar{t-1}$ that was fit on the same dataset used to form $\thetahat{}$. In other words, even though on adaptively sampled data the estimated policy parameters affect the \textit{data collection}, they can be handled analogously to plug-in estimators for nuisance parameters that are used only in the \textit{data analysis}.

The greatest limitation of this work is that it does not apply to adaptively sampled batch datasets in which the policies used to collect the data are (i) not smooth in their parameters or (ii) allow the amount of exploration to go to zero. As discussed in Section \ref{sec:policyConditions}, many common RL algorithms for bandit and Markov decision process settings do not satisfy our smoothness and exploration conditions. However, the studies in digital interventions motivating this work will use adaptive sampling algorithms that do satisfy these conditions. 

Future work includes deriving efficient estimators based on adaptively sampled data and designing algorithms that are able to effectively pool over heterogeneous users. Another direction for future work will be to allow the estimating function itself, $\est{}$ to include the adaptive sampling action selection probabilities. For example, estimating functions used in off-policy analyses commonly include the action selection probabilities used to collect the data \cite{jiang2016doubly,kallus2020double,thomas2016data}. More generally, it would be of interest to extend this work to allow the estimating function $\est{}$ to include different types of plug-in estimators, e.g., plug-in estimates of the Q-function which are also often used in off-policy analyses \cite{jiang2016doubly,kallus2020double,thomas2016data}. There are also open questions about how to incorporate estimates formed by high-dimensional machine learning models to potentially increase the efficiency of estimators in these settings.

%\input{01_introduction}
%\input{01a_preliminaries}
%\input{01b_preliminaries}
%\input{02_related_work}
%\input{03_asymptotic_results}
%\input{03b_proof_approach}
%\input{03c_extensions}
%\input{05_simulation_results}
%\input{06_discussion}

%%%%%%%%%%%%%%%%%%%%%%%%%%%%%%%%%%%%%%%%%%%%%%
%% Funding information, if any,             %%
%% should be provided in the                %%
%% funding section.                         %%
%%%%%%%%%%%%%%%%%%%%%%%%%%%%%%%%%%%%%%%%%%%%%%
\subsection*{Acknowledgements}
%The first author was supported by NSF Grant DMS-??-??????.
%The second author was supported in part by NIH Grant ???????????.
Research reported in this paper was supported
%National Institute on Alcohol Abuse and Alcoholism (NIAAA) of the National Institutes of Health under award number R01AA23187, 
%National Institute on Drug Abuse (NIDA) of the National Institutes of Health under award number P50DA039838, 
%National Cancer Institute (NCI) of the National Institutes of Health under award number U01CA229437, 
%/NIBIB and OD
by NIH grants numbers P50DA05403, P41EB028242, and UG3DE028723. The content is solely the responsibility of the authors and does not necessarily represent the official views of the National Institutes of Health.

This material is based upon work supported by the National Science Foundation grant number NSF CBET–2112085. KWZ is also supported by the National Science Foundation Graduate Research Fellowship Program under Grant No. DGE1745303 and by a Siebel Scholars grant from the Siebel Foundation. %Any opinions, findings, and conclusions or recommendations expressed in this material are those of the author(s) and do not necessarily reflect the views of the National Science Foundation.

\subsection*{Overview of Appendices}
\begin{itemize}
    \item \bo{Appendix \ref{app:simulations}: Examples and Simulation Details.} 
    This appendix includes formulas for variance estimators, example RL algorithms that satisfy Conditions \ref{cond:exploration} and \ref{cond:lipschitzPolicy}, and Lemmas pertaining to bracketing numbers.
    %%%%%%%%%%%%%%%%%%%%
    \item \bo{Appendix \ref{app:algConditions}: Policy Parameter Results.} This appendix includes consistency and stochastic equicontinuity results for policy parameters $\betahat{1:T-1}$.
    %%%%%%%%%%%%%%%%%%%%
    \item \bo{Appendix \ref{app:mainResults}: Main Asymptotic Results.} 
    This appendix includes proofs of consistency of $\thetahat{}$ (Theorem \ref{thm:consistency}) and asymptotic normality of $\sqrt{n} \big( \thetahat{} - \thetastar{} \big)$ (Theorem \ref{thm:normality}).
    %%%%%%%%%%%%%%%%%%%%
    \item \bo{Appendix \ref{app:supporting}: Limit Theorems for Adaptively Collected Data.} 
    This appendix builds up results to show functional weak Law of Large Number and functional asymptotic normality results for Radon-Nikodym weighted empirical processes on adaptively collected data.
    %%%%%%%%%%%%%%%%%%%%
    \item \bo{Appendix \ref{app:maximalInequalities}: Maximal Inequalities for Adaptively Collected Data.} 
    This appendix builds up results to show a maximal inequality for Radon-Nikodym weighted empirical processes on adaptively collected data for function classes with finite bracketing integrals.
\end{itemize}

\bibliographystyle{plainnat}
%% if your bibliography is in bibtex format, uncomment commands:
%\bibliographystyle{imsart-number} % Style BST file (imsart-number.bst or imsart-nameyear.bst)
\bibliography{aos-pooling}       % Bibliography file (usually '*.bib')

\newpage
\begin{appendix}
%\addtocontents{toc}{\protect\setcounter{tocdepth}{1}}

%%%%%%%%%%%%%%%%%%%%%%%%%%%%%%%%%%%%%%%%%%%%
\section{Examples and Simulation Details}
\label{app:simulations}
%%%%%%%%%%%%%%%%%%%%%%%%%%%%%%%%%%%%%%%%%%%%
%%%%%%%%%%%%%%%%%%%%%%%%%%%%%%%%%%%%%%%%%%%%

\startOverview{Overview of Appendix \ref{app:simulations}}

\begin{itemize}
    \item \bo{Section \ref{app:simDetails}:} Simulation Details
    \begin{itemize}
        \item \bo{Section \ref{simDetail:rewardGen}:} Additional Information on Reward Generation
        \item \bo{Section \ref{simDetail:sandwich}:} Sandwich Variance Estimator
        \item \bo{Section \ref{app:simulationsEstimators}:} Adaptive Sandwich Variance Estimator
    \end{itemize}
    %%%%%%%%%%%%%
    \item \bo{Section \ref{appEx:examples}:} Example Algorithms that Satisfy Conditions \ref{cond:exploration} and \ref{cond:lipschitzPolicy}
    \begin{itemize}
        \item \bo{Section \ref{appEx:boltzmannExample}:} Boltzmann Exploration Algorithm (Lemma \ref{lemma:algExampleBoltzman})
        \item \bo{Section \ref{appEx:stochasticMirrorExample}:} Stochastic Mirror Descent Algorithm (Lemma \ref{lemma:stochasticMirrorExample})
    \end{itemize}
    %%%%%%%%%%%%%
    \item \bo{Section \ref{appsupport:RN}:} Radon-Nikodym Derivatives (Lemma \ref{lemma:radonNikodym})
    %%%%%%%%%%%%%
    \item \bo{Section \ref{mainapp:bracketingComplexity}:} Bracketing Numbers
    \begin{itemize}
        \item \bo{Section \ref{sec:bracketingNumberDef}:} Definition of Bracketing Numbers
        \item \bo{Section \ref{app:productLipschitz}:} Product of Lipschitz Policy Functions are Lipschitz (Lemma \ref{lemma:lipschitzPolicyProduct})
        \item \bo{Section \ref{app:productBracketing}:} Bracketing Number for Product of Function Classes (Lemma \ref{lemma:bracketingProduct})
    \end{itemize}
    %%%%%%%%%%%%%%%%%%%%%%%%%%%%%%%%%%%%%%%%%%%%%%%%%%%%%%%%%%%%
%%%%%%%%%%%%%%%%%%%%%%%%%%%%%%%%%%%%%%%%%%%%%%%%%%%%%%%%%%%%
\end{itemize}

%%%%%%%%%%%%%%%%%%%%%%%%%%%%%%%%%%%%%%
%%%%%%%%%%%%%%%%%%%%%%%%%%%%%%%%%%%%%%
\subsection{Simulation Details}
\label{app:simDetails}

\subsubsection{Additional Information on Reward Generation}
\label{simDetail:rewardGen}

In the reward generation formula from display \eqref{eqn:rewardGen}, for each user $i \in [1 \colon n]$, the errors $\epsilon_t^{(i)} \sim \N(0,1)$ marginally for each $t \in [1 \colon T]$; however, $\TN{Corr}(\epsilon_t^{(i)}, \epsilon_s^{(i)}) = 0.5^{|t-s|/2}$, which means the reward errors within a user are correlated over time.
Additionally, we set the parameters from display \eqref{eqn:rewardGen} to the following values: $\kappa_0 = 0$, $\kappa_2 = 0$, and we consider simulations with both $\kappa_1 = 1$ and $\kappa_1 = 5$.

\subsubsection{Sandwich Variance Estimator} %%%%%%%%%%%%%%%%%%%%%%%%%%%%%%%%%%%%%%
\label{simDetail:sandwich}

Recall from display \eqref{eqn:normalityStandardSandwich} that the sandwich variance is $[ \EstDotStar{} ]^{-1} \Sigma [ \EstDotStar{} ]^{-1, \top}$, where 
\begin{equation}
    \label{eqnapp:PsiSTarDef}
    \EstDotStar{} \triangleq \fracpartial{}{\theta} \Estar{2:T} \left[ \est{} \big( \history{i}{T}; \theta \big) \right] \bigg|_{\theta = \thetastar{}}
    \TN{~~~~~~and~~~~~~}
    \Sigma \triangleq \Estar{2:T} \left[ \est{} \big( \history{i}{T}; \thetastar{} \big)^{\otimes 2} \right].
\end{equation}
Above, we use the notation $x^{\otimes 2} \triangleq x x^\top$.

The sandwich variance \textit{estimator} we use is $\under{ \dot{\Psi} }^{-1,\top} \hat{\Sigma} \under{ \dot{\Psi} }^{-1,\top}$, where
\begin{equation}
    \label{eqnapp:PsiHatDef}
   \under{\dot{\Psi}} \triangleq \frac{1}{n} \sum_{i=1}^n \fracpartial{}{\theta}  \est{} \big( \history{i}{T}; \theta \big) \bigg|_{\theta = \thetahat{}}
    \TN{~~~~~~and~~~~~~}
    \hat{\Sigma} \triangleq \frac{1}{n} \sum_{i=1}^n \est{} \big( \history{i}{T}; \thetahat{} \big)^{\otimes 2}.
\end{equation}

\subsubsection{Adaptive Sandwich Variance Estimator} %%%%%%%%%%%%%%%%%%%%%%%%%%%%%%%%%%%%%%
\label{app:simulationsEstimators}

Recall from display \eqref{eqn:normalityresult} that the sandwich variance is $[ \EstDotStar{} ]^{-1} \Madapt{} [ \EstDotStar{} ]^{-1, \top}$, where $\EstDotStar{}$ is defined as in display \eqref{eqnapp:PsiSTarDef} and 
\begin{equation*}
    \Madapt{} \triangleq \Estar{2:T} \bigg[ \bigg\{ \est{} \big(\history{i}{T}; \thetastar{} \big) + \EstDotStar{} \sum_{t=1}^{T-1} M_t ~ \piest{t} \big( \history{i}{t} ; \betastar{t} \big) \bigg\}^{\otimes 2} \bigg].
\end{equation*}
By Lemma \ref{lemma:normalityAdaptiveUnpack}, the adaptive sandwich variance $[ \EstDotStar{} ]^{-1} \Madapt{} [ \EstDotStar{} ]^{-1, \top}$ equals the lower-right $\TN{dim}(\theta) \by \TN{dim}(\theta)$ block of the following matrix:
\begin{equation}
    \label{eqnapp:varianceStacked}
    \begin{bmatrix} 
        \piEstDotStar{1:T-1} && \bs{0} \\
        \bs{V}_{T,1:T-1} && \EstDotStar{}
    \end{bmatrix}^{-1} \Sigma_{1:T} \begin{bmatrix} 
        \piEstDotStar{1:T-1} && \bs{0} \\
        \bs{V}_{T,1:T-1} && \EstDotStar{}
    \end{bmatrix}^{-1, \top}
\end{equation}
where
\begin{equation*}
    \Sigma_{1:T} \triangleq 
    \Estar{2:T} \left[ \begin{pmatrix}
        \piest1(\history{i}{1}; \betastar1) \\
        \piest2(\history{i}{2}; \betastar2) \\
        \vdots \\
        \piest{T-1}(\history{i}{T-1}; \betastar{T-1}) \\
        \est{}(\history{i}{T}; \thetastar{}) 
    \end{pmatrix}^{\otimes2} \right].
\end{equation*}
and
\begin{equation*}
    \begin{bmatrix}
    \piEstDotStar{1:T-1} && \bs{0} \\
        \bs{V}_{T,1:T-1} && \EstDotStar{}
    \end{bmatrix}
    \triangleq \begin{bmatrix}
        \fracpartial{}{\beta_{1:T-1}} \piEst{1:T-1}(\beta_{1:T-1}) && \fracpartial{}{\theta} \piEst{1:T-1}(\beta_{1:T-1}) \\
        \fracpartial{}{\beta_{1:T-1}} \Est{}(\beta_{1:T-1}, \theta) && \fracpartial{}{\theta} \Est{}(\beta_{1:T-1}, \theta)
    \end{bmatrix} \bigg|_{( \beta_{1:T-1}, \theta ) = ( \betastar{1:T-1}, \thetastar{} )}.
\end{equation*}
Note above that $\fracpartial{}{\theta} \piEst{1:T-1}(\beta_{1:T-1}) = 0$ since $\piEst{1:T-1}(\beta_{1:T-1})$ is not a function of $\theta$. We estimate the entire matrix from display \eqref{eqnapp:varianceStacked} as follows:
\begin{equation*}
    \begin{bmatrix}
        \under{\dot{\Phi}}_{1:T-1} && \bs{0} \\
        \hat{\bs{V}}_{T,1:T-1} && \under{\dot{\Psi}}
    \end{bmatrix}^{-1}
    \hat{\Sigma}_{1:T} \begin{bmatrix}
        \under{\dot{\Phi}}_{1:T-1} && \bs{0} \\
        \hat{\bs{V}}_{T,1:T-1} && \under{\dot{\Psi}}
    \end{bmatrix}^{-1, \top}
\end{equation*}
where $\under{\dot{\Psi}}$ is defined in display \eqref{eqnapp:PsiHatDef},
\begin{equation*}
    \under{\dot{\Phi}}_{1:T-1} \triangleq \frac{1}{n} \sum_{i=1}^n \fracpartial{}{\beta_{1:T-1}} \begin{bmatrix}
        \piest1(\history{i}{1}; \beta_1) \\
        \WW{i}{2}(\beta_1, \betahat{1}) \piest2(\history{i}{2}; \beta_2) \\
        \WW{i}{2:3}(\beta_{1:2}, \betahat{1:2}) \piest3(\history{i}{3}; \beta_3) \\
        \vdots \\
        \WW{i}{2:T-1}(\beta_{1:T-2}, \betahat{1:T-2}) \piest{T-1}(\history{i}{T-1}; \beta_{T-1}) 
    \end{bmatrix} \bigg|_{\beta_{1:T-1} = \betahat{1:T-1}},
\end{equation*}
\begin{equation*}
    \hat{\bs{V}}_{T,1:T-1} \triangleq \frac{1}{n} \sum_{i=1}^n \bigg\{ \fracpartial{}{\beta_{1:T-1}} \WW{i}{2:T}(\beta_{1:T-1}, \betahat{1:T-1}) \bigg\} \bigg|_{\beta_{1:T-1} = \betahat{1:T-1}} \est{}(\history{i}{T}; \thetahat{}),
\end{equation*}
and
\begin{equation*}
    \hat{\Sigma}_{1:T} \triangleq \frac{1}{n} \sum_{i=1}^n \begin{pmatrix}
        \piest1(\history{i}{1}; \betahat1) \\
        \piest2(\history{i}{2}; \betahat2) \\
        \vdots \\
        \piest{T-1}(\history{i}{T-1}; \betahat{T-1}) \\
        \est{}(\history{i}{T}; \thetahat{}) 
    \end{pmatrix}^{\otimes2}.
\end{equation*}

%%%%%%%%%%%%%%%%%%%%%%%%%%%%%%%%%%%%%%
%%%%%%%%%%%%%%%%%%%%%%%%%%%%%%%%%%%%%%
\subsection{Example Algorithms that Satisfy Conditions \ref{cond:exploration} and \ref{cond:lipschitzPolicy}}
\label{appEx:examples}

\subsubsection{Boltzmann Exploration Algorithm (Lemma \ref{lemma:algExampleBoltzman})} %%%%%%%%%%%%%%%%%%%%%%%%%%%%
\label{appEx:boltzmannExample}

We consider the Boltzmann (or Softmax) exploration type adaptive sampling algorithm \cite{asadi2017alternative,cesa2017boltzmann,sutton2018reinforcement} as described in Section \ref{sec:simulations} (Simulation Results). Specifically, we consider a binary action setting ($\MC{A} = \{0, 1 \}$) and a Boltzmann sampling algorithm that forms action selection probabilities as follows:
\begin{equation}
    \label{eqn:piSimulationsBoltzmann}
    \pi_t \big(1, \state{i}{t}; \beta_{t-1} \big) = \TN{Clip}_{\pi_{\min}} \left[ \TN{expit} \big( \rho \cdot \beta_{t-1,1}^\top \state{i}{t} \big) \right],
\end{equation}
where $\TN{Clip}_{\pi_{\min}}[x] \triangleq \min( \max(x, \pi_{\min}), 1-\pi_{\min} )$. Above the parameter $\rho$ is a positive constant that controls the steepness of the Softmax function; larger values of $\rho$ make the Softmax function steeper. 
The policy parameters $\betahat{t} = \big[ \betahat{t,0}, \betahat{t,1} \big]$ are those from the least squares example defined earlier in display \eqref{eqn:psiPolicyParameter}. 

Note that exploration Condition \ref{cond:exploration} is satisfied because the action selection probabilities are constrained between $[\pi_{\min}, 1-\pi_{\min}]$. In Lemma \ref{lemma:algExampleBoltzman} below, we show that Condition \ref{cond:lipschitzPolicy} holds under the assumption that $\Estar{2:t} \big[ \big\| \state{i}{t} \big\|_2^{2+\alpha} \big] < \infty$ for all $t \in [2 \colon T]$.

\begin{lemma}[Boltzmann Exploration Algorithm]
    \label{lemma:algExampleBoltzman}
    We consider the Boltzmann algorithm example that selects actions as described in display \eqref{eqn:piSimulationsBoltzmann}. We show that Condition \ref{cond:lipschitzPolicy} holds under the condition that $\Estar{2:t} \big[ \big\| \state{i}{t} \big\|_2^{2+\alpha} \big] < \infty$ for all $t \in [2 \colon T]$.
\end{lemma}

%%%%%%%%%%%%%%%%%%%%%%%
\startproof{Lemma \ref{lemma:algExampleBoltzman}}
Note that for any $\beta_{1:t-1} \in \real^{d_{1:t-1}}$,
\begin{equation*}
    \big| \pi_t \big( 1, \state{i}{t}; \beta_{1:t-1} \big)
	- \pi_t \big( 1, \state{i}{t}; \beta_{1:t-1}^* \big) \big|
\end{equation*}
\begin{equation}
    \label{eqnexample:clipInequality}
    = \bigg| \TN{Clip}_{\pi_{\min}} \left[ \TN{expit} \big( \rho \cdot \beta_{t-1,1}^\top \state{i}{t} \big) \right]
    - \TN{Clip}_{\pi_{\min}} \left[ \TN{expit} \big( \rho \cdot (\betastar{t-1,1})^\top \state{i}{t} \big) \right] \bigg|.
\end{equation}
Note that for any real numbers $x,y$ that $\big| \TN{Clip}_{\pi_{\min}}(x) - \TN{Clip}_{\pi_{\min}}(y) \big| \leq | x-y |$. This is because 
\begin{itemize}
    \item If $x, y \in [\pi_{\min}, 1-\pi_{\min}]$, then $\big| \TN{Clip}_{\pi_{\min}}(x) - \TN{Clip}_{\pi_{\min}}(y) \big| = | x-y |$.
    \item If $x, y < \pi_{\min}$ or $x, y > 1-\pi_{\min}$, then $0 = \big| \TN{Clip}_{\pi_{\min}}(x) - \TN{Clip}_{\pi_{\min}}(y) \big| \leq | x-y |$.
    \item If $x > \pi_{\min}$ and $y < \pi_{\min}$, then $\big| \TN{Clip}_{\pi_{\min}}(x) - \TN{Clip}_{\pi_{\min}}(y) \big| \leq x-\TN{Clip}_{\pi_{\min}}(y) < x - y = |x-y|$. Same argument goes for the case that $y > \pi_{\min}$ and $x < \pi_{\min}$.
    \item If $x < 1-\pi_{\min}$ and $y > 1-\pi_{\min}$, then $\big| \TN{Clip}_{\pi_{\min}}(x) - \TN{Clip}_{\pi_{\min}}(y) \big| \leq \TN{Clip}_{\pi_{\min}}(y)-x 
    < y-x = |x-y|$. Same argument goes for the case that $y < 1-\pi_{\min}$ and $x > 1-\pi_{\min}$.
\end{itemize}
Thus,
\begin{equation}
    \label{eqnexample:boltzmanExpitDiff}
	\leq \left| \TN{expit} \big( \rho \cdot \beta_{t-1,1}^\top \state{i}{t} \big)
    - \TN{expit} \big( \rho \cdot (\betastar{t-1,1})^\top \state{i}{t} \big) \right|
\end{equation}

Note that for any function $f : \MC{X} \mapsto [0,1]$, for any $x, x' \in \MC{X}$,
\begin{equation*}
    |f(x) - f(x')| \leq \sup_{x_0 \in \MC{X}} \bigg| \fracpartial{}{x} f(x) \big|_{x = x_0} \bigg| | x - x' |.
\end{equation*}
Above $\sup_{x_0 \in \MC{X}} \big| \fracpartial{}{x} f(x) \big|_{x = x_0} \big|$ is the maximum absolute value of the derivative of $f$.
We can use the above observation to upper bound display \eqref{eqnexample:boltzmanExpitDiff} as follows:
\begin{equation}
    \label{eqnexample:boltzmannDerivative}
    \leq \sup_{x_0 \in \real} \bigg| \fracpartial{}{x} \TN{expit}(x) \big|_{x = x_0} \bigg|
    \cdot \big| \rho \cdot \beta_{t-1,1}^\top \state{i}{t} - \rho \cdot (\betastar{t-1,1})^\top \state{i}{t} \big|
\end{equation}
Note that $\fracpartial{}{x} \TN{expit}(x) = \TN{expit}(x) \big\{ 1 - \TN{expit}(x) \big\}$ and that $\sup_{p \in [0,1]} p (1-p) = 0.25$. Thus,
\begin{equation*}
    \sup_{x_0 \in \real} \bigg| \fracpartial{}{x} \TN{expit}(x) \big|_{x = x_0} \bigg|
    = \sup_{x_0 \in \real} \left| \TN{expit}(x_0) \big\{ 1 - \TN{expit}(x_0) \big\} \right|
    \leq \sup_{p \in [0,1]} \big| p (1-p) \big|
    \leq 0.25.
\end{equation*}

Thus, we can upper bound display \eqref{eqnexample:boltzmannDerivative} with the following
\begin{equation*}
    \leq 0.25 \big| \rho \cdot \beta_{t-1,1}^\top \state{i}{t} - \rho \cdot (\betastar{t-1,1})^\top \state{i}{t} \big|
\end{equation*}
\begin{equation*}
    = 0.25 \rho \big| \beta_{t-1,1}^\top \state{i}{t} - (\betastar{t-1,1})^\top \state{i}{t} \big|
\end{equation*}
By Cauchy Schwartz inequality,
\begin{equation*}
    \leq 0.25 \rho \big\| \beta_{t-1,1} - \betastar{t-1,1} \big\|_2 \big\| \state{i}{t} \big\|_2.
\end{equation*}
By the above result, Condition \ref{cond:lipschitzPolicy} holds since $\E\left[ \big\| \state{i}{t} \big\|_2^{2+\alpha} \right] < \infty$. ~~~$\blacksquare$

\subsubsection{Stochastic Mirror Descent Algorithm (Lemma \ref{lemma:stochasticMirrorExample})} %%%%%%%%%%%%%%%%%%%%%%%%%%%%
\label{appEx:stochasticMirrorExample}

We now give an example of an online stochastic mirror descent algorithm, based on those from 
\cite[pg 361]{lattimore2020bandit} and 
\cite{bubeck2012towards}, whose policy class satisfies Conditions \ref{cond:exploration} and \ref{cond:lipschitzPolicy}. We assume a  binary action setting with $\MC{A} = \{0, 1\}$ and assume that $\hat{\beta}_{t-1}^{(n)} = [\hat{\beta}_{t-1,0}^{(n)}, \hat{\beta}_{t-1,1}^{(n)}]$ below are estimated using the least squares criterion from display \eqref{eqn:psiPolicyParameter}. 

Note that for online stochastic mirror descent algorithms, $\pihat{t}$ is an updated version of $\pihat{t-1}$, which itself is an updated version of $\pihat{t-2}$, and so on. This means that parameters of the class $\pi_t$ must include those of $\pi_{t-1}, \pi_{t-2}, \dots, \pi_2$. We will use slightly non-standard notation to represent this, $\pihat{t}(\cdotspace) = \pi_t(\cdotspace; \betahat{1:t-1})$, where each $\betahat{t-1} = [\hat{\beta}_{t-1,0}^{(n)}, \hat{\beta}_{t-1,1}^{(n)}]$ is estimated using the least squares criterion from display \eqref{eqn:psiPolicyParameter}. Since we consider a binary action setting, to characterize a policy it is sufficient to define the probability that action $1$ is selected in each state.
%\sam{why did we include the term $\hat{\beta}_{t-1,0}^{(n), \top} \state{i}{t}$ in the criterion below?} \kwz{Usually for these stochastic mirror descent algorithms, they have a separate model for each action. Due to the way we've parameterized it, the term $\hat{\beta}_{t-1,0}^{(n), \top} \state{i}{t}$ is not necessary. But I include it so that it looks like more standard model formulation.}
\begin{multline}
	\label{eqn:stochasticMirrorDescent}
	\pihat{t} \big( 1, \state{i}{t} \big) 
	= \pi_t \big( 1, \state{i}{t}; \betahat{1:t-1} \big) \\
	= \TN{argmin}_{p \in [\pi_{\min}, 1-\pi_{\min}]} 
	\left\{ \eta_t \big( - \hat{\beta}_{t-1,0}^{(n), \top} \state{i}{t} - p ~ \hat{\beta}_{t-1,1}^{(n), \top} \state{i}{t} \big) 
	+ \big( \hat{\pi}_{t-1} \big( 1, \state{i}{t} \big) - p \big)^2 \right\}.
\end{multline}
Above, $\eta_t > 0$ is a learning rate and $\pi_{\min} \in (0, 0.5]$ is the minimum exploration rate. Note that $\hat{\beta}_{t-1,0}^{(n),\top} \state{i}{t} + p ~ \hat{\beta}_{t-1,1}^{(n), \top} \state{i}{t}$ is an estimate of the expectation of $R^{(i)}_{t}$ given $\state{i}{t}, \history{i}{t-1}$ when $\action{i}{t}$ is selected with probability $p$. Since the algorithm is designed to minimize a loss, we multiply $\hat{\beta}_{t-1,0}^{(n),\top} \state{i}{t} + p ~ \hat{\beta}_{t-1,1}^{(n), \top} \state{i}{t}$ by minus $1$ to ensure the algorithm is maximizing the reward (i.e., minimizing the negative reward). The term $\big( \hat{\pi}_{t-1} ( 1, \state{i}{t} ) - p \big)^2$ is a Bregman divergence and can be replaced by other Bregman divergences, e.g., KL-divergence. 

By display \eqref{eqn:stochasticMirrorDescent} above, we can derive
\begin{equation}
	\label{eqn:stochasticMirrorDescentExplicit}
	\pi_t \big( 1, \state{i}{t}; \betahat{1:t-1} \big)
	 = \TN{Clip}_{\pi_{\min}} \left( \pi_{t-1} \big( 1, \state{i}{t} ; \hat{\beta}_{1:t-2}^{(n)} \big) + \frac{1}{2} \eta_t \hat{\beta}_{t-1,1}^{(n), \top} \state{i}{t} \right),
\end{equation}
where $\TN{Clip}_{\pi_{\min}}(x) \triangleq \min \big( \max\big( x, \pi_{\min}\big), 1-\pi_{\min} \big)$; see Lemma \ref{lemma:stochasticMirrorExample} below for proof. 

Note that exploration Condition \ref{cond:exploration} is satisfied because the action selection probabilities are constrained between $[\pi_{\min}, 1-\pi_{\min}]$.
We can also show that Condition \ref{cond:lipschitzPolicy} holds because 
\begin{equation}
    \label{eqnApp:SGDlipschitz}
    \big| \pi_t \big( 1, \state{i}{t}; \beta_{1:t-1} \big)
	- \pi_t \big( 1, \state{i}{t}; \beta_{1:t-1}^* \big) \big|
	 \leq \frac{1}{2} \eta_t \big\| \state{i}{t} \big\| 
	 ~ \big\| \beta_{t-1,1} - \beta_{t-1,1}^* \big\|
\end{equation}
for any $\beta_{1:t-1} \in \real^{d_{1:t-1}}$, where $d_{1:t-1} \triangleq \sum_{t'=1}^{t-1} d_{t'}$; we also show this in Lemma \ref{lemma:stochasticMirrorExample} below.

\begin{lemma}[Stochastic Mirror Descent Algorithm]
    \label{lemma:stochasticMirrorExample}
    We consider the stochastic mirror descent algorithm example that selects actions as described in display \eqref{eqn:stochasticMirrorDescent}. We show that display \eqref{eqn:stochasticMirrorDescentExplicit} holds. We also show that Condition \ref{cond:lipschitzPolicy} holds under the conditions that \\
    (a) $\Estar{2:t} \big[ \big\| \state{i}{t} \big\|_2^{2+\alpha} \big] < \infty$ for all $t \in [2 \colon T]$ (the constant $\alpha > 0$ is the same as that from Condition \ref{cond:lipschitzPolicy}), and \\
    (b) the learning rates are bounded, i.e., for a constant $\eta_{\max}$, $\eta_t \leq \eta_{\max} < \infty$ for all $t \in [1 \colon T]$.
\end{lemma}

%%%%%%%%%%%%%%%%%%%%%%%
\startproof{Lemma \ref{lemma:stochasticMirrorExample}}

\proofSubsection{Showing display \eqref{eqn:stochasticMirrorDescentExplicit} holds}
Recall from display \eqref{eqn:stochasticMirrorDescent} that the stochastic mirror descent algorithm uses the following action selection probabilities:
\begin{multline*}
	\pihat{t} \big( 1, \state{i}{t} \big) 
	= \pi_t \big( 1, \state{i}{t}; \betahat{1:t-1} \big) \\
	= \TN{argmin}_{p \in [\pi_{\min}, 1-\pi_{\min}]} 
	\left\{ \eta_t \big( - \hat{\beta}_{t-1,0}^{(n), \top} \state{i}{t} - p ~ \hat{\beta}_{t-1,1}^{(n), \top} \state{i}{t} \big) 
	+ \big( \hat{\pi}_{t-1} \big( 1, \state{i}{t} \big) - p \big)^2 \right\}.
\end{multline*}

By taking the derivative of the following criterion with respect to $p$,
\begin{equation}
    \label{stochasticg:criterion}
    - \eta_t \big( \hat{\beta}_{t-1,0}^{(n), \top} \state{i}{t} + p ~ \hat{\beta}_{t-1,1}^{(n), \top} \state{i}{t} \big) 
	+ \big( \hat{\pi}_{t-1} \big( 1, \state{i}{t} \big) - p \big)^2,
\end{equation}
we have
\begin{equation*}
	- \eta_t ~ \hat{\beta}_{t-1,1}^{(n), \top} \state{i}{t}
	- 2 \big\{ \hat{\pi}_{t-1} \big( 1, \state{i}{t} \big) - p \big\}.
\end{equation*}

Since the second derivative of the criterion from display \eqref{stochasticg:criterion} with respect to $p$ is $2 > 0$, the global minimizer of the criterion (not restricted to $[\pi_{\min}, 1-\pi_{\min}]$) is $p = \hat{\pi}_{t-1} \big( 1, \state{i}{t} \big) + \frac{1}{2} \eta_t ~ \hat{\beta}_{t-1,1}^{(n), \top} \state{i}{t}$. Also note that the criterion from Equation \eqref{stochasticg:criterion} is convex because its derivative is strictly increasing in $p$. Note that the constrained minimizer of a convex function either equals the global minimizer or is on the boundary of the constraint space. Thus we have that the constrained minimizer, $\hat{\pi}_t \big( 1, \state{i}{t} \big)$, equals the following:
\begin{equation*}
	\pi_t \big( 1, \state{i}{t}; \hat{\beta}_{1:t-1}^{(n)} \big)
	= \TN{Clip}_{\pi_{\min}} \bigg( \hat{\pi}_{t-1} ( 1, \state{i}{t} ) + \frac{1}{2} \eta_t \hat{\beta}_{t-1,1}^{(n), \top} \state{i}{t} \bigg),
\end{equation*}
where $\TN{Clip}_{\pi_{\min}}(x) \triangleq \min \big( \max( x, \pi_{\min}), 1-\pi_{\min} \big)$. Thus, we have shown that display \eqref{eqn:stochasticMirrorDescentExplicit} holds.

%%%%%%%%%%%%%%%%%%%%%%%%%%
\proofSubsection{Showing Condition \ref{cond:lipschitzPolicy} holds}
Note that for any $\beta_{1:t-1} \in \real^{d_{1:t-1}}$,
\begin{equation*}
    \big| \pi_t \big( 1, \state{i}{t}; \beta_{1:t-1} \big)
	- \pi_t \big( 1, \state{i}{t}; \beta_{1:t-1}^* \big) \big|
\end{equation*}
\begin{equation*}
    %\label{eqnexample:clipInequality}
    = \bigg| \TN{Clip}_{\pi_{\min}} \bigg( \hat{\pi}_{t-1} ( 1, \state{i}{t} ) + \frac{1}{2} \eta_t \beta_{t-1,1}^\top \state{i}{t} \bigg)
    - \TN{Clip}_{\pi_{\min}} \bigg( \hat{\pi}_{t-1} ( 1, \state{i}{t} ) + \frac{1}{2} \eta_t \beta_{t-1,1}^{*, \top} \state{i}{t} \bigg) \bigg|.
\end{equation*}
Note that for any real numbers $x,y$ that $\big| \TN{Clip}_{\pi_{\min}}(x) - \TN{Clip}_{\pi_{\min}}(y) \big| \leq | x-y |$; the justification for this is discussed below display \eqref{eqnexample:clipInequality}. 
Thus, we have that display \eqref{eqnexample:clipInequality} can be upper bounded by the following:
\begin{equation*}
	\leq \left| \hat{\pi}_{t-1} ( 1, \state{i}{t} ) + \frac{1}{2} \eta_t \beta_{t-1,1}^\top \state{i}{t}
    - \hat{\pi}_{t-1} ( 1, \state{i}{t} ) - \frac{1}{2} \eta_t \beta_{t-1,1}^{*, \top} \state{i}{t} \right|
\end{equation*}
\begin{equation*}
	= \left| \frac{1}{2} \eta_t \left( \beta_{t-1,1} - \beta_{t-1,1}^* \right)^\top \state{i}{t} \right|
	\leq \frac{1}{2} \eta_t \big\| \state{i}{t} \big\|_2 
	 ~ \big\| \beta_{t-1,1} - \beta_{t-1,1}^* \big\|_2
\end{equation*}
The last inequality above holds by Cauchy-Schwartz.
By the above, Condition \ref{cond:lipschitzPolicy} holds since $\E\left[ \big\| \state{i}{t} \big\|_2^{2+\alpha} \right] < \infty$ and $\eta_t \leq \eta_{\max} < \infty$ for all $t \in [1 \colon T]$. $\blacksquare$

%%%%%%%%%%%%%%%%%%%%%%%%%%%%%%%%%%%%%%%%%%%%%%%%%%%%%%%%%%%%
%%%%%%%%%%%%%%%%%%%%%%%%%%%%%%%%%%%%%%%%%%%%%%%%%%%%%%%%%%%%
\subsection{Radon-Nikodym Derivatives (Lemma \ref{lemma:radonNikodym})}
\label{appsupport:RN}
\begin{lemma}[Radon-Nikodym Derivatives]
    \label{lemma:radonNikodym} 
    For any $\beta_{t-1} \in \real^{d_{t-1}}$, conditional on any $\state{i}{t}$, ~ $\mu(\cdotspace) \triangleq \pi_{t} \big( \cdotspace, \state{i}{t}; \beta_{t-1} \big)$ defines a probability measure on the sigma-algebra $\sigma(\action{i}{t})$. Also conditional on any $\history{1:n}{t-1}$ and $\state{i}{t}$, the policy $\nu(\cdotspace) \triangleq \pihat{t} \big( \cdotspace, \state{i}{t} \big)$ defines a probability measure on the sigma-algebra $\sigma(\action{i}{t})$. 
    
    Under Condition \ref{cond:exploration}, conditionally on almost every $\history{1:n}{t-1}$ and $\state{i}{t}$, $g(\cdotspace) \triangleq \frac{ \pi_t ( \cdotspace, \state{i}{t} ; \beta_{t-1} ) }{ \pihat{t} ( \cdotspace, \state{i}{t} ) }$ is a Radon-Nikodym derivative, i.e., for any measurable subset $\bar{A} \subseteq \MC{A}$, conditionally on almost every $\history{1:n}{t-1}$ and $\state{i}{t}$, $\mu(\bar{A}) = \int_{\bar{A}} g ~ d \nu$.
\end{lemma}

\startproof{Lemma \ref{lemma:radonNikodym}} 
We first show that that conditionally on almost every $\history{1:n}{t-1}$ and $\state{i}{t}$, $\mu(\cdotspace) = \pi_{t} \big( \cdotspace, \state{i}{t}; \beta_{t-1} \big)$ is absolutely continuous with respect to $\nu(\cdotspace) = \pihat{t} \big( \cdotspace, \state{i}{t} \big)$.

By exploration Condition \ref{cond:exploration} we have that for any measurable subset measurable subset $\bar{A} \subseteq \MC{A}$, $\pihat{t} \big( \bar{A}, \state{i}{t} \big) \geq \pi_{\min} > 0$ a.s. This means that for any measurable subset $\bar{A} \subseteq \MC{A}$, $\pihat{t} \big( \bar{A}, \state{i}{t} \big) \geq \pi_{\min} > 0$  conditionally on almost every $\history{1:n}{t-1}$ and $\state{i}{t}$. Thus we have that conditional on almost every $\history{1:n}{t-1}$ and $\state{i}{t}$, $\mu(\cdotspace) = \pi_{t} \big( \cdotspace, \state{i}{t}; \beta_{t-1} \big)$ is absolutely continuous with respect to $\nu(\cdotspace) = \pihat{t} \big( \cdotspace, \state{i}{t} \big)$. 

\vspace{0.5mm}
Thus, for some function $g: \MC{A} \mapsto [0, \infty)$, conditionally on almost every $\history{1:n}{t-1}$ and $\state{i}{t}$, $\mu(\bar{A}) = \int_{\bar{A}} g ~ d \nu$ for any measurable subset $\bar{A} \subseteq \sigma(\MC{A})$. This means that conditionally on almost every $\history{1:n}{t-1}$ and $\state{i}{t}$, 
\begin{equation*}
    \pi_{t} \big( \bar{A}, \state{i}{t}; \beta_{t-1} \big) = \int_{\bar{A}} g ~ d \pihat{t} ( \cdotspace, \state{i}{t} ).
\end{equation*}
For $g(\cdotspace) = \frac{ \pi_t ( \cdotspace, \state{i}{t} ; \beta_{t-1} ) }{ \pihat{t} ( \cdotspace, \state{i}{t} ) }$, the above equality is satisfied. $\blacksquare$

%%%%%%%%%%%%%%%%%%%%%%%%%%%%%%%%%%%%%%%%%%%%%%%%%%%%%%%%%%%%
%%%%%%%%%%%%%%%%%%%%%%%%%%%%%%%%%%%%%%%%%%%%%%%%%%%%%%%%%%%%
\subsection{Bracketing Numbers}
\label{mainapp:bracketingComplexity}

\subsubsection{Definition of Bracketing Numbers} %%%%%%%%%%%%%%%%%%%%%%%%%%%%%%%%%
\label{sec:bracketingNumberDef}

Following the notation used in Chapter 19 of \cite{van2000asymptotic}, for any function class $\F$ of real-valued functions of $\history{i}{T}$, we use $N_{[~]} \big( \epsilon, \F, L_p(\Pstar) \big)$ to denote the number of brackets of size $\epsilon$ in $L_P(\Pstar)$ norm needed to cover $\F$. Formally, this means we can find $N_\epsilon \triangleq N_{[~]} \big( \epsilon, \F, L_p(\Pstar) \big)$ number of brackets or pairs of real-valued functions of $\history{i}{T}$, $\big\{ (l_k, u_k) \big\}_{k = 1}^{N_\epsilon}$, such that (i) all brackets together cover $\F$, i.e., for any $f \in \F$ we can find some bracket $(l_k, u_k)$ such that $l_k(\history{i}{T}) \leq f(\history{i}{T}) \leq u_k(\history{i}{T})$ a.s., and (ii) the brackets have size less than $\epsilon$, i.e., $\Estar{2:T} \big[ \big| u_k(\history{i}{T}) - l_k(\history{i}{T}) \big|^p \big]^{1/p} < \epsilon$. As done in Chapter 19 of \cite{van2000asymptotic}, we assume that the bracketing functions themselves also have finite $L_p(\Pstar)$ norm, i.e., for any $\epsilon > 0$ and $ k \in [1 \colon N_\epsilon]$, $\Estar{2:T} \big[ | u_k(\history{i}{T}) |^p \big]^{1/p} < \infty$ and $\Estar{2:T} \big[ | l_k(\history{i}{T}) |^p \big]^{1/p} < \infty$.

\subsubsection{Product of Lipschitz Policy Functions are Lipschitz (Lemma \ref{lemma:lipschitzPolicyProduct})} %%%%%%%%%%%%%%%%%%%%%%%%%%%%%%%%%
\label{app:productLipschitz}

\begin{lemma}[Product of Lipschitz Policy Functions are Lipschitz]
    \label{lemma:lipschitzPolicyProduct}
    Let $t \in [3 \colon T-1]$. Under Condition \ref{cond:lipschitzPolicy} (Lipschitz Policy Function), for any $\beta_{1:t-1}, \tilde{\beta}_{1:t-1} \in B_{1:t-1}$,
    \begin{multline}
        \label{eqn:productLipschitzPolicy}
        \bigg| \prod_{t'=2}^t \pi_{t'} \big( \action{i}{t'}, \state{i}{t'}; \beta_{t'-1} \big)
        - \prod_{t'=2}^t \pi_{t'} \big( \action{i}{t'}, \state{i}{t'}; \tilde{\beta}_{t'-1} \big) \bigg| \\
        \leq \bigg\{ \sum_{t'=2}^t \dot{\pi}_{t'} \big( \action{i}{t'}, \state{i}{t'} \big) \bigg\}
        \big\| \beta_{1:t-1} - \tilde{\beta}_{1:t-1} \big\|_2 \TN{~~~a.s.} 
    \end{multline}
\end{lemma}

\startproof{Lemma \ref{lemma:lipschitzPolicyProduct}}
Note that by telescoping series,
\begin{equation*}
    \prod_{t'=2}^t \pi_{t'} \big( \action{i}{t'}, \state{i}{t'}; \beta_{t'-1} \big) - \prod_{t'=2}^t \pi_{t'} \big(\action{i}{t'}, \state{i}{t'}; \betatilde{t'-1} \big)
\end{equation*}
\begin{multline*}
    = \left[ \pi_2(\action{i}{2}, \state{i}{2}; \beta_1) - \pi_2(\action{i}{2}, \state{i}{2}; \betatilde1) \right]
    \bigg\{ \prod_{t'=3}^t \pi_{t'}(\action{i}{t'}, \state{i}{t'}; \beta_{t'-1}) \bigg\} \\ 
    + \left\{ \pi_2(\action{i}{2}, \state{i}{2}; \betatilde1) \right\} \left[ \pi_3(\action{i}{3}, \state{i}{3}; \beta_2) - \pi_3(\action{i}{3}, \state{i}{3}; \betatilde2) \right] 
    \bigg\{ \prod_{t'=4}^t \pi_{t'}(\action{i}{t'}, \state{i}{t'}; \beta_{t'-1}) \bigg\} \\
    + \hdots \\
    + \bigg\{ \prod_{t'=2}^{t-1} \pi_{t'}(\action{i}{t'}, \state{i}{t'}; \betatilde{t'-1}) \bigg\} \left[ \pi_{t}(\action{i}{t}, \state{i}{t}; \beta_{t-1}) - \pi_{t}(\action{i}{t}, \state{i}{t}; \betatilde{t-1}) \right]
\end{multline*}
\begin{multline*}
    = \sum_{t'=2}^{t} \bigg\{ \prod_{k=2}^{t'-1} \pi_k(\action{i}{k}, \state{i}{k}; \betatilde{k-1}) \bigg\} \left[ \pi_{t'}(\action{i}{t'}, \state{i}{t'}; \beta_{t'-1}) - \pi_{t'}(\action{i}{t'}, \state{i}{t'}; \betatilde{t'-1}) \right] \\
    \bigg\{ \prod_{k=t'+1}^{t} \pi_k(\action{i}{k}, \state{i}{k}; \beta_{k-1}) \bigg\}
\end{multline*}
By slight abuse of notation, above we use $\prod_{k=2}^{1} \pi_k(\action{i}{k}, \state{i}{k}; \betatilde{k-1}) = 1$ and \\
$\prod_{k=t+1}^{t} \pi_k(\action{i}{k}, \state{i}{k}; \beta_{k-1}) = 1$.

Using the above result and triangle inequality,
\begin{equation*}
    \bigg| \prod_{t'=2}^t \pi_{t'} \big( \action{i}{t'}, \state{i}{t'}; \beta_{t'-1} \big) - \prod_{t'=2}^t \pi_{t'} \big(\action{i}{t'}, \state{i}{t'}; \betatilde{t'-1} \big) \bigg|
\end{equation*}
\begin{multline*}
    \leq \sum_{t'=2}^{t} \bigg| \prod_{k=2}^{t'-1} \pi_k(\action{i}{k}, \state{i}{k}; \betatilde{k-1}) \bigg| \left| \pi_{t'}(\action{i}{t'}, \state{i}{t'}; \beta_{t'-1}) - \pi_{t'}(\action{i}{t'}, \state{i}{t'}; \betatilde{t'-1}) \right| \\
    \bigg| \prod_{k=t'+1}^{t} \pi_k(\action{i}{k}, \state{i}{k}; \beta_{k-1}) \bigg|
\end{multline*}
Since the terms $\big| \pi_k(\action{i}{k}, \state{i}{k}; \beta_{k-1}) \big|$ and $\big|\pi_k(\action{i}{k}, \state{i}{k}; \betatilde{k-1}) \big|$ are less than or equal to $1$ a.s.,
\begin{equation*}
    \leq \sum_{t'=2}^{t} \left| \pi_{t'}(\action{i}{t'}, \state{i}{t'}; \beta_{t'-1}) - \pi_{t'}(\action{i}{t'}, \state{i}{t'}; \betatilde{t'-1}) \right|
\end{equation*}
By Condition \ref{cond:lipschitzPolicy} (Lipschitz Policy Function),
\begin{equation*}
    \leq \sum_{t'=2}^{t} \dot{\pi}_{t'}(\action{i}{t'}, \state{i}{t'}) \big\| \beta_{t'-1} - \betatilde{t'-1} \big\|_2
\end{equation*}
\begin{equation*}
    \leq \bigg\{ \sum_{t'=2}^{t} \dot{\pi}_{t'}(\action{i}{t'}, \state{i}{t'}) \bigg\} \big\| \beta_{1:t-1} - \betatilde{1:t-1} \big\|_2
\end{equation*}
By the above argument, we have that
\begin{multline}
    \label{eqnapp:consistencyIntermed}
    \bigg| \prod_{t'=2}^t \pi_t(\action{i}{t'}, \state{i}{t'}; \beta_{t'-1}) - \prod_{t'=2}^t \pi_t(\action{i}{t'}, \state{i}{t'}; \betatilde{t'-1}) \bigg| \\
    \leq \bigg\{ \sum_{t'=2}^{t} \dot{\pi}_{t'}(\action{i}{t'}, \state{i}{t'}) \bigg\} \big\| \beta_{1:t-1} - \betatilde{1:t-1} \big\|_2 \TN{~~~a.s.} ~\blacksquare
\end{multline}

\subsubsection{Bracketing Number for Product of Function Classes (Lemma \ref{lemma:bracketingProduct})} %%%%%%%%%%%%%%%%%%%%%%%%%%%%%%%%%
\label{app:productBracketing}

\begin{lemma}[Bracketing Number for Product of Function Classes]
    \label{lemma:bracketingProduct}
    Let $t \in [2 \colon T-1]$. Let $\F$ be a class of real-valued functions of $\history{i}{t}$ indexed by $\lambda \in L \subseteq \real^{d_L}$, i.e., $\F \triangleq \big\{ f( \cdotspace; \lambda) : \lambda \in L \big\}$. Also let $\Pi_{2:t} \triangleq \big\{ \prod_{t'=2}^t \pi_{t'} \big( \cdotspace; \beta_{t'-1} \big) \TN{~~s.t.~~} \beta_{1:t-1} \in B_{1:t-1} \big\}$. Finally, also let
    \begin{equation*}
        \Pi_{2:t} \cdot \F \triangleq \bigg\{ \bigg[ \prod_{t'=2}^t \pi_{t'} \big( \cdotspace; \beta_{t'-1} \big) \bigg]  f(\cdotspace; \lambda) \TN{~~s.t.~~} \beta_{1:t-1} \in B_{1:t-1}, \lambda \in L \bigg\}.
    \end{equation*}
    For any $p \geq 1$, $N_{[~]} \big( \epsilon, \Pi_{2:t} \cdot \F, L_{p}(\Pstar) \big) < \infty$ for all $\epsilon > 0$ under the following conditions:
    \begin{enumerate}[label=(\roman*)]
        \item Condition \ref{cond:lipschitzPolicy} (Lipschitz Policy Function) holds; if $p > 2$, then Condition \ref{cond:lipschitzPolicy} must hold for $\alpha = p-2$.
        \label{bracketcond:condLipschitzPolicy}
        \item $N_{[~]} \big( \epsilon, \F, L_{p}(\Pstar) \big) < \infty$ for all $\epsilon > 0$.
        \label{bracketcond:finite}
        \item There exists a real-valued, measurable function $F$ of $\history{i}{t}$ such that (a) $\big| f(\history{i}{t}) \big| \leq F(\history{i}{t})$ a.s. for all $f \in \F$, and (b) $\Estar{2:t} \big[ \big| F(\history{i}{t}) \dot{\pi}_{t'}(\history{i}{t'}) \big|^p \big] < \infty$ for all $t' \in [2 \colon t]$, where the functions $\dot{\pi}_{t'}$ are from Condition \ref{cond:lipschitzPolicy}.
        \label{bracketcond:productMoment}
    \end{enumerate}
    Furthermore, $\int_0^1 \sqrt{ \log N_{[~]} \big( \epsilon, \Pi_{2:t} \cdot \F, L_{p}(\Pstar) \big) } d \epsilon < \infty$ under the additional condition that
    \begin{enumerate}[label=(\roman*)]
        \setcounter{enumi}{3}
        \item $\int_0^1 \sqrt{ \log N_{[~]} \big( \epsilon, \F, L_{p}(\Pstar) \big) } d \epsilon < \infty$
        \label{bracketcond:finiteIntegral}
    \end{enumerate}
    Note that assumption \ref{bracketcond:finiteIntegral} above implies that assumption \ref{bracketcond:finite} holds.
\end{lemma}

\begin{remark}[Bracketing Number for the Product of Policy and Estimating Functions]
    \label{remark:bracketingProduct}
    Let $\F_{ \Pi c^\top \est{}}\big( B_{1:T-1}, K_\theta) \triangleq \left\{ \big[ \prod_{t=2}^T \pi_{t} ( \cdotspace; \beta_{t-1} ) \big] c^\top \est{}(\cdotspace; \theta \big) \TN{~s.t.~} \beta_{1:T-1} \in B_{1:T-1}, \theta \in K_\theta \right\}$ for any compact set $K_\theta \subseteq \real^{d_\theta}$ and any $c \in \real^{d_\theta}$. By Lemma \ref{lemma:bracketingProduct}, we have the following results:
    \begin{enumerate}[label=(a\arabic*)] 
        \item Under Condition \ref{cond:lipschitzPolicy} and assumption \ref{consistency:finiteBracketing}, $\F_{ \Pi c^\top \est{}}\big( B_{1:T-1}, K_\theta)$ has a finite bracketing number, i.e., that $N_{[~]} \left( \epsilon, ~ \F_{\Pi c^\top\est{} }( B_{1:T-1}, K_\theta), ~ L_{1+\alpha}(\Pstar) \right) < \infty$ for any $\epsilon > 0$.
        \label{productBracket:finitePsi}
        \smallskip
        \item Under Condition \ref{cond:lipschitzPolicy} and assumption \ref{normality:bracketing}, $\F_{ \Pi c^\top \est{}}\big( B_{1:T-1}, \Theta)$ has a finite bracketing integral, i.e., that $\int_0^1 \sqrt{ \log N_{[~]} \left( \epsilon, ~ \F_{\Pi c^\top \est{} }( B_{1:T-1}, \Theta), ~ L_{2+\alpha}(\Pstar) \right) } d \epsilon < \infty$ for any $\epsilon > 0$.
        \label{productBracket:integralPsi}
    \end{enumerate}
    Let $\F_{\Pi c^\top \phi_t}(B_{1:t-1}, K_t) \triangleq \left\{ \big[ \prod_{t'=2}^t \pi_{t'} ( \cdotspace; \beta_{t'-1} ) \big] c^\top \piest{t}(\cdotspace; \beta_t \big) \TN{~s.t.~} \beta_{1:t-1} \in B_{1:t-1}, \beta_t \in K_t \right\}$ for any compact set $K_t \subseteq \real^{d_t}$ and any $c \in \real^{d_t}$. By Lemma \ref{lemma:bracketingProduct}, we have the following results:
    \begin{enumerate}[label=(b\arabic*)] 
        \item Under Condition \ref{cond:lipschitzPolicy} and assumption \ref{consistencyBeta:finiteBracketing}, $\F_{ \Pi c^\top \piest{t}}\big( B_{1:t-1}, K_t)$ has a finite bracketing number, i.e., that $N_{[~]} \left( \epsilon, ~ \F_{\Pi c^\top \piest{t} }( B_{1:t-1}, K_t), ~ L_{1+\alpha}(\Pstar) \right) < \infty$ for any $\epsilon > 0$.
        \label{productBracket:finitePhi}
        \smallskip
        \item Under Condition \ref{cond:lipschitzPolicy} and assumption \ref{normalityBeta:bracketing}, $\F_{ \Pi c^\top \piest{t}}\big( B_{1:t})$ has a finite bracketing integral, i.e., that $\int_0^1 \sqrt{ \log N_{[~]} \left( \epsilon, ~ \F_{\Pi c^\top \piest{t} }( B_{1:t}), ~ L_{2+\alpha}(\Pstar) \right) } d \epsilon < \infty$ for any $\epsilon > 0$.
        \label{productBracket:integralPhi}
    \end{enumerate}
\end{remark}

\startproof{Lemma \ref{lemma:bracketingProduct}}
For notational convenience, let
\begin{equation*}
    \pi_{2:t}(\history{i}{t}; \beta_{1:t-1}) \triangleq \prod_{t'=2}^t \pi_{t'} \big( \action{i}{t'}, \state{i}{t'}; \beta_{t'-1} \big).
\end{equation*}
By Lemma \ref{lemma:lipschitzPolicyProduct} (Product of Lipschitz Policy Function are Lipschitz) for any $\beta_{1:t-1}, \betatilde{1:t-1} \in B_{1:t-1}$,
\begin{equation}
    \label{eqnapp:zlipschitz}
	\big| \pi_{2:t}(\history{i}{t}; \beta_{1:t-1}) - \pi_{2:t}(\history{i}{t}; \beta_{1:t-1}) \big| \leq \dot{\pi}_{2:t}(\history{i}{t}) \big\| \beta_{1:t-1} - \betatilde{1:t-1} \big\|_2
\end{equation}
where $\dot{\pi}_{2:t}(\history{i}{t}) \triangleq \sum_{t'=2}^{t} \dot{\pi}_{t'} \big( \action{i}{t'}, \state{i}{t'} \big)$. Note that by assumption \ref{bracketcond:condLipschitzPolicy} (that Condition \ref{cond:lipschitzPolicy} holds), $\| \dot{\pi}_{2:t} \|_{\Pstar, p} \triangleq \Estar{2:t} \big[ | \dot{\pi}_{2:t}(\history{i}{t}) |^{p} \big]^{1/p} < \infty$.

\proofSubsection{Constructing bracketing functions that cover $\Pi_{2:t}$}
Let $\epsilon > 0$. We now use the approach from Lemma 19.7 of \cite{van2000asymptotic}. Since $B_{1:t-1}$ is compact, the size of $B_{1:t-1}$ in every fixed dimension is at most $\TN{diam}(B_{1:t-1}) < \infty$. We can cover $B_{1:t-1}$ with $\big[ \TN{diam}(B_{1:t-1}) / \epsilon \big]^{d_{1:t-1}}$ or fewer cubes of with edges of size $\epsilon$; recall that $d_{1:t-1} \triangleq \sum_{t'=1}^{t-1} d_{t'}$. Let the projection of the centers of each of these finitely many cubes onto $B_{1:t-1}$ be the points $\MC{B}_{1:t-1} \subset B_{1:t-1}$. Note that $| \MC{B}_{1:t-1} | = \big[ \TN{diam}(B_{1:t-1}) / \epsilon \big]^{d_{1:t-1}}$. For each of these cubes, consider the circumscribed ball that contains the cube; each of these balls has radius of $c\epsilon$ for some constant $0 < c < \infty$. 

We now construct a collection of bracketing functions that cover $\Pi$. These bracketing functions are
\begin{equation}
    \label{eqnapp:Pibrackets}
    \left\{ \left[  \pi_{2:t} \big( \cdotspace; \beta_{1:t-1} \big) - \epsilon \dot{\pi}_{2:t}(\cdotspace), 
    ~ \pi_{2:t} \big( \cdotspace; \beta_{1:t-1} \big) + \epsilon \dot{\pi}_{2:t}(\cdotspace) \right] \right\}_{\beta_{1:t-1} \in \MC{B}_{1:t-1}}.
\end{equation}
The above brackets are of size at most $2 \epsilon \| \dot{\pi}_{2:t} \|_{\Pstar, p}$ in $L_{p}(\Pstar)$ norm by assumption \ref{bracketcond:condLipschitzPolicy} (that Condition \ref{cond:lipschitzPolicy} holds).

We now discuss why the brackets from display \eqref{eqnapp:Pibrackets} cover $\Pi_{2:t}$. Consider any $\beta_{1:t-1} \in B_{1:t-1}$. Since the grid of cubes that cover $B_{1:t-1}$ whose projected centers form the collection of points $\MC{B}_{1:t-1}$ have edges of length $\epsilon$, thus there must exists some $\beta_{1:t-1}^{(k)} \in \MC{B}_{1:t-1}$ such that $\big\| \beta_{1:t-1} - \beta_{1:t-1}^{(k)} \big\|_2 \leq \epsilon$. By display \eqref{eqnapp:zlipschitz}, 
\begin{equation*}
    \pi_{2:t} \big( \history{i}{t}; \beta_{1:t-1} \big) \in \big[  \pi_{2:t} \big( \history{i}{t}; \beta_{1:t-1}^{(k)} \big) - \epsilon \dot{\pi}_{2:t}(\history{i}{t}),  \pi_{2:t} \big( \history{i}{t}; \beta_{1:t-1}^{(k)} \big) + \epsilon \dot{\pi}_{2:t}(\history{i}{t}) \big] \TN{~~a.s.}
\end{equation*}
Thus, we have that
\begin{equation}
    \label{app:PiBracketingUpperBound}
     N_{[~]} \left( 2 \epsilon \| \dot{\pi}_{2:t} \|_{\Pstar, p}, \Pi_{2:t} \cdot \F, L_p(\Pstar) \right) \leq |\MC{B}_{1:t-1}|
     = \big[ \TN{diam}(B_{1:t-1}) / \epsilon \big]^{d_{1:t-1}}.
\end{equation}

Additionally, note that $0 \leq \pi_{2:t}(\history{i}{t}; \beta_{1:t-1}) \leq 1$ a.s. since this function is a product of probabilities. Thus the brackets from display \eqref{eqnapp:Pibrackets} can be modified such that the bracketing functions are in $[0, 1]$ w.p. $1$ while maintaining coverage of $\Pi_{2:t}$ and not increasing the size of the brackets. Specifically, these brackets are:
\begin{multline}
    \label{eqnapp:PibracketsModified}
    \big\{ \big( l_{\Pi,k}, u_{\Pi, k} \big) \big\}_{k=1}^{ |\MC{B}_{1:t-1}| } \triangleq \\
    \bigg\{ \bigg[ \max \big\{ 0, \pi_{2:t} \big( \cdotspace; \beta_{1:t-1} \big) - \epsilon \pi_{2:t}(\cdotspace) \big\},  \min \big\{ 1,  \pi_{2:t} \big( \cdotspace; \beta_{1:t-1} \big) + \epsilon \pi_{2:t}(\cdotspace) \big\} \bigg] \bigg\}_{\beta_{1:t-1} \in \MC{B}_{1:t-1}}.
\end{multline}

\proofSubsection{Constructing bracketing functions that cover $\Pi_{2:t} \cdot \F$}
By assumption \ref{bracketcond:finite}, we can find $N_{\F,\epsilon} \triangleq N_{[~]} \big( \epsilon, \F, L_{p}(\Pstar) \big) < \infty$ bracketing functions which cover $\F$. We will call these bracketing functions $\big\{ \big[ l_{\F,k}, u_{\F, k} \big] \big\}_{k=1}^{N_{\F,\epsilon}}$. 
We now show that we can construct a finite collection of bracketing functions which cover $\Pi_{2:t} \cdot \F$ using the bracketing functions for $\F$ and $\Pi_{2:t}$. 

Consider any function $\pi_{2:t}(\cdotspace; \beta_{1:t-1}) f(\cdotspace) \in \Pi_{2:t} \cdot \F$. 
\begin{itemize}
    \item From display \eqref{eqnapp:PibracketsModified}, we can find some bracket $\big[ l_{\Pi,k}, u_{\Pi,k}\big]$ for $k \in [1 \colon |\MC{B}_{1:t-1}|]$ such that $0 \leq l_{\Pi,k}(\history{i}{t}) \leq \pi_{2:t}(\history{i}{t}; \beta_{1:t-1}) \leq u_{\Pi,k}(\history{i}{t}) \leq 1$ a.s. 
    \item Additionally, we can find some bracket $\big[ l_{\F,j}, u_{\F,j} \big]$ for $k \in [1 \colon N_{\F,\epsilon}]$ such that \\
    $l_{\F,j}(\history{i}{t}) \leq f(\history{i}{t}) \leq u_{\F,j}(\history{i}{t})$ a.s.
\end{itemize}
Now note the following observations for any particular $\history{i}{t}$:
\begin{itemize}
    \item If for a particular $\history{i}{t}$, $f(\history{i}{t}) \geq 0$, then since $0 \leq l_{\Pi,k}(\history{i}{t}) \leq u_{\Pi,k}(\history{i}{t}) \leq 1$, then
    \begin{equation*}
        l_{\Pi,k}(\history{i}{t}) \cdot l_{\F,j}(\history{i}{t}) 
        \leq \pi_{2:t}(\history{i}{t}; \beta_{1:t-1}) f(\history{i}{t})
        \leq u_{\Pi,k}(\history{i}{t}) \cdot u_{\F,j}(\history{i}{t}).
    \end{equation*}
    \item If for a particular $\history{i}{t}$, $f(\history{i}{t}) < 0$, then since $0 \leq l_{\Pi,k}(\history{i}{t}) \leq u_{\Pi,k}(\history{i}{t}) \leq 1$, then
    \begin{equation*}
        u_{\Pi,k}(\history{i}{t}) \cdot l_{\F,j}(\history{i}{t}) 
        \leq \pi_{2:t}(\history{i}{t}; \beta_{1:t-1}) f(\history{i}{t})
        \leq l_{\Pi,k}(\history{i}{t}) \cdot u_{\F,j}(\history{i}{t}).
    \end{equation*}
\end{itemize}
By the above two observations, we have that
\begin{multline*}
    \min \left\{ l_{\Pi,k}(\history{i}{t}) \cdot l_{\F,j}(\history{i}{t}), u_{\Pi,k}(\history{i}{t}) \cdot l_{\F,j}(\history{i}{t}) \right\} 
    \leq \pi_{2:t}(\history{i}{t}; \beta_{1:t-1}) f(\history{i}{t}) \\
    \leq \max \left\{ u_{\Pi, k}(\history{i}{t}) \cdot u_{\F, j}(\history{i}{t}), l_{\Pi, k}(\history{i}{t}) \cdot u_{\F, j}(\history{i}{t}) \right\} \TN{~~a.s.}
\end{multline*}
Thus, the following bracketing functions cover $\Pi_{2:t} \cdot \F$:
\begin{multline}
    \label{eqnapp:PiFbrackets}
    \big\{ \big( l_{\Pi_{2:t} \cdot \F,k}, u_{\Pi_{2:t} \cdot \F, k} \big) \big\}_{k=1}^{ |\MC{B}_{1:t-1}| \cdot N_{\F,\epsilon} } \\
    \triangleq \bigg\{ \bigg[ \min \big( l_{\Pi,k} \cdot l_{\F,j}, u_{\Pi,k} \cdot l_{\F,j} \big) , \max \big( u_{\Pi, k} \cdot u_{\F, j}, l_{\Pi, k} \cdot u_{\F, j} \big) \bigg] \bigg\}_{k=1;j=1}^{k=|\MC{B}_{1:t-1}|;j=N_{\F,\epsilon}}.
\end{multline}
Note that there are $|\MC{B}_{1:t-1}| \cdot N_{\F,\epsilon}$ brackets above.

We now derive the size of the above brackets for $\Pi_{2:t} \cdot \F$. 
\begin{equation*}
    \left| \max \big( u_{\Pi, k} \cdot u_{\F, j}, l_{\Pi, k} \cdot u_{\F, j} \big) - \min \big( l_{\Pi,k} \cdot l_{\F,j}, u_{\Pi,k} \cdot l_{\F,j} \big) \right|
\end{equation*}
Since $\big| \max(a, b) - c \big| \leq | a - c | + |b - c|$,
\begin{multline*}
    \leq \left| u_{\Pi, k} \cdot u_{\F, j} - \min \big( l_{\Pi,k} \cdot l_{\F,j}, u_{\Pi,k} \cdot l_{\F,j} \big) \right| \\
    + \left| l_{\Pi, k} \cdot u_{\F, j} - \min \big( l_{\Pi,k} \cdot l_{\F,j}, u_{\Pi,k} \cdot l_{\F,j} \big) \right| \TN{~~a.s.}
\end{multline*}
\begin{multline*}
    \leq \left| u_{\Pi, k} \cdot u_{\F, j} - l_{\Pi,k} \cdot l_{\F,j} \right| + \left| u_{\Pi, k} \cdot u_{\F, j} - u_{\Pi,k} \cdot l_{\F,j} \right| \\
    + \left| l_{\Pi, k} \cdot u_{\F, j} - l_{\Pi,k} \cdot l_{\F,j} \right| + \left| l_{\Pi, k} \cdot u_{\F, j} - u_{\Pi,k} \cdot l_{\F,j} \right| \TN{~~a.s.}
\end{multline*}
Since $0 \leq u_{\Pi, k}(\history{i}{t}) \leq 1$ a.s. and $0 \leq l_{\Pi, k}(\history{i}{t}) \leq 1$ a.s.,
\begin{multline*}
    = \left| u_{\Pi, k} \cdot u_{\F, j} - l_{\Pi,k} \cdot l_{\F,j} \right| + \left| l_{\Pi, k} \cdot u_{\F, j} - u_{\Pi,k} \cdot l_{\F,j} \right| \\
    +  u_{\Pi, k} \left| u_{\F, j} - l_{\F,j} \right| + l_{\Pi, k} \left| u_{\F, j} - l_{\F,j} \right|
\end{multline*}
Again since $0 \leq u_{\Pi, k}(\history{i}{t}) \leq 1$ a.s. and $0 \leq l_{\Pi, k}(\history{i}{t}) \leq 1$,
\begin{equation*}
    \leq \left| u_{\Pi, k} \cdot u_{\F, j} - l_{\Pi,k} \cdot l_{\F,j} \right| + \left| l_{\Pi, k} \cdot u_{\F, j} - u_{\Pi,k} \cdot l_{\F,j} \right| 
    +  2 \left| u_{\F, j} - l_{\F,j} \right| \TN{~~a.s.}
\end{equation*}
By triangle inequality,
\begin{multline*}
    \leq \left| u_{\Pi, k} \cdot u_{\F, j} - u_{\Pi, k} \cdot l_{\F,j} \right| + \left| u_{\Pi, k} \cdot l_{\F,j} - l_{\Pi,k} \cdot l_{\F,j} \right| \\
    + \left| l_{\Pi, k} \cdot u_{\F, j} - l_{\Pi, k} \cdot l_{\F,j} \right| + \left| l_{\Pi, k} \cdot l_{\F,j} - u_{\Pi,k} \cdot l_{\F,j} \right| 
    +  2 \left| u_{\F, j} - l_{\F,j} \right| \TN{~~a.s.}
\end{multline*}
Using the same arguments as used above,
\begin{equation*}
    \leq \left| u_{\F, j} - l_{\F,j} \right| + \left| u_{\Pi, k} - l_{\Pi,k}  \right| | l_{\F,j} | 
    + \left| u_{\F, j} - l_{\F,j} \right| + \left| l_{\Pi, k} - u_{\Pi,k} \right| |l_{\F,j} |
    +  2 \left| u_{\F, j} - l_{\F,j} \right| \TN{~a.s.}
\end{equation*}
\begin{equation*}
    = 2 \left| u_{\Pi, k} - l_{\Pi,k}  \right| | l_{\F,j} | 
    +  4 \left| u_{\F, j} - l_{\F,j} \right| 
\end{equation*}
Thus, 
\begin{multline*}
    \Estar{2:t} \bigg[ \bigg| \max \big\{ u_{\Pi, k}(\history{i}{t}) \cdot u_{\F, j}(\history{i}{t}), l_{\Pi, k}(\history{i}{t}) \cdot u_{\F, j}(\history{i}{t}) \big\} \\
    - \min \big\{ l_{\Pi,k}(\history{i}{t}) \cdot l_{\F,j}(\history{i}{t}), u_{\Pi,k}(\history{i}{t}) \cdot l_{\F,j}(\history{i}{t}) \big\} \bigg|^p \bigg]
\end{multline*}
\begin{equation*}
    \leq \Estar{2:t} \bigg[ \bigg| 2 \big| u_{\Pi, k}(\history{i}{t}) - l_{\Pi,k}(\history{i}{t}) \big| | l_{\F,j}(\history{i}{t}) | +  4 \big| u_{\F, j}(\history{i}{t}) - l_{\F,j}(\history{i}{t}) \big|  \bigg|^p \bigg]
\end{equation*}
By Lemma \ref{lemma:binomialBound} (Inequality using Binomial Theorem), for some constant $c_p < \infty$,
\begin{multline*}
    \leq 2^p c_p \Estar{2:t} \left[ \left| u_{\Pi, k}(\history{i}{t}) - l_{\Pi,k}(\history{i}{t}) \right|^p | l_{\F,j}(\history{i}{t}) |^p \right]  \\
    + 4^p c_p \Estar{2:t} \left[ \left| u_{\F, j}(\history{i}{t}) - l_{\F,j}(\history{i}{t}) \right|^p \right]
\end{multline*}
Since brackets $[l_{\F,j}, u_{\F, j}]$ are of size $\epsilon$ or less in $L_p(\Pstar)$ norm by construction,
\begin{equation*}
    \leq 2^p c_p \Estar{2:t} \left[ \left| u_{\Pi, k}(\history{i}{t}) - l_{\Pi,k}(\history{i}{t}) \right|^p \big| l_{\F,j}(\history{i}{t}) \big|^p \right] 
    + 4^p c_p \epsilon^p
\end{equation*}
By the definition of brackets $[ l_{\Pi,k}, u_{\Pi, k} ]$ from display \eqref{eqnapp:PibracketsModified},
\begin{equation*}
    \leq 2^p c_p \Estar{2:t} \left[ \left| 2 \epsilon \dot{\pi}_{2:t}(\history{i}{t}) \right|^p | l_{\F,j}(\history{i}{t}) |^p \right] 
    + 4^p c_p \epsilon^p
\end{equation*}
Recall that by assumption of the Lemma, $F$ is a function such that $| f(\history{i}{t})| \leq F(\history{i}{t})$ a.s. for all $f \in \F$. Thus, the brackets $\big[ l_{\F,k}, u_{\F, k} \big]$ can always be chosen such that $| l_{\F,j}(\history{i}{t})| \leq F(\history{i}{t})$ a.s. Thus,
\begin{equation*}
    \leq 2^{p} c_p \Estar{2:t} \left[ \left| 2 \epsilon \dot{\pi}_{2:t}(\history{i}{t}) \right|^p | F(\history{i}{t}) |^p \right] 
    + 4^p c_p \epsilon^p
\end{equation*}
Since $\dot{\pi}_{2:t}(\history{i}{t}) \geq 0$ and $F(\history{i}{t}) \geq 0$ by definition,
\begin{equation*}
    = 2^{2p} c_p \epsilon^p \Estar{2:t} \left[ \left| \dot{\pi}_{2:t}(\history{i}{t}) F(\history{i}{t}) \right|^p \right] 
    + 4^p c_p \epsilon^p
\end{equation*}
Since $\dot{\pi}_{2:t}(\history{i}{t}) = \sum_{t'=2}^{t} \dot{\pi}_{t'} \big( \action{i}{t'}, \state{i}{t'} \big)$ by definition,
\begin{equation*}
    = 2^{2p} c_p \epsilon^p \Estar{2:t} \bigg[ \bigg| \sum_{t'=2}^{t} \dot{\pi}_{t'} \big( \action{i}{t'}, \state{i}{t'} \big) F(\history{i}{t}) \bigg|^p \bigg] 
    + 4^p c_p \epsilon^p
\end{equation*}
By repeatedly applying Lemma \ref{lemma:binomialBound} (Inequality using Binomial Theorem), for some positive constant $k_p$,
\begin{equation*}
    = \epsilon^p \bigg\{ 2^{2p} c_p k_p^t \sum_{t'=2}^{t} \Estar{2:t} \left[ \big| \dot{\pi}_{t'} \big( \action{i}{t'}, \state{i}{t'} \big) F(\history{i}{t}) \big|^p \right] 
    + 4^p c_p \bigg\}.
\end{equation*}
The term $2^{2p} c_p k_p^t \sum_{t'=2}^{t} \Estar{2:t} \left[ \big| \dot{\pi}_{t'} \big( \action{i}{t'}, \state{i}{t'} \big) F(\history{i}{t}) \big|^p \right] + 4^p c_p$ above is bounded by assumption \ref{bracketcond:productMoment}.

Let $c_{\Pi \cdot \F} \triangleq \left\{ 2^{2p} c_p k_p^t \sum_{t'=2}^{t} \Estar{2:t} \left[ \left| \dot{\pi}_{t'} \big( \action{i}{t'}, \state{i}{t'} \big) F(\history{i}{t}) \right|^p \right]  + 4^p c_p \right\}^{1/p}$.
By the above result, the brackets for $\Pi_{2:t} \cdot \F$ from display \eqref{eqnapp:PiFbrackets} have size at most $\epsilon c_{\Pi \cdot \F}$ in $L_p(\Pstar)$ norm. Thus,
\begin{multline*}
    N_{[~]} \left( \epsilon c_{\Pi \cdot \F}, \Pi_{2:t} \cdot \F, L_p(\Pstar) \right) 
    \leq \underbrace{ |\MC{B}_{1:t-1}| }_{\TN{Upper~bounds~}N_{[~]} \left( \epsilon, \Pi_{2:t}, L_p(\Pstar) \right) } \cdot N_{\F,\epsilon} \\
    = \big[ \TN{diam}(B_{1:t-1}) / \epsilon \big]^{d_{1:t-1}} N_{[~]} \big( \epsilon, \F, L_{p}(\Pstar) \big).
\end{multline*}
The final equality above holds by display \eqref{app:PiBracketingUpperBound}.

The above implies that
\begin{equation}
    \label{eqnapp:piFproductBracket}
    N_{[~]} \left( \epsilon, \Pi_{2:t} \cdot \F, L_p(\Pstar) \right) 
    \leq \big[ \TN{diam}(B_{1:t-1}) c_{\Pi \cdot \F} / \epsilon \big]^{d_{1:t-1}}  \cdot N_{[~]} \big( \epsilon / c_{\Pi \cdot \F}, \F, L_{p}(\Pstar) \big).
\end{equation}
Note that by assumption \ref{bracketcond:finite}, $N_{[~]} \big( \epsilon / c_{\Pi \cdot \F}, \F, L_{p}(\Pstar) \big) < \infty$, so the above implies that $N_{[~]} \big( \epsilon, \Pi_{2:t} \cdot \F, L_{p}(\Pstar) \big) < \infty$ for all $\epsilon > 0$.

%%%%%%%%%%%%%%%%%%%%%%%%%%%%%
\proofSubsection{Bracketing integral result} We now work on showing the second part of the Lemma, i.e., that $\int_0^1 \sqrt{ \log N_{[~]} \big( \epsilon, \Pi_{2:t} \cdot \F, L_{p}(\Pstar) \big) } d \epsilon < \infty$.
\begin{equation*}
    \int_0^1 \sqrt{ \log N_{[~]} \left( \epsilon, \Pi_{2:t} \cdot \F, L_{p}(\Pstar) \right) } d \epsilon
\end{equation*}
By display \eqref{eqnapp:piFproductBracket},
\begin{equation*}
    \leq \int_0^1 \sqrt{ \log \left\{ \big[ \TN{diam}(B_{1:t-1}) c_{\Pi \cdot \F} / \epsilon \big]^{d_{1:t-1}} N_{[~]} \big( \epsilon / c_{\Pi \cdot \F}, \F, L_{p}(\Pstar) \big) \right\} } d \epsilon
\end{equation*}
Using properties of $\log$,
\begin{equation*}
    = \int_0^1 \sqrt{ d_{1:t-1} \log \big[ \TN{diam}(B_{1:t-1}) c_{\Pi \cdot \F} / \epsilon \big] 
    + \log N_{[~]} \big( \epsilon / c_{\Pi \cdot \F}, \F, L_{p}(\Pstar) \big) } d \epsilon
\end{equation*}
Note that $\TN{diam}(B_{1:t-1}) c_{\Pi \cdot \F} / \epsilon \geq 1$ since $\TN{diam}(B_{1:t-1}) c_{\Pi \cdot \F} / \epsilon$ upper bounds the bracketing number $N_{[~]} \left( \epsilon, \Pi_{2:t}, L_p(\Pstar) \right)$ (which must be at least $1$). Thus, $\log \big[ \TN{diam}(B_{1:t-1}) c_{\Pi \cdot \F} / \epsilon \big] \geq 0$. Since $\sqrt{a + b} \leq \sqrt{a} + \sqrt{b}$ for any numbers $a, b > 0$ (to see this, square both sides of the inequality),
\begin{equation}
    \label{eqnapp:bracketingLast}
    = \sqrt{ d_{1:t-1} } \int_0^1 \sqrt{ \log \big[ \TN{diam}(B_{1:t-1}) c_{\Pi \cdot \F} / \epsilon \big] } d \epsilon
    + \int_0^1 \sqrt{ \log N_{[~]} \big( \epsilon / c_{\Pi \cdot \F}, \F, L_{p}(\Pstar) \big) } d \epsilon.
\end{equation}
We now discuss why the quantity in the display above is finite:
\begin{itemize}
    \item Regarding the first term in display \eqref{eqnapp:bracketingLast}, note that since $\log \big[ \TN{diam}(B_{1:t-1}) c_{\Pi \cdot \F} / \epsilon \big] \geq 0$, we have that $\log \big[ \TN{diam}(B_{1:t-1}) c_{\Pi \cdot \F} / \epsilon \big] \leq \TN{diam}(B_{1:t-1}) c_{\Pi \cdot \F} / \epsilon$. Thus,
    \begin{multline*}
        \int_0^1 \sqrt{ \log \big[ \TN{diam}(B_{1:t-1}) c_{\Pi \cdot \F} / \epsilon \big] } d \epsilon 
        \leq \int_0^1 \sqrt{ \TN{diam}(B_{1:t-1}) c_{\Pi \cdot \F} / \epsilon  } d \epsilon \\
        = \sqrt{ \TN{diam}(B_{1:t-1}) c_{\Pi \cdot \F} } \int_0^1 \epsilon^{-2} d \epsilon < \infty.
    \end{multline*}
    %%%%%%%%%%
    \item For the second term in display \eqref{eqnapp:bracketingLast} above, note that
    \begin{equation*}
        \int_0^1 \sqrt{ \log N_{[~]} \big( \epsilon / c_{\Pi \cdot \F}, \F, L_{p}(\Pstar) \big) } d \epsilon
        = c_{\Pi \cdot \F} \int_0^1 \sqrt{ \log N_{[~]} \big( \epsilon / c_{\Pi \cdot \F}, \F, L_{p}(\Pstar) \big) } c_{\Pi \cdot \F}^{-1} d \epsilon
    \end{equation*}
    By integration by substitution, for $u = \epsilon / c_{\Pi \cdot \F}$,
    \begin{equation}
        \label{eqnapp:bracketingInequalityIntegral}
        = c_{\Pi \cdot \F} \int_0^{c_{\Pi \cdot \F}^{-1}} \sqrt{ \log N_{[~]} \big( u, \F, L_{p}(\Pstar) \big) } d u  < \infty.
    \end{equation}
    The above term is finite since $\int_0^1 \sqrt{ \log N_{[~]} \big( \epsilon, \F, L_{p}(\Pstar) \big) } d \epsilon  < \infty$ by assumption \ref{bracketcond:finiteIntegral}. If $c_{\Pi \cdot \F}^{-1} \leq 1$ display \eqref{eqnapp:bracketingInequalityIntegral} holds straightforwardly. If $c_{\Pi \cdot \F}^{-1} \geq 1$, display \eqref{eqnapp:bracketingInequalityIntegral} holds because $N_{[~]} \big( 1, \F, L_{p}(\Pstar) \big) \geq N_{[~]} \big( u, \F, L_{p}(\Pstar) \big)$ for all $u \geq 1$ by the definition of bracketing numbers.
\end{itemize}
We have shown that $\int_0^1 \sqrt{ \log N_{[~]} \left( \epsilon, \Pi_{2:t} \cdot \F, L_{p}(\Pstar) \right) } d \epsilon < \infty$, so we have now shown the second result of the Lemma. $~~\blacksquare$
\clearpage

%%%%%%%%%%%%%%%%%%%%%%%%%%%%%%%%%%%%%%%%%%%%
\section{Policy Parameter Results}
\label{app:algConditions}

\startOverview{Overview of Supplement \ref{app:algConditions} Results}
\begin{itemize}
    \item \bo{Section \ref{app:algConditionsConsistency}:} Consistency of Policy Parameters (Sufficient Assumptions for Condition \ref{cond:consistencyPolicy}; Theorem \ref{thm:consistencyBeta})
    %%%%%%%%%%%%%
    \item \bo{Section \ref{mainapp:binomialBound}:} Inequality Using Binomial Theorem (Helper Lemma \ref{lemma:binomialBound})
    %%%%%%%%%%%%%
    \item \bo{Section \ref{mainapp:equicontinuityPolicy}:} Stochastic Equicontinuity for Policy Parameters (Theorem \ref{thm:asymptoticEquicontinuityPolicy})
    %%%%%%%%%%%%%
    \item \bo{Section \ref{mainapp:lemmaInvertibility}:} Invertibility of $\piEstDotStar{1:t}$ (Lemma \ref{lemma:invertibilityPiDotStar})
\end{itemize}

%%%%%%%%%%%%%%%%%%%%%%%%%%%%%%%%%%%%%%%%%%%%%%%%%%%%%%%%%%%%%%%%%%%%%%%%%%%%%%
%%%%%%%%%%%%%%%%%%%%%%%%%%%%%%%%%%%%%%%%%%%%%%%%%%%%%%%%%%%%%%%%%%%%%%%%%%%%%%
\subsection{Consistency of Policy Parameters (Theorem \ref{thm:consistencyBeta})}
\label{app:algConditionsConsistency}

\begin{theorem}[Consistency of Policy Parameters]
    \label{thm:consistencyBeta}
    We assume Conditions \ref{cond:exploration} (Minimum Exploration) and \ref{cond:lipschitzPolicy} (Lipschitz Policy Function) hold.
    Condition \ref{cond:consistencyPolicy} holds (i.e., $\betahat{t} \Pto \betastar{t}$ for each $t \in [1 \colon T-1]$), 
    under the following additional assumptions:
    \begin{enumerate}[label=(CP\arabic*)] %[label=(\roman*)]
        \item \bo{Well-Separated Solutions:} 
        \label{consistencyBeta:wellSeparated}
        For each $t \in [1 \colon T-1]$, for any $\epsilon > 0$, there exists some $\eta > 0$ such that
        \begin{equation*}
        	\label{eqn:consistencySeparatedBeta}
            \inf_{ \beta_t \in \real^{d_t} \TN{~s.t.~} \| \beta_t - \betastar{t} \|_1 > \epsilon} \left\| \Estar{2:t} \big[ \piest{t}(\history{i}{t}; \beta_t) \big]  \right\|_1 > \eta > 0.
        \end{equation*}
        %\smallskip %%%%%%%%%%%%%%%%%%
        \item \bo{Asymptotically Tight:} 
        \label{consistencyBeta:tight}
        For each $t \in [1 \colon T-1]$, for any $\epsilon > 0$, there exists some $k < \infty$ such that
         \begin{equation*}
         	\label{eqn:consistencyTightBeta}
            \limsup_{n \to \infty} \PP \big( \big\| \betahat{t} \big\|_1 > k \big) \leq \epsilon.
        \end{equation*}
        %%%%%%%%%%%%%%%%%%%%%%%%%%%%%%%%%%%
        \item \bo{Finite Bracketing Number:}
        \label{consistencyBeta:finiteBracketing}
        Let $\alpha > 0$ be a constant. For each $t \in [1 \colon T-1]$ and any compact subset $K_t \subset \real^{d_t}$, 
        \begin{enumerate}[label=(\roman*)]
            \item For any $\epsilon > 0$ and any vector $c \in \real^{d_t}$, the bracketing number 
            \begin{equation*}
                N_{[~]} \left( \epsilon, ~ \big\{ c^\top \piest{t}(\cdotspace; \beta_t) \big\}_{\beta_t \in K_t}, ~ L_{1+\alpha}(\Pstar) \right) < \infty.
            \end{equation*}
            %%%%%%%%%%%%%%%%%%%%
            \item There exists a function $F_{\piest{t}}$ such that for all $\beta_t \in K_t$, $\big\| \piest{t}(\history{i}{t}; \beta_t ) \big\|_1 \leq F_{\piest{t}}(\history{i}{t})$ a.s. and 
            \begin{equation*}
                \Estar{2:t} \left[ \big| F_{\piest{t}}(\history{i}{t}) \dot{\pi}_{t'}(\action{i}{t'}, \state{i}{t'}) \big|^{1+\alpha} 
                \right] < \infty
            \end{equation*}
            for all $t' \in [2 \colon t]$; the functions $\dot{\pi}_{t'}$ are from Condition \ref{cond:lipschitzPolicy}.
        \end{enumerate}
    \end{enumerate}
\end{theorem}

\startproof{Theorem \ref{thm:consistencyBeta} (Consistency of Policy Parameters)} %%%%%%%%%%%%%%%%%%%%%%%%%%%%%%%%%%%%%%%%%%
We use an induction-based argument. For the base case, we show that $\betahat{1} \Pto \betastar1$. For the induction step, we show that $\betahat{t} \Pto \betastar{t}$, given that $\betahat{1:t-1} \Pto \betastar{1:t-1}$. 

\proofSubsection{Base Case}
Note that $\history{1}{1}, \history{2}{1}, \history{3}{1}, \dots, \history{n}{1}$ are i.i.d. (data from the first time-step; no adaptive sampling yet). This type of consistency proof is standard for Z-estimators; we include it for completeness since the induction step uses similar techniques. 

We first define the following useful functions:
\begin{equation*}
    \piEst{1} \big( \beta_1 \big) \triangleq \E \big[ \piest{1}\big( \history{i}{1}; \beta_1 \big) \big]
    ~~~~~~\TN{and}~~~~~~
    \piEstHat{1} \big( \beta_1 \big) \triangleq \frac{1}{n} \sum_{i=1}^n \piest{1} \big( \history{i}{1}; \beta_1 \big).
\end{equation*}

Let $\epsilon > 0$. By assumption \ref{consistencyBeta:wellSeparated}, there exists some $\eta > 0$ such that if $\beta_1 \in \real^{d_1}$ satisfies $\big\| \beta_1 - \betastar1 \big\|_1 > \epsilon$, then $\big\| \piEst{1} \big( \beta_1 \big) \big\|_1 = \big\| \E \big[ \piest{1}\big( \history{i}{1}; \beta_1 \big) \big] \big\|_1 > \eta > 0$. Thus, $\II_{ \| \betahat{1} - \betastar{1} \|_1 > \epsilon } \leq \II_{ \| \piEst{1} ( \betahat1 ) \|_1 > \eta }$, so
\begin{equation*}
    \PP \left( \big\| \betahat{1} - \betastar{1} \big\|_1 > \epsilon \right)
    \leq \PP \left( \big\| \piEst{1} \big( \betahat1 \big) \big\|_1 > \eta \right).
\end{equation*}

By the definition of $\betahat1$ from display \eqref{eqn:betahatDef}, $\piEstHat{1} \big( \betahat{1} \big) = \frac{1}{n} \sum_{i=1}^n \piest{1}\big( \history{i}{1}; \betahat1 \big) \\
= o_P(1/\sqrt{n}) = o_P(1)$. Thus,
\begin{equation*}
    = \PP \left( \big\| \piEstHat{1} \big( \betahat1 \big) - \piEst{1} \big( \betahat1 \big) \big\|_1 > \eta - o_P(1) \right).
\end{equation*}

By assumption \ref{consistencyBeta:tight}, for any $\delta > 0$, there exists some $k < \infty$ such that \\
$\limsup_{n \to \infty} \PP \big( \big\| \betahat{1} \big\|_1 > k \big) \leq \delta$. We use this $k$ below:
\begin{equation*}
     = \PP \left( \big\| \piEstHat{1} \big( \betahat{1} \big) - \piEst{1}\big( \betahat{1} \big) \big\|_1 
     \left\{ \II_{ \| \betahat{1} \|_1 > k } + \II_{ \| \betahat{1} \|_1 \leq k } \right\} > \eta - o_P(1) \right)
\end{equation*}
\begin{multline*}
     \leq \PP \left( \big\| \piEstHat{1} \big( \betahat{1} \big) - \piEst{1}\big( \betahat{1} \big) \big\|_1 
     ~\II_{ \| \betahat{1} \|_1 \leq k } > \eta/2 - o_P(1) \right) \\
     + \PP \left( \big\| \piEstHat{1} \big( \betahat{1} \big) - \piEst{1}\big( \betahat{1} \big) \big\|_1 
     ~\II_{ \| \betahat{1} \|_1 > k } > \eta/2 - o_P(1) \right)
\end{multline*}
\begin{equation*}
     \leq \PP \left( \big\| \piEstHat{1} \big( \betahat{1} \big) - \piEst{1}\big( \betahat{1} \big) \big\|_1 
     ~\II_{ \| \betahat{1} \|_1 \leq k } > \eta/2 - o_P(1) \right) 
     + \PP \left( \big\| \betahat{1} \big\|_1 > k \right) + o(1)
\end{equation*}
\begin{equation}
    \label{consistency:appT1last}
     \leq \underbrace{ \PP \bigg( \sup_{ \beta_1 \in \real^{d_1} \TN{~s.t.~} \| \beta_1 \|_1 \leq k} \big\| \piEstHat{1} \big( \beta_{1} \big) - \piEst{1}\big( \beta_{1} \big) \big\|_1 > \eta/2 - o_P(1) \bigg) }_{= o(1) } \\
     + \underbrace{ \PP \left( \big\| \betahat{1} \big\|_1 > k \right) }_{\leq \delta} + o(1).
\end{equation}

For the second term above, by assumption \ref{consistencyBeta:tight}, $\limsup_{n \to \infty} \PP \big( \big\| \betahat{1} \big\|_1 > k \big) \leq \delta$ and $\delta$ can be made arbitrarily small. 

For the first term above, we can apply the Uniform Weak Law of Large Numbers result for i.i.d. data \cite[Theorem 2.4.1]{van1996weak}
%19.4 of \cite{van2000asymptotic} 
to get that it converges to zero as $n \to \infty$. Specifically, note that 
\begin{equation*}
    N_{[~]} \left( \epsilon, ~ \big\{ c^\top \piest{1}(\cdotspace; \beta_1) \big\}_{\beta_1 \in \real^{d_1} \TN{~s.t.~} \| \beta_1 \|_1 \leq k}, ~ L_{1+\alpha}(\Pstar) \right) < \infty,
\end{equation*}
for any $c \in \real^{d_1}$ and any $\epsilon > 0$ by assumption  \ref{consistencyBeta:finiteBracketing}. Thus, by the Uniform Weak Law of Large Numbers, we have that
\begin{equation*}
    \sup_{\beta_1 \in \real^{d_1} \TN{~s.t.~} \| \beta_1 \|_1 \leq k} \left| c^\top \piEstHat{1} \big( \beta_{1} \big) - c^\top \piEst{1}\big( \beta_{1} \big) \right| \Pto 0,
\end{equation*}
for any $c \in \real^{d_1}$. Thus, by Cramer Wold device we have that
\begin{equation*}
    \sup_{\beta_1 \in \real^{d_1} \TN{~s.t.~} \| \beta_1 \|_1 \leq k} \big\| \piEstHat{1} \big( \beta_{1} \big) - \piEst{1}\big( \beta_{1} \big) \big\|_1 \Pto 0.
\end{equation*}
Thus, the expression in display \eqref{consistency:appT1last} above converges to zero.

\proofSubsection{Induction Step}
For our induction assumption, we assume that $\betahat{1:t-1} \Pto \betastar{1:t-1}$. Given this assumption, we will show that $\betahat{t} \Pto \betastar{t}$.

We now define the following useful functions:
\begin{equation}
    \label{eqn:policyPhiStar}
    \piEst{t} \big( \beta_{1:t} \big) \triangleq \E_{\pi(\beta_{1:t-1})} \left[ \piest{t}\big( \history{i}{t}; \beta_t \big) \right]
    = \E \left[ \WW{i}{2:t} \big(\beta_{1:t-1}, \betahat{1:t-1} \big) \piest{t}\big( \history{i}{t}; \beta_t \big) \right]
\end{equation}
and
\begin{equation}
    \label{eqn:policyPhiHat}
    \piEstHat{t} \big( \beta_{1:t} \big) \triangleq \frac{1}{n} \sum_{i=1}^n \WW{i}{2:t} \big(\beta_{1:t-1}, \betahat{1:t-1} \big) \piest{t} \big( \history{i}{t}; \beta_t \big).
\end{equation}

For now, we take as given that display \eqref{eqn:PhiBetahatAssumption} below holds; we will show this result holds at the end of this proof.
\begin{equation}
    \label{eqn:PhiBetahatAssumption}
    \left\| \piEst{t} \big( \betahat{1:t-1}, \betastar{t} \big) - \piEst{t} \big( \betastar{1:t-1}, \betastar{t} \big) \right\|_1
    = o_P(1).
\end{equation}

Let $\epsilon > 0$. By assumption \ref{consistencyBeta:wellSeparated}, there exists some $\eta > 0$ such that if $\beta_t \in \real^{d_t}$ satisfies $\big\| \beta_{t} - \betastar{t} \big\|_1 > \epsilon$, then $\big\| \piEst{t} \big( \betastar{1:t-1}, \beta_{t} \big) \big\|_1 = \big\| \Estar{2:t} \big[ \piest{t}(\history{i}{t}; \beta_t) \big] \big\|_1 > \eta > 0$. Thus,
\begin{equation*}
    \PP \left( \big\| \betahat{t} - \betastar{t} \big\|_1 > \epsilon \right)
    \leq \PP \left( \big\| \piEst{t} \big( \betastar{1:t-1}, \betahat{t} \big) \big\|_1 > \eta \right)
\end{equation*}

Note $\big\| \piEst{t} \big( \betastar{1:t-1}, \betahat{t} \big) \big\|_1 
= \big\| \piEst{t} \big( \betastar{1:t-1}, \betahat{t} \big) - \piEst{t} \big( \betahat{1:t-1}, \betahat{t} \big) + \piEst{t} \big( \betahat{1:t-1}, \betahat{t} \big) \big\|_1 \\
\leq \big\| \piEst{t} \big( \betastar{1:t-1}, \betahat{t} \big) - \piEst{t} \big( \betahat{1:t-1}, \betahat{t} \big) \big\|_1 + \big\| \piEst{t} \big( \betahat{1:t-1}, \betahat{t} \big) \big\|_1 
= \big\| \piEst{t} \big( \betahat{1:t} \big) \big\|_1 + o_P(1)$; the last equality holds by display \eqref{eqn:PhiBetahatAssumption} and the definition of $\betahat{t}$ from display \eqref{eqn:betahatDef}. Thus,
\begin{equation*}
    \leq \PP \left( \big\| \piEst{t} \big( \betahat{1:t} \big) \big\|_1 > \eta  - o_P(1) \right)
\end{equation*}

Note $\piEstHat{t} \big( \betahat{1:t} \big)
= \frac{1}{n} \sum_{i=1}^n \WW{i}{2:t} \big(\betahat{1:t-1}, \betahat{1:t-1} \big) \piest{t} \big( \history{i}{t}; \betahat{t} \big) 
= \frac{1}{n} \sum_{i=1}^n \piest{t} \big( \history{i}{t}; \betahat{t} \big) \\
= o_P(1/\sqrt{n}) = o_P(1)$; the second to last equality holds by the definition of $\betahat{t}$ from display \eqref{eqn:betahatDef}. Thus,
\begin{equation*}
    = \PP \left( \big\| \piEstHat{t} \big( \betahat{1:t} \big) - \piEst{t} \big( \betahat{1:t} \big) \big\|_1 > \eta - o_P(1) \right)
\end{equation*}

By assumption \ref{consistencyBeta:tight}, for any $\delta > 0$, there exists some $k < \infty$ such that \\
$\limsup_{n \to \infty} \PP \big( \big\| \betahat{t} \big\|_1 > k \big) \leq \delta$. We use this $k$ below:
\begin{equation*}
     = \PP \left( \big\| \piEstHat{t} \big( \betahat{1:t} \big) - \piEst{t}\big( \betahat{1:t} \big) \big\|_1 
     \left\{ \II_{ \| \betahat{t} \|_1 > k } + \II_{ \| \betahat{t} \|_1 \leq k } \right\} > \eta - o_P(1) \right)
\end{equation*}
\begin{multline*}
     \leq \PP \left( \big\| \piEstHat{t} \big( \betahat{1:t} \big) - \piEst{t}\big( \betahat{1:t} \big) \big\|_1 
     ~\II_{ \| \betahat{t} \|_1 \leq k } > \eta/2 - o_P(1) \right) \\
     + \PP \left( \big\| \piEstHat{t} \big( \betahat{1:t} \big) - \piEst{t}\big( \betahat{1:t} \big) \big\|_1 
     ~\II_{ \| \betahat{t} \|_1 > k } > \eta/2 - o_P(1) \right)
\end{multline*}
\begin{multline*}
     \leq \PP \left( \big\| \piEstHat{t} \big( \betahat{1:t} \big) - \piEst{t}\big( \betahat{1:t} \big) \big\|_1 
     ~\II_{ \| \betahat{t} \|_1 \leq k } > \eta/2 - o_P(1) \right) 
     + \PP \left( \big\| \betahat{t} \big\|_1 > k \right) + o(1)
\end{multline*}
\begin{multline}
     \leq \PP \bigg( \sup_{\beta_t \in \real^{d_t} \TN{~s.t.~} \| \beta_t \|_1 \leq k} \big\| \piEstHat{t} \big( \betahat{1:t-1}, \beta_{t} \big) - \piEst{t}\big( \betahat{1:t-1}, \beta_{t} \big) \big\|_1 > \eta/2 - o_P(1) \bigg) \\
     + \PP \left( \big\| \betahat{t} \big\|_1 > k \right) + o(1)
\end{multline}

Recall $\betahat{1:t-1} \Pto \betastar{1:t-1}$ by our induction assumption; thus, $\II_{ \betahat{1:t-1} \in B_{1:t-1} } \Pto 1$, where recall that $B_{1:t-1} \subset \real^{d_{1:t-1}}$ is a compact subset whose interior contains $\betastar{1:t-1}$. Thus,
\begin{multline*}
     \leq \underbrace{ \PP \bigg( \sup_{\beta_t \in \real^{d_t} \TN{~s.t.~} \| \beta_t \|_1 \leq k} ~ \sup_{\beta_{1:t-1} \in B_{1:t-1}} \big\| \piEstHat{t} \big( \beta_{1:t-1}, \beta_{t} \big) - \piEst{t}\big( \beta_{1:t-1}, \beta_{t} \big) \big\|_1 > \eta/2 - o_P(1) \bigg) }_{=o(1)} \\
     + \underbrace{ \PP \left( \big\| \betahat{t} \big\|_1 > k \right) }_{\leq \delta} + o(1)
\end{multline*}
Note that the above converges to zero as $n \to \infty$ for the following reasons:
\begin{itemize}
    \item By assumption \ref{consistencyBeta:tight}, $\limsup_{n \to \infty} \PP \big( \big\| \betahat{t} \big\|_1 > k \big) \leq \delta$ and $\delta$ can be made arbitrarily small.
    %%%%%%%%%%%%%
    \item Note that by Cramer Wold device, to show that
    \begin{equation*}
        \sup_{\beta_t \in \real^{d_t} \TN{~s.t.~} \| \beta_t \|_1 \leq k} ~ \sup_{\beta_{1:t-1} \in B_{1:t-1}} \left\| \piEstHat{t} \big( \beta_{1:t-1}, \beta_{t} \big) - \piEst{t}\big( \beta_{1:t-1}, \beta_{t} \big) \right\|_1 \Pto 0,
    \end{equation*}
    it is sufficient to show that for any vector $c \in \real^{d_t}$,
    \begin{equation*}
        \sup_{\beta_t \in \real^{d_t} \TN{~s.t.~} \| \beta_t \|_1 \leq k} ~ \sup_{\beta_{1:t-1} \in B_{1:t-1}} c^\top \left\{ \piEstHat{t} \big( \beta_{1:t-1}, \beta_{t} \big) - \piEst{t}\big( \beta_{1:t-1}, \beta_{t} \big) \right\} \Pto 0.
    \end{equation*}
    Also, note that
    \begin{equation*}
        \sup_{\beta_t \in \real^{d_t} \TN{~s.t.~} \| \beta_t \|_1 \leq k} ~ \sup_{\beta_{1:t-1} \in B_{1:t-1}} c^\top \left\{ \piEstHat{t} \big( \beta_{1:t-1}, \beta_{t} \big) - \piEst{t}\big( \beta_{1:t-1}, \beta_{t} \big) \right\}
    \end{equation*}
    \begin{multline*}
        = \sup_{\beta_t \in \real^{d_t} \TN{~s.t.~} \| \beta_t \|_1 \leq k} ~ \sup_{\beta_{1:t-1} \in B_{1:t-1}} \frac{1}{n} \sum_{i=1}^n \bigg\{ \WW{i}{2:t} \big(\beta_{1:t-1}, \betahat{1:t-1} \big) c^\top \piest{t}(\history{i}{t}; \beta_t) \\
        - \E \left[ \WW{i}{2:t} \big(\beta_{1:t-1}, \betahat{1:t-1} \big) c^\top \piest{t}(\history{i}{t}; \beta_t) \right] \bigg\} \Pto 0.
    \end{multline*}
    The above convergence result holds by Theorem \ref{thm:weightedUWLLN} (Weighted Martingale Triangular Array Uniform Weak Law of Large Numbers).
    Specifically we are able to apply Theorem \ref{thm:weightedUWLLN} because Condition \ref{cond:exploration} holds and $N_{[~]} \left( \epsilon, \F_{\Pi c^\top \phi_t}(B_{1:t-1}, k), L_{1+\alpha}(\Pstar) \right) < \infty$ for all $c \in \real^{d_t}$ and all $\epsilon > 0$, where
    \begin{equation*}
        \F_{\Pi c^\top \phi_t}(B_{1:t-1}, k) 
        \triangleq \bigg\{ \bigg[ \prod_{t'=2}^t \pi_{t'} ( \cdotspace; \beta_{t'-1} ) \bigg] c^\top \piest{t}(\cdotspace; \beta_t \big) \bigg\}_{\beta_{1:t-1} \in B_{1:t-1}, ~ \beta_t \in \real^{d_t} \TN{~s.t.~} \| \beta_t \|_1 \leq k }.
    \end{equation*}
    The above finite bracketing number result holds since using assumption \ref{consistencyBeta:finiteBracketing} and Condition \ref{cond:lipschitzPolicy}, we can apply by Lemma \ref{lemma:bracketingProduct} (specifically see Remark \ref{remark:bracketingProduct} part \ref{productBracket:finitePhi}).
\end{itemize}

\proofSubsection{\noindent We now show that display \eqref{eqn:PhiBetahatAssumption} holds}
Let $\beta_{1:t-1} \in B_{1:t-1}$.
\begin{equation*}
    \left\| \piEst{t} \big( \beta_{1:t-1}, \betastar{t} \big) - \piEst{t} \big( \betastar{1:t-1}, \betastar{t} \big) \right\|_1
\end{equation*}
\begin{equation*}
    = \bigg\| \E \bigg[ \left\{ \WW{i}{2:t} \big(\beta_{1:t-1}, \betahat{1:t-1} \big) - \WW{i}{2:t} \big(\betastar{1:t-1}, \betahat{1:t-1} \big) \right\} \piest{t}\big( \history{i}{t}; \betastar{t} \big) \bigg] \bigg\|_1
\end{equation*}
\begin{equation*}
    = \bigg\| \Estar{2:t} \bigg[ \left\{ \WW{i}{2:t} \big(\beta_{1:t-1}, \betastar{1:t-1} \big) - \WW{i}{2:t} \big(\betastar{1:t-1}, \betastar{1:t-1} \big) \right\} \piest{t}\big( \history{i}{t}; \betastar{t} \big) \bigg] \bigg\|_1
\end{equation*}
By Jensen's inequality,
\begin{equation*}
    \leq \Estar{2:t} \bigg[ \left| \WW{i}{2:t} \big(\beta_{1:t-1}, \betastar{1:t-1} \big) - \WW{i}{2:t} \big(\betastar{1:t-1}, \betastar{1:t-1} \big) \right| \big\| \piest{t}\big( \history{i}{t}; \betastar{t} \big) \big\|_1 \bigg]
\end{equation*}
By definition of $\WW{i}{t'} \big(\beta_{t'-1}, \betastar{t'-1} \big)$ from display \eqref{eqn:weightsDef},
\begin{multline*}
    = \Estar{2:t} \bigg[ \bigg| \prod_{t'=2}^t \pi_{t'} \big( \action{i}{t'}, \state{i}{t'}; \beta_{t'-1} \big) - \prod_{t'=2}^t \pi_{t'} \big( \action{i}{t'}, \state{i}{t'}; \betastar{t'-1} \big) \bigg| \\
    \bigg\{ \prod_{t'=2}^t \frac{1}{ \pi_{t'} \big( \action{i}{t'}, \state{i}{t'}; \betastar{t'-1} \big) } \bigg\} \big\| \piest{t}\big( \history{i}{t}; \betastar{t} \big) \big\|_1 \bigg]
\end{multline*}
By Condition \ref{cond:exploration} (Minimum Exploration), $\pi_{t'} \big( \action{i}{t'}, \state{i}{t'}; \betastar{t'-1} \big)^{-1} \leq \pi_{\min}^{-1}$ a.s. Thus,
\begin{equation}
    \label{eqn:upperBoundConsistencyDiff}
    \leq \pi_{\min}^{-(t-1)} \Estar{2:t} \bigg[ \bigg| \prod_{t'=2}^t \pi_{t'}(\action{i}{t'}, \state{i}{t'}; \beta_{t'-1}) - \prod_{t'=2}^t \pi_{t'}(\action{i}{t'}, \state{i}{t'}; \betastar{t'-1}) \bigg| \big\| \piest{t}\big( \history{i}{t}; \betastar{t} \big) \big\|_1 \bigg]
\end{equation}

By Condition \ref{cond:lipschitzPolicy}, we can apply Lemma \ref{lemma:lipschitzPolicyProduct} (Product of Lipschitz Policy Functions are Lipschitz) to get that
\begin{equation*}
    \leq \pi_{\min}^{-(t-1)} \Estar{2:t} \bigg[ \big\| \piest{t}\big( \history{i}{t}; \betastar{t} \big) \big\|_1 \bigg\{ \sum_{t'=2}^{t} \dot{\pi}_{t'}(\action{i}{t'}, \state{i}{t'}) \bigg\} \big\| \beta_{1:t-1} - \betastar{1:t-1} \big\|_2 \bigg] 
\end{equation*}
By linearity of expectations,
\begin{equation*}
    = \pi_{\min}^{-(t-1)} \bigg\{ \sum_{t'=2}^{t} \Estar{2:t} \left[ \big\| \piest{t}\big( \history{i}{t}; \betastar{t} \big) \big\|_1 \dot{\pi}_{t'}(\action{i}{t'}, \state{i}{t'}) \right] \bigg\} \big\| \beta_{1:t-1} - \betastar{1:t-1} \big\|_2.
\end{equation*}
Thus, by consolidating the above results, we have that
\begin{multline*}
    \big\| \piEst{t} \big( \betahat{1:t-1}, \betastar{t} \big) - \piEst{t} \big( \betastar{1:t-1}, \betastar{t} \big) \big\|_1 \\
    \leq \pi_{\min}^{-(t-1)} \bigg\{ \sum_{t'=2}^{t} \Estar{2:t} \left[ \big\| \piest{t}\big( \history{i}{t}; \betastar{t} \big) \big\|_1 \dot{\pi}_{t'}(\action{i}{t'}, \state{i}{t'}) \right] \bigg\} \big\| \betahat{1:t-1} - \betastar{1:t-1} \big\|_2
    = o_P(1).
\end{multline*}
The last limit above holds because
\begin{itemize}
    \item $\big\| \betahat{1:t-1} - \betastar{1:t-1} \big\|_2 = o_P(1)$ since $\betahat{1:t-1} \Pto \betastar{1:t-1}$ by our induction assumption.
    %%%%%%%%%%
    \item By assumption \ref{consistencyBeta:finiteBracketing} (Finite Bracketing Number for Policy Functions), there exists a function $F_{\piest{t}}$ such that $\big\| \piest{t}\big( \history{i}{t}; \beta_t \big) \big\|_1 \leq F_{\piest{t}}(\history{i}{t})$ a.s. and for all $t' \in [2 \colon t]$,
    $\Estar{2:t} \big[ F_{\piest{t}} \big( \history{i}{t} \big) \dot{\pi}_{t'}(\action{i}{t'}, \state{i}{t'}) \big] < \infty$. Thus,
    \begin{equation*}
        \Estar{2:t} \left[ \big\| \piest{t}\big( \history{i}{t}; \betastar{t} \big) \big\|_1 \dot{\pi}_{t'}(\action{i}{t'}, \state{i}{t'}) \right] 
        \leq \Estar{2:t} \left[ F_{\piest{t}} \big( \history{i}{t} \big) \dot{\pi}_{t'}(\action{i}{t'}, \state{i}{t'}) \right] < \infty.
    \end{equation*}
\end{itemize}
We have now shown that display \eqref{eqn:PhiBetahatAssumption} holds. ~ $\blacksquare$

%%%%%%%%%%%%%%%%%%%%%%%%%%%%%%%%%%%%%%%%%%%%
\subsection{Inequality Using Binomial Theorem (Helper Lemma \ref{lemma:binomialBound})} %%%%%%%%%%%%%%%%%%%%%%%%%%%%%%%%%%%%%%%%%%%%
\label{mainapp:binomialBound}

\begin{lemma}[Inequality Using Binomial Theorem]
    \label{lemma:binomialBound}
     For any $\eta \geq 1$ and any $a, b \in \real$, we have that $|a+b|^\eta \leq c_\eta ( |a|^\eta + |b|^\eta )$ for some constant $c_\eta < \infty$.
\end{lemma}

\startproof{Lemma \ref{lemma:binomialBound}}
Note that
\begin{equation*}
	| a + b |^\eta
	\leq \begin{cases}
			|a-b|^{\floor{\eta}} & \TN{ if } |a + b| < 1 \\
			|a-b|^{\ceil{\eta}} & \TN{ if } |a + b| \geq 1 \\
		\end{cases}
\end{equation*}
Above we use $\floor{\eta}$ to round $\eta$ down to the nearest integer and we use $\ceil{\eta}$ to round $\eta$ up to the nearest integer. 
Let
\begin{equation*}
    k \triangleq \begin{cases}
        \floor{\eta} & \TN{ if } |a + b| < 1 \\
	   \ceil{\eta} & \TN{ if } |a + b| \geq 1
    \end{cases}.
\end{equation*}

Note $k \geq 1$ since that $\ceil{\eta} \geq \floor{\eta} \geq 1$ because $\eta \geq 1$.
Since $k$ is a positive integer, by the Binomial theorem,
\begin{equation*}
    | a + b |^k 
    = \big| ( a + b )^k \big| 
    = \bigg| \sum_{j=0}^k {k \choose j} a^j ~ b^{k-j} \bigg| 
    = \sum_{j=0}^k {k \choose j} |a|^j ~ |b|^{k-j}
\end{equation*}
Note that $|a|^j ~ |b|^{k-j} \leq \max \big( |a|^k, |b|^k \big)$ for all $j \in [0 \colon k]$. Thus,
\begin{equation*}
    \leq \bigg\{ \sum_{j=0}^{k} {k \choose j} \bigg\} \max \big( |a|^k, |b|^k \big)
    \leq \bigg\{ \sum_{j=0}^{k} {k \choose j} \bigg\} \big( |a|^k + |b|^k \big).
\end{equation*}
Thus we can choose $c_{\eta} = \left\{ \sum_{j=0}^{k} {k \choose j} \right\}$. $\blacksquare$

%%%%%%%%%%%%%%%%%%%%%%%%%%%%%%%%%%%%%
%%%%%%%%%%%%%%%%%%%%%%%%%%%%%%%%%%%%%%
%%%%%%%%%%%%%%%%%%%%%%%%%%%%%%%%%%%%%%%%%%
\subsection{Asymptotic Equicontinuity for Policy Parameters (Theorem \ref{thm:asymptoticEquicontinuityPolicy})} %%%%%%%%%%%%%%%%%%%%%%%%%%%%%%%%%%%
\label{mainapp:equicontinuityPolicy}

\begin{theorem}[Asymptotic Equicontinuity for Policy Parameters]
    \label{thm:asymptoticEquicontinuityPolicy}
    We assume that Conditions \ref{cond:consistencyPolicy}-\ref{cond:differentiabilityPolicy} on the adaptive sampling algorithm hold. %\ref{cond:exploration} (Minimum Exploration), \ref{cond:lipschitzPolicy} (Lipschitz Policy Functions), \ref{cond:differentiabilityPolicy} (Differentiability of Policy Parameter Estimating Functions), and  (Consistency of Policy Parameters) hold. 
    Consider the following results:
    \begin{multline}
        \label{eqnApp:policyEquicontinuity}
        \sqrt{n} \left\{ \piEstHat{1:T-1}(\betahat{1:T-1}) - \piEst{1:T-1}(\betahat{1:T-1}) \right\} \\
        = \sqrt{n} \left\{ \piEstHat{1:T-1}(\betastar{1:T-1}) - \piEst{1:T-1}(\betastar{1:T-1}) \right\} + o_P(1)
    \end{multline}
    and 
    \begin{equation*}
        \sqrt{n} \big( \betahat{1:T-1} - \betastar{1:T-1} \big) = O_P(1).
    \end{equation*}
    The above results hold under the following additional assumptions:
    \begin{enumerate}[label=(NP\arabic*)] 
        %%%%%%%%%%%%%%%%%%%%%%%%%%%%%%%
        \item \bo{Finite Bracketing Integral:} 
        \label{normalityBeta:bracketing}
        Let $\alpha > 0$ be a constant.  For each $t \in [1 \colon T-1]$, for any vector $c \in \real^{d_t}$,
        \begin{equation}
            \label{eqn:bracketingIntegralPiest}
            \int_0^1 \sqrt{ \log N_{[~]} \left( \epsilon, ~ \big\{ c^\top \piest{t}(\cdotspace; \beta_t) \big\}_{\beta_t \in B_t}, ~ L_{2+\alpha}(\Pstar) \right) } d \epsilon < \infty.
        \end{equation}
        Additionally, there exists a function $F_{\piest{t}}$ such that for all $\beta_t \in B_t$, $\big\| \piest{t}(\history{i}{t}; \beta_t ) \big\|_1 \leq F_{\piest{t}}(\history{i}{t})$ a.s. and 
        \begin{equation}
            \label{eqn:bracketingEnvelopePiest}
             \Estar{2:T} \left[ \big| F_{ \piest{}, t}(\history{i}{t}) \dot{\pi}_{t'}(\action{i}{t'}, \state{i}{t'}) \big|^{2+\alpha} \right] < \infty
        \end{equation}
        for all $t' \in [2 \colon T]$; the functions $\dot{\pi}_{t'}$ above are from Condition \ref{cond:lipschitzPolicy}.
        \smallskip
        %%%%%%%%%%%%%%%%%%%%%%%%%%%%%%%
        \item \bo{Continuity Condition:}
        \label{normalityBeta:L2convergence}
        For each $t \in [1 \colon T-1]$, the following mapping is continuous at $\beta_t = \betastar{t}$:
        \begin{equation*}
            \beta_t \mapsto 
            \Estar{2:t} \left[ \big\| \piest{t}\big(\history{i}{t}; \beta_t \big)
            - \piest{t} \big(\history{i}{t}; \betastar{t} \big) \big\|_2^2 \right].
        \end{equation*}
    \end{enumerate}
\end{theorem}

\begin{remark}[Condition \ref{cond:lipschitzPolicyEstimatingFunction} implies assumptions \ref{normalityBeta:bracketing} and \ref{normalityBeta:L2convergence} above hold]
    \label{remark:lipschitzSuffient}
    As we discussed in the Remark below Condition \ref{cond:lipschitzPolicyEstimatingFunction} (Lipschitz Policy Estimating Function), Condition \ref{cond:lipschitzPolicyEstimatingFunction} can be replaced by more general assumptions; these are assumptions \ref{normalityBeta:bracketing} and  \ref{normalityBeta:L2convergence} above.
    \begin{itemize}
    \item \textit{Condition \ref{cond:lipschitzPolicyEstimatingFunction} implies assumption \ref{normalityBeta:bracketing} because}
    %%%%%%%%%%%%%%%%%%%%%%
    \begin{itemize}
        \item Example 19.7 of \cite{van2000asymptotic} shows that Lipschitz property of Condition \ref{cond:lipschitzPolicyEstimatingFunction} and the compactness of $B_t$ implies that the bracketing integral condition from display \eqref{eqn:bracketingIntegralPiest} holds.
        %%%%%%%%%%%%%%%%%%%%%%
        \item We now show why Condition \ref{cond:lipschitzPolicyEstimatingFunction} implies that display \eqref{eqn:bracketingEnvelopePiest} holds. Since $B_t$ is compact, let $\TN{diam}(B_t) < \infty$ be the diameter of $B_t$ in Euclidian distance. Note the following inequality for all $\beta_t \in B_t$,
        \begin{multline*}
            \big\| \piest{t}(\history{i}{t}; \beta_t ) \big\|_1 
            \leq \sqrt{d_t} \big\| \piest{t}(\history{i}{t}; \beta_t ) \big\|_2 \\
            \leq \sqrt{d_t} \left\{ \big\| \piest{t}(\history{i}{t}; \betastar{t} ) \big\|_2 +  \dot{\phi}_t(\history{i}{t}) \TN{diam}(B_t) \right\} \TN{~~a.s.}
        \end{multline*}
        The first inequality above holds by property of norms and the second inequality above holds by display \eqref{eqn:piestLipschitz} of Condition \ref{cond:lipschitzPolicy}. Furthermore, note that
        \begin{equation*}
            \Estar{2:T} \left[ \left| \sqrt{d_t} \left\{ \big\| \piest{t}(\history{i}{t}; \betastar{t} ) \big\|_2 +  \dot{\phi}_t(\history{i}{t}) \TN{diam}(B_t) \right\} \dot{\pi}_{t'}(\action{i}{t'}, \state{i}{t'}) \right|^{2+\alpha} \right].
        \end{equation*}
        By Lemma \ref{lemma:binomialBound} (Inequality Using Binomial Theorem), for some positive constant $c_{2+\alpha} < \infty$,
        \begin{multline*}
            \leq c_{2+\alpha} \Estar{2:T} \left[ \left| \sqrt{d_t} \big\| \piest{t}(\history{i}{t}; \betastar{t} ) \big\|_2 \dot{\pi}_{t'}(\action{i}{t'}, \state{i}{t'}) \right|^{2+\alpha} \right] \\
            + c_{2+\alpha} \Estar{2:T} \left[ \left| \sqrt{d_t} \dot{\phi}_t(\history{i}{t}) \TN{diam}(B_t) \dot{\pi}_{t'}(\action{i}{t'}, \state{i}{t'}) \right|^{2+\alpha} \right] < \infty.
        \end{multline*}
        The final inequality holds by Condition \ref{cond:lipschitzPolicyEstimatingFunction}. The above implies that display \eqref{eqn:bracketingEnvelopePiest} holds.
    \end{itemize}
    \smallskip
    %%%%%%%%%%%%%%%%%%%%%%%%%%%%%%%%%%%%%%%%%%%%
    \item We now discuss why Condition \ref{cond:lipschitzPolicyEstimatingFunction} implies assumption \ref{normalityBeta:L2convergence} holds. Let $\epsilon > 0$. We want to show that there exists some $\delta > 0$ such that $\Estar{2:t} \left[ \big\| \piest{t} \big(\history{i}{t}; \beta_{t} \big) - \piest{t} \big( \history{i}{t}; \betastar{t} \big) \big\|_2^2 \right] \leq \epsilon$ whenever $\big\| \beta_{t} - \betastar{t} \big\|_2 \leq \delta$.
    By Condition \ref{cond:lipschitzPolicyEstimatingFunction}, for any $\beta_t \in B_t$,
    \begin{equation*}
        \Estar{2:t} \left[ \big\| \piest{t} \big(\history{i}{t}; \beta_{t} \big) - \piest{t} \big( \history{i}{t}; \betastar{t} \big) \big\|_2^2 \right]
        \leq \Estar{2:t} \big[ \dot{\phi}_{t}(\history{i}{t})^2 \big] \big\| \beta_{t} - \betastar{t} \big\|_2^2.
    \end{equation*}
    Recall $\Estar{2:t} \big[ \dot{\phi}_{t}(\history{i}{t})^2 \big] < \infty$ by Condition \ref{cond:lipschitzPolicyEstimatingFunction}. Thus, $ \delta = \epsilon^{1/2} \Estar{2:t} \big[ \dot{\phi}_{t}(\history{i}{t})^2 \big]^{-1/2}$ is sufficient.
\end{itemize}
\end{remark}

\startproof{Theorem \ref{thm:asymptoticEquicontinuityPolicy}}
Recall in displays \eqref{eqn:diffPolicy} and \eqref{proofSketch:piEstHat}, we defined the functions $\piEst{1:T-1} ( \beta_{1:T-1} )$ and $\piEstHat{1:T-1} ( \beta_{1:T-1} )$ respectively. More generally, we now define for \textit{any} $t \in [1 \colon T-1]$ the following functions:
\begin{equation*}
    \piEst{1:t}(\beta_{1:t}) \triangleq \E \begin{bmatrix}
        \piest1(\history{i}{1}; \beta_1) \\
        \WW{i}{2}(\beta_1, \betahat{1}) \piest2(\history{i}{2}; \beta_2) \\
        \WW{i}{2:3}(\beta_{1:2}, \betahat{1:2}) \piest3(\history{i}{3}; \beta_3) \\
        \vdots \\
        \WW{i}{2:t}(\beta_{1:t-1}, \betahat{1:t-1}) \piest{t}(\history{i}{t}; \beta_t) 
    \end{bmatrix}
\end{equation*}
and
\begin{equation*}
    \piEstHat{1:t}(\beta_{1:t}) \triangleq \frac{1}{n} \sum_{i=1}^n \begin{bmatrix}
        \piest1(\history{i}{1}; \beta_1) \\
        \WW{i}{2}(\beta_1, \betahat{1}) \piest2(\history{i}{2}; \beta_2) \\
        \WW{i}{2:3}(\beta_{1:2}, \betahat{1:2}) \piest3(\history{i}{3}; \beta_3) \\
        \vdots \\
        \WW{i}{2:t}(\beta_{1:t-1}, \betahat{1:t-1}) \piest{t}(\history{i}{t}; \beta_t) 
    \end{bmatrix}.
\end{equation*}
We use an induction argument. 
\begin{itemize}
    \item For the base case, $t=1$, we will show that $\sqrt{n} \big( \betahat{1} - \betastar1 \big) = O_P(1)$ and that
    \begin{equation*}
        \sqrt{n} \left\{ \piEstHat{1}(\betahat1) - \piEst{1}(\betahat1)  \right\}
        = \sqrt{n} \left\{ \piEstHat{1}(\betastar1) - \piEst{1}(\betastar1) \right\} + o_P(1).
    \end{equation*}
    %%%%%%%%%%%%%%%%%
    \item For the induction step, $t > 1$, we assume that $\sqrt{n} \big( \betahat{1:t-1} - \betastar{1:t-1} \big) = O_P(1)$ and will show that $\sqrt{n} \big( \betahat{1:t} - \betastar{1:t} \big) = O_P(1)$ and that
    \begin{equation*}
        \sqrt{n} \left\{ \piEstHat{1:t}(\betahat{1:t}) -  \piEst{1:t}(\betahat{1:t}) \right\} 
        = \sqrt{n} \left\{ \piEstHat{1:t}(\betastar{1:t}) -  \piEst{1:t}(\betastar{1:t}) \right\} + o_P(1).
    \end{equation*}
\end{itemize}
To show that the Theorem holds, it is sufficient to show the above results hold.

\proofSubsection{Base Case} %%%%%%%%%%%%%%%%%%%%%%%%%%%%%%%%%%
Note that since $\history{1}{1}, \history{2}{1}, \history{3}{1}, \dots, \history{n}{1}$ are i.i.d., we can use an asymptotic normality argument for Z-estimators on i.i.d. data.

\proofSubsubsection{Stochastic Equicontinuity Result} First, we will apply Lemma \ref{lemma:stochasticEquicontinuity} (Stochastic Equicontinuity) to get that for any fixed vector $c \in \real^{d_1}$,
\begin{equation}
    \label{norm:equicontinuityPreBase}
    \sqrt{n} c^\top \left\{ \piEstHat{1}(\betahat{1}) - \piEst{1}(\betahat{1}) \right\}
    = \sqrt{n} c^\top \left\{ \piEstHat{1}(\betastar{1}) - \piEst{1}(\betastar{1}) \right\} + o_P(1).
\end{equation}
We are able to apply Lemma \ref{lemma:stochasticEquicontinuity} (Stochastic Equicontinuity) because the following assumptions hold:
\begin{itemize}
    \item Conditions \ref{cond:exploration} (Minimum Exploration) and \ref{cond:lipschitzPolicy} (Lipschitz Policy Functions) hold.
    %%%%%%%%%%%%%%%
    \item By assumption \ref{normalityBeta:bracketing} (Finite Bracketing Integral), for any vector $c \in \real^{d_1}$,
    \begin{equation*}
            \int_0^1 \sqrt{ \log N_{[~]} \left( \epsilon, ~ \big\{ c^\top \piest{1}(\cdotspace; \beta_1) \big\}_{\beta_1 \in B_1}, ~ L_{2+\alpha}(\Pstar) \right) } d \epsilon < \infty.
        \end{equation*}
    %%%%%%%%%%%%%%%
    \item $\betahat{1} \Pto \betastar{1}$ by Condition \ref{cond:consistencyPolicy}. Thus, by assumption \ref{normalityBeta:L2convergence} and continuous mapping theorem we have that for any vector $c \in \real^{d_1}$, $\nu \big( c^\top \piest{t}(\cdotspace; \beta_t), ~ c^\top \piest{t}(\cdotspace; \betastar{t}) \big) \Pto 0$, where
    \begin{equation*}
        \nu \bigg( c^\top \piest{t}(\cdotspace; \beta_t), ~ c^\top \piest{t}(\cdotspace; \beta_t') \bigg) 
        \triangleq \Estar{2:t} \left[ \left\{ c^\top \piest{t}\big(\history{i}{t}; \beta_t \big)
        - c^\top \piest{t} \big(\history{i}{t}; \beta_t' \big) \right\}^2 \right]^{1/2}.
    \end{equation*}
\end{itemize}

\noindent By display \eqref{norm:equicontinuityPreBase} and Cramer Wold device, we have that
\begin{equation}
    \label{norm:equicontinuityBase}
    \sqrt{n} \left\{ \piEstHat{1}(\betahat{1}) - \piEst{1}(\betahat{1}) \right\}
    = \sqrt{n} \left\{ \piEstHat{1}(\betastar{1}) - \piEst{1}(\betastar{1}) \right\} + o_P(1).
\end{equation}
With the above result, for the base case of the induction argument, we now just need to show that $\sqrt{n} \big( \betahat{1} - \betastar1 \big) = O_P(1)$.

\proofSubsubsection{Showing that $\sqrt{n} \big( \betahat{1} - \betastar1 \big) = O_P(1)$} By the definitions of $\betahat{1}$ and $\betastar{1}$ from displays \eqref{eqn:betahatDef} and \eqref{eqn:betastarDef} respectively, we can rewrite the left-hand side of the display \eqref{norm:equicontinuityBase} above:
\begin{equation*}
    \sqrt{n} \big\{ \underbrace{ \piEstHat{1}(\betahat{1} ) }_{=o_P(1/\sqrt{n})} - \piEst{1}(\betahat{1} )  \big\}
    = \sqrt{n} \big\{ \underbrace{ \piEst{1}(\betastar{1} ) }_{=0} - \piEst{1}(\betahat{1} ) \big\} + o_P(1).
\end{equation*}

By Condition \ref{cond:differentiabilityPolicy} (Differentiability of Policy Parameter Estimating Functions), the mapping $\beta_{1} \mapsto \piEst{1}(\beta_{1} )$ is differentiable at $\beta_{1} = \betastar{1}$ with derivative matrix $\piEstDotStar{1} \triangleq \fracpartial{}{\beta_1} \piEst{1}(\beta_{1} ) \big|_{\beta_1 = \betastar1}$. So,
\begin{equation*}
    = - \sqrt{n} \piEstDotStar{1} \big( \betahat{1} - \betastar{1} \big) + \sqrt{n} o_P \big( \big\| \betahat{1} - \betastar{1} \big\|_2 \big) + o_P(1).
\end{equation*}

In summary, we have that
\begin{equation*}
    \sqrt{n} \left\{ \piEstHat{1}(\betastar{1}) - \piEst{1}(\betastar{1}) \right\} + o_P(1)
    = - \sqrt{n} \piEstDotStar{1} \big( \betahat{1} - \betastar{1} \big) + \sqrt{n} o_P \big( \big\| \betahat{1} - \betastar{1} \big\|_2 \big).
\end{equation*}

By Condition \ref{cond:differentiabilityPolicy} (Differentiability of Policy Parameter Estimating Functions), $\piEstDotStar{1}$ is invertible, so $\big[ \piEstDotStar{1} \big]^{-1} = O(1)$ and
\begin{multline}
    \label{eqn:betahatRemainderAlgBase}
    \sqrt{n} \big[ \piEstDotStar{1} \big]^{-1} \left\{ \piEstHat{1}(\betastar{1}) - \piEst{1}(\betastar{1}) \right\} + o_P(1) \\
    = - \sqrt{n} \big( \betahat{1} - \betastar{1} \big) + \sqrt{n} O(1) o_P \big( \big\| \betahat{1} - \betastar{1} \big\|_2 \big).
\end{multline}

By the central limit theorem (for i.i.d. data), 
\begin{multline}
    \label{eqn:phiAlgOp1Base}
    \sqrt{n} \big[ \piEstDotStar{1} \big]^{-1} \big\{ \piEstHat{1}(\betastar{1}) - \underbrace{ \piEst{1}(\betastar{1}) }_{=0} \big\}  
    = \frac{1}{\sqrt{n}} \big[ \piEstDotStar{1} \big]^{-1} \sum_{i=1}^n \piest{1} \big( \history{i}{1}; \betastar1 \big) + o_P(1) \\
    \Dto \N \left( 0, ~\big[ \piEstDotStar{1} \big]^{-1} \Sigma_1 \big[ \piEstDotStar{1} \big]^{-1, \top} \right),
\end{multline}
where $\Sigma_1 = \E \big[ \piest{1} \big( \history{i}{1}; \betastar1 \big)^{\otimes 2} \big]$.

By displays \eqref{eqn:betahatRemainderAlgBase} and \eqref{eqn:phiAlgOp1Base} we have that $- \sqrt{n} \big( \betahat{1} - \betastar{1} \big) + \sqrt{n} O(1) o_P \big( \big\| \betahat{1} - \betastar{1} \big\|_2 \big) = O_P(1)$. This implies that $\sqrt{n} \big( \betahat{1} - \betastar{1} \big) = O_P(1)$; thus, $\sqrt{n} O(1) o_P \big( \big\| \betahat{1} - \betastar{1} \big\|_2 \big) = o_P(1)$. Using these results, we have that
\begin{equation*}
    \sqrt{n} \big( \betahat{1} - \betastar{1} \big)
    = - \sqrt{n} \big[ \piEstDotStar{1} \big]^{-1} \left\{ \piEstHat{1}(\betastar{1}) - \piEst{1}(\betastar{1}) \right\} + o_P(1)
    = O_P(1).
\end{equation*}

\proofSubsection{Induction Step} %%%%%%%%%%%%%%%%%%%%%%%%%%%%%%%%%%
For the induction step, for a given $t > 1$, we make the induction assumption that $\sqrt{n} \big( \betahat{1:t-1} - \betastar{1:t-1} \big) = O_P(1)$. We then show that $\sqrt{n} \big( \betahat{1:t} - \betastar{1:t} \big) = O_P(1)$ and $\sqrt{n} \big\{ \piEstHat{1:t}(\betahat{1:t}) -  \piEst{1:t}(\betahat{1:t}) \big\} 
= \sqrt{n} \big\{ \piEstHat{1:t}(\betastar{1:t}) -  \piEst{1:t}(\betastar{1:t}) \big\} + o_P(1)$.

\proofSubsubsection{Stochastic Equicontinuity Result} First, we will apply Lemma \ref{lemma:stochasticEquicontinuity} (Stochastic Equicontinuity) to get that for any fixed vector $c \in \real^{d_{1:t}}$ (we use $d_{1:t} \triangleq \sum_{t'=1}^t d_{t'}$),
\begin{equation}
    \label{norm:equicontinuityPre}
    \sqrt{n} c^\top \left\{ \piEstHat{1:t}(\betahat{1:t}) - \piEst{1:t}(\betahat{1:t}) \right\}
    = \sqrt{n} c^\top \left\{ \piEstHat{1:t}(\betastar{1:t}) - \piEst{1:t}(\betastar{1:t}) \right\} + o_P(1).
\end{equation}
We are able to apply Lemma \ref{lemma:stochasticEquicontinuity} (Stochastic Equicontinuity) because the following assumptions hold:
\begin{itemize}
    \item Conditions \ref{cond:exploration} (Minimum Exploration) and \ref{cond:lipschitzPolicy} (Lipschitz Policy Functions) hold.
    %%%%%%%%%%%%%%%
    \item $\betahat{1:t-1} - \betastar{1:t-1} = O_P(1/\sqrt{n})$ by our induction assumption.
    %%%%%%%%%%%%%%%
    \item Since assumption \ref{normalityBeta:bracketing} (Finite Bracketing Integral) and Condition \ref{cond:lipschitzPolicy} (Lipschitz Policy Function) hold, we can apply Lemma \ref{lemma:bracketingProduct} (Finite Bracketing Integral) to get that for any vector $c \in \real^{d_t}$,
    \begin{equation*}
        \int_0^1 \sqrt{ \log N_{[~]} \left( \epsilon, ~ \big\{ \pi_{2:t}(\cdotspace; \beta_{1:t-1} ) c^\top \piest{t}(\cdotspace; \beta_t) \big\}_{\beta_{1:t} \in B_{1:t}}, ~ L_{2+\alpha}(\Pstar) \right) } d \epsilon < \infty,
    \end{equation*}
    where $\pi_{2:t}(\cdotspace; \beta_{1:t-1} ) \triangleq \prod_{t'=2}^t \pi_{t'}(\cdotspace; \beta_{t'-1} )$; specifically see the result in Remark \ref{remark:bracketingProduct} part \ref{productBracket:integralPhi} below the statement of Lemma \ref{lemma:bracketingProduct}.
    \smallskip
    %%%%%%%%%%%%%%%
    \item $\betahat{1:t} \Pto \betastar{1:t}$ by Condition \ref{cond:consistencyPolicy}. Thus, by assumption \ref{normalityBeta:L2convergence} and continuous mapping theorem we have that for any vector $c \in \real^{d_t}$, \\
    $\nu \left( \pi_{2:t}(\cdotspace; \betahat{1:t-1} ) c^\top \piest{t}(\cdotspace; \betahat{t}), ~ \pi_{2:t}(\cdotspace; \betastar{1:t-1} ) c^\top \piest{t}(\cdotspace; \betastar{t}) \right) \Pto 0$, where
    \begin{multline*}
        \nu \bigg( \pi_{2:t}(\cdotspace; \beta_{1:t-1} ) c^\top \piest{t}(\cdotspace; \beta_t), ~ \pi_{2:t}(\cdotspace; \beta_{1:t-1}' ) c^\top \piest{t}(\cdotspace; \beta_t') \bigg) \\
        \triangleq \Estar{2:t} \left[ \left\{ \pi_{2:t}(\cdotspace; \beta_{1:t-1} ) c^\top \piest{t}(\cdotspace; \beta_t)
        - \pi_{2:t}(\cdotspace; \beta_{1:t-1}' ) c^\top \piest{t}(\cdotspace; \beta_t') \right\}^2 \right]^{1/2}.
    \end{multline*}
\end{itemize}

\medskip
By display \eqref{norm:equicontinuityPre} and Cramer Wold device, we have that
\begin{equation}
    \label{norm:equicontinuity}
    \sqrt{n} \left\{ \piEstHat{1:t}(\betahat{1:t}) - \piEst{1:t}(\betahat{1:t}) \right\}
    = \sqrt{n} \left\{ \piEstHat{1:t}(\betastar{1:t}) - \piEst{1:t}(\betastar{1:t}) \right\} + o_P(1).
\end{equation}
With the above result, for the induction step of the induction argument, we now just need to show that $\sqrt{n} \big( \betahat{1:t} - \betastar{1:t} \big) = O_P(1)$.

\proofSubsubsection{Showing that $\sqrt{n} \big( \betahat{1:t} - \betastar{1:t} \big) = O_P(1)$} 
By the definitions of $\betahat{1:t}$ and $\betastar{1:t}$ from displays \eqref{eqn:betahatDef} and \eqref{eqn:betastarDef} respectively, we can rewrite the left-hand side of the display \eqref{norm:equicontinuity} above:
\begin{equation*}
    \sqrt{n} \big[ \underbrace{ \piEstHat{1:t}(\betahat{1:t} ) }_{=o_P(1/\sqrt{n})} - \piEst{1:t}(\betahat{1:t} )  \big]
    = \sqrt{n} \big[ \underbrace{ \piEst{1:t}(\betastar{1:t} ) }_{=0} - \piEst{1:t}(\betahat{1:t} )  \big] + o_P(1).
\end{equation*}

By Condition \ref{cond:differentiabilityPolicy} (Differentiability of Policy Parameter Estimating Functions), the mapping $\beta_{1:t} \mapsto \piEst{1:t}(\beta_{1:t} )$ is differentiable at $\beta_{1:t} = \betastar{1:t}$ with the derivative matrix \\
$\piEstDotStar{1:t} \triangleq \fracpartial{}{\beta_{1:t}} \piEst{1:t}(\beta_{1:t} ) \big|_{\beta_{1:t} = \betastar{1:t}}$. So,
\begin{equation*}
    = - \sqrt{n} \piEstDotStar{1:t} \big( \betahat{1:t} - \betastar{1:t} \big) + \sqrt{n} o_P \big( \big\| \betahat{1:t} - \betastar{1:t} \big\|_2 \big) + o_P(1).
\end{equation*}

In summary, we have that
\begin{equation*}
    \sqrt{n} \left\{ \piEstHat{1:t}(\betastar{1:t}) - \piEst{1:t}(\betastar{1:t}) \right\} + o_P(1)
    = - \sqrt{n} \piEstDotStar{1:t} \big( \betahat{1:t} - \betastar{1:t} \big) + \sqrt{n} o_P \big( \big\| \betahat{1:t} - \betastar{1:t} \big\|_2 \big).
\end{equation*}

By Condition \ref{cond:differentiabilityPolicy}, we can apply Lemma \ref{lemma:invertibilityPiDotStar} (Invertibility of $\piEstDotStar{1:t}$) to get that $\piEstDotStar{1:t}$ is invertible; so $\big[ \piEstDotStar{1:t} \big]^{-1} = O(1)$ and
\begin{multline}
    \label{eqn:betahatRemainderAlg}
    \sqrt{n} \big[ \piEstDotStar{1:t} \big]^{-1} \left\{ \piEstHat{1:t}(\betastar{1:t}) - \piEst{1:t}(\betastar{1:t}) \right\} + o_P(1) \\
    = - \sqrt{n} \big( \betahat{1:t} - \betastar{1:t} \big) + \sqrt{n} O(1) o_P \big( \big\| \betahat{1:t} - \betastar{1:t} \big\|_2 \big).
\end{multline}

For now, we take as given that the following result in display \eqref{eqn:phiAlgOp1} holds; we prove this at the end of this proof.
\begin{equation}
    \label{eqn:phiAlgOp1}
    \sqrt{n} \big[ \piEstDotStar{1:t} \big]^{-1} \left\{ \piEstHat{1:t}(\betastar{1:t}) - \piEst{1:t}(\betastar{1:t}) \right\} = O_P(1).
\end{equation}

By displays \eqref{eqn:betahatRemainderAlg} and \eqref{eqn:phiAlgOp1} we have that $- \sqrt{n} \big( \betahat{1:t} - \betastar{1:t} \big) + \sqrt{n} O(1) o_P \big( \big\| \betahat{1:t} - \betastar{1:t} \big\|_2 \big) = O_P(1)$. This implies that $\sqrt{n} \big( \betahat{1:t} - \betastar{1:t} \big) = O_P(1)$; thus, $\sqrt{n} O(1) o_P \big( \big\| \betahat{1:t} - \betastar{1:t} \big\|_2 \big) = o_P(1)$. Using these results, we have that
\begin{equation*}
    \sqrt{n} \big( \betahat{1:t} - \betastar{1:t} \big)
    = - \sqrt{n} \big[ \piEstDotStar{1:t} \big]^{-1} \left\{ \piEstHat{1:t}(\betastar{1:t}) - \piEst{1:t}(\betastar{1:t}) \right\} + o_P(1) = O_P(1).
\end{equation*}

\proofSubsection{We now show that display \eqref{eqn:phiAlgOp1} holds.} 
 For any fixed vector $c = [c_1, c_2, \cdots, c_{t} ] \in \real^{d_{1:t}}$, 
\begin{multline}
    \label{eqn:phiAlgOp1Full}
    \sqrt{n} c^\top \big\{ \piEstHat{1:t-1}(\betastar{1:t}) - \underbrace{ \piEst{1:t}(\betastar{1:t}) }_{=0} \big\} \\
    = \frac{1}{\sqrt{n}} \sum_{i=1}^n \sum_{k=1}^t c_{k}^\top \WW{i}{2:k}(\betastar{1:k-1}, \betahat{1:k-1}) \piest{k}(\history{i}{k}; \betastar{k}) 
    \Dto \N \left( 0, c^\top \Sigma_{1:t} c \right),
\end{multline}
where
\begin{equation*}
    \Sigma_{1:t} \triangleq \Estar{2} \left[ \begin{pmatrix}
        \piest1(\history{i}{1}; \betastar1) \\
        \piest2(\history{i}{2}; \betastar2) \\
        \vdots \\
        \piest{t}(\history{i}{t}; \betastar{t})
    \end{pmatrix}^{\otimes 2} \right].
\end{equation*}
The asymptotic normality result above holds by Theorem \ref{thm:weightedCLT} (Weighted Martingale Triangular Array Central Limit Theorem). Specifically we can apply Theorem \ref{thm:weightedCLT} because:
\begin{itemize}
    \item Conditions \ref{cond:exploration} (Minimum Exploration) and \ref{cond:lipschitzPolicy} (Lipschitz Policy Functions) hold.
    \item By our induction assumption, $\betahat{t'} - \betastar{t'} = O_P( 1/\sqrt{n} )$ for all $t' \in [1 \colon t-1]$.
    \item $\Estar{2:t} \big[ \big| \sum_{t'=1}^t \piest{t'}(\history{i}{t'}; \betastar{t'}) \big|^{2+\alpha} \big] < \infty$ by the Finite Bracketing Integral assumption \ref{normalityBeta:bracketing}.
\end{itemize}

Thus, by Cramer-Wold device and display \eqref{eqn:phiAlgOp1Full} we have that
\begin{equation*}
    \sqrt{n} \left\{ \piEstHat{1:t}(\betastar{1:t}) - \piEst{1:t}(\betastar{1:t}) \right\} 
    \Dto \N( 0, \Sigma_{1:t} ).
\end{equation*}
By Condition \ref{cond:differentiabilityPolicy}, we can apply Lemma \ref{lemma:invertibilityPiDotStar} (Invertibility of $\piEstDotStar{1:t}$) to get that $\piEstDotStar{1:t}$ is invertible. Thus, by continuous mapping theorem,
\begin{equation*}
    \sqrt{n} \big[ \piEstDotStar{1:t} \big]^{-1} \left\{ \piEstHat{1:t}(\betastar{1:t}) - \piEst{1:t}(\betastar{1:t}) \right\} 
    \Dto \N \left( 0, \big[ \piEstDotStar{1:t} \big]^{-1} \Sigma_{1:t} \big[ \piEstDotStar{1:t} \big]^{-1, \top} \right).
\end{equation*}
The above implies that display \eqref{eqn:phiAlgOp1} holds, i.e., that $\sqrt{n} \big[ \piEstDotStar{1:t} \big]^{-1} \big\{ \piEstHat{1:t}(\betastar{1:t}) - \piEst{1:t}(\betastar{1:t}) \big\} = O_P(1)$. ~$\blacksquare$

\subsection{Invertibility of $\piEstDotStar{1:t}$ (Lemma \ref{lemma:invertibilityPiDotStar})} %%%%%%%%%%%%%%%%%%%%%%%%%%%%%%%%%%%
\label{mainapp:lemmaInvertibility}

\begin{lemma}[Invertibility of $\piEstDotStar{1:t}$]
    \label{lemma:invertibilityPiDotStar}
    Under Condition \ref{cond:differentiabilityPolicy} (Differentiability of Policy Parameter Estimating Functions), for each $t \in [1 \colon T-1]$, $\piEstDotStar{1:t}$ is invertible.
\end{lemma}

\noindent In the proof of Lemma \ref{lemma:invertibilityPiDotStar} we will use the following proposition:
\begin{proposition}[Blockwise Inversion of Matrix]
    \label{prop:blockInversion}
    Let $A \in \real^{k \by k}$, $B \in \real^{k \by j}$, $C \in \real^{j \by k}$, and $D \in \real^{j \by j}$. If $A$ and $D - C A^{-1} B$ are invertible then
    \begin{equation*}
        \begin{bmatrix}
            A & & B \\
            C & & D
        \end{bmatrix}^{-1}
        = \begin{bmatrix}
            A^{-1} + A^{-1} B (D - C A^{-1} B)^{-1} C A^{-1} & & - A^{-1} B (D - C A^{-1} B)^{-1} \\
            -(D - C A^{-1} B)^{-1} C A^{-1} & & (D - C A^{-1} B )^{-1}
        \end{bmatrix}.
    \end{equation*}
    Furthermore, in the special case that $B = 0$,
    \begin{equation*}
        \begin{bmatrix}
            A & & 0 \\
            C & & D
        \end{bmatrix}^{-1}
        = \begin{bmatrix}
            A^{-1} & & 0 \\
            -D^{-1} C A^{-1} & & D^{-1}
        \end{bmatrix}.
    \end{equation*}
    This is a result is proved in Proposition 3.9.7 of \cite{bernstein2018matrix}.
\end{proposition}

\startproof{Lemma \ref{lemma:invertibilityPiDotStar}}
By Condition \ref{cond:differentiabilityPolicy} (Differentiability of Policy Parameter Estimating Functions), the mapping $\beta_{1:t} \mapsto \piEst{1:t}(\beta_{1:t})$ is differentiable at $\beta_{1:t} = \betastar{1:t}$.
Let $\piEstDotStar{1:t} \triangleq \fracpartial{}{\beta_{1:t}} \piEst{t}(\beta_{1:t}) \big|_{\beta_{1:t} = \betastar{1:t}}$ Specifically,
\begin{equation*}
    \piEstDotStar{1:t}
    = \fracpartial{}{\beta_{1:t}} \E_{\pi(\beta_{1:t-1})} \begin{bmatrix}
            \piest1(\history{i}{1}; \beta_1) \\
            \piest2(\history{i}{2}; \beta_2) \\
            \vdots \\
            \piest{t}(\history{i}{t}; \beta_{t})
        \end{bmatrix} \bigg|_{\beta_{1:t} = \betastar{1:t}}
    = \begin{bmatrix}
        \piEstDotStar{1} && 0 && 0 && \hdots && 0 \\
        V_{2,1} && \piEstDotStar{2} && 0 && \hdots && 0 \\
        V_{3,1} && V_{3,2} && \piEstDotStar{3} && \hdots && 0 \\
        \vdots && \vdots && \vdots && \ddots && \vdots \\
        V_{t,1} && V_{t,2} && V_{t,3} && \hdots && \piEstDotStar{t}
    \end{bmatrix},
\end{equation*}
where $V_{t,s} \triangleq \fracpartial{}{\beta_{s}} \Estar{2:t} \left[ \piest{t}(\history{i}{t}; \betastar{t}) \WW{i}{s+1}(\beta_{s}, \betastar{s}) \right] \big|_{\beta_s = \betastar{s}} \in \real^{d_t \by d_s}$ and \\ 
$\piEstDotStar{t} \triangleq \fracpartial{}{\beta_{t}} \piEst{t}(\betastar{1:t-1}, \beta_{t}) \big|_{\beta_{t} = \betastar{t}} \triangleq \fracpartial{}{\beta_{t}} \Estar{2:t} \big[ \piest{t}(\history{i}{t}; \beta_{t}) \big] \big|_{\beta_{t} = \betastar{t}}$.

Note that by repeatedly applying Proposition \ref{prop:blockInversion} we can show that $\piEstDotStar{1:t}$ is invertible. Moreover, we will show that the inverse of $\piEstDotStar{1:t}$ is lower block triangular. To see this, consider the following induction argument:
\begin{itemize}
    \item \textit{Base case:} By Condition \ref{cond:differentiabilityPolicy} (Differentiability of Policy Parameter Estimating Functions), $\piEstDotStar{1}$ and $\piEstDotStar{2}$ are both invertible. Thus, by Proposition \ref{prop:blockInversion}, the matrix $\begin{bmatrix}
        \piEstDotStar{1} & 0 \\
        V_{2,1} & \piEstDotStar{2} 
    \end{bmatrix}$ is invertible; moreover, its inverse is lower block triangular.
    \smallskip %%%%%%%%%%%%%%%%%%%%%
    \item \textit{Induction step:} Suppose we know that for some $t \geq 2$, the inverse of the following lower block triangular matrix is invertible:
    \begin{equation}
        \label{eqn:blockInversionInduction}
        \begin{bmatrix}
            \piEstDotStar{1} & 0 & 0 & \hdots & 0 \\
            V_{2,1} & \piEstDotStar{2} & 0 & \hdots & 0 \\
            V_{3,1} & V_{3,2} & \piEstDotStar{3} & \hdots & 0 \\
            \vdots & \vdots & \vdots & \ddots & \vdots \\
            V_{t-1,1} & V_{t-1,2} & V_{t-1,3} & \hdots & \piEstDotStar{t-1} 
        \end{bmatrix}.
    \end{equation}
    By Condition \ref{cond:differentiabilityPolicy} (Differentiability of Policy Parameter Estimating Functions), $\piEstDotStar{t}$ is invertible. By Proposition \ref{prop:blockInversion} we can conclude that the following matrix is invertible and lower block triangular:
    \begin{equation*}
        \begin{bmatrix}
            \piEstDotStar{1} & 0 & 0 & \hdots & 0 & 0 \\
            V_{2,1} & \piEstDotStar{2} & 0 & \hdots & 0 & 0 \\
            V_{3,1} & V_{3,2} & \piEstDotStar{3} & \hdots & 0 & 0 \\
            \vdots & \vdots & \vdots & \ddots & \vdots & \vdots \\
            V_{t-1,1} & V_{t-1,2} & V_{t-1,3} & \hdots & \piEstDotStar{t-1} & 0 \\
            V_{t,1} & V_{t,2} & V_{t,3} & \hdots & V_{t,t-1} & \piEstDotStar{t} 
        \end{bmatrix}.
    \end{equation*}
    When applying Proposition \ref{prop:blockInversion}, take the matrix from display \eqref{eqn:blockInversionInduction} (upper left-hand side of the matrix above) to be matrix $A$; take $\piEstDotStar{t}$ to be matrix $D$; take $\begin{bmatrix} V_{t,1} & V_{t,2} & V_{t,3} & \hdots & V_{t,t-1} \end{bmatrix}$ to be matrix $C$; and take the block of zeros above $\piEstDotStar{t}$ to be matrix $B$.
\end{itemize}
By the above argument, $\piEstDotStar{1:t}$ is invertible for any $t \in [1 \colon T-1]$. $~\blacksquare$

\clearpage
%%%%%%%%%%%%%%%%%%%%%%%%%%%%%%%%%%%%%%%%%%%%
\section{Main Asymptotic Results}
\label{app:mainResults}

\startOverview{Overview of Supplement \ref{app:mainResults} Results}
\begin{itemize}
	%%%%%%%%%%%%%%%
    \item \bo{Section \ref{mainapp:consistency}:} Consistency of $\thetahat{}$ (Theorem \ref{thm:consistency})
    %%%%%%%%%%%%%%%%%%
    \item \bo{Section \ref{mainapp:normalityAdaptiveUnpack}:} Equivalent Formulations for the Adaptive Sandwich Variance (Lemma \ref{lemma:normalityAdaptiveUnpack})
    %%%%%%%%%%%%%%%%%%
    \item \bo{Section \ref{mainapp:normalityProof}:} Asymptotic Normality of $\thetahat{}$ (Theorem \ref{thm:normality})
    %%%%%%%%%%%%%%%%%%
\end{itemize}

%%%%%%%%%%%%%%%%%%%%%%%%%%%%%%%%%%%%%%%%%%
%%%%%%%%%%%%%%%%%%%%%%%%%%%%%%%%%%%%%%%%%%
\subsection{Consistency of $\thetahat{}$ (Theorem \ref{thm:consistency})}
\label{mainapp:consistency}

~ \\

\startproof{Theorem \ref{thm:consistency}}
Note that this argument is extremely similar to the proof for Theorem \ref{thm:consistencyBeta}. The proof will use the estimating functions $\Est \big( \beta_{1:T-1}, \theta \big)$ and $\EstHat \big( \beta_{1:T-1}, \theta \big)$ defined earlier in displays \eqref{eqn:estDefWeights} and \eqref{proofSktech:EstHat} respectively.

%the following functions which were first defined in display \eqref{proofSktech:EstHat}:
%\begin{equation}
%    \EstHat \big( \beta_{1:T-1}, \theta \big) \triangleq \frac{1}{n} \sum_{i=1}^n \bigg\{ \prod_{t=2}^T \WW{i}{t} \big( \beta_{t-1}, \betahat{t-1} \big) \bigg\} \est{} \big( \history{i}{t}; \theta \big)
%\end{equation}
%and
%\begin{equation*}
%    \Est \big( \beta_{1:T-1}, \theta \big) 
%    \triangleq \E_{\pi(\beta_{1:T-1})} \left[ \est{} \big( \history{i}{t}; \theta \big) \right]
%    = \Estar{2:T} \bigg[ \bigg\{ \prod_{t=2}^T \WW{i}{t} \big( \beta_{t-1}, \betastar{t-1} \big) \bigg\} \est{} \big( \history{i}{t}; \theta \big) \bigg].
%\end{equation*}

For now, we take as given that display \eqref{eqn:PhiBetahatAssumption} below holds; we will show this result holds at the end of this proof.
\begin{equation}
    \label{eqn:PsiBetahatAssumption}
    \left\| \Est{} \big( \betahat{1:T-1}, \thetastar{} \big) - \Est{} \big( \betastar{1:T-1}, \thetastar{} \big) \right\|_1
    = o_P(1).
\end{equation}

Let $\epsilon > 0$. By assumption \ref{consistency:wellSeparated}, there exists some $\eta > 0$ such that if $\theta \in \real^{d_\theta}$ satisfies $\big\| \theta - \thetastar{} \big\|_1 > \epsilon$, then $\big\| \Est{} \big( \betastar{1:T-1}, \theta \big) \big\|_1 = \big\| \Estar{2:T} \big[ \est{}(\history{i}{T}; \theta) \big] \big\|_1 > \eta > 0$. Thus,
\begin{equation*}
    \PP \left( \big\| \thetahat{} - \thetastar{} \big\|_1 > \epsilon \right)
    \leq \PP \left( \big\| \Est{} \big( \betastar{1:T-1}, \thetahat{} \big) \big\|_1 > \eta \right).
\end{equation*}

Note $\big\| \Est{} \big( \betastar{1:T-1}, \thetahat{} \big) \big\|_1 
= \big\| \Est{} \big( \betastar{1:T-1}, \thetahat{} \big) - \Est{} \big( \betahat{1:T-1}, \thetahat{} \big) + \Est{} \big( \betahat{1:T-1}, \thetahat{} \big) \big\|_1 \\
\leq \big\| \Est{} \big( \betastar{1:T-1}, \thetahat{} \big) - \Est{} \big( \betahat{1:T-1}, \thetahat{} \big) \big\|_1 + \big\| \Est{} \big( \betahat{1:T-1}, \thetahat{} \big) \big\|_1 \\ 
= \big\| \Est{} \big( \betahat{1:T-1}, \thetahat{} \big) \big\|_1 + o_P(1)$; the last equality holds by display \eqref{eqn:PsiBetahatAssumption}. Thus,
\begin{equation*}
    \leq \PP \left( \big\| \Est{} \big( \betahat{1:T-1}, \thetahat{} \big) \big\|_1 > \eta  - o_P(1) \right).
\end{equation*}

Note that $\Est{} \big( \betahat{1:T-1}, \thetahat{} \big)
= \frac{1}{n} \sum_{i=1}^n \WW{i}{2:T} \big(\betahat{1:T-1}, \betahat{1:T-1} \big) \est{} \big( \history{i}{T}; \thetahat{} \big) \\
= \frac{1}{n} \sum_{i=1}^n \est{} \big( \history{i}{T}; \thetahat{} \big) = o_P(1/\sqrt{n}) = o_P(1)$; the second to last equality holds by the definition of $\thetahat{}$ from display \eqref{eqn:ZestimatorDef}. Thus,
\begin{equation*}
    = \PP \left( \big\| \EstHat{} \big( \betahat{1:T-1}, \thetahat{} \big) - \Est{} \big( \betahat{1:T-1}, \thetahat{} \big) \big\|_1 > \eta - o_P(1) \right).
\end{equation*}

By assumption \ref{consistency:tight}, for any $\delta > 0$, there exists some $k < \infty$ such that \\
$\limsup_{n \to \infty} \PP \big( \big\| \thetahat{} \big\|_1 > k \big) \leq \delta$. We use this $k$ below:
\begin{equation*}
     = \PP \left( \big\| \EstHat{} \big( \betahat{1:T-1}, \thetahat{} \big) - \Est{} \big( \betahat{1:T-1}, \thetahat{} \big) \big\|_1 
     \left\{ \II_{ \| \thetahat{}  \|_1 > k } + \II_{ \| \thetahat{} \|_1 \leq k } \right\} > \eta - o_P(1) \right)
\end{equation*}
\begin{multline*}
     \leq \PP \left( \big\| \EstHat{} \big( \betahat{1:T-1}, \thetahat{} \big) - \Est{} \big( \betahat{1:T-1}, \thetahat{} \big) \big\|_1 
     ~\II_{ \| \thetahat{} \|_1 \leq k } > \eta/2 - o_P(1) \right) \\
     + \PP \left( \big\| \EstHat{} \big( \betahat{1:T-1}, \thetahat{} \big) - \Est{} \big( \betahat{1:T-1}, \thetahat{} \big) \big\|_1 
     ~\II_{ \| \thetahat{} \|_1 > k } > \eta/2 - o_P(1) \right)
\end{multline*}
\begin{multline*}
     \leq \PP \left( \big\| \EstHat{} \big( \betahat{1:T-1}, \thetahat{} \big) - \Est{} \big( \betahat{1:T-1}, \thetahat{} \big) \big\|_1 
     ~\II_{ \| \thetahat{} \|_1 \leq k } > \eta/2 - o_P(1) \right) \\
     + \PP \left( \big\| \thetahat{} \big\|_1 > k \right) + o(1)
\end{multline*}
\begin{multline*}
     \leq \PP \bigg( \sup_{\theta \in \real^{d_\theta} \TN{~s.t.~} \| \theta \|_1 \leq k} \big\| \EstHat{} \big( \betahat{1:T-1}, \theta \big) - \Est{} \big( \betahat{1:T-1}, \theta \big) \big\|_1 > \eta/2 - o_P(1) \bigg) \\
     + \PP \left( \big\| \thetahat{} \big\|_1 > k \right) + o(1)
\end{multline*}

Since $\betahat{1:T-1} \Pto \betastar{1:T-1}$ by Condition \ref{cond:consistencyPolicy}, thus $\II_{ \betahat{1:T-1} \in B_{1:T-1} } \Pto 1$; recall that $B_{1:T-1} \subset \real^{d_{1:T-1}}$ is a compact subset whose interior contains $\betastar{1:T-1}$. Thus,
\begin{multline*}
     \leq \underbrace{ \PP \bigg( \sup_{\theta \in \real^{d_\theta} \TN{~s.t.~} \| \theta \|_1 \leq k} ~ \sup_{\beta_{1:t-1} \in B_{1:t-1}} \big\| \EstHat{} \big( \beta_{1:T-1}, \theta \big) - \Est{}\big( \beta_{1:T-1}, \theta \big) \big\|_1 > \eta/2 - o_P(1) \bigg) }_{=o(1)} \\
     + \underbrace{ \PP \left( \big\| \thetahat{} \big\|_1 > k \right) }_{\leq \delta} + o(1).
\end{multline*}
Note that the above converges to zero as $n \to \infty$ for the following reasons:
\begin{itemize}
    \item By assumption \ref{consistencyBeta:tight}, $\limsup_{n \to \infty} \PP \big( \big\| \thetahat{} \big\|_1 > k \big) \leq \delta$ and $\delta$ can be made arbitrarily small.
    %%%%%%%%%%%%%
    \item Note that by Cramer Wold device, to show that
    \begin{equation*}
        \sup_{\theta \in \real^{d_\theta} \TN{~s.t.~} \| \theta \|_1 \leq k} ~ \sup_{\beta_{1:t-1} \in B_{1:t-1}} \left\| \EstHat{} \big( \beta_{1:T-1}, \theta \big) - \Est{} \big( \beta_{1:T-1}, \theta \big) \right\|_1 \Pto 0,
    \end{equation*}
    it is sufficient to show that for any vector $c \in \real^{d_\theta}$,
    \begin{equation*}
        \sup_{\theta \in \real^{d_\theta} \TN{~s.t.~} \| \theta \|_1 \leq k} ~ \sup_{\beta_{1:t-1} \in B_{1:t-1}} c^\top \left\{ \EstHat{} \big( \beta_{1:T-1}, \theta \big) - \Est{} \big( \beta_{1:T-1}, \theta \big) \right\} \Pto 0.
    \end{equation*}
    Also note that
    \begin{equation*}
        \sup_{\theta \in \real^{d_\theta} \TN{~s.t.~} \| \theta \|_1 \leq k} ~ \sup_{\beta_{1:t-1} \in B_{1:t-1}} c^\top \left\{ \EstHat{} \big( \beta_{1:T-1}, \theta \big) - \Est{} \big( \beta_{1:T-1}, \theta \big) \right\}
    \end{equation*}
    \begin{multline*}
        = \sup_{\theta \in \real^{d_\theta} \TN{~s.t.~} \| \theta \|_1 \leq k} ~ \sup_{\beta_{1:t-1} \in B_{1:t-1}} \frac{1}{n} \sum_{i=1}^n \bigg\{ \WW{i}{2:T}(\beta_{1:T-1}, \betahat{1:T-1}) c^\top \est{}(\history{i}{T}; \theta) \\
        - \E \left[ \WW{i}{2:T}(\beta_{1:T-1}, \betahat{1:T-1}) c^\top \est{} (\history{i}{T}; \theta) \right] \bigg\} \Pto 0.
    \end{multline*}
    The above convergence result holds by Theorem \ref{thm:weightedUWLLN} (Weighted Martingale Triangular Array Uniform Weak Law of Large Numbers).
    Specifically we are able to apply Theorem \ref{thm:weightedUWLLN} because Condition \ref{cond:exploration} holds and $N_{[~]} \left( \epsilon, \F_{\Pi c^\top \est{}}(B_{1:T-1}, k), L_{1+\alpha}(\Pstar) \right) < \infty$ for any $c \in \real^{d_\theta}$ and any $\epsilon > 0$, where
    \begin{equation*}
        \F_{\Pi c^\top \est{}}(B_{1:T-1}, k) 
        \triangleq \bigg\{ \bigg[ \prod_{t=2}^T \pi_{t} ( \cdotspace; \beta_{t-1} ) \bigg] c^\top \est{}(\cdotspace; \theta \big) \bigg\}_{\beta_{1:T-1} \in B_{1:T-1}, ~ \theta \in \real^{d_\theta} \TN{~s.t.~} \| \theta \|_1 \leq k }.
    \end{equation*}
    The above finite bracketing number result holds since using assumption \ref{consistencyBeta:finiteBracketing} and Condition \ref{cond:lipschitzPolicy}, we can apply by Lemma \ref{lemma:bracketingProduct} (specifically see Remark \ref{remark:bracketingProduct} part \ref{productBracket:finitePsi}).
\end{itemize}

\proofSubsection{We now show that display \eqref{eqn:PsiBetahatAssumption} holds} Let $\beta_{1:T-1} \in B_{1:T-1}$.
\begin{equation*}
    \left\| \Est{} \big( \beta_{1:T-1}, \thetastar{} \big) - \Est{} \big( \betastar{1:T-1}, \thetastar{} \big) \right\|_1
\end{equation*}
\begin{equation*}
    = \bigg\| \E \bigg[ \bigg\{ \prod_{t=2}^T \WW{i}{t} \big(\beta_{t-1}, \betahat{t-1} \big) - \WW{i}{t} \big(\betastar{t-1}, \betahat{t-1} \big) \bigg\} \est{} \big( \history{i}{T}; \thetastar{} \big) \bigg] \bigg\|_1
\end{equation*}
\begin{equation*}
    = \bigg\| \Estar{2:T} \bigg[ \bigg\{ \prod_{t=2}^T \WW{i}{t} \big(\beta_{t-1}, \betastar{t-1} \big) - \WW{i}{t} \big(\betastar{t-1}, \betastar{t-1} \big) \bigg\} \est{} \big( \history{i}{T}; \thetastar{} \big) \bigg] \bigg\|_1
\end{equation*}
By Jensen's inequality,
\begin{equation*}
    \leq \Estar{2:T} \bigg[ \bigg| \prod_{t=2}^T \WW{i}{t} \big(\beta_{t-1}, \betastar{t-1} \big) - \WW{i}{t} \big(\betastar{t-1}, \betastar{t-1} \big) \bigg| \big\| \est{} \big( \history{i}{T}; \thetastar{} \big) \big\|_1 \bigg]
\end{equation*}
By definition of $\WW{i}{t} \big(\beta_{t-1}, \betastar{t-1} \big)$ from display \eqref{eqn:weightsDef},
\begin{multline*}
    = \Estar{2:T} \bigg[ \bigg| \prod_{t=2}^T \pi_{t} \big( \action{i}{t}, \state{i}{t}; \beta_{t-1} \big) - \prod_{t=2}^T \pi_{t} \big( \action{i}{t}, \state{i}{t}; \betastar{t-1} \big) \bigg| \\
    \bigg\{ \prod_{t=2}^T \frac{1}{ \pistar{t}(\action{i}{t}, \state{i}{t}) } \bigg\} \big\| \est{} \big( \history{i}{T}; \thetastar{} \big) \big\|_1 \bigg]
\end{multline*}
By Condition \ref{cond:exploration} (Minimum Exploration), $\pistar{t} \big( \action{i}{t}, \state{i}{t} \big)^{-1} \leq \pi_{\min}^{-1}$ a.s. Thus,
\begin{equation*}
    %\label{eqn:upperBoundConsistencyDiff}
    \leq \pi_{\min}^{-(T-1)} \Estar{2:T} \bigg[ \bigg| \prod_{t=2}^T \pi_{t} \big( \action{i}{t}, \state{i}{t}; \beta_{t-1} \big) - \prod_{t=2}^T \pi_{t} \big( \action{i}{t}, \state{i}{t}; \betastar{t-1} \big) \bigg| \big\| \est{} \big( \history{i}{T}; \thetastar{} \big) \big\|_1 \bigg]
\end{equation*}
By Condition \ref{cond:lipschitzPolicy} and Lemma \ref{lemma:lipschitzPolicyProduct} (Product of Lipschitz Policy Functions are Lipschitz),
\begin{equation*}
    \leq \pi_{\min}^{-(T-1)} \Estar{2:T} \bigg[ \big\| \est{} \big( \history{i}{T}; \thetastar{} \big) \big\|_1 \bigg\{ \sum_{t=2}^T \dot{\pi}_{t}(\action{i}{t}, \state{i}{t}) \bigg\} \big\| \beta_{1:T-1} - \betastar{1:T-1} \big\|_2 \bigg] 
\end{equation*}
By linearity of expectations,
\begin{equation*}
    = \pi_{\min}^{-(T-1)} \bigg\{ \sum_{t=2}^T \Estar{2:T} \left[ \big\| \est{} \big( \history{i}{T}; \thetastar{} \big) \big\|_1 \dot{\pi}_{t}(\action{i}{t}, \state{i}{t}) \right] \bigg\} \big\| \beta_{1:T-1} - \betastar{1:T-1} \big\|_2.
\end{equation*}
Thus, by consolidating the above results, we have that
\begin{multline*}
    \big\| \Est{} \big( \betahat{1:T-1}, \thetastar{} \big) - \Est{} \big( \betastar{1:T-1}, \thetastar{} \big) \big\|_1 \\
    \leq \pi_{\min}^{-(T-1)} \bigg\{ \sum_{t=2}^T \Estar{2:T} \left[ \big\| \est{} \big( \history{i}{T}; \thetastar{} \big) \big\|_1 \dot{\pi}_{t}(\action{i}{t}, \state{i}{t}) \right] \bigg\} \big\| \betahat{1:T-1} - \betastar{1:T-1} \big\|_2
    = o_P(1).
\end{multline*}
The last limit above holds because
\begin{itemize}
    \item $\big\| \betahat{1:T-1} - \betastar{1:T-1} \big\|_2 = o_P(1)$ since $\betahat{1:T-1} \Pto \betastar{1:T-1}$ by Condition \ref{cond:consistencyPolicy}.
    %%%%%%%%%%
    \item By assumption \ref{consistency:finiteBracketing} (Finite Bracketing Number for Policy Functions), there exists a function $F_{\est{}}$ such that $\big\| \est{} \big( \history{i}{T}; \thetastar{} \big) \big\|_1 \leq F_{\est{}}(\history{i}{t})$ a.s. and for all $t \in [2 \colon T]$, \\
    $\Estar{2:T} \big[ F_{\est{}} \big( \history{i}{T} \big) \dot{\pi}_{t}(\action{i}{t}, \state{i}{t}) \big] < \infty$. Thus,
    \begin{equation*}
        \Estar{2:T} \left[ \big\| \est{} \big( \history{i}{T}; \thetastar{} \big) \big\|_1 \dot{\pi}_{t}(\action{i}{t}, \state{i}{t}) \right] 
        \leq \Estar{2:T} \left[ F_{\est{}} \big( \history{i}{T} \big) \dot{\pi}_{t}(\action{i}{t}, \state{i}{t}) \right] < \infty.
    \end{equation*}
\end{itemize}
We have now shown that display \eqref{eqn:PsiBetahatAssumption} holds. ~ $\blacksquare$

%%%%%%%%%%%%%%%%%%%%%%%%%%%%%%%%%%%%%%%%%%%%
\subsection{Equivalent Formulations for the Adaptive Sandwich Variance (Lemma \ref{lemma:normalityAdaptiveUnpack})} %%%%%%%%%%%%%%%%%%%%%%%%%%%%%%%%%%%%%%%%%%%%
%%%%%%%%%%%%%%%%%%%%%%%%%%%%%%%%%%%%%%%%%%%%
\label{mainapp:normalityAdaptiveUnpack}

\begin{lemma}[Equivalent Formulations for the Adaptive Sandwich Variance]
    \label{lemma:normalityAdaptiveUnpack}
    Let Condition \ref{cond:differentiabilityPolicy} (Differentiability of Policy Parameter Estimating Functions) and assumptions \ref{normality:bread} and \ref{normality:differentiable} from Theorem \ref{thm:normality} (Asymptotic Normality hold). Also let $\Sigma_{1:T}$ as defined in display \eqref{eqnMain:SigmaStackedDef} be finite. 
    
    Then, the lower-right $d_\theta \by d_\theta$ block of limiting variance from display \eqref{mainNorm:limit}, i.e.,
    \begin{equation}
        \label{appeqn:stackedVar}
        \begin{bmatrix} 
        \piEstDotStar{1:T-1} && \bs{0} \\
        \bs{V}_{T,1:T-1} && \EstDotStar{}
    \end{bmatrix}^{-1} \Sigma_{1:T} \begin{bmatrix} 
        \piEstDotStar{1:T-1} && \bs{0} \\
        \bs{V}_{T,1:T-1} && \EstDotStar{}
    \end{bmatrix}^{-1, \top},
    \end{equation}
    equals the adaptive sandwich variance, i.e., $[\EstDotStar{}]^{-1} \Madapt{} [\EstDotStar{}]^{-1, \top}$ from display \eqref{eqn:Madapt}.
\end{lemma}

%%%%%%%%%%%%%%%%%%%%%%%%%%%%%%%%%%%%%
\startproof{Lemma \ref{lemma:normalityAdaptiveUnpack}}
By the definition of $\Madapt{}$ from display \eqref{eqn:Madapt}, it is sufficient to show that the lower-right $d_\theta \by d_\theta$ block of the the limiting variance from display \eqref{appeqn:stackedVar} above equals the following
\begin{multline*}
    [\EstDotStar{}]^{-1} \Madapt{} [\EstDotStar{}]^{-1, \top} \\
    = [\EstDotStar{}]^{-1}
    \Estar{2:T} \bigg[ \bigg\{ \est{} \big(\history{i}{T}; \thetastar{} \big) + \EstDotStar{} \sum_{t=1}^{T-1} M_t \piest{t} \big( \history{i}{t} ; \betastar{t} \big) \bigg\}^{\otimes 2} \bigg]
    [\EstDotStar{}]^{-1, \top}.
\end{multline*}

Consider the following matrix from display \eqref{appeqn:stackedVar} (the terms in the matrix below are derivatives that exist by Condition \ref{cond:differentiabilityPolicy}, and assumptions \ref{normality:bread} and \ref{normality:differentiable}):
\begin{equation}
    \label{eqn:brackCompressedStack}
    \begin{bmatrix} 
        \piEstDotStar{1:T-1} && \bs{0} \\
        \bs{V}_{T,1:T-1} && \EstDotStar{}
    \end{bmatrix}.
\end{equation}
By Proposition \ref{prop:blockInversion} (Blockwise Inversion of Matrix), we have that for a block matrix $\begin{bmatrix}
    A & & 0 \\
    C & & D
\end{bmatrix}$, if square matrices $A$ and $D$ are invertible, then the whole matrix is invertible and
\begin{equation*}
    \begin{bmatrix}
        A & & 0 \\
        C & & D
    \end{bmatrix}^{-1}
        = \begin{bmatrix}
        A^{-1} & & 0 \\
        -D^{-1} C A^{-1} & & D^{-1}
    \end{bmatrix}.
\end{equation*}

By Condition \ref{cond:differentiabilityPolicy} and Lemma \ref{lemma:invertibilityPiDotStar} (Invertibility of $\piEstDotStar{1:t}$), $\piEstDotStar{1:t}$ is invertible and $\EstDotStar{}$ is invertible by assumption \ref{normality:bread}. Thus the matrix from display \eqref{eqn:brackCompressedStack} is invertible and
\begin{equation*}
    \begin{bmatrix} 
        \piEstDotStar{1:T-1} && \bs{0} \\
        \bs{V}_{T,1:T-1} && \EstDotStar{}
    \end{bmatrix}^{-1}
    = \begin{bmatrix} 
        \big\{ \piEstDotStar{1:T-1} \big\}^{-1} && \bs{0} \\
        - \big\{ \EstDotStar{} \big\}^{-1}\bs{V}_{T,1:T-1} \big\{ \piEstDotStar{1:T-1} \big\}^{-1} && \big\{ \EstDotStar{} \big\}^{-1}
    \end{bmatrix}.
\end{equation*}
Recall in display \eqref{eqn:Qmatrices} we defined the following matrices:
\begin{equation}
    \label{app:Qdef}
    M_{1:T-1} \triangleq \big[ M_1, M_2, \dots, M_{T-1} \big] 
    \triangleq -\big\{ \EstDotStar{} \big\}^{-1} \bs{V}_{T,1:T-1} \big\{ \piEstDotStar{1:T-1} \big\}^{-1} \in \real^{d_\theta \by d_{1:T-1}}.
\end{equation}
Above we use $d_{1:T-1} \triangleq \sum_{t=1}^{T-1} d_t$. Thus,
\begin{equation*}
    \begin{bmatrix} 
        \piEstDotStar{1:T-1} && \bs{0} \\
        \bs{V}_{T,1:T-1} && \EstDotStar{}
    \end{bmatrix}^{-1}
    \Sigma_{1:T} 
    \begin{bmatrix} 
        \piEstDotStar{1:T-1} && \bs{0} \\
        \bs{V}_{T,1:T-1} && \EstDotStar{}
    \end{bmatrix}^{-1, \top}
\end{equation*}
\begin{equation}
    \label{appeqn:stackedVariance}
    = \begin{bmatrix} 
        \big\{ \piEstDotStar{1:T-1} \big\}^{-1} && \bs{0} \\
        M_{1:T-1} && \big\{ \EstDotStar{} \big\}^{-1}
    \end{bmatrix}
    \begin{bmatrix}
        \Sigma_{1:T-1} & U_{1:T-1,T} \\
        U_{T,1:T-1} & \Sigma
    \end{bmatrix}
    \begin{bmatrix} 
        \big\{ \piEstDotStar{1:T-1} \big\}^{-1} && \bs{0} \\
        M_{1:T-1} && \big\{ \EstDotStar{} \big\}^{-1}
    \end{bmatrix}^{\top}.
\end{equation}
Above $\Sigma \triangleq \Estar{2:T-1} \left[ \est{}(\history{i}{T}; \thetastar{})^{\otimes 2} \right]$, $\Sigma_{1:T-1} \triangleq \Estar{2:T-1} \left[ \piest{1:T-1}(\history{i}{T-1}; \betastar{1:T-1})^{\otimes 2} \right]$, \\
$U_{T,1:T-1} \triangleq \Estar{2:T} \left[ \est{}(\history{i}{T}; \thetastar{}) \piest{1:T-1}(\history{i}{T-1}; \betastar{1:T-1})^{\top} \right]$, and $U_{1:T-1,T} \triangleq U_{T,1:T-1}^\top$, where
\begin{equation}
    \label{app:stackedPhiFuncUser}
    \piest{1:T-1}(\history{i}{T-1}; \betastar{1:T-1})
    \triangleq \begin{pmatrix}
        \piest1(\history{i}{1}; \betastar1) \\
        \piest2(\history{i}{2}; \betastar2) \\
        \vdots \\
        \piest{T-1}(\history{i}{T-1}; \betastar{T-1})
    \end{pmatrix}.
\end{equation}
Thus display \eqref{appeqn:stackedVariance} equals the following:
\begin{multline*}
    = \begin{bmatrix} 
        \big\{ \piEstDotStar{1:T-1} \big\}^{-1} \Sigma_{1:T-1} &&~ \big\{ \piEstDotStar{1:T-1} \big\}^{-1} U_{1:T-1,T} \\
        M_{1:T-1}  \Sigma_{1:T-1} + \big\{ \EstDotStar{} \big\}^{-1} U_{T,1:T-1} &&~ M_{1:T-1}  U_{1:T-1,T} + \big\{ \EstDotStar{} \big\}^{-1} \Sigma
    \end{bmatrix} \\
    \begin{bmatrix} 
        \big\{ \piEstDotStar{1:T-1} \big\}^{-1, \top} && M_{1:T-1}^\top \\
        \bs{0} && \big\{ \EstDotStar{} \big\}^{-1, \top}
    \end{bmatrix}
\end{multline*}
Thus, the lower-right $d_\theta \by d_\theta$ block of the product of matrices above equals the following:
\begin{multline*}
    M_{1:T-1}  \Sigma_{1:T-1} M_{1:T-1}^\top + \big\{ \EstDotStar{} \big\}^{-1} U_{T,1:T-1} M_{1:T-1}^\top \\
    + M_{1:T-1}  U_{1:T-1,T} \big\{ \EstDotStar{} \big\}^{-1, \top} + \big\{ \EstDotStar{} \big\}^{-1} \Sigma \big\{ \EstDotStar{} \big\}^{-1, \top}
\end{multline*}
\begin{multline*}
    = \big\{ \EstDotStar{} \big\}^{-1} 
    \bigg[ \EstDotStar{} M_{1:T-1}  \Sigma_{1:T-1} M_{1:T-1}^\top \big\{ \EstDotStar{} \}^\top + U_{T,1:T-1} M_{1:T-1}^\top \big\{ \EstDotStar{} \}^\top \\
    + \EstDotStar{} M_{1:T-1}  U_{1:T-1,T} + \Sigma \bigg] 
    \big\{ \EstDotStar{} \big\}^{-1, \top}
\end{multline*}
\begin{equation*}
    = \big\{ \EstDotStar{} \big\}^{-1} 
    \Estar{2:T} \bigg[ \bigg\{ \est{} \big( \history{i}{T}; \thetastar{} \big) + \EstDotStar{} M_{1:T-1} \piest{1:T-1} \big( \history{i}{T-1}; \betastar{1:T-1} \big) \bigg\}^{\otimes 2} \bigg]
     \big\{ \EstDotStar{} \big\}^{-1, \top}
\end{equation*}
By the definition of $M_{1:T-1}$ from display \eqref{app:Qdef} and the definition of $\piest{1:T-1}$ from display \eqref{app:stackedPhiFuncUser},
\begin{equation*}
    = \big\{ \EstDotStar{} \big\}^{-1} 
    \Estar{2:T} \bigg[ \bigg\{ \est{} \big( \history{i}{T}; \thetastar{} \big) + \EstDotStar{} \sum_{t=1}^{T-1} M_t \piest{t} \big( \history{i}{t}; \betastar{t} \big) \bigg\}^{\otimes 2} \bigg]
     \big\{ \EstDotStar{} \big\}^{-1, \top}.
\end{equation*}
We have now shown the desired result. ~ $\blacksquare$

%%%%%%%%%%%%%%%%%%%%%%%%%%%%%%%%%%%%%%%%%%
%%%%%%%%%%%%%%%%%%%%%%%%%%%%%%%%%%%%%%%%%%
\subsection{Asymptotic Normality of $\thetahat{}$ (Theorem \ref{thm:normality})} %%%%%%%%%%%%%%%%%%%%%%%%%%%%%%%%%%%
\label{mainapp:normalityProof}

~ \\

\startproof{Theorem \ref{thm:normality}}
By Lemma \ref{lemma:normalityAdaptiveUnpack} (Equivalent Formulations for the Adaptive Sandwich Variance) above, it is sufficient to show that
\begin{equation*}
    \sqrt{n} \begin{pmatrix}
        \betahat{1:T-1} - \betastar{1:T-1} \\
        \thetahat{} - \thetastar{}
    \end{pmatrix} 
    \Dto \N \left( 0, \begin{bmatrix} 
        \piEstDotStar{1:T-1} && \bs{0} \\
        \bs{V}_{T,1:T-1} && \EstDotStar{}
    \end{bmatrix}^{-1} \Sigma_{1:T} \begin{bmatrix} 
        \piEstDotStar{1:T-1} && \bs{0} \\
        \bs{V}_{T,1:T-1} && \EstDotStar{}
    \end{bmatrix}^{-1, \top} \right).
\end{equation*}

%First note that by Conditions \ref{cond:exploration}-\ref{cond:lipschitzPolicyEstimatingFunction} and Remark \ref{remark:lipschitzSuffient} (which shows that Condition \ref{cond:lipschitzPolicyEstimatingFunction} implies that assumptions \ref{normalityBeta:bracketing} and \ref{normalityBeta:L2convergence} hold), we can apply Theorem \ref{thm:asymptoticEquicontinuityPolicy} (Asymptotic Equicontinuity for Policy Parameters) to get the following results:
%\begin{equation}
%    \label{appnorm:squarootNratePolicy}
%    \betahat{1:T-1} - \betastar{1:T-1} = O_P(1/\sqrt{n})
%\end{equation}
%and
%\begin{multline}
%    \label{appnorm:equicontinuity}
%    \sqrt{n} \left\{ \piEstHat{1:T-1}(\betahat{1:T-1}) - \piEst{1:T-1}(\betahat{1:T-1}) \right\} \\
%    = \sqrt{n} \left\{ \piEstHat{1:T-1}(\betastar{1:T-1}) - \piEst{1:T-1}(\betastar{1:T-1}) \right\} + o_P(1).
%\end{multline}

This proof will use the estimating functions $\Est \big( \beta_{1:T-1}, \theta \big)$, $\EstHat \big( \beta_{1:T-1}, \theta \big)$, \\
$\piEst{1:T-1} ( \beta_{1:T-1} )$, and $\piEstHat{1:T-1} ( \beta_{1:T-1} )$ defined earlier in displays \eqref{eqn:estDefWeights}, \eqref{proofSktech:EstHat}, \eqref{eqn:diffPolicy}, and \eqref{proofSketch:piEstHat} respectively.

\bigskip
\noindent We now state several equalities and discuss why they hold below:
\begin{multline}
    \label{app:proofMain1}
    - \sqrt{n} \begin{bmatrix}
        \piEstHat{1:T-1} \big( \betahat{1:T-1} \big) - \piEst{1:T-1} \big(\betahat{1:T-1} \big) \\
        \undermat{= o_P(1/\sqrt{n})}{ \EstHat \big( \betahat{1:T-1}, \thetahat{} \big) } - \Est \big(\betahat{1:T-1}, \thetahat{} \big)
    \end{bmatrix} \\
    ~~ \\
    \underbrace{=}_{(a)} - \sqrt{n} \begin{bmatrix}
        \piEst{1:T-1} \big( \betastar{1:T-1} \big) - \piEst{1:T-1} \big(\betahat{1:T-1} \big) \\
        \undermat{= 0}{ \Est \big( \betastar{1:T-1}, \thetastar{} \big) } - \Est \big(\betahat{1:T-1}, \thetahat{} \big)
    \end{bmatrix} + o_P(1) \\
    ~~ \\
    \underbrace{=}_{(b)} \sqrt{n} \begin{bmatrix}
    \piEstDotStar{1:T-1} && \bs{0} \\
        \bs{V}_{T,1:T-1} && \EstDotStar{}
    \end{bmatrix} \begin{bmatrix}
        \betahat{1:T-1} - \betastar{1:T-1} \\
        \thetahat{} - \thetastar{}
    \end{bmatrix} 
    + \sqrt{n} o_P \left( \left\| \begin{matrix}
        \betahat{1:T-1} - \betastar{1:T-1} \\
        \thetahat{} - \thetastar{}
    \end{matrix} \right\|_2 \right)
    + o_P(1).
\end{multline}

\proofSubsection{Equality (a)} Equality (a) above holds since $\EstHat \big( \betahat{1:T-1}, \thetahat{} \big) = o_P(1/\sqrt{n})$  and $\Est \big( \betastar{1:T-1}, \thetastar{} \big) = 0$ by the definitions of $\thetahat{}$ and $\thetastar{}$ from displays \eqref{eqn:ZestimatorDef} and \eqref{eqn:ZestimandDef} respectively; 
also since $\piEstHat{1:T-1} \big( \betahat{1:T-1} \big) = o_P(1/\sqrt{n})$ and $\piEst{1:T-1} \big( \betastar{1:T-1} \big) = 0$ by the definitions of $\betahat{t}$ and $\betastar{t}$ from displays \eqref{eqn:betahatDef} and \eqref{eqn:betastarDef}.

\proofSubsection{Equality (b)} Equality (b) above holds by a Taylor series expansion. Specifically by assumptions \ref{normality:bread} and \ref{normality:differentiable}, the mapping $\big( \beta_{1:T-1}, \theta \big) \mapsto \Est(\beta_{1:T-1}, \theta )$ is differentiable at $\big( \beta_{1:T-1}, \theta \big) = \big( \betastar{1:T-1}, \thetastar{} \big)$. Additionally  the mapping $\beta_{1:T-1} \mapsto \piEst{1:T-1}(\beta_{1:T-1})$ is differentiable at $\beta_{1:T-1} = \betastar{1:T-1}$ by Condition \ref{cond:differentiabilityPolicy}. %As mentioned below display \eqref{eqnMain:stackedBread},  $\fracpartial{}{\theta} \piEst{1:T-1}(\betastar{1:T-1}) \big|_{ \theta = \thetastar{} } = \bs{0}$ since $\piEst{1:T-1}(\betastar{1:T-1})$ is not a function of $\theta$. \\
%By assumptions \ref{normality:bread} and \ref{normality:differentiable}, the mapping $\big( \beta_{1:T-1}, \theta \big) \mapsto \Est(\beta_{1:T-1}, \theta )$ is differentiable at $\big( \beta_{1:T-1}, \theta \big) = \big( \betastar{1:T-1}, \thetastar{} \big)$. Moreover the mapping $\beta_{1:T-1} \mapsto \piEst{1:T-1}(\beta_{1:T-1})$ is differentiable at $\beta_{1:T-1} = \betastar{1:T-1}$ by Condition \ref{cond:differentiabilityPolicy}.
Thus,
\begin{multline*}
    \fracpartial{}{(\beta_{1:T-1}, \theta)} \begin{bmatrix}
        \piEst{1:T-1}(\beta_{1:T-1}) \\
        \Est{}(\beta_{1:T-1}, \theta)
    \end{bmatrix} \bigg|_{( \beta_{1:T-1}, \theta ) = ( \betastar{1:T-1}, \thetastar{} )} \\
    = \begin{bmatrix}
        \fracpartial{}{\beta_{1:T-1}} \piEst{1:T-1}(\beta_{1:T-1}) \big|_{ \beta_{1:T-1} = \betastar{1:T-1} } && \fracpartial{}{\theta} \piEst{1:T-1}(\betastar{1:T-1}) \big|_{ \theta = \thetastar{} } \\
        \fracpartial{}{\beta_{1:T-1}} \Est{}(\beta_{1:T-1}, \thetastar{}) \big|_{ \beta_{1:T-1} = \betastar{1:T-1} } && \fracpartial{}{\theta} \Est{}(\betastar{1:T-1}, \theta) \big|_{\theta = \thetastar{}}
    \end{bmatrix}
    = \begin{bmatrix}
    \piEstDotStar{1:T-1} && \bs{0} \\
        \bs{V}_{T,1:T-1} && \EstDotStar{}
    \end{bmatrix}.
\end{multline*}
As mentioned below display \eqref{eqnMain:stackedBread},  $\fracpartial{}{\theta} \piEst{1:T-1}(\betastar{1:T-1}) \big|_{ \theta = \thetastar{} } = \bs{0}$ since $\piEst{1:T-1}(\betastar{1:T-1})$ is not a function of $\theta$.

\bigskip
\noindent We now state the next set of results and discuss why they hold below:
\begin{multline}
    \label{app:proofMain2}
    -\sqrt{n} \begin{bmatrix}
        \piEstHat{1:T-1} \big( \betahat{1:T-1} \big) - \piEst{1:T-1} \big(\betahat{1:T-1} \big) \\
        \EstHat \big( \betahat{1:T-1}, \thetahat{} \big) - \Est \big(\betahat{1:T-1}, \thetahat{} \big)
    \end{bmatrix} \\
    \underbrace{=}_{(c)} -\sqrt{n} \begin{bmatrix}
        \piEstHat{1:T-1} \big( \betastar{1:T-1} \big) - \piEst{1:T-1} \big(\betastar{1:T-1} \big) \\
        \EstHat \big( \betastar{1:T-1}, \thetastar{} \big) - \Est \big(\betastar{1:T-1}, \thetastar{} \big)
    \end{bmatrix} + o_P(1)
    \underbrace{\Dto}_{(d)} \N \big( 0, \Sigma_{1:T} \big).
\end{multline}

\proofSubsection{Equality (c)} Equality (c) above is an asymptotic equicontinuity result. We now discuss why equality (c) holds. First note that by Conditions \ref{cond:consistencyPolicy}-\ref{cond:lipschitzPolicyEstimatingFunction} and Remark \ref{remark:lipschitzSuffient} (which shows that Condition \ref{cond:lipschitzPolicyEstimatingFunction} implies that assumptions \ref{normalityBeta:bracketing} and \ref{normalityBeta:L2convergence} hold), we can apply Theorem \ref{thm:asymptoticEquicontinuityPolicy} (Asymptotic Equicontinuity for Policy Parameters) to get the following results:
\begin{equation}
    \label{appnorm:squarootNratePolicy}
    \betahat{1:T-1} - \betastar{1:T-1} = O_P(1/\sqrt{n})
\end{equation}
and
\begin{multline}
    \label{appnorm:equicontinuity}
    \sqrt{n} \left\{ \piEstHat{1:T-1}(\betahat{1:T-1}) - \piEst{1:T-1}(\betahat{1:T-1}) \right\} \\
    = \sqrt{n} \left\{ \piEstHat{1:T-1}(\betastar{1:T-1}) - \piEst{1:T-1}(\betastar{1:T-1}) \right\} + o_P(1).
\end{multline}
We now apply Lemma \ref{lemma:stochasticEquicontinuity} (Stochastic Equicontinuity) to get that
\begin{multline}
    \label{normTheta:equicontinuity}
    \sqrt{n} \left\{ \EstHat \big( \betahat{1:T-1}, \thetahat{} \big) - \Est \big(\betahat{1:T-1}, \thetahat{} \big) \right\} \\
    = \sqrt{n} \left\{ \EstHat \big(\betastar{1:T-1}, \thetastar{} \big) - \Est \big(\betastar{1:T-1}, \thetastar{} \big) \right\} + o_P(1),
\end{multline}
We are able to apply Lemma \ref{lemma:stochasticEquicontinuity} (Stochastic Equicontinuity) because the following assumptions hold:
\begin{itemize}
    \item Conditions \ref{cond:exploration} (Minimum Exploration) and \ref{cond:lipschitzPolicy} (Lipschitz Policy Functions) hold.
    \smallskip
    %%%%%%%%%%%%%%%
    \item Note that $\betahat{1:T-1} - \betastar{1:T-1} = O_P(1/\sqrt{n})$ by display \eqref{appnorm:squarootNratePolicy}.
    \smallskip
    %%%%%%%%%%%%%%%
    \item Since assumption \ref{normality:bracketing} (Finite Bracketing Integral) and Condition \ref{cond:lipschitzPolicy} (Lipschitz Policy Function) hold, we can apply Lemma \ref{lemma:bracketingProduct} (specifically see Remark \ref{remark:bracketingProduct} part \ref{productBracket:integralPsi}) to get that for any $c \in \real^{d_\theta}$,
    \begin{equation*}
        \int_0^1 \sqrt{ \log N_{[~]} \left( \epsilon, ~ \big\{ \pi_{2:T}(\cdotspace; \beta_{1:T-1} ) c^\top \est{}(\cdotspace; \theta) \big\}_{\beta_{1:T-1} \in B_{1:T-1}, \theta \in \Theta}, ~ L_{2+\alpha}(\Pstar) \right) } d \epsilon < \infty,
    \end{equation*}
    where $\pi_{2:T}(\cdotspace; \beta_{1:T-1} ) \triangleq \prod_{t'=2}^{T} \pi_{t'}(\cdotspace; \beta_{t'-1} )$.
    \smallskip
    %%%%%%%%%%%%%%%
    \item $\big( \betahat{1:T-1}, \thetahat{} \big) \Pto \big( \betastar{1:T-1}, \thetastar{} \big)$ by Slutsky's theorem since $\betahat{1:T-1} \Pto \betastar{1:T-1}$ by Condition \ref{cond:consistencyPolicy} and since $\thetahat{} \Pto \thetastar{}$ by assumption. Thus, by assumption \ref{normality:L2convergence} and continuous mapping theorem we have that for any $c \in \real^{d_\theta}$, \\
    $\nu \left( c^\top \pi_{2:T}(\cdotspace; \betahat{1:T-1} ) \est{}(\cdotspace; \thetahat{}),
    ~ c^\top \pi_{2:T}(\cdotspace; \betastar{1:T-1} ) \est{}(\cdotspace; \thetastar{}) \right) \Pto 0$, where
    \begin{multline*}
        \nu \bigg( \pi_{2:T}(\cdotspace; \beta_{1:T-1} ) c^\top \est{}(\cdotspace; \theta), 
        ~ \pi_{2:T}(\cdotspace; \beta_{1:T-1}' ) c^\top \est{}(\cdotspace; \theta') \bigg) \\
        \triangleq \Estar{2:T} \left[ \big\{ \pi_{2:T}(\cdotspace; \beta_{1:T-1} ) c^\top \est{}(\cdotspace; \theta)
        - \pi_{2:T}(\cdotspace; \beta_{1:T-1}' ) c^\top \est{}(\cdotspace; \theta') \big\}^2 \right]^{1/2}.
    \end{multline*}
\end{itemize}
Equality (c) holds since by Slutsky's Theorem, we can combine the results above from displays \eqref{appnorm:equicontinuity} and \eqref{normTheta:equicontinuity}.

\proofSubsection{Asymptotic normality result (d)} 
Equality (d) above holds by Theorem \ref{thm:weightedCLT} (Weighted Martingale Triangular Array Central Limit Theorem). Specifically note that for any fixed vector $c = [c_1, c_2, \dots, c_{T} ] \in \real^{d_{1:T-1}+d_\theta}$, 
\begin{equation*}
    -\sqrt{n} c^\top \begin{bmatrix}
        \piEstHat{1:T-1} \big( \betastar{1:T-1} \big) - \piEst{1:T-1} \big(\betastar{1:T-1} \big) \\
        \EstHat \big( \betastar{1:T-1}, \thetastar{} \big) - \Est \big(\betastar{1:T-1}, \thetastar{} \big)
    \end{bmatrix}
\end{equation*}
\begin{multline}
    \label{eqnnorm:phiAlgOp1Full}
    = -\frac{1}{\sqrt{n}} \sum_{i=1}^n \sum_{t=1}^{T-1} c_{t}^\top \WW{i}{2:t}(\betastar{1:t-1}, \betahat{1:t-1}) \piest{t}(\history{i}{t}; \betastar{t}) \\
    - \frac{1}{\sqrt{n}} \sum_{i=1}^n c_{T}^\top \WW{i}{2:T}(\betastar{1:T-1}, \betahat{1:T-1}) \est{}(\history{i}{T}; \thetastar{}) 
    \Dto \N \big( 0, c^\top \Sigma_{1:T} c \big)
\end{multline}
where 
\begin{equation*}
    \Sigma_{1:T} \triangleq 
    \Estar{2:T} \left[ \begin{pmatrix}
        \piest1(\history{i}{1}; \betastar1) \\
        \piest2(\history{i}{2}; \betastar2) \\
        \vdots \\
        \piest{T-1}(\history{i}{T-1}; \betastar{T-1}) \\
        \est{}(\history{i}{T}; \thetastar{}) 
    \end{pmatrix}^{\otimes2} \right].
\end{equation*}
The final asymptotic normality result above holds by Theorem \ref{thm:weightedCLT} (Weighted Martingale Triangular Array Central Limit Theorem). Specifically we can apply Theorem \ref{thm:weightedCLT} because:
\begin{itemize}
    \item Conditions \ref{cond:exploration} (Minimum Exploration) and \ref{cond:lipschitzPolicy} (Lipschitz Policy Functions) hold.
    \item $\betahat{t'} - \betastar{t'} = O_P( 1/\sqrt{n} )$ for all $t' \in [1 \colon T-1]$ by display \eqref{appnorm:squarootNratePolicy}.
    \item $\Estar{2:T} \left[ \big| c_T^\top \est{}(\history{i}{T}; \thetastar{}) \big|^{2+\alpha} \right] < \infty$ by Finite Bracketing Integral assumption \ref{normality:bracketing}. Also, $\Estar{2:t} \left[ \big| c_t^\top \piest{t}(\history{i}{t}; \betastar{t}) \big|^{2+\alpha} \right] < \infty$ for each $t \in [1 \colon T-1]$ by Condition \ref{cond:lipschitzPolicyEstimatingFunction}.
\end{itemize}

\smallskip
Thus, by Cramer-Wold device and display \eqref{eqnnorm:phiAlgOp1Full} we have that
\begin{equation}
    \label{eqnnorm:stackedNormality}
    -\sqrt{n} \begin{bmatrix}
        \piEstHat{1:T-1} \big( \betastar{1:T-1} \big) - \piEst{1:T-1} \big(\betastar{1:T-1} \big) \\
        \EstHat \big( \betastar{1:T-1}, \thetastar{} \big) - \Est \big(\betastar{1:T-1}, \thetastar{} \big)
    \end{bmatrix}
    \Dto \N( 0, \Sigma_{1:T} ).
\end{equation}
Thus, we have shown that asymptotic normality result (d) holds.

\proofSubsection{Consolidating Results} By consolidating the results from displays \eqref{app:proofMain1} and \eqref{app:proofMain2} above, and applying Slutsky's theorem we get the following result:
\begin{equation}
    \label{app:proofMainConsolidate}
    \sqrt{n} \begin{bmatrix}
    \piEstDotStar{1:T-1} && \bs{0} \\
        \bs{V}_{T,1:T-1} && \EstDotStar{}
    \end{bmatrix} \begin{bmatrix}
        \betahat{1:T-1} - \betastar{1:T-1} \\
        \thetahat{} - \thetastar{}
    \end{bmatrix} 
    + \sqrt{n} o_P \left( \left\| \begin{matrix}
        \betahat{1:T-1} - \betastar{1:T-1} \\
        \thetahat{} - \thetastar{}
    \end{matrix} \right\|_2 \right)
    \Dto \N \big( 0, \Sigma_{1:T} \big).
\end{equation}
Note that $\EstDotStar{}$ is invertible by assumption \ref{normality:bread} and $\piEstDotStar{1:T-1}$ is invertible by Condition \ref{cond:differentiabilityPolicy} and Lemma \ref{lemma:invertibilityPiDotStar}. By Proposition \ref{prop:blockInversion} (Block Inversion of Matrices), this is sufficient for $\begin{bmatrix}
    \piEstDotStar{1:T-1} && \bs{0} \\
        \bs{V}_{T,1:T-1} && \EstDotStar{}
    \end{bmatrix}$ to be invertible.
Thus, by continuous mapping theorem, and display \eqref{app:proofMainConsolidate},
\begin{multline}
    \label{mainApp:stackedNormalityResult}
    \sqrt{n} \begin{bmatrix}
        \betahat{1:T-1} - \betastar{1:T-1} \\
        \thetahat{} - \thetastar{}
    \end{bmatrix} 
    + \sqrt{n} O(1) o_P \left( \left\| \begin{matrix}
        \betahat{1:T-1} - \betastar{1:T-1} \\
        \thetahat{} - \thetastar{}
    \end{matrix} \right\|_2 \right)
    + o_P(1) \\
    \Dto \N \left( 0, \begin{bmatrix}
    \piEstDotStar{1:T-1} && \bs{0} \\
        \bs{V}_{T,1:T-1} && \EstDotStar{}
    \end{bmatrix}^{-1} \Sigma_{1:T} \begin{bmatrix}
    \piEstDotStar{1:T-1} && \bs{0} \\
        \bs{V}_{T,1:T-1} && \EstDotStar{}
    \end{bmatrix}^{-1, \top} \right).
\end{multline}
The asymptotic normality result above in display \eqref{mainApp:stackedNormalityResult} implies that 
\begin{equation*}
    \sqrt{n} \begin{bmatrix}
        \betahat{1:T-1} - \betastar{1:T-1} \\
        \thetahat{} - \thetastar{}
    \end{bmatrix} + \sqrt{n} O(1) o_P \left( \left\| \begin{matrix}
        \betahat{1:T-1} - \betastar{1:T-1} \\
        \thetahat{} - \thetastar{}
    \end{matrix} \right\|_2 \right) = O_P(1).
\end{equation*}
This implies that $\sqrt{n} \begin{bmatrix}
        \betahat{1:T-1} - \betastar{1:T-1} \\
        \thetahat{} - \thetastar{}
    \end{bmatrix} = O_P(1)$.
Thus, we have that
\begin{equation*}
    \sqrt{n} O(1) o_P \left( \left\| \begin{matrix}
        \betahat{1:T-1} - \betastar{1:T-1} \\
        \thetahat{} - \thetastar{}
    \end{matrix} \right\|_2 \right) = o_P(1).
\end{equation*}
So, by display \eqref{mainApp:stackedNormalityResult} and Slutsky's Theorem we have that
\begin{equation*}
    \sqrt{n} \begin{bmatrix}
        \betahat{1:T-1} - \betastar{1:T-1} \\
        \thetahat{} - \thetastar{}
    \end{bmatrix} 
    \Dto \N \left( 0, \begin{bmatrix}
    \piEstDotStar{1:T-1} && \bs{0} \\
        \bs{V}_{T,1:T-1} && \EstDotStar{}
    \end{bmatrix}^{-1} \Sigma_{1:T} \begin{bmatrix}
    \piEstDotStar{1:T-1} && \bs{0} \\
        \bs{V}_{T,1:T-1} && \EstDotStar{}
    \end{bmatrix}^{-1, \top} \right). ~\blacksquare
\end{equation*}

\clearpage
\section{Limit Theorems for Adaptively Sampled Data}
\label{app:supporting}

%%%%%%%%%%%%%%%%%%%%%%%%%%%%%%%%%%%%%%%%%%%%%%%%%%%%%%%%%%%%
%%%%%%%%%%%%%%%%%%%%%%%%%%%%%%%%%%%%%%%%%%%%%%%%%%%%%%%%%%%%

\startOverview{Overview of Appendix \ref{app:supporting} Results}

\begin{itemize}
    \item \bo{Section \ref{appsupport:weightedWLLN}:} Weighted Martingale Triangular Array Law of Large Numbers (Theorem \ref{thm:weightedWLLN})
    %%%%%%%%%%%%%%%%%%%
    \item \bo{Section \ref{appsupport:weightedUWLLN}:} Weighted Martingale Triangular Array Uniform Law of Large Numbers (Theorem \ref{thm:weightedUWLLN})
    %%%%%%%%%%%%%%%%%%%
    \item \bo{Section \ref{appsupport:OpOne}:} Showing Terms are $O_P(1)$ (Lemma \ref{lemma:OpOne})
    %%%%%%%%%%%%%%%%%%%
    \item \bo{Section \ref{appsupport:weightedCLT}:} Weighted Martingale Triangular Array Central Limit Theorem (Theorem \ref{thm:weightedCLT})
    %%%%%%%%%%%%%%%%%%%
    \item \bo{Section \ref{mainapp:functionalNormality}:} Functional Asymptotic Normality under Finite Bracketing Integral (Theorem \ref{thm:functionalAsymptoticNormality})
    %%%%%%%%%%%%%%%%%%%
    \item \bo{Section \ref{mainapp:stochasticEquicontinuity}:} Stochastic Equicontinuity (Lemma \ref{lemma:stochasticEquicontinuity})
\end{itemize}

%%%%%%%%%%%%%%%%%%%%%%%%%%%%%%%%%%%%%%%%%%%%
%%%%%%%%%%%%%%%%%%%%%%%%%%%%%%%%%%%%%%%%%%%%
%\subsection{Appendix \ref{app:supporting} Notation}
%\label{app:supportingNotation}
%For notational convenience we will use
%\begin{equation*}
%    \WW{i}{1:t}(\betastar{}, \betahat{}) \triangleq \WW{i}{2:t}(\betastar{}, \betahat{}).
%\end{equation*}
%Above we can think of $\WW{i}{1} = 1$.

%%%%%%%%%%%%%%%%%%%%%%%%%%%%%%%%%%%%%%%%%%%%
%%%%%%%%%%%%%%%%%%%%%%%%%%%%%%%%%%%%%%%%%%%%
\subsection{Weighted Martingale Triangular Array Law of Large Numbers (Theorem \ref{thm:weightedWLLN})}
\label{appsupport:weightedWLLN}

%%%%%%%%%%%%%%%%%%%%%%%%%%%%%%%%%%%%%%%%%%%%
\begin{theorem}[Weighted Martingale Triangular Array Weak Law of Large Numbers]
	\label{thm:weightedWLLN}
    Let $f$ be a real-valued function of $\history{i}{t}$ such that for some $\alpha > 0$, $\Estar{2:t} \left[ \big| f \big( \history{i}{t} \big) \big|^{1+\alpha} \right] < \infty$. Under Condition \ref{cond:exploration} (Minimum Exploration), 
	\begin{equation}
	    \label{WLLN:result}
		\frac{1}{n} \sum_{i=1}^n \WW{i}{2:t}(\betastar{}, \betahat{}) f(\history{i}{t}) 
		\Pto \Estar{2:t} \left[ f(\history{i}{t}) \right].
	\end{equation}
    Moreover, %for any fixed $\betatilde{1:t-1} \in \real^{d_{1:t-1}}$ where $d_{1:t-1} \triangleq \sum_{t'=1}^{t-1} d_{t'}$,
    \begin{equation}
	    \label{WLLN:eqnTilde}
		\frac{1}{n} \sum_{i=1}^n \rhohat{i}{2:t} f(\history{i}{t}) 
		\Pto \Estar{2:t} \left[ \rhostar{i}{2:t} f(\history{i}{t}) \right].
	\end{equation}
    %\WW{i}{2:t}(\betatilde{}, \betahat{}) \Pto \E_{\pi_{2:t}(\betatilde{1:t-1})} \left[ f(\history{i}{t}) \right].
\end{theorem}

\startproof{Theorem \ref{thm:weightedWLLN}}
%%%%%%%%%%%%%%%%%%%%%%%%%%%%%%%%%%%

\proofSubsection{Proving display \eqref{WLLN:result} holds} 
To show display \eqref{WLLN:result} holds it is sufficient to show that for any $t \in [1 \colon T]$,
\begin{equation}
    \label{app:WLLNmainResult}
	\frac{1}{n} \sum_{i=1}^n \left\{ \WW{i}{1:t}(\betastar{}, \betahat{}) f(\history{i}{t}) - \Estar{2:t} \left[ f(\history{i}{t}) \right] \right\} \Pto 0.
\end{equation}
Above we use $\WW{i}{1} = 1$ and $\WW{i}{1:t}(\betastar{}, \betahat{}) \triangleq \WW{i}{2:t}(\betastar{}, \betahat{})$.

For the $t = 1$ case, $\history{1}{1}, \history{2}{1}, \history{3}{1}, \dots, \history{n}{1}$ are i.i.d.; in this case, display \eqref{app:WLLNmainResult} holds by the Weak Law of Large numbers for i.i.d. data.

Note that for $t \geq 2$, the distribution of $\history{i}{t}$ is changing with the number of users $n$, since it is depends on policy parameter $\betahat{t-1}$. Thus, we need to consider triangular array asymptotics.

The first task is to rewrite the left-hand side of display \eqref{app:WLLNmainResult} as a sum of triangular array martingale differences. 
Note that we can rewrite the left-hand side of display \eqref{app:WLLNmainResult} as follows:
\begin{equation*}
	\frac{1}{n} \sum_{i=1}^n \left\{ \WW{i}{1:t}( \betastar{}, \betahat{} ) f(\history{i}{t}) 
	- \E \left[ \WW{i}{1:t}( \betastar{}, \betahat{} ) f(\history{i}{t}) \right] \right\}
\end{equation*}
Above the expectation $\E$ is with respect to the data distribution used to collect the data, thus, $\E \left[ \WW{i}{1:t}( \betastar{}, \betahat{} ) f(\history{i}{t}) \right] = \Estar{2:t} \left[ f(\history{i}{t}) \right]$.

Let $\Xvar{i}{t} \triangleq \WW{i}{1:t}( \betastar{}, \betahat{} ) f(\history{i}{t})$. Also let $\history{1:n}{0} \triangleq \emptyset$ and $\state{1:n}{T+1} \triangleq \emptyset$ (the second definition is only used for the $t = T$ case). By telescoping series, 
\begin{equation*}
	= \frac{1}{n} \sum_{i=1}^n \left\{ \Xvar{i}{t} - \E \left[ \Xvar{i}{t} \right] \right\}
\end{equation*}
\begin{multline}
    \label{app:WLLNintermedDifferencesZ}
	= \frac{1}{n} \sum_{i=1}^n \underbrace{ \left\{ \E \left[ \Xvar{i}{t} \big| \history{1:n}{0}, \state{1:n}{1}  \right] - \E \left[ \Xvar{i}{t} \right] \right\} }_{ \triangleq \Zvar{i}{0} } \\
	+ \sum_{t'=1}^{t} \bigg[ \frac{1}{n} \sum_{i=1}^n \underbrace{ \left\{ \E \left[ \Xvar{i}{t} \big| \history{1:n}{t'}, \state{1:n}{t'+1} \right] 
	- \E \left[ \Xvar{i}{t} \big| \history{1:n}{t'-1}, \state{1:n}{t'} \right] \right\} }_{ \triangleq \Zvar{i}{t'} } \bigg].
\end{multline}
Note above that $\Xvar{i}{t} = \E \left[ \Xvar{i}{t} \big| \history{1:n}{t}, \state{1:n}{t+1} \right]$, since $\Xvar{i}{t}$ is a constant given $\history{1:n}{t}$. 

Using the $\Zvar{i}{t'}$ notation defined in display \eqref{app:WLLNintermedDifferencesZ} above,
\begin{equation}
    \label{app:ZsumWLLN}
    = \underbrace{ \frac{1}{n} \sum_{i=1}^n \Zvar{i}{0} }_{= o_P(1)}
    + \frac{1}{n} \sum_{i=1}^n \sum_{t'=1}^t \Zvar{i}{t'}
\end{equation}
Note above that $\frac{1}{n} \sum_{i=1}^n \Zvar{i}{0} 
= \frac{1}{n} \sum_{i=1}^n \left\{ \E \big[ \Xvar{i}{t} \big| \history{1:n}{0}, \state{1:n}{1}  \big] - \E \big[ \Xvar{i}{t} \big] \right\} \\
= \frac{1}{n} \sum_{i=1}^n \left\{ \E \big[ f(\history{i}{1}) \big| \state{1:n}{1}  \big] - \E \big[ f(\history{i}{1}) \big] \right\} \\
= \frac{1}{n} \sum_{i=1}^n \left\{ \E \big[ f(\history{i}{1}) \big| \state{i}{1}  \big] - \E \big[ f(\history{i}{1}) \big] \right\} \Pto 0$ by the weak law of large numbers for i.i.d. random variables.

Regarding the second summation in display \eqref{app:ZsumWLLN}, note that $\big\{ \Zvar{i}{t'} \big\}_{i=1;t'=1}^{i=n;t'=t}$ is a martingale difference triangular array with respect to the filtration $\big\{ \sigma \big( \history{1:n}{t'-1}, \state{1:n}{t'} \big) \big\}_{t'=1}^t$. This is the case because for any $i \in [1 \colon n]$ and $t' \in [0 \colon t]$,
\begin{equation}
    \label{app:WLLNZzeroFirst}
    \E \left[ \Zvar{i}{t'} \big| \history{1:n}{t'-1}, \state{1:n}{t'} \right]
    = \E \left[ \E \left[ \Xvar{i}{t} \big| \history{1:n}{t'}, \state{1:n}{t'+1} \right] 
	- \E \left[ \Xvar{i}{t} \big| \history{1:n}{t'-1}, \state{1:n}{t'} \right] \bigg| \history{1:n}{t'-1}, \state{1:n}{t'} \right]
\end{equation}
\begin{equation*}
    = \E \left[ \E \left[ \Xvar{i}{t} \big| \history{1:n}{t'}, \state{1:n}{t'+1} \right] 
	   \bigg| \history{1:n}{t'-1}, \state{1:n}{t'} \right]
    - \E \left[ \Xvar{i}{t} \big| \history{1:n}{t'-1}, \state{1:n}{t'} \right] = 0.
\end{equation*}
The final equality above holds by the law of iterated expectations.

The next step will be to apply Theorem 2(a) of \cite{andrews1988laws} (Weak Law of Large Numbers for Triangular Array Mixingales). Theorem 2(a) ensures that $\frac{1}{n} \sum_{i=1}^n \sum_{t'=1}^t \Zvar{i}{t'} \Pto 0$ if we can show the following hold:
\begin{enumerate}[label=(\roman*)]
    \item $\E \big[ \Zvar{i}{t'} \big| \history{1:n}{t'-1}, \state{1:n}{t'} \big] = 0$ for all $i \in [1 \colon n]$ and $t' \in [1 \colon t]$. (Note that since we have a martingale difference triangular array, the mixing constant $c_{n,i}$ in Theorem 2(a) is satisfied for $c_{n,i} = 0$.) 
    \label{app:wllnCondMartingale}
    %%%%%%%%%%%%%%%%%
    \item For some $\alpha > 0$, $\E \big[ | \Zvar{i}{t'} |^{1+\alpha} \big] < \infty$ for all $i \in [1 \colon n]$ and $t' \in [1 \colon t]$. Note that by Exercise 5.5.1 of \cite{durrett2019probability} this implies $\Zvar{i}{t'}$ are uniformly integrable.
    \label{app:wllnCondUI}
\end{enumerate}
We already showed that property \ref{app:wllnCondMartingale} above holds earlier in display \eqref{app:WLLNZzeroFirst}. All that remains is to show that property \ref{app:wllnCondUI} above holds.

Consider any $t' \in [1 \colon t]$ and any $i \in [1 \colon n]$,
\begin{equation*}
    \E \big[ \big| \Zvar{i}{t'} \big|^{1+\alpha} \big]
    = \E \bigg[ \left| \E \big[ \Xvar{i}{t} \big| \history{1:n}{t'}, \state{1:n}{t'+1} \big] 
	- \E \big[ \Xvar{i}{t} \big| \history{1:n}{t'-1}, \state{1:n}{t'} \big] \right|^{1+\alpha} \bigg]
\end{equation*}
By Lemma \ref{lemma:binomialBound}, for some positive constant $c_{1+\alpha} < \infty$,
\begin{equation*}
    \leq c_{1+\alpha} \bigg\{ \E \bigg[ \left| \E \big[ \Xvar{i}{t} \big| \history{1:n}{t'}, \state{1:n}{t'+1} \big] \right|^{1+\alpha} \bigg]
	+ \E \bigg[ \left| \E \big[ \Xvar{i}{t} \big| \history{1:n}{t'-1}, \state{1:n}{t'} \big] \right|^{1+\alpha} \bigg] \bigg\}
\end{equation*}
By Jensen's inequality
\begin{equation*}
    \leq c_{1+\alpha} \bigg\{ \E \bigg[  \E \left[ \big| \Xvar{i}{t} \big|^{1+\alpha} \bigg| \history{1:n}{t'}, \state{1:n}{t'+1} \right] \bigg]
	+ \E \bigg[ \E \left[ \big| \Xvar{i}{t} \big|^{1+\alpha} \bigg| \history{1:n}{t'-1}, \state{1:n}{t'} \right] \bigg] \bigg\}
\end{equation*}
By the law of iterated expectations,
\begin{equation*}
    = c_{1+\alpha} \bigg\{ \E \left[ \big| \Xvar{i}{t} \big|^{1+\alpha} \right] 
	+ \E \left[ \big| \Xvar{i}{t} \big|^{1+\alpha} \right] \bigg\}
    = c_{1+\alpha} 2 \E \left[ \big| \Xvar{i}{t} \big|^{1+\alpha} \right]
\end{equation*}
Since $\Xvar{i}{t} \triangleq \WW{i}{1:t}( \betastar{}, \betahat{} ) f(\history{i}{t})$,
\begin{equation*}
    = c_{1+\alpha} 2 \E \left[ \big| \WW{i}{1:t}( \betastar{}, \betahat{} ) f(\history{i}{t}) \big|^{1+\alpha} \right]
\end{equation*}
Since $\WW{i}{1:t}( \betastar{}, \betahat{} ) \leq \pi_{\min}^{-(t-1)}$ by Condition \ref{cond:exploration} (Minimum Exploration),
\begin{equation*}
    \leq c_{1+\alpha} 2 \pi_{\min}^{-(t-1) (1+\alpha)} \E \left[ \big| f(\history{i}{t}) \big|^{1+\alpha} \right] < \infty.
\end{equation*}
The final inequality holds since $\E \left[ \big| f(\history{i}{t}) \big|^{1+\alpha} \right] < \infty$ by assumption of this Theorem. We have now shown that property \ref{app:wllnCondUI} holds. We have now shown that display \eqref{WLLN:result} holds.

%%%%%%%%%%%%%%%%%%%%%
\proofSubsection{Proving Display \eqref{WLLN:eqnTilde} holds}

\begin{equation}
	\frac{1}{n} \sum_{i=1}^n \rhohat{i}{2:t} f(\history{i}{t}) 
	\Pto \Estar{2:t} \left[ \rhostar{i}{2:t} f(\history{i}{t}) \right].
\end{equation}

Note that
\begin{equation*}
    \rhohat{i}{2:t} f(\history{i}{t}) 
    = \WW{i}{2:t}(\betastar{}, \betahat{}) \rhostar{i}{2:t} f(\history{i}{t}).
\end{equation*}

By Condition \ref{cond:exploration}, $\rhostar{i}{2:t} \leq \pi_{\min}^{-(t-1)}$ a.s. Thus, $\Estar{2:t} \left[ \big| \rhostar{i}{2:t} f(\history{i}{t}) \big|^{1+\alpha} \right] \\
\leq \pi_{\min}^{-(t-1)} \Estar{2:t} \left[ \big| f(\history{i}{t}) \big|^{1+\alpha} \right] < \infty$; the final inequality holds by the assumption that \\
$\Estar{2:t} \left[ \big| f \big( \history{i}{t} \big) \big|^{1+\alpha} \right] < \infty$. Thus, using the result from display \eqref{WLLN:result},
\begin{multline*}
    \frac{1}{n} \sum_{i=1}^n \rhohat{i}{2:t} f(\history{i}{t}) 
    = \frac{1}{n} \sum_{i=1}^n \WW{i}{2:t}(\betastar{}, \betahat{}) \rhostar{i}{2:t} f(\history{i}{t}) \\
    \Pto \Estar{2:t} \left[ \rhostar{i}{2:t} f(\history{i}{t}) \right].  ~~~ \blacksquare
\end{multline*}

\subsection{Weighted Martingale Triangular Array Uniform Law of Large Numbers (Theorem \ref{thm:weightedUWLLN})}
\label{appsupport:weightedUWLLN}

%%%%%%%%%%%%%%%%%%%%%%%%%%%%%%%%%%%%%%%%%%%%
\begin{theorem}[Weighted Martingale Triangular Array Uniform Weak Law of Large Numbers]
	\label{thm:weightedUWLLN}
    Let $\F$ be a class of real-valued, measurable functions such that for some $\alpha > 0$, $N_{[~]}\big(\epsilon, \F, L_{1+\alpha}(\Pstar) \big) < \infty$ for any $\epsilon > 0$. 
	Under Condition \ref{cond:exploration} (Minimum Exploration), 
	\begin{equation}
	    \label{UWLLN:eqn1}
		\sup_{f \in \F} \bigg| \frac{1}{n} \sum_{i=1}^n \left\{ \WW{i}{2:t}(\betastar{}, \betahat{}) f(\history{i}{t}) 
		- \Estar{2:t} \big[ f(\history{i}{t}) \big] \right\} \bigg| \Pto 0.
	\end{equation}
    Moreover, %for any $\betatilde{1:t-1} \in \real^{d_{1:t-1}}$,
    \begin{equation}
	    \label{UWLLN:eqnTilde}
        \sup_{f \in \F} \bigg| \frac{1}{n} \sum_{i=1}^n \left( \rhohat{i}{2:t} f(\history{i}{t}) - \Estar{2:t} \big[ \rhostar{i}{2:t} f(\history{i}{t}) \big] \right) \bigg| \Pto 0.
        %\sup_{f \in \F} \bigg| \frac{1}{n} \sum_{i=1}^n \left\{ \WW{i}{2:t}(\betatilde{}, \betahat{}) f(\history{i}{t}) - \E_{\pi_{2:t}(\betatilde{1:t-1})} \big[ f(\history{i}{t}) \big] \right\} \bigg| \Pto 0.
    \end{equation}
\end{theorem}

\startproof{Theorem \ref{thm:weightedUWLLN}}
%%%%%%%%%%%%%%%%%%%%%%%%%%%%%%%%%%%

\proofSubsection{Showing display \eqref{UWLLN:eqn1} holds}
Let $\epsilon > 0$. For convenience, let $N_\epsilon \triangleq N_{[~]}\big(\epsilon, \F, L_{1+\alpha}(\Pstar) \big)$. Since $N_{[~]}\big(\epsilon, \F, L_{1+\alpha}(\Pstar) \big) < \infty$ by assumption, we can finitely many brackets $\big\{ (l_k, u_k) \big\}_{k=1}^{N_\epsilon}$ that cover $\F$ with $\Estar{2:t} \big[ \big| u_k(\history{i}{t}) - l_k(\history{i}{t}) \big|^{1+\alpha} \big]^{1/(1+\alpha)} \leq \epsilon$. Thus,
\begin{equation*}
	\sup_{f \in \F} \frac{1}{n} \sum_{i=1}^n \left\{ \WW{i}{2:t}(\betastar{}, \betahat{}) f(\history{i}{t}) 
	- \Estar{2:t} \big[ f(\history{i}{t}) \big] \right\}
\end{equation*}
\begin{equation*}
	\leq \max_{k \in [1 \colon N_\epsilon]} \frac{1}{n} \sum_{i=1}^n \left\{ \WW{i}{2:t}(\betastar{}, \betahat{}) u_k(\history{i}{t}) 
	- \Estar{2:t} \big[ l_k(\history{i}{t}) \big] \right\}
\end{equation*}
By Theorem \ref{thm:weightedWLLN} (Weighted Martingale Triangular Array Weak Law of Large Numbers), for any $k \in [1 \colon N_\epsilon]$, $\frac{1}{n} \sum_{i=1}^n \WW{i}{2:t}(\betastar{}, \betahat{}) u_k(\history{i}{t}) \Pto \Estar{2:t} \big[ u_k(\history{i}{t}) \big]$. Since there are finitely many brackets, by Slutsky's theorem this result holds simultaneously for all brackets $k \in [1 \colon N_\epsilon]$.
\begin{equation*}
	= o_P(1) + \max_{k \in [1 \colon N_\epsilon]} \left\{ \Estar{2:t} \big[ u_k(\history{i}{t}) \big]
	- \Estar{2:t} \big[ l_k(\history{i}{t}) \big] \right\}
\end{equation*}
By Jensen's inequality
\begin{equation*}
    \leq o_P(1) + \max_{k \in [1 \colon N_\epsilon]}\Estar{2:t} \big[ | u_k(\history{i}{t}) - l_k(\history{i}{t}) |^{1+\alpha} \big]^{1/(1+\alpha)}
    \leq o_P(1) + \epsilon.
\end{equation*}
The last inequality above holds because our brackets were chosen to be at most of size $\epsilon$, in $L_{1+\alpha}(\Pstar)$ norm. The above converges to zero because $\epsilon$ can be chosen to be arbitrarily small. 

The final result display \eqref{UWLLN:eqn1} holds by using the same argument above to show that $\sup_{f \in \F} \frac{1}{n} \sum_{i=1}^n \left\{ \Estar{2:t} \big[ f(\history{i}{t}) \big] - \WW{i}{2:t}(\betastar{}, \betahat{}) f(\history{i}{t}) \right\} \Pto 0$. 

\proofSubsection{Showing display \eqref{UWLLN:eqnTilde} holds}
Note that
\begin{multline*}
    \rhohat{i}{2:t} f(\history{i}{t}) 
    = \WW{i}{2:t}(\betastar{}, \betahat{}) \rhostar{i}{2:t} f(\history{i}{t}) \\
    = \WW{i}{2:t}(\betastar{}, \betahat{}) \bigg\{ \prod_{t'=2}^t \frac{1}{\pi_{t'}(\action{i}{t'}, \state{i}{t'}; \betastar{t'-1})} \bigg\} f(\history{i}{t}).
\end{multline*}
By the result from display \eqref{UWLLN:eqn1}, to show display \eqref{UWLLN:eqnTilde} holds it is sufficient to show that the function class
\begin{equation*}
    \F_{\pi} \triangleq \bigg\{ \bigg( \prod_{t'=2}^t \frac{1}{\pi_{t'}(\cdotspace; \betastar{t'-1})} \bigg) f(\cdotspace) \TN{~~s.t.~~} f \in \F \bigg\}
\end{equation*}
is such that $N_{[~]}\big(\epsilon, \F_\pi, L_{1+\alpha}(\Pstar) \big) < \infty$ for all $\epsilon > 0$.

Since $N_\epsilon \triangleq N_{[~]}\big(\epsilon, \F, L_{1+\alpha}(\Pstar) \big) < \infty$ by conditions of the theorem, we can find brackets $\big\{ (l_k, u_k) \big\}_{k=1}^{N_\epsilon}$ that cover $\F$ and $\Estar{2:t} \big[ \big| u_k(\history{i}{t}) - l_k(\history{i}{t}) \big|^{1+\alpha} \big]^{1/(1+\alpha)} \leq \epsilon$ for each bracket $(l_k, u_k)$. Let $f \in \F$. We can find one of these brackets $(l_k, u_k)$ such that $l_k(\history{i}{t}) \leq f(\history{i}{t}) \leq u_k(\history{i}{t})$ a.s.

Since $\rhostar{i}{2:t} > 0$ a.s., thus,
\begin{equation*}
    \rhostar{i}{2:t} l_k(\history{i}{t}) 
    \leq \rhostar{i}{2:t} f(\history{i}{t}) 
    \leq \rhostar{i}{2:t} u_k(\history{i}{t}).
\end{equation*}
So the brackets $\big\{ \big(\rhostar{i}{2:t} l_k, \rhostar{i}{2:t} u_k \big) \big\}_{k=1}^{N_\epsilon}$ cover $\F_\pi$.

Moreover, since $\rhostar{i}{2:t} \leq \pi_{\min}^{-(t-1)}$ a.s. by Condition \ref{cond:exploration}, thus
\begin{equation*}
    \Estar{2:t} \big[ \big| \rhostar{i}{2:t} u_k(\history{i}{t}) - \rhostar{i}{2:t} l_k(\history{i}{t}) \big|^{1+\alpha} \big]^{1/(1+\alpha)} \leq \pi_{\min}^{-(t-1)} \epsilon.
\end{equation*}

Thus, we have that $N_{[~]}\big(\pi_{\min}^{-(t-1)} \epsilon, \F_W, L_{1+\alpha}(\Pstar) \big) = N_{[~]}\big(\epsilon, \F, L_{1+\alpha}(\Pstar) \big)$. Thus, since $N_{[~]}\big(\epsilon, \F, L_{1+\alpha}(\Pstar) \big) < \infty$ for any $\epsilon > 0$ (by assumption of the Theorem), thus we have that $N_{[~]}\big(\epsilon, \F_W, L_{1+\alpha}(\Pstar) \big) < \infty$ for any $\epsilon > 0$. ~$\blacksquare$

\subsection{Showing Terms are $O_P(1)$ (Helper Lemma \ref{lemma:OpOne})} %%%%%%%%%%%%%%%%%%%%%%%%%%%%%%%%%%%%%%%%%%%%
\label{appsupport:OpOne}

\begin{lemma}[Showing Terms are $O_P(1)$ (Helper Lemma)]
    \label{lemma:OpOne}
    Let $g$ be a real-valued, function of $\history{i}{t}$ such that $\Estar{2:t} \big[ \big| g(\history{i}{t}) \big| \big] < \infty$. Under Condition \ref{cond:exploration},
    we have that $g(\history{i}{t}) = O_P(1)$. 
\end{lemma}

\startproof{Lemma \ref{lemma:OpOne}}
Let $\epsilon > 0$. By Condition \ref{cond:exploration} (Minimum Exploration),
\begin{equation*}
    \WW{i}{t'}\big( \betastar{t'-1}, \betahat{t'-1} \big) = \frac{\pistar{t'}(\action{i}{t'}, \state{i}{t'})}{\pihat{t'}(\action{i}{t'}, \state{i}{t'})}
    \geq \frac{\pistar{t'}(\action{i}{t'}, \state{i}{t'})}{1}
    \geq \pi_{\min} ~~~ \TN{a.s.}
\end{equation*}
By the above result and Markov inequality, for any $c_\epsilon > 0$,
\begin{multline}
    \label{eqn:Op1terms}
    \PP \big( \big| g(\history{i}{t}) \big| > c_{\epsilon} \big)
    \leq c_{\epsilon}^{-1} \E \big[ \big| g(\history{i}{t}) \big| \big] \\
    \leq c_{\epsilon}^{-1} \pi_{\min}^{-(t-1)} \E \left[ \WW{i}{2:t} \big( \betastar{}, \betahat{} \big) \big| g(\history{i}{t}) \big| \right] 
    = c_{\epsilon}^{-1} \pi_{\min}^{-(t-1)} \Estar{2:t} \left[ \big| g(\history{i}{t}) \big| \right].
\end{multline}
The above is less than or equal to $\epsilon$ by choosing $c_\epsilon > \epsilon^{-1} \pi_{\min}^{-(t-1)} \Estar{2:t} \big[ \big| g(\history{i}{t}) \big| \big]$. This is sufficient by the definition of $O_P(1)$. $\blacksquare$

%%%%%%%%%%%%%%%%%%%%%%%%%%%%%%%%%%%%%%%%%%%%
\subsection{Weighted Martingale Triangular Array Central Limit Theorem (Theorem \ref{thm:weightedCLT})} %%%%%%%%%%%%%%%%%%%%%%%%%%%%%%%%%%%%%%%%%%%%
\label{appsupport:weightedCLT}

%%%%%%%%%%%%%%%%%%%%%%%%%%%%%%%%%%%%%%%%%%%%
%%%%%%%%%%%%%%%%%%%%%%%%%%%%%%%%%%%%%%%%%%%%
\begin{theorem}[Weighted Martingale Triangular Array Central Limit Theorem]
	\label{thm:weightedCLT}
	Let $f_1, f_2,  ..., f_t$ be real-valued, measurable functions of $\history{i}{1}, \history{i}{2}, ..., \history{i}{t}$ respectively such that \\
    $\Estar{2:t'} \left[ \big| f_{t'} \big( \history{i}{t'} \big) \big|^{2+\alpha} \right] < \infty$ for some $\alpha > 0$, for each $t' \in [1 \colon t]$. We show that (below we use $\WW{i}{1} = 1$)
	\begin{multline}
	    \label{eqn:WMCLTresult}
		\frac{1}{\sqrt{n}} \sum_{i=1}^n \bigg\{ \sum_{t'=1}^t \WW{i}{1:t'}(\betastar{}, \betahat{}) f_{t'}(\history{i}{t'}) - \Estar{2:t} \bigg[ \sum_{t'=1}^t f_{t'}(\history{i}{t'}) \bigg] \bigg\} \\
		\Dto \N \bigg( 0, \Var_{\pistar{2:t}} \bigg( \sum_{t'=1}^t f_{t'}(\history{i}{t'}) \bigg) \bigg)
	\end{multline}	
	for $\Var_{\pistar{2:t}} \left( \sum_{t'=1}^t f_{t'}(\history{i}{t'}) \right) \triangleq \Estar{2:t} \left[ \big\{ \sum_{t'=1}^t f_{t'}(\history{i}{t'}) \big\}^2 \right] - \Estar{2:t} \left[ \sum_{t'=1}^t f_{t'}(\history{i}{t'}) \right]^2$ under the following conditions:
	\begin{enumerate}[label=(\Alph*)]
	    \item Conditions \ref{cond:exploration} (Minimum Exploration) and \ref{cond:lipschitzPolicy} (Lipschitz Policy Functions) hold.
	    %%%%%%%%%%%%%%%%%%%%
	    \item $\betahat{t'} - \betastar{t'} = O_P( 1/\sqrt{n} )$ for all $t' \in [1 \colon t-1]$.
	\end{enumerate}
\end{theorem}

\startproof{Theorem \ref{thm:weightedCLT}} %%%%%%%%%%%%%%%%%%%%%%%%%%%%%%%%%%%
We want to show that display \eqref{eqn:WMCLTresult} holds for any $t \in [2 \colon T]$. For notational convenience we consider the $t$ set to $T$ case; the argument holds by the same argument for any $t \in [2 \colon T]$.

The first task is to rewrite the left-hand side of display \eqref{eqn:WMCLTresult} as a sum of triangular array martingale differences; we take an approach similar to that we used in the proof of Theorem \ref{thm:weightedWLLN}.

Note that the left-hand side of display \eqref{eqn:WMCLTresult} can be rewritten as follows:
\begin{multline*}
    \frac{1}{\sqrt{n}} \sum_{i=1}^n \bigg\{ \sum_{t'=1}^t \WW{i}{1:t'}(\betastar{}, \betahat{}) f_{t'}(\history{i}{t'}) - \Estar{2:t} \bigg[ \sum_{t'=1}^t f_{t'}(\history{i}{t'}) \bigg] \bigg\} \\
	= \frac{1}{\sqrt{n}} \sum_{i=1}^n \sum_{t=1}^T  \bigg\{ \WW{i}{1:t}( \betastar{}, \betahat{} ) f_t(\history{i}{t}) 
	- \E \left[ \WW{i}{1:t}( \betastar{}, \betahat{} ) f_t(\history{i}{t}) \right] \bigg\}
\end{multline*} 
Above we use $\WW{i}{1} = 1$ and $\WW{i}{1:t}(\betastar{}, \betahat{}) \triangleq \WW{i}{2:t}(\betastar{}, \betahat{})$.
Additionally, above the expectation $\E$ is with respect to the data distribution used to collect the data, thus, $\E \left[ \WW{i}{1:t}( \betastar{}, \betahat{} ) f(\history{i}{t}) \right] = \Estar{2:t} \left[ f(\history{i}{t}) \right]$.

Let $\Xvar{i}{t} \triangleq \WW{i}{1:t}( \betastar{}, \betahat{} ) f_t(\history{i}{t})$ and $\Xvar{i}{1:T} \triangleq \sum_{t=1}^T \Xvar{i}{t}$.
\begin{equation*}
	= \frac{1}{\sqrt{n}} \sum_{i=1}^n \sum_{t=1}^T \left\{ \Xvar{i}{t} - \E \left[ \Xvar{i}{t} \right] \right\}
	= \frac{1}{\sqrt{n}} \sum_{i=1}^n \left\{ \Xvar{i}{1:T} - \E \left[ \Xvar{i}{1:T} \right] \right\}
\end{equation*}

Let $\history{1:n}{0} \triangleq \emptyset$ and $\state{1:n}{T+1} \triangleq \emptyset$. Note that $\Xvar{i}{1:T} = \E \left[ \Xvar{i}{1:T} \big| \history{1:n}{T}, \state{1:n}{T+1} \right]$ since $\Xvar{i}{1:T}$ is known given $\history{1:n}{T}$. By telescoping series, 
\begin{multline*}
	= \frac{1}{\sqrt{n}} \sum_{i=1}^n \underbrace{ \left\{ \E \left[ \Xvar{i}{1:T} \big| \history{1:n}{0}, \state{1:n}{1}  \right] - \E \left[ \Xvar{i}{1:T} \right] \right\} }_{ \triangleq \Zvar{i}{0} } \\
	+ \sum_{t=1}^{T} \bigg[ \frac{1}{\sqrt{n}} \sum_{i=1}^n \underbrace{ \left\{ \E \left[ \Xvar{i}{1:T} \big| \history{1:n}{t}, \state{1:n}{t+1} \right] 
	- \E \left[ \Xvar{i}{1:T} \big| \history{1:n}{t-1}, \state{1:n}{t} \right] \right\} }_{ \triangleq \Zvar{i}{t} } \bigg].
\end{multline*}
Note that the terms $\Zvar{i}{t}$ above are different from those we defined in the proof of Theorem \ref{thm:weightedWLLN}, since here the terms $\Xvar{i}{1:T}$ are a sum over $T$ terms.

Note that $\E \left[ \Zvar{i}{t} \big| \history{1:n}{t-1}, \state{1:n}{t} \right] = 0$ for all $i \in [1 \colon n]$ and $t \in [1 \colon T]$.
This is the case because for any $i \in [1 \colon n]$ and $t \in [1 \colon T]$,
\begin{equation*}
    %\label{app:WLLNZzero}
    \E \left[ \Zvar{i}{t} \big| \history{1:n}{t-1}, \state{1:n}{t} \right]
    = \E \left[ \E \left[ \Xvar{i}{1:T} \big| \history{1:n}{t}, \state{1:n}{t+1} \right] 
	- \E \left[ \Xvar{i}{1:T} \big| \history{1:n}{t-1}, \state{1:n}{t} \right] \bigg| \history{1:n}{t-1}, \state{1:n}{t} \right]
\end{equation*}
\begin{equation*}
    = \E \left[ \E \left[ \Xvar{i}{1:T} \big| \history{1:n}{t}, \state{1:n}{t+1} \right] 
	   \bigg| \history{1:n}{t-1}, \state{1:n}{t} \right]
    - \E \left[ \Xvar{i}{1:T} \big| \history{1:n}{t-1}, \state{1:n}{t} \right] = 0.
\end{equation*}
The final equality above holds by the law of iterated expectations.

\medskip
\noindent In the next two subsections we will show the following two results: 

\paragraph*{(i) Convergence of conditional variance}
\begin{equation}
	\label{mclt:condvarResult}
	\frac{1}{n} \sum_{i=1}^n \E \left[ \big( \Zvar{i}{0} \big)^2 \right] 
	+ \sum_{t=1}^T \frac{1}{n} \sum_{i=1}^n \E \left[ \big( \Zvar{i}{t} \big)^2 \big| \history{1:n}{t-1}, \state{1:n}{t} \right] \\
	\Pto \Var_{\pistar{2:T}} \bigg( \sum_{t=1}^T f_t(\history{i}{t}) \bigg).
\end{equation}

\paragraph*{(ii) Conditional Lindeberg} For any $\epsilon > 0$,
\begin{equation}
	\label{mclt:lindebergResult}
	\frac{1}{n} \sum_{i=1}^n \E \left[ \big( \Zvar{i}{0} \big)^2 \II_{|\Zvar{i}{0}| / \sqrt{n} > \epsilon } \right] 
	+ \sum_{t=1}^T \frac{1}{n} \sum_{i=1}^n \E \left[ \big( \Zvar{i}{t} \big)^2 \II_{|\Zvar{i}{t}| / \sqrt{n} > \epsilon } \big| \history{1:n}{t-1}, \state{1:n}{t} \right] \\
	\Pto 0.
\end{equation}

\noindent With the above two results we can apply Theorem 2.2 %for dependent random variables 
of \cite{dvoretzky1972asymptotic} (a martingale central limit theorem) to conclude that our desired result holds, i.e., 
\begin{equation*}
	\frac{1}{ \sqrt{n} } \sum_{t=0}^T \sum_{i=1}^n \Zvar{i}{t} \Dto \N \bigg( 0, \Var_{\pistar{2:T}} \bigg( \sum_{t=1}^T f_t(\history{i}{t}) \bigg) \bigg).
\end{equation*}
We first show that display \eqref{mclt:lindebergResult} holds. We then show that display \eqref{mclt:condvarResult} holds.

%%%%%%%%%%%%%%%%%%%%%%%%%%%%%%%%%%%%%%%%%%%%%%%%%%%%%%%%%%%%
%%%%%%%%%%%%%%%%%%%%%%%%%%%%%%%%%%%%%%%%%%%%%%%%%%%%%%%%%%%%
\proofSubsection{\bo{(ii) Conditional Lindeberg; Display \eqref{mclt:lindebergResult}}} For any $\epsilon > 0$, we show that the following conditional Lindeberg term is $o_P(1)$:
\begin{equation*}
	\frac{1}{n} \sum_{i=1}^n \E \left[ \big( \Zvar{i}{0} \big)^2 \II_{|\Zvar{i}{0}| / \sqrt{n} > \epsilon } \right] 
	+ \sum_{t=1}^T \frac{1}{n} \sum_{i=1}^n \E \left[ \big( \Zvar{i}{t} \big)^2 \II_{|\Zvar{i}{t}| / \sqrt{n} > \epsilon } \big| \history{1:n}{t-1}, \state{1:n}{t} \right]
\end{equation*}

Note that for $\alpha > 0$, 
$\II_{|Z|/\sqrt{n}> \epsilon} 
= \II_{|Z|/(\epsilon \sqrt{n}) > 1}
= \II_{|Z|^\alpha/(\epsilon \sqrt{n})^\alpha > 1} 
\leq |Z|^\alpha/(\epsilon \sqrt{n})^\alpha$. \\
Thus we can upper-bound the previous display as follows:
\begin{equation}
	\label{mclt:lindeberg1}
	\leq \frac{1}{n (\epsilon \sqrt{n})^{\alpha}} \sum_{i=1}^n \E \left[ \big| \Zvar{i}{0} \big|^{2+\alpha} \right] 
	+ \sum_{t=1}^T \frac{1}{n (\epsilon \sqrt{n})^{\alpha}} \sum_{i=1}^n\E \left[ \big| \Zvar{i}{t} \big|^{2+\alpha} \big| \history{1:n}{t-1}, \state{1:n}{t} \right].
\end{equation}

Note that for any $t' \in [0 \colon T]$,
\begin{equation*}
	\big| Z_{t'}^{(i)} \big|^{2+\alpha}
	= \left| \E \left[ \Xvar{i}{1:T} \big| \history{1:n}{t'}, \state{1:n}{t'+1} \right] 
	- \E \left[ \Xvar{i}{1:T} \big| \history{1:n}{t'-1}, \state{1:n}{t'} \right] \right|^{2+\alpha} 
\end{equation*}
Above, by slight abuse of notation, if $t' = 0$, we use $\history{1:n}{-1} \triangleq \emptyset$ and $\state{1:n}{0} \triangleq \emptyset$.
\begin{equation*}
	= \left| \sum_{s=1}^T \left\{ \E \left[ \Xvar{i}{s} \big| \history{1:n}{t'}, \state{1:n}{t'+1} \right] 
	- \E \left[ \Xvar{i}{s} \big| \history{1:n}{t'-1}, \state{1:n}{t'} \right] \right\} \right|^{2+\alpha} 
\end{equation*}

By repeatedly applying Lemma \ref{lemma:binomialBound} (Inequality Using Binomial Theorem) for any numbers $a_1, a_2, ..., a_K$, $| \sum_{k=1}^K a_k |^{\eta} \leq c_{2+\alpha}^K \sum_{k=1}^K | a_k |^{2+\alpha}$ for some constant $c_{2+\alpha} > 0$.
\begin{equation*}
	\leq c_{2+\alpha}^{T} \sum_{s=1}^T \bigg\{ \bigg| \E \left[ \Xvar{i}{s} \big| \history{1:n}{t'}, \state{1:n}{t'+1} \right] \bigg|^{2+\alpha}
	+ \bigg| \E \left[ \Xvar{i}{s} \big| \history{1:n}{t'-1}, \state{1:n}{t'} \right] \bigg|^{2+\alpha} \bigg\}
\end{equation*}
By Jensen's Inequality,
\begin{equation*}
	\leq c_{2+\alpha}^{T} \sum_{s=1}^T \bigg\{ \E \left[ \big| \Xvar{i}{s} \big|^{2+\alpha} \big| \history{1:n}{t'}, \state{1:n}{t'+1} \right]
	+ \E \left[ \big| \Xvar{i}{s} \big|^{2+\alpha}  \big| \history{1:n}{t'-1}, \state{1:n}{t'} \right] \bigg|\bigg\}
\end{equation*}

\medskip
\noindent Thus, we can upper-bound display \eqref{mclt:lindeberg1} as follows:
\begin{equation*}
	\frac{1}{n (\epsilon \sqrt{n})^{\alpha}} \sum_{i=1}^n \E \left[ \big| \Zvar{i}{0} \big|^{2+\alpha} \right] 
	+ \sum_{t=1}^T \frac{1}{n (\epsilon \sqrt{n})^{\alpha}} \sum_{i=1}^n\E \left[ \big| \Zvar{i}{t} \big|^{2+\alpha} \big| \history{1:n}{t-1}, \state{1:n}{t} \right]
\end{equation*}
\begin{multline*}
	\leq \frac{c_{2+\alpha}^{T}}{n (\epsilon \sqrt{n})^{\alpha}} \sum_{i=1}^n \sum_{s=1}^T \bigg\{ \E \bigg[ \E \left[ \big| \Xvar{i}{s} \big|^{2+\alpha} \big| \history{1:n}{0}, \state{1:n}{1}  \right] + \E \left[ \big| \Xvar{i}{s} \big|^{2+\alpha} \right] \bigg] \bigg\} \\
	+ \sum_{t=1}^{T} \frac{c_{2+\alpha}^{T}}{n (\epsilon \sqrt{n})^{\alpha}} \sum_{i=1}^n \sum_{s=1}^T \bigg\{ \E \bigg[ \E \left[ \big| \Xvar{i}{s} \big|^{2+\alpha} \big| \history{1:n}{t}, \state{1:n}{t+1} \right]  \\
	+ \E \left[ \big| \Xvar{i}{s} \big|^{2+\alpha} \big| \history{1:n}{t-1}, \state{1:n}{t} \right] \bigg| \history{1:n}{t-1}, \state{1:n}{t} \bigg] \bigg\}
\end{multline*}
By the law of itereated expectations,
\begin{equation*}
	= \frac{c_{2+\alpha}^{T}}{n (\epsilon \sqrt{n})^{\alpha}} \sum_{i=1}^n \sum_{s=1}^T 2 \E \left[ \big| \Xvar{i}{s} \big|^{2+\alpha} \right] \\
	+ \sum_{t=1}^{T} \frac{c_{2+\alpha}^{T}}{n (\epsilon \sqrt{n})^{\alpha}} \sum_{i=1}^n \sum_{s=1}^T 
	2 \E \left[ \big| \Xvar{i}{s} \big|^{2+\alpha} \big| \history{1:n}{t-1}, \state{1:n}{t} \right]
\end{equation*}

To show that the above is $o_P(1)$, it is sufficient to show that $\E \left[ \big| \Xvar{i}{s} \big|^{2+\alpha} \big| \history{1:n}{t-1}, \state{1:n}{t} \right]$ and $\E \left[ \big| \Xvar{i}{s} \big|^{2+\alpha} \right]$ for all $t, s \in [1 \colon T]$ are all $O_P(1)$. By Lemma \ref{lemma:OpOne}, it is sufficient to show that $\Estar{2:s} \left[ \big| \Xvar{i}{s} \big|^{2+\alpha} \right]$ and $\Estar{2:t} \left[ \left| \E \left[ \big| \Xvar{i}{s} \big|^{2+\alpha} \big| \history{1:n}{t-1}, \state{1:n}{t} \right] \right| \right]$ are bounded.

By Jensen's inequality,
\begin{equation*}
	\Estar{2:t} \bigg[ \bigg| \E \left[ \big| \Xvar{i}{s} \big|^{2+\alpha} \bigg| \history{1:n}{t-1}, \state{1:n}{t} \right] \bigg| \bigg]
	\leq \Estar{2:t} \left[ \big| \Xvar{i}{s} \big|^{2+\alpha} \right].
\end{equation*}
Thus, it is sufficient to show that $\Estar{2:t} \left[ \big| \Xvar{i}{s} \big|^{2+\alpha} \right] < \infty$ for all $s \in [1 \colon T]$.

By Condition \ref{cond:exploration}, $\WW{i}{2:s}(\betastar{}, \betahat{}) \leq \pi_{\min}^{-(s-1)}$ a.s. so, 
\begin{equation*}
	\Estar{2:s} \left[ \big| \Xvar{i}{s} \big|^{2+\alpha} \right]
	= \Estar{2:s} \left[ \big| \WW{i}{2:s} ( \betastar{}, \betahat{} ) f_s(\history{i}{s}) \big|^{2+\alpha} \right]
\end{equation*}
\begin{equation*}
	\leq \pi_{\min}^{-(s-1)(2+\alpha)} \Estar{2:s} \left[ \WW{i}{2:s} ( \betastar{}, \betahat{} ) \big| f_s(\history{i}{s}) \big|^{2+\alpha} \right]
\end{equation*}
\begin{equation*}
	= \pi_{\min}^{-(s-1)(2+\alpha)} \Estar{2:s} \left[ \big| f_s(\history{i}{s}) \big|^{2+\alpha} \right]
	< \infty.
\end{equation*}

%%%%%%%%%%%%%%%%%%%%%%%%%%%%%%%%%%%%%%%%%%%%%%%%%%%%%%%%%%%%
%%%%%%%%%%%%%%%%%%%%%%%%%%%%%%%%%%%%%%%%%%%%
\proofSubsection{\bo{(i) Convergence of conditional variance; Display \eqref{mclt:condvarResult}:}} We now show that display \eqref{mclt:condvarResult} holds.
Using the definition of $\Zvar{i}{t}$,
\begin{equation*}
	\frac{1}{n} \sum_{i=1}^n \E \left[ \big( \Zvar{i}{0} \big)^2 \right] 
	+ \sum_{t=1}^T \frac{1}{n} \sum_{i=1}^n \E \left[ \big( \Zvar{i}{t} \big)^2 \big| \history{1:n}{t-1}, \state{1:n}{t} \right]
\end{equation*}

\vspace{-3mm}
\begin{multline*}
	= \frac{1}{n} \sum_{i=1}^n \E \bigg[ \bigg( \E \left[ \Xvar{i}{1:T} \big| \history{1:n}{0}, \state{1:n}{1}  \right] - \E \left[ \Xvar{i}{1:T} \right] \bigg)^2 \bigg] \\
	+ \sum_{t=1}^{T} \frac{1}{n} \sum_{i=1}^n \E \bigg[ \bigg( \E \left[ \Xvar{i}{1:T} \big| \history{1:n}{t}, \state{1:n}{t+1} \right] 
	- \E \left[ \Xvar{i}{1:T} \big| \history{1:n}{t-1}, \state{1:n}{t} \right] \bigg)^2 \bigg| \history{1:n}{t-1}, \state{1:n}{t} \bigg].
\end{multline*}

\vspace{-3mm}
\begin{multline}
    \label{app:WCLTreindexingPre}
	= \frac{1}{n} \sum_{i=1}^n \bigg\{ \E \bigg[ \E \left[ \Xvar{i}{1:T} \big| \state{1:n}{1}  \right]^2 \bigg] - \E \left[ \Xvar{i}{1:T} \right]^2 \bigg\} \\
	+ \sum_{t=1}^{T} \frac{1}{n} \sum_{i=1}^n \bigg\{ \E \bigg[ \E \left[ \Xvar{i}{1:T} \big| \history{1:n}{t}, \state{1:n}{t+1} \right]^2 \bigg| \history{1:n}{t-1}, \state{1:n}{t} \bigg] 
	- \E \left[ \Xvar{i}{1:T} \big| \history{1:n}{t-1}, \state{1:n}{t} \right]^2 \bigg\}.
\end{multline}

\noindent Note by re-indexing, $- \sum_{t=1}^{T} \E \left[ \Xvar{i}{1:T} \big| \history{1:n}{t-1}, \state{1:n}{t} \right]^2
= - \sum_{t=0}^{T-1} \E \left[ \Xvar{i}{1:T} \big| \history{1:n}{t}, \state{1:n}{t+1} \right]^2 \\
= - \sum_{t=1}^{T} \underbrace{ \E \left[ \Xvar{i}{1:T} \big| \history{1:n}{t}, \state{1:n}{t+1} \right]^2 }_{(a)}
- \underbrace{ \E \left[ \Xvar{i}{1:T} \big| \state{1:n}{1} \right]^2 }_{(b)}
+ \underbrace{ \big( \Xvar{i}{1:T} \big)^2 }_{(c)}$.
By rearranging terms the terms in display \eqref{app:WCLTreindexingPre},
\begin{multline}
	\label{mclteqn:YdiffRaw}
	= \frac{1}{n} \sum_{i=1}^n \bigg\{ \underbrace{ \big( \Xvar{i}{1:T} \big)^2 }_{(c)} - \E \left[ \Xvar{i}{1:T} \right]^2 \bigg\} \\
	+ \frac{1}{n} \sum_{i=1}^n \bigg\{ \E \bigg[ \E \left[ \Xvar{i}{1:T} \big| \state{1:n}{1}  \right]^2 \bigg] - \underbrace{ \E \left[ \Xvar{i}{1:T} \big| \state{1:n}{1} \right]^2 }_{(b)} \bigg\} \\
	+ \sum_{t=1}^{T} \frac{1}{n} \sum_{i=1}^n \bigg\{ \E \bigg[ \E \left[ \Xvar{i}{1:T} \big| \history{1:n}{t}, \state{1:n}{t+1} \right]^2 \bigg| \history{1:n}{t-1}, \state{1:n}{t} \bigg] 
	- \underbrace{ \E \left[ \Xvar{i}{1:T} \big| \history{1:n}{t}, \state{1:n}{t+1} \right]^2 }_{(a)} \bigg\}.
\end{multline}
For the remainder of the proof we will show the following results, which combined with display \eqref{mclteqn:YdiffRaw} above are sufficient for display \eqref{mclt:condvarResult}:
\begin{equation}
	\label{mclteqn:YdiffRaw1}
	\frac{1}{n} \sum_{i=1}^n \bigg\{ \big( \Xvar{i}{1:T} \big)^2 - \E \left[ \Xvar{i}{1:T} \right]^2 \bigg\} 
	\Pto \Var_{\pistar{2:T}} \bigg( \sum_{t=1}^T f_t( \history{i}{t}) \bigg)
\end{equation}
\begin{equation}
	\label{mclteqn:YdiffRaw2}
	\frac{1}{n} \sum_{i=1}^n \bigg\{ \E \bigg[ \E \left[ \Xvar{i}{1:T} \big| \state{1:n}{1}  \right]^2 \bigg] - \E \left[ \Xvar{i}{1:T} \big| \state{1:n}{1} \right]^2 \bigg\} 	
	\Pto 0
\end{equation}
\begin{equation}
	\label{mclteqn:YdiffRaw3}
	\sum_{t=1}^{T} \frac{1}{n} \sum_{i=1}^n \bigg\{ \E \bigg[ \E \left[ \Xvar{i}{1:T} \big| \history{1:n}{t}, \state{1:n}{t+1} \right]^2 \bigg| \history{1:n}{t-1}, \state{1:n}{t} \bigg] 
	- \E \left[ \Xvar{i}{1:T} \big| \history{1:n}{t}, \state{1:n}{t+1} \right]^2 \bigg\}
	\Pto 0
\end{equation}

\medskip
\noindent Before showing the above three results, first note the following observations:
\begin{itemize}
	%%%%%%%%%%%%%%%%%%%%%%%%%%%%%%%%%%%%%%%%%%%%%%%%
	\item By Condition \ref{cond:exploration} (Minimum Exploration),
    \begin{equation}
        \label{eqn:MCLTWbound}
        \pihat{t}(\action{i}{t}, \state{i}{t}) \geq \pi_{\min}
        ~~~\TN{a.s.}
        ~~~~~~\TN{and}~~~~~~
        \WW{i}{2:t}(\betastar{}, \betahat{}) \leq \pi_{\min}^{-(t-1)}~~~\TN{a.s.}
    \end{equation}
	
	\item Additionally note that
	\begin{multline}
	    \label{mclteqn:WdiffInequality}
	    \left| \WW{i}{t}(\betastar{t-1}, \betahat{t-1}) - 1 \right|
	    = \left| \WW{i}{t}(\betastar{t-1}, \betahat{t-1}) - \WW{i}{t}(\betastar{t-1}, \betastar{t-1}) \right|  \\
	    \leq \left| \pihat{t} \big( \action{i}{t}, \state{i}{t} \big)^{-1} - \pistar{t} \big( \action{i}{t}, \state{i}{t} \big)^{-1} \right| 
	    \leq \max_{a \in \MC{A}} \left| \pihat{t} \big( a, \state{i}{t} \big)^{-1}
	    - \pistar{t} \big( a, \state{i}{t} \big)^{-1} \right| \\
	    \underbrace{\leq}_{(i)} \pi_{\min}^{-2} \max_{a \in \MC{A}} \left| \pihat{t} \big( a, \state{i}{t} \big)
	    - \pistar{t} \big( a, \state{i}{t} \big) \right| 
	    \underbrace{\leq}_{(ii)} \pi_{\min}^{-2} \max_{a \in \MC{A}}\pidot{t} \big( a, \state{i}{t} \big) \big\| \betahat{t-1} - \beta_{t-1}^* \big\|_2.
    \end{multline}

    Inequality (i) above holds because by Taylor Series expansion, $\hat{\pi}^{-1} - \pi^{*,-1} = (-1) \tilde{\pi}^{-2} (\hat{\pi}-\pi^*)$ for some $\tilde{\pi}$ between $\hat{\pi}$ and $\pi^*$. By Condition \ref{cond:exploration} (Minimum Exploration) and display \eqref{eqn:MCLTWbound}, $\tilde{\pi} \geq \min(\hat{\pi}, \pi^*) \geq \pi_{\min} > 0$ a.s.

    Inequality (ii) holds by Condition \ref{cond:lipschitzPolicy} (Lipschitz Policy Functions). 
    
    \item Since the action space $\MC{A}$ is finite $\Estar{2:t} \big[ \max_{a \in \MC{A}}\pidot{t} \big( a, \state{i}{t} \big) \big] 
    \leq \sum_{a \in \MC{A}} \Estar{2:t} \big[ \pidot{t} \big( a, \state{i}{t} \big) \big] < \infty$; the last inequality holds by Condition \ref{cond:lipschitzPolicy}. Thus, by Lemma \ref{lemma:OpOne}, $\max_{a \in \MC{A}}\pidot{t} \big( a, \state{i}{t} \big) = O_P(1)$.
    Also since $\big\| \betahat{t-1} - \betastar{t-1} \big\|_2 = O_P(1 / \sqrt{n} )$ by assumption, by display \eqref{mclteqn:WdiffInequality}, we have that $\WW{i}{t}(\betastar{}, \betahat{}) = 1 + O_P( 1/ \sqrt{n} )$. 
    Moreover, note that
    \begin{equation}
	\label{mclteqn:Wlimit}
	    \WW{i}{2:t} \big( \betastar{t-1}, \betahat{t-1} \big)
	    = \left( 1 + O_P( 1/ \sqrt{n} ) \right)^{t-1}
	    = 1 + O_P( 1/ \sqrt{n} ).
    \end{equation}
\end{itemize}

%%%%%%%%%%%%%%%%%%%%%%%%%%%%%%%%%%%%%%%%%%
\proofSubsubsection{1. Showing Display \eqref{mclteqn:YdiffRaw1} holds}
Since $\Xvar{i}{1:T} = \sum_{t=1}^T \Xvar{i}{t} = \sum_{t=1}^T \WW{i}{2:t}(\betastar{}, \betahat{}) f_t(\history{i}{t})$,
\begin{equation*}
	\frac{1}{n} \sum_{i=1}^n \big( \Xvar{i}{1:T} \big)^2
	= \frac{1}{n} \sum_{i=1}^n \bigg\{ \sum_{t=1}^T \WW{i}{2:t}(\betastar{}, \betahat{}) f_t(\history{i}{t}) \bigg\}^2
\end{equation*}
\begin{equation*}
	= \frac{1}{n} \sum_{i=1}^n \sum_{t=1}^T \sum_{s=1}^T \WW{i}{2:t}(\betastar{}, \betahat{}) f_t(\history{i}{t}) 
	\WW{i}{2:s}(\betastar{}, \betahat{}) f_s(\history{i}{s})
\end{equation*}
\begin{equation*}
	= \frac{1}{n} \sum_{i=1}^n \sum_{t=1}^T \sum_{s=1}^T \WW{i}{2:\max(t,s)}(\betastar{}, \betahat{}) f_t(\history{i}{t}) 
	f_s(\history{i}{s}) \WW{i}{2:\min(t,s)}(\betastar{}, \betahat{})
\end{equation*}

Note that since $\Estar{2:t} \big[ f_t(\history{i}{t})^2 \big]$ and $\Estar{2:s} \big[ f_s(\history{i}{s})^2 \big]$ is bounded by assumption, by Lemma \ref{lemma:OpOne}, $f_t(\history{i}{t}) = O_P(1)$ and $f_s(\history{i}{s}) = O_P(1)$.

Moreover, by display \eqref{mclteqn:Wlimit}, we have that $\WW{i}{2:\min(t,s)}(\betastar{}, \betahat{}) = 1 + O_P( 1 / \sqrt{n} )$ and $\WW{i}{2:\max(t,s)}(\betastar{}, \betahat{}) = 1 + O_P( 1 / \sqrt{n} )$. 

Thus, $\WW{i}{2:\max(t,s)}(\betastar{}, \betahat{}) f_t(\history{i}{t}) f_s(\history{i}{s}) \WW{i}{2:\min(t,s)}(\betastar{}, \betahat{}) \\
= \WW{i}{2:\max(t,s)}(\betastar{}, \betahat{}) f_t(\history{i}{t}) f_s(\history{i}{s}) + O_P( 1 / \sqrt{n} )$. So,
\begin{equation*}
	= \frac{1}{n} \sum_{i=1}^n \sum_{t=1}^T \bigg\{ \sum_{s=1}^T \WW{i}{2:\max(t,s)}(\betastar{}, \betahat{}) f_t(\history{i}{t}) f_s(\history{i}{s})
	+ O_P( 1 / \sqrt{n} ) \bigg\}
\end{equation*}
\begin{equation*}
	= \frac{1}{n} \sum_{i=1}^n \sum_{t=1}^T \sum_{s=1}^T \WW{i}{2:\max(t,s)}(\betastar{}, \betahat{}) f_t(\history{i}{t}) f_s(\history{i}{s})
	+ o_P(1).
\end{equation*}
\medskip

\noindent Note that by display \eqref{eqn:MCLTWbound},
\begin{equation*}
	\Estar{2:T} \left[ \big| \WW{i}{2:\max(t,s)}(\betastar{}, \betahat{}) f_t(\history{i}{t}) f_s(\history{i}{s}) \big|^{1+\alpha/2} \right]
\end{equation*}
\begin{equation*}
	\leq \pi_{\min}^{-\{ \max(t,s)-1\}} \Estar{2:T} \left[ \big| f_t(\history{i}{t}) f_s(\history{i}{s}) \big|^{1+\alpha/2} \right]
\end{equation*}
\begin{equation*}
	\leq \pi_{\min}^{-\{ \max(t,s)-1\}} \Estar{2:T} \bigg[  \max\left\{ \big| f_t(\history{i}{t}) \big|^2, ~ \big| f_s(\history{i}{s}) \big|^2 \right\}^{1+\alpha/2} \bigg]
\end{equation*}
\begin{equation*}
	\leq \pi_{\min}^{-\{ \max(t,s)-1\}} \Estar{2:T} \left[  \left\{ \big| f_t(\history{i}{t}) \big|^2 + \big| f_s(\history{i}{s}) \big|^2 \right\}^{1+\alpha/2} \right]
\end{equation*}
By Lemma \ref{lemma:binomialBound}, for some positive constant $c_{1+\alpha/2}$,
\begin{equation*}
	\leq \pi_{\min}^{-\{ \max(t,s)-1\}} c_{1+\alpha/2} \Estar{2:T} \left[ \big| f_t(\history{i}{t}) \big|^{2+\alpha} + \big| f_s(\history{i}{s}) \big|^{2+\alpha} \right] < \infty.
\end{equation*}
The above is bounded because of our assumption that $\Estar{2:T} \big[  \big| f_t(\history{i}{t}) \big|^{2+\alpha} \big] < \infty$ and \\
$\Estar{2:T} \big[  \big| f_s(\history{i}{s}) \big|^{2+\alpha} \big] < \infty$.
\medskip
\noindent Thus, we can apply the Weighted Martingale Triangular Array Weak Law of Large Numbers (Theorem \ref{thm:weightedWLLN}) to get that
\begin{multline*}
	\frac{1}{n} \sum_{i=1}^n \big( \Xvar{i}{1:T} \big)^2
    = \frac{1}{n} \sum_{i=1}^n \sum_{t=1}^T \sum_{s=1}^T \WW{i}{2:\max(t,s)}(\betastar{}, \betahat{}) f_t(\history{i}{t}) f_s(\history{i}{s})
	+ o_P(1) \\
    \Pto \sum_{t=1}^T \sum_{s=1}^T \Estar{2:T} \left[ f_t(\history{i}{t}) f_s(\history{i}{s}) \right]
    = \Estar{2:T} \bigg[ \bigg\{ \sum_{t=1}^T f_t(\history{i}{t}) \bigg\}^2 \bigg]
\end{multline*}

\noindent Additionally, note that
\begin{equation*}
	\frac{1}{n} \sum_{i=1}^n \E \left[ \Xvar{i}{1:T} \right]^2
	= \frac{1}{n} \sum_{i=1}^n \Estar{2:T} \bigg[ \sum_{t=1}^T f_t(\history{i}{t}) \bigg]^2
	= \Estar{2:T} \bigg[ \sum_{t=1}^T f_t(\history{i}{t}) \bigg]^2.
\end{equation*}

\noindent Thus,
\begin{multline*}
    %\label{mclteqn:YconvVar}
	\frac{1}{n} \sum_{i=1}^n \left\{ \big( \Xvar{i}{1:T} \big)^2 - \E \left[ \Xvar{i}{1:T} \right]^2 \right\}
	\Pto \Estar{2:T} \bigg[ \bigg\{ \sum_{t=1}^T f_t(\history{i}{t}) \bigg\}^2 \bigg] - \Estar{2:T} \bigg[ \sum_{t=1}^T f_t(\history{i}{t}) \bigg]^2 \\
    = \Var_{\pistar{2:T}} \bigg( \sum_{t=1}^T f_t( \history{i}{t}) \bigg).
\end{multline*}
The last equality above holds since $\Var_{\pistar{2:T}} \left( \sum_{t=1}^T f( \history{i}{t}) \right)
= \Estar{2:T} \left[ \left\{ \sum_{t=1}^T f_t(\history{i}{t}) \right\}^2 \right] - \Estar{2:T} \left[ \sum_{t=1}^T f_t(\history{i}{t}) \right]^2$.

\medskip
%%%%%%%%%%%%%%%%%%%%%%%%%%%%%%%%%%%%%%%%%%
\proofSubsubsection{2. Showing Display \eqref{mclteqn:YdiffRaw2} holds}

\noindent Note that $\E \left[ \E \big[ \Xvar{i}{1:T} \big| \state{1:n}{1} \big]^2 \right]
    = \E \left[ \Estar{2:T} \left[ \sum_{t=1}^T  f_t(\history{i}{t}) \big| \state{i}{1} \right]^2 \right]$. Also, note the following:
\begin{equation*}
	\frac{1}{n} \sum_{i=1}^n \E \left[ \Xvar{i}{1:T} \big| \state{1:n}{1} \right]^2
	= \frac{1}{n} \sum_{i=1}^n \bigg\{ \sum_{t=1}^T \E \left[ \WW{i}{2:t}(\betastar{}, \betahat{}) f_t(\history{i}{t}) \big| \state{1:n}{1} \right] \bigg\}^2
\end{equation*}
\begin{equation*}
	= \frac{1}{n} \sum_{i=1}^n \bigg\{ \sum_{t=1}^T \Estar{2:t} \left[ f_t(\history{i}{t}) \big| \state{1:n}{1} \right] \bigg\}^2
	= \frac{1}{n} \sum_{i=1}^n \bigg\{ \sum_{t=1}^T \Estar{2:t} \left[ f_t(\history{i}{t}) \big| \state{i}{1} \right] \bigg\}^2
\end{equation*}

Since $\Estar{2:t} \big[ f_t(\history{i}{t})^2 \big] < \infty$ by assumption, thus by Lemma \ref{lemma:binomialBound}, \\
$\Estar{2:t} \big[ \big\{ \sum_{t'=1}^t f_t(\history{i}{t}) \big\}^2 \big] < \infty$. Since $\state{1}{1}, \state{2}{1}, \state{3}{1}, \dots \state{n}{1}$ are i.i.d., by the weak law of large numbers,
\begin{equation*}
	\frac{1}{n} \sum_{i=1}^n \bigg( \bigg\{ \sum_{t=1}^T \Estar{2:t} \left[ f_t(\history{i}{t}) \big| \state{i}{1} \right] \bigg\}^2
	- \E \bigg[ \bigg\{ \sum_{t=1}^T \Estar{2:T} \left[  f_t(\history{i}{t}) \big| \state{i}{1} \right] \bigg\}^2 \bigg] \bigg) \Pto 0
\end{equation*}
Thus, we have that
\begin{equation*}
	%\label{mclteqn:remainderPzeroT0}
	\frac{1}{n} \sum_{i=1}^n \bigg\{ \E \bigg[ \E \left[ \Xvar{i}{1:T} \big| \state{1:n}{1}  \right]^2 \bigg] - \E \left[ \Xvar{i}{1:T} \big| \state{1:n}{1} \right]^2 \bigg\} \Pto 0.
\end{equation*}

\medskip
%%%%%%%%%%%%%%%%%%%%%%%%%%%%%%%%%%%%%%%%%%
\proofSubsubsection{3. Showing Display \eqref{mclteqn:YdiffRaw3} holds}
Note that for any $t \in [1 \colon T]$,
\begin{equation*}
	\E \left[ \Xvar{i}{1:T} \big| \history{1:n}{t}, \state{1:n}{t+1} \right]^2 
	= \frac{1}{n} \sum_{i=1}^n \E \bigg[ \sum_{t'=1}^{t} \Xvar{i}{t'}
	+ \sum_{t'=t+1}^T \Xvar{i}{t'} \bigg| \history{1:n}{t}, \state{1:n}{t+1} \bigg]^2
\end{equation*}
Recall $\Xvar{i}{t'} \triangleq \WW{i}{2:t'}(\betastar{}, \betahat{}) f_{t'}(\history{i}{t'})$. 
\begin{equation*}
	= \bigg\{ \sum_{t'=1}^{t} \WW{i}{2:t'}(\betastar{}, \betahat{}) f_{t'}(\history{i}{t'})
	+ \E \bigg[ \sum_{t'=t+1}^T \WW{i}{2:t'}(\betastar{}, \betahat{}) f_{t'}(\history{i}{t'}) \bigg| \history{1:n}{t}, \state{1:n}{t+1} \bigg] \bigg\}^2
\end{equation*}
\begin{multline*}
	= \bigg\{ \sum_{t'=1}^{t} \WW{i}{2:t'}(\betastar{}, \betahat{}) f_{t'}(\history{i}{t'}) \\
	+ \WW{i}{2:t}(\betastar{}, \betahat{}) \E \bigg[ \sum_{t'=t+1}^T \WW{i}{t+1:t'}(\betastar{}, \betahat{}) f_{t'}(\history{i}{t'}) \bigg| \history{1:n}{t}, \state{1:n}{t+1} \bigg] \bigg\}^2
\end{multline*}
\begin{equation*}
	= \bigg\{ \sum_{t'=1}^{t} \WW{i}{2:t'}(\betastar{}, \betahat{}) f_{t'}(\history{i}{t'}) 
	+ \WW{i}{2:t}(\betastar{}, \betahat{}) \Estar{t+1:T} \bigg[ \sum_{t'=t+1}^T f_{t'}(\history{i}{t'}) \bigg| \history{1:n}{t}, \state{1:n}{t+1} \bigg] \bigg\}^2
\end{equation*}

Note $\Estar{t+1:T} \left[ \sum_{t'=t+1}^T f_{t'}(\history{i}{t'}) \big| \history{1:n}{t}, \state{1:n}{t+1} \right] 
= \Estar{t+1:T} \left[ \sum_{t'=t+1}^T f_{t'}(\history{i}{t'}) \big| \history{i}{t}, \state{i}{t+1} \right]$, since in the expectation actions are selected using fixed target policies $\pistar{t+1:T}$ (the first expectation conditions on $\history{1:n}{t}, \state{1:n}{t+1}$ and the second conditions on $\history{i}{t}, \state{i}{t+1}$).
\begin{equation*}
	= \bigg\{ \sum_{t'=1}^{t} \WW{i}{2:t'}(\betastar{}, \betahat{}) f_{t'}(\history{i}{t'}) 
	+ \WW{i}{2:t}(\betastar{}, \betahat{}) \Estar{t+1:T} \bigg[ \sum_{t'=t+1}^T f_{t'}(\history{i}{t'}) \bigg| \history{i}{t}, \state{i}{t+1} \bigg] \bigg\}^2
\end{equation*}

For convenience, let $\tilde{f}_{t'}(\history{i}{t'}, \state{i}{t'+1}) \triangleq f_{t'}(\history{i}{t'})$ for all $t'\in [1 \colon t-1]$ and let $\tilde{f}_{t}(\history{i}{t}, \state{i}{t+1}) \triangleq f_{t}(\history{i}{t}) + \Estar{t+1:T} \left[ \sum_{t'=t+1}^T f_{t'}(\history{i}{t'}) \big| \history{i}{t}, \state{i}{t+1} \right]$.
\begin{equation*}
	= \bigg\{ \sum_{t'=1}^{t} \WW{i}{2:t'}(\betastar{}, \betahat{}) \tilde{f}_{t'} (\history{i}{t'}, \state{i}{t'+1}) \bigg\}^2
\end{equation*}
\begin{multline}
    \label{app:WCLTthreeTermsSquare}
	= \underbrace{ \bigg\{ \sum_{t'=1}^{t-1} \WW{i}{2:t'}(\betastar{}, \betahat{}) \tilde{f}_{t'} (\history{i}{t'}, \state{i}{t'+1}) \bigg\}^2 }_{ \triangleq ( U_{1:t-1}^{(i)} )^2 }
    + \underbrace{ \bigg\{ \WW{i}{2:t}(\betastar{}, \betahat{}) \tilde{f}_{t} (\history{i}{t}, \state{i}{t+1}) \bigg\}^2 }_{ \triangleq ( U_{t}^{(i)} )^2 } \\
    + 2 \underbrace{ \bigg\{ \sum_{t'=1}^{t-1} \WW{i}{2:t'}(\betastar{}, \betahat{}) \tilde{f}_{t'} (\history{i}{t'}, \state{i}{t'+1}) \bigg\} \WW{i}{2:t}(\betastar{}, \betahat{}) \tilde{f}_{t} (\history{i}{t}, \state{i}{t+1})  }_{ \triangleq U_{1:t-1}^{(i)} U_{t}^{(i)} }
\end{multline}

Using the above result to rewrite $\E \left[ \Xvar{i}{1:T} \big| \history{1:n}{t}, \state{1:n}{t+1} \right]^2$, we can rewrite the left-hand side of display \eqref{mclteqn:YdiffRaw3}:
\begin{equation*}
	\sum_{t=1}^{T} \frac{1}{n} \sum_{i=1}^n \bigg\{ \E \bigg[ \E \left[ \Xvar{i}{1:T} \big| \history{1:n}{t}, \state{1:n}{t+1} \right]^2 \bigg| \history{1:n}{t-1}, \state{1:n}{t} \bigg] 
	- \E \left[ \Xvar{i}{1:T} \big| \history{1:n}{t}, \state{1:n}{t+1} \right]^2 \bigg\}
\end{equation*}
\begin{multline*}
	= \sum_{t=1}^{T} \frac{1}{n} \sum_{i=1}^n \bigg\{ \E \left[ (U_{1:t-1}^{(i)})^2 + (U_{t}^{(i)})^2 + 2 U_{1:t-1} U_{t}^{(i)} \big| \history{1:n}{t-1}, \state{1:n}{t} \right] \\
	- (U_{1:t-1}^{(i)})^2 - (U_{t}^{(i)})^2 - 2 U_{1:t-1}^{(i)} U_{t}^{(i)} \bigg\}
\end{multline*}

Note that $(U_{1:t-1}^{(i)})^2$ is a constant given $\history{1:n}{t-1}, \state{1:n}{t}$ and cancel out in the display above. Thus,
\begin{multline}
    \label{app:CLTtwoUs}
	= \frac{1}{n} 2 \sum_{i=1}^n \bigg\{ \E \left[ U_{1:t-1}^{(i)} U_{t}^{(i)} \big| \history{1:n}{t-1}, \state{1:n}{t} \right] 
	- U_{1:t-1}^{(i)} U_{t}^{(i)} \bigg\} \\
	+ \frac{1}{n} \sum_{i=1}^n \bigg\{ \E \left[ (U_{t}^{(i)})^2 \big| \history{1:n}{t-1}, \state{1:n}{t} \right] 
	- (U_{t}^{(i)})^2 \bigg\}
\end{multline}

Before we show that display \eqref{app:CLTtwoUs} converges in probability to zero, first note the following useful results:
\begin{itemize}
    \item We first show that for all $t' \in [1 \colon T]$,
    \begin{equation}
        \label{app:CLTftilde2alpha}
        \Estar{2:t'}\left[ \big| \tilde{f}_{t'} (\history{i}{t'}, \state{i}{t'+1}) \big|^{2+\alpha} \right] < \infty.
    \end{equation}
    For $t' \in [1 \colon t-1]$, $\tilde{f}_{t'} (\history{i}{t'}, \state{i}{t'+1}) = f_{t'} (\history{i}{t'})$. Since $\Estar{2:t'} \big[ | f_{t'} (\history{i}{t'}) |^{2+\alpha} \big] < \infty$ by assumption.

    For the $t'=t$ case, recall $\tilde{f}_{t}(\history{i}{t}, \state{i}{t+1}) = f_{t}(\history{i}{t}) + \sum_{t'=t+1}^T \Estar{t+1:T} \left[ f_{t'}(\history{i}{t'}) \big| \history{i}{t}, \state{i}{t+1} \right]$. By repeatedly applying Lemma \ref{lemma:binomialBound} (Inequality Using Binomial Theorem) for any numbers $a_1, a_2, ..., a_K$, $| \sum_{k=1}^K a_k |^{2+\alpha} \leq c_{2+\alpha}^K \sum_{k=1}^K | a_k |^{2+\alpha}$ for some constant $c_{2+\alpha} > 0$.
    Thus,
    \begin{multline*}
        \Estar{2:t} \bigg[ \bigg| f_{t}(\history{i}{t}) + \sum_{t'=t+1}^T \Estar{t+1:T} \left[ f_{t'}(\history{i}{t'}) \big| \history{i}{t}, \state{i}{t+1} \right] \bigg|^{2+\alpha} \bigg] \\
        \leq c_{2+\alpha}^{T-t} \Estar{2:t} \bigg[ \big| f_{t}(\history{i}{t}) \big|^{2+\alpha} + \sum_{t'=t+1}^T \bigg| \Estar{t+1:T} \left[ f_{t'}(\history{i}{t'}) \big| \history{i}{t}, \state{i}{t+1} \right] \bigg|^{2+\alpha} \bigg] \\
        \leq c_{2+\alpha}^{T-t} \sum_{t'=t}^T \Estar{2:T} \left[ \big| f_{t'}(\history{i}{t'}) \big|^{2+\alpha} \right] < \infty.
    \end{multline*}
    The second to last inequality above holds by Jensen's inequality. The last inequality holds since $\Estar{2:t'} \big[ | f_{t'} (\history{i}{t'}) |^{2+\alpha} \big] < \infty$ by assumption.
    %%%%%%%%%%%%%%%%%%%%%%%%%%%%%%%%%%%%
    \item We now show that for all $t', s \in [1 \colon T]$,
    \begin{equation}
        \label{app:CLTftildeProduct}
        \Estar{2:T}\left[ \big| \tilde{f}_{t'} (\history{i}{t'}, \state{i}{t'+1}) \tilde{f}_{s} (\history{i}{s}, \state{i}{s+1}) \big|^{1+\alpha/2} \right] < \infty.
    \end{equation}
    Note that for any real numbers $a, b$, that $ab \leq \frac{1}{2} (a^2 + b^2)$. Thus,
    \begin{multline*}
        \Estar{2:T}\left[ \big| \tilde{f}_{t'} (\history{i}{t'}, \state{i}{t'+1}) \tilde{f}_{s} (\history{i}{s}, \state{i}{s+1}) \big|^{1+\alpha/2} \right] \\
        \leq \frac{1}{2} \Estar{2:T}\left[ \big| \tilde{f}_{t'} (\history{i}{t'}, \state{i}{t'+1}) \big|^{2+\alpha} 
        + \big| \tilde{f}_{s} (\history{i}{s}, \state{i}{s+1}) \big|^{2+\alpha} \right] < \infty.
    \end{multline*}
    The last inequality holds by display \eqref{app:CLTftilde2alpha}.
\end{itemize}

\noindent We now show that $\tilde{f}_{t'} (\history{i}{t'}, \state{i}{t'+1}) = O_P(1)$. 
\begin{itemize}
    \item For $t' \in [1 \colon t-1]$, $\tilde{f}_{t'} (\history{i}{t'}, \state{i}{t'+1}) = f_{t'} (\history{i}{t'})$ by definition. Since $\Estar{2:t'} \big[ | f_{t'} (\history{i}{t'}) | \big] < \infty$ by assumption, by Lemma \ref{lemma:OpOne}, $\tilde{f}_{t'}(\history{i}{t'}, \state{i}{t'+1}) = f_{t'} (\history{i}{t'}) = O_P(1)$.
    %%%%%%%%%%%%%%%%%%
    \item Recall that $\tilde{f}_{t}(\history{i}{t}, \state{i}{t+1}) = f_{t}(\history{i}{t}) + \sum_{t'=t+1}^T \Estar{t+1:T} \left[ f_{t'}(\history{i}{t'}) \big| \history{i}{t}, \state{i}{t+1} \right]$ by definition. 
    By Lemma \ref{lemma:OpOne}, $f_{t} (\history{i}{t}) = O_P(1)$ since $\Estar{2:t} \big[ | f_{t} (\history{i}{t}) | \big] < \infty$ by assumption. 
    Also note that by Jensen's inequality,
    \begin{multline*}
        \Estar{2:t} \left[ \bigg| \Estar{t+1:T} \left[ f_{t'}(\history{i}{t'}) \big| \history{i}{t}, \state{i}{t+1} \right] \bigg| \right]
        \leq \Estar{2:t} \left[  \Estar{t+1:T} \left[ \big| f_{t'}(\history{i}{t'}) \big| \big| \history{i}{t}, \state{i}{t+1} \right] \right] \\
        = \Estar{2:T} \left[ \big| f_{t'}(\history{i}{t'}) \big| \right] < \infty.
    \end{multline*}
    Thus by Lemma \ref{lemma:OpOne}, we have that $\Estar{t+1:T} \left[ f_{t'}(\history{i}{t'}) \big| \history{i}{t}, \state{i}{t+1} \right] = O_P(1)$.
    Combining the above results we have that $\tilde{f}_{t}(\history{i}{t}, \state{i}{t+1}) = O_P(1)$.
\end{itemize}

\medskip
\noindent \under{3a. First Summation in Display \eqref{app:CLTtwoUs}}%%%%%%%%%%%%%%%%%%%
\begin{equation*}
	\frac{1}{n} \sum_{i=1}^n \bigg\{ \E \left[ U_{1:t-1}^{(i)} U_{t}^{(i)} \big| \history{1:n}{t-1}, \state{1:n}{t} \right] 
	- U_{1:t-1}^{(i)} U_{t}^{(i)} \bigg\}
\end{equation*}
\begin{multline*}
	= \frac{1}{n} \sum_{i=1}^n \bigg( \E \bigg[ \bigg\{ \sum_{t'=1}^{t-1} \WW{i}{2:t'}(\betastar{}, \betahat{}) \tilde{f}_{t'} (\history{i}{t'}, \state{i}{t'+1}) \bigg\} \WW{i}{2:t}(\betastar{}, \betahat{}) \tilde{f}_{t} (\history{i}{t}, \state{i}{t+1})  \bigg| \history{1:n}{t-1}, \state{1:n}{t} \bigg] \\
	- \bigg\{ \sum_{t'=1}^{t-1} \WW{i}{2:t'}(\betastar{}, \betahat{}) \tilde{f}_{t'} (\history{i}{t'}, \state{i}{t'+1}) \bigg\} \WW{i}{2:t}(\betastar{}, \betahat{}) \tilde{f}_{t} (\history{i}{t}, \state{i}{t+1}) \bigg)
\end{multline*}
\begin{multline*}
	= \sum_{t'=1}^{t-1} \frac{1}{n} \sum_{i=1}^n \bigg( \E \bigg[ \WW{i}{2:t'}(\betastar{}, \betahat{}) \tilde{f}_{t'} (\history{i}{t'}, \state{i}{t'+1}) \WW{i}{2:t}(\betastar{}, \betahat{}) \tilde{f}_{t} (\history{i}{t}, \state{i}{t+1})  \bigg| \history{1:n}{t-1}, \state{1:n}{t} \bigg] \\
	- \WW{i}{2:t'}(\betastar{}, \betahat{}) \tilde{f}_{t'} (\history{i}{t'}, \state{i}{t'+1}) \WW{i}{2:t}(\betastar{}, \betahat{}) \tilde{f}_{t} (\history{i}{t}, \state{i}{t+1}) \bigg)
\end{multline*}

\begin{itemize}
    \item Note the following for $t' \in [1 \colon t-1]$:
    \begin{multline}
        \label{app:firstUbullets}
        \E \left[ \WW{i}{2:t'}(\betastar{}, \betahat{}) \tilde{f}_{t'} (\history{i}{t'}, \state{i}{t'+1}) \WW{i}{2:t}(\betastar{}, \betahat{}) \tilde{f}_{t} (\history{i}{t}, \state{i}{t+1})  \big| \history{1:n}{t-1}, \state{1:n}{t} \right] \\
        \underbrace{=}_{(a)} \WW{i}{2:t'}(\betastar{}, \betahat{}) \WW{i}{2:t-1}(\betastar{}, \betahat{}) \tilde{f}_{t'} (\history{i}{t'}, \state{i}{t'+1})  \E \left[ \WW{i}{t}(\betastar{}, \betahat{}) \tilde{f}_{t} (\history{i}{t}, \state{i}{t+1})  \big| \history{1:n}{t-1}, \state{1:n}{t} \right] \\
        \underbrace{=}_{(b)} \WW{i}{2:t'}(\betastar{}, \betahat{}) \WW{i}{2:t-1}(\betastar{}, \betahat{}) \tilde{f}_{t'} (\history{i}{t'}, \state{i}{t'+1}) \Estar{t} \left[ \tilde{f}_{t} (\history{i}{t}, \state{i}{t+1})  \big| \history{1:n}{t-1}, \state{1:n}{t} \right] \\
        \underbrace{=}_{(c)} \WW{i}{2:t'}(\betastar{}, \betahat{}) \WW{i}{2:t-1}(\betastar{}, \betahat{}) \tilde{f}_{t'} (\history{i}{t'}, \state{i}{t'+1}) \Estar{t} \left[ \tilde{f}_{t} (\history{i}{t}, \state{i}{t+1}) \big| \history{i}{t-1}, \state{i}{t} \right]
    \end{multline}
    \item Equality (a) above holds since $t' < t$, so $\WW{i}{2:t'}(\betastar{}, \betahat{})$ and $\tilde{f}_{t'} (\history{i}{t'}, \state{i}{t'+1})$ are constants given $\history{1:n}{t-1}$.
    \item Equality (b) above holds since the Radon-Nikodym weighting changes the policy with which actions are chosen with in the expectation.
    \item Regarding equality (c), note that in the expectation indexed by $\pistar{t}$ that conditions on $\history{1:n}{t-1}, \state{1:n}{t}$, the only thing that is integrated over is the distribution of $(\action{i}{t}, \reward{i}{t})$. Given $\history{i}{t-1}, \state{i}{t}$, when actions are selected using the policy $\pistar{t}$, the distribution of $(\action{i}{t}, \reward{i}{t})$ does not depend on the data of other users, i.e., $\history{j}{t-1}, \state{j}{t}$ for $j \not = i$.
\end{itemize}
By display \eqref{app:firstUbullets},
\begin{multline*}
	= \sum_{t'=1}^{t-1} \frac{1}{n} \sum_{i=1}^n \bigg( \WW{i}{2:t-1}(\betastar{}, \betahat{}) \WW{i}{2:t'}(\betastar{}, \betahat{}) \tilde{f}_{t'} (\history{i}{t'}, \state{i}{t'+1}) 
    \Estar{t} \left[ \tilde{f}_{t} (\history{i}{t}, \state{i}{t+1}) \big| \history{i}{t-1}, \state{i}{t} \right] \\
	- \WW{i}{2:t'}(\betastar{}, \betahat{}) \WW{i}{2:t}(\betastar{}, \betahat{}) \tilde{f}_{t'} (\history{i}{t'}, \state{i}{t'+1}) \tilde{f}_{t} (\history{i}{t}, \state{i}{t+1}) \bigg)
\end{multline*}

By display \eqref{app:CLTftilde2alpha}, Jensen's inequality, and Lemma \ref{lemma:OpOne}, we have that $\tilde{f}_{t'} (\history{i}{t'}, \state{i}{t'+1}) = O_P(1)$, $\tilde{f}_{t} (\history{i}{t}, \state{i}{t+1}) = O_P(1)$ and $\Estar{t} \left[ \tilde{f}_{t} (\history{i}{t}, \state{i}{t+1})  \big| \history{i}{t-1}, \state{i}{t} \right] = O_P(1)$.
Additionally, by display \eqref{mclteqn:Wlimit}, $\WW{i}{2:t'}(\betastar{}, \betahat{}) = 1 + O_P(1/\sqrt{n})$. Thus, 
\begin{multline*}
	= o_P(1) + \sum_{t'=1}^{t-1} \frac{1}{n} \sum_{i=1}^n \bigg( \WW{i}{2:t-1}(\betastar{}, \betahat{}) \tilde{f}_{t'} (\history{i}{t'}, \state{i}{t'+1}) \Estar{t} \left[ \tilde{f}_{t} (\history{i}{t}, \state{i}{t+1})  \big| \history{i}{t-1}, \state{i}{t} \right] \\
	- \WW{i}{2:t}(\betastar{}, \betahat{}) \tilde{f}_{t'} (\history{i}{t'}, \state{i}{t'+1}) \tilde{f}_{t} (\history{i}{t}, \state{i}{t+1}) \bigg) \Pto 0.
\end{multline*}

The final limit above holds by the Weighted Martingale Triangular Array Weak Law of Large Numbers (Theorem \ref{thm:weightedWLLN}); note we can apply Theorem \ref{thm:weightedWLLN} because we assume that Condition \ref{cond:exploration} holds and because $\Estar{2:t} \left[ \big| \tilde{f}_{t'} (\history{i}{t'}, \state{i}{t'+1}) \tilde{f}_{t} (\history{i}{t}, \state{i}{t+1}) \big|^{1+\alpha/2} \right] < \infty$ and \\
$\Estar{2:t} \left[ \left| \Estar{t} \left[ \tilde{f}_{t'} (\history{i}{t'}, \state{i}{t'+1}) \tilde{f}_{t} (\history{i}{t}, \state{i}{t+1})  \big| \history{i}{t-1}, \state{i}{t} \right] \right|^{1+\alpha/2} \right] < \infty$ by display \eqref{app:CLTftildeProduct} and Jensen's inequality. 

\bigskip
\noindent \under{3b. Second Summation in Display \eqref{app:CLTtwoUs}}%%%%%%%%%%%%%%%%%%%
\begin{equation*}
	\frac{1}{n} \sum_{i=1}^n \bigg\{ \E \left[ ( U_{t}^{(i)} )^2 \big| \history{1:n}{t-1}, \state{1:n}{t} \right] 
	- ( U_{t}^{(i)} )^2 \bigg\}
\end{equation*}
\begin{multline*}
	= \frac{1}{n} \sum_{i=1}^n \bigg( \E \left[ \left\{ \WW{i}{2:t}(\betastar{}, \betahat{}) \tilde{f}_{t} (\history{i}{t}, \state{i}{t+1}) \right\}^2 \bigg| \history{1:n}{t-1}, \state{1:n}{t} \right] \\ 
	- \left\{ \WW{i}{2:t}(\betastar{}, \betahat{}) \tilde{f}_{t} (\history{i}{t}, \state{i}{t+1}) \right\}^2 \bigg)
\end{multline*}

Since $\WW{i}{2:t-1}(\betastar{}, \betahat{})$ is a constant $\history{1:n}{t-1}$ and since by the Radon-Nikodym weighting, 
$\E \left[ \WW{i}{t}(\betastar{}, \betahat{})^2 \tilde{f}_{t} (\history{i}{t}, \state{i}{t+1})^2 \big| \history{1:n}{t-1}, \state{1:n}{t} \right] \\
= \Estar{t} \left[ \WW{i}{t}(\betastar{}, \betahat{}) \tilde{f}_{t} (\history{i}{t}, \state{i}{t+1})^2 \big| \history{1:n}{t-1}, \state{1:n}{t} \right]$,
\begin{multline*}
    %\label{app:CLTreplaceInsideExpectationPRE}
	= \frac{1}{n} \sum_{i=1}^n \bigg( \WW{i}{2:t-1}(\betastar{}, \betahat{})^2 \Estar{t} \left[ \WW{i}{t}(\betastar{}, \betahat{}) \tilde{f}_{t} (\history{i}{t}, \state{i}{t+1})^2 \big| \history{1:n}{t-1}, \state{1:n}{t} \right] \\
	- \WW{i}{2:t}(\betastar{}, \betahat{})^2 \tilde{f}_{t} (\history{i}{t}, \state{i}{t+1})^2 \bigg).
\end{multline*}

\noindent Note the following observations:
\begin{itemize}
    \item By display \eqref{mclteqn:Wlimit}, $\WW{i}{2:t}(\betastar{}, \betahat{}) = 1 + O_P(1/\sqrt{n})$. By display \eqref{app:CLTftildeProduct} and Lemma \ref{lemma:OpOne}, we have that $\tilde{f}_{t} (\history{i}{t}, \state{i}{t+1})^2 = O_P(1)$. Thus,
    \begin{equation}
        \label{mcltapp:error1}
        \WW{i}{2:t}(\betastar{}, \betahat{})^2 \tilde{f}_{t} (\history{i}{t}, \state{i}{t+1})^2 = \WW{i}{2:t}(\betastar{}, \betahat{}) \tilde{f}_{t} (\history{i}{t}, \state{i}{t+1})^2 + O_P(1/\sqrt{n}).
    \end{equation}
    %%%%%%%%%%%%%%%%%%%%%%%%%%%%%%
    \item Additionally, for now, we take as given that
    \begin{multline}
        \label{mcltapp:error2}
        \Estar{t} \left[ \WW{i}{t}(\betastar{}, \betahat{}) \tilde{f}_{t} (\history{i}{t}, \state{i}{t+1})^2 \big| \history{1:n}{t-1}, \state{1:n}{t} \right] \\
        = \Estar{t} \left[ \tilde{f}_{t} (\history{i}{t}, \state{i}{t+1})^2 \big| \history{i}{t-1}, \state{i}{t} \right] + O_P(1/\sqrt{n}).
    \end{multline}
    For now, we take as given that display \eqref{mcltapp:error2} holds; we prove this at the end of this proof.
\end{itemize}

\noindent By displays \eqref{mcltapp:error1} and \eqref{mcltapp:error2} above,
\begin{multline*}
	= \frac{1}{n} \sum_{i=1}^n \bigg( \WW{i}{2:t-1}(\betastar{}, \betahat{}) \Estar{t} \left[ \tilde{f}_{t} (\history{i}{t}, \state{i}{t+1})^2 \big| \history{i}{t-1}, \state{i}{t} \right] \\
	- \WW{i}{2:t}(\betastar{}, \betahat{}) \tilde{f}_{t} (\history{i}{t}, \state{i}{t+1})^2 + O_P(1/\sqrt{n}) \bigg)
\end{multline*}
\begin{multline*}
	= o_P(1) + \frac{1}{n} \sum_{i=1}^n \bigg( \WW{i}{2:t-1}(\betastar{}, \betahat{}) \Estar{t} \left[ \tilde{f}_{t} (\history{i}{t}, \state{i}{t+1})^2 \big| \history{i}{t-1}, \state{i}{t} \right] \\
	- \WW{i}{2:t}(\betastar{}, \betahat{}) \tilde{f}_{t} (\history{i}{t}, \state{i}{t+1})^2 \bigg) \Pto 0.
\end{multline*}
The final limit above holds by the Weighted Martingale Triangular Array Weak Law of Large Numbers (Theorem \ref{thm:weightedWLLN}); note we can apply Theorem \ref{thm:weightedWLLN} because we assume that Condition \ref{cond:exploration} holds and because by display \eqref{app:CLTftildeProduct} and Jensen's inequality, $\Estar{2:t} \left[ \big| \tilde{f}_{t} (\history{i}{t}, \state{i}{t+1})^2 \big|^{1+\alpha/2} \right] < \infty$ and $\Estar{2:t} \left[ \left| \Estar{t} \left[ \tilde{f}_{t} (\history{i}{t}, \state{i}{t+1})^2  \big| \history{i}{t-1}, \state{i}{t} \right] \right|^{1+\alpha/2} \right] < \infty$.

Now all that remains is to show that display \eqref{mcltapp:error2} holds. We do this below:
\begin{itemize}
    \item Note the following:
    \begin{multline}
        \label{WCLT:eqntmp}
        \Estar{t} \left[ \WW{i}{t}(\betastar{}, \betahat{}) \tilde{f}_{t} (\history{i}{t}, \state{i}{t+1})^2 \big| \history{1:n}{t-1}, \state{1:n}{t} \right] \\
        = \Estar{t} \left[ \tilde{f}_{t} (\history{i}{t}, \state{i}{t+1})^2 \big| \history{1:n}{t-1}, \state{1:n}{t} \right]
        + \Estar{t} \left[ \big\{ \WW{i}{t}(\betastar{}, \betahat{}) -1 \big\} \tilde{f}_{t} (\history{i}{t}, \state{i}{t+1})^2 \big| \history{1:n}{t-1}, \state{1:n}{t} \right] \\
        = \Estar{t} \left[ \tilde{f}_{t} (\history{i}{t}, \state{i}{t+1})^2 \big| \history{i}{t-1}, \state{i}{t} \right]
        + \Estar{t} \left[ \big\{ \WW{i}{t}(\betastar{}, \betahat{}) -1 \big\} \tilde{f}_{t} (\history{i}{t}, \state{i}{t+1})^2 \big| \history{1:n}{t-1}, \state{1:n}{t} \right]
    \end{multline}
    The first equality above holds by adding and subtracting $\Estar{t} \left[ \tilde{f}_{t} (\history{i}{t}, \state{i}{t+1})^2 \big| \history{1:n}{t-1}, \state{1:n}{t} \right]$.
    
    The second equality above holds because the expectation $\Estar{t} \left[ \tilde{f}_{t} (\history{i}{t}, \state{i}{t+1})^2 \big| \history{1:n}{t-1}, \state{1:n}{t} \right]$ only integrates over is the distribution of $(\action{i}{t}, \reward{i}{t})$; given $\history{i}{t-1}, \state{i}{t}$, when actions are selected using the policy $\pistar{t}$, the distribution of $(\action{i}{t}, \reward{i}{t})$ does not depend on the data of other users, i.e., $\history{j}{t-1}, \state{j}{t}$ for $j \not = i$.

    \item Now consider just the second term in the last line of display \eqref{WCLT:eqntmp} above:
    \begin{multline*}
        \Estar{t} \left[ \big\{ \WW{i}{t}(\betastar{}, \betahat{}) -1 \big\} \tilde{f}_{t} (\history{i}{t}, \state{i}{t+1})^2 \big| \history{1:n}{t-1}, \state{1:n}{t} \right] \\
        \leq \Estar{t} \left[ \big| \WW{i}{t}(\betastar{}, \betahat{}) -1 \big| \tilde{f}_{t} (\history{i}{t}, \state{i}{t+1})^2 \bigg| \history{1:n}{t-1}, \state{1:n}{t} \right] \\
        \underbrace{\leq}_{(a)} \pi_{\min}^{-2} \max_{a \in \MC{A}}\pidot{t} \big( a, \state{i}{t} \big) \Estar{t} \left[ \tilde{f}_{t} (\history{i}{t}, \state{i}{t+1})^2 \big| \history{1:n}{t-1}, \state{1:n}{t} \right] \big\| \betahat{t-1} - \beta_{t-1}^* \big\|_2  \\
        \underbrace{=}_{(b)} O_P(1/\sqrt{n}).
    \end{multline*}
    Above inequality (a) holds by display \eqref{mclteqn:WdiffInequality}; we are able to move the terms $\max_{a \in \MC{A}}\pidot{t} \big( a, \state{i}{t} \big)$ and $\big\| \betahat{t-1} - \beta_{t-1}^* \big\|_2$ out of the conditional expectation they are known given $\history{1:n}{t-1}, \state{1:n}{t}$.

    Above limit (b) holds because (i) $\betahat{t-1} = O_P(1/\sqrt{n})$ by assumption, \\
    (ii) $\Estar{t} \left[ \tilde{f}_{t} (\history{i}{t}, \state{i}{t+1})^2 \big| \history{1:n}{t-1}, \state{1:n}{t} \right] = \Estar{t} \left[ \tilde{f}_{t} (\history{i}{t}, \state{i}{t+1})^2 \big| \history{i}{t-1}, \state{i}{t} \right] = O_P(1)$ by display \eqref{app:CLTftildeProduct}, Jensen's inequality, and Lemma \ref{lemma:OpOne}, and \\
    (iii) $\max_{a \in \MC{A}}\pidot{t} \big( a, \state{i}{t} \big) = O_P(1)$; this holds by Lemma \ref{lemma:OpOne} since
    \begin{multline*}
        \Estar{2:t-1} \left[ \big| \max_{a \in \MC{A}}\pidot{t} \big( a, \state{i}{t} \big) \big| \right] 
        \underbrace{\leq}_{(a)} \sum_{a \in \MC{A}} \Estar{2:t-1} \left[ \big| \pidot{t} \big( a, \state{i}{t} \big) \big| \right] \\
        = |\MC{A}| \sum_{a \in \MC{A}} \Estar{2:t-1} \left[ |\MC{A}|^{-1} \big| \pidot{t} \big( a, \state{i}{t} \big) \big| \right] 
        \underbrace{=}_{(b)}  |\MC{A}| \E_{\pistar{2:t-1}, \pi_t^{\TN{uniform}}} \left[ \big| \pidot{t} \big( \action{i}{t}, \state{i}{t} \big) \big| \right] \\
        \underbrace{=}_{(c)} |\MC{A}| \E_{\pistar{2:t}} \left[ \frac{\pi_t^{\TN{uniform}}(\action{i}{t}, \state{i}{t})}{\pistar{t}(\action{i}{t}, \state{i}{t})} \big| \pidot{t} \big( a, \state{i}{t} \big) \big| \right] \\
        \underbrace{\leq}_{(d)} |\MC{A}| \pi_{\min}^{-1} \Estar{2:t} \left[ \big| \pidot{t} \big( \action{i}{t}, \state{i}{t} \big) \big| \right] 
        \underbrace{\leq}_{(e)} \infty.
    \end{multline*}
    Inequality (a) above holds since the action space $\MC{A}$ is a finite set. 
    
    Equality (b) holds for $\pi_t^{\TN{uniform}}(\action{i}{t}, \state{i}{t}) \triangleq |\MC{A}|^{-1}$, i.e., the policy that selects action uniformly over the action space $\MC{A}$ for the $t^{\TN{th}}$ action.

    Equality (c) uses Radon-Nikodym derivative weights.

    Inequality (d) above holds by exploration Condition \ref{cond:exploration}.
    
    Inequality (e) holds by Condition \ref{cond:lipschitzPolicy}. $\blacksquare$
\end{itemize}

%%%%%%%%%%%%%%%%%%%%%%%%%%%%%%%%%%%%%%%%%%%%
\subsection{Functional Asymptotic Normality under Finite Bracketing Integral (Theorem \ref{thm:functionalAsymptoticNormality})} %%%%%%%%%%%%%%%%%%%%%%%%%%%%%%%%%%%%%%%%%%%%
%%%%%%%%%%%%%%%%%%%%%%%%%%%%%%%%%%%%%%%%%%%%
\label{mainapp:functionalNormality}

\begin{theorem}[Functional Asymptotic Normality under Finite Bracketing Integral for Adaptively Sampled Data]
    \label{thm:functionalAsymptoticNormality}
    Let $\F$ be any class of real-valued measurable functions of $\history{i}{t}$ such that for all $f \in \F$, for some $\alpha > 0$, $\int_{0}^1 \sqrt{ \log N_{[~]} \big( \epsilon, \F, L_{2+\alpha} ( \Pstar ) \big) } d \epsilon < \infty$.
	Let Conditions \ref{cond:exploration} (Minimum Exploration) and \ref{cond:lipschitzPolicy} (Lipschitz Policy Functions) hold and also let $\betahat{t'} - \betastar{t'} = O_P( 1/\sqrt{n} )$ for all $t' \in [1 \colon t-1]$.
	Then for 
    \begin{equation*}
        \GG_{\F}^{(n)} (f) \triangleq \frac{1}{ \sqrt{n} } \sum_{i=1}^n \left( \rhohat{i}{2:t} f( \history{i}{t} ) - \E\big[ \rhohat{i}{2:t} f( \history{i}{t} ) \big] \right),
    \end{equation*}
    the empirical process $\big\{ \GG_{\F}^{(n)} (f) : f \in \F \big\}$ converges in distribution to $\GG_{\F}$ a mean-zero Gaussian process in $\ell^\infty(\F)$ (the collection of all bounded functions from $\mathcal{F}$ to $\real$) with the following covariance function:
	 \vspace{-2mm}
	 \begin{multline}
        \label{eqn:functionalNormalityVariance}
	 	\E \big[ \GG_{\F}(f) \GG_{\F}(g) \big] \triangleq \Estar{2:t} \bigg[ \left( \rhostar{i}{2:t} f( \history{i}{t} ) - \Estar{2:t} \big[ \rhostar{i}{2:t} f( \history{i}{t} ) ] \right) \\
	 	\left( \rhostar{i}{2:t} g( \history{i}{t} ) - \Estar{2:t} \big[ \rhostar{i}{2:t} g( \history{i}{t} ) \big] \right) \bigg].
	 \end{multline}
  Above we use $\pi_{2:t}^{*,(i)} \triangleq \prod_{t'=2}^t \pistar{t'}(\action{i}{t'}, \state{i}{t'})$.
\end{theorem}

\startproof{Theorem \ref{thm:functionalAsymptoticNormality}} %%%%%%%%%%%%%%%%%%%%%%%%%%%%%%%%%%%%%
By \cite[Theorem 18.14]{van2000asymptotic}, to show the desired result it is sufficient to show that the following two properties hold: 
\begin{enumerate}[label=(\alph*)]
	\item \bo{Joint Convergence of Marginals}
	\label{fnormality:joint}	
	For any finite number of functions $f_1, f_2, ..., f_K \in \F$, 
	\begin{equation*}
		\bigg( \GG_{\F}^{(n)}(f_1), \GG_{\F}^{(n)}(f_2), ..., \GG_{\F}^{(n)}(f_K) \bigg)
		\Dto \bigg( \GG_{\F}(f_1), \GG_{\F}(f_2), ..., \GG_{\F}(f_K) \bigg)
	\end{equation*}
	\item \bo{Asymptotically Tight} 
	\label{fnormality:tight}		
	For any $\epsilon, \eta > 0$, there exists a partition of $\F$ into finitely many sets $\F_1, \F_2, ..., \F_J$ such that 
	\begin{equation*}
		\limsup_{n \to \infty} \PP^* \bigg( \sup_{j \in [1 \colon J]} \sup_{f, f' \in \F_j} \left| \GG_{\F}^{(n)}(f) - \GG_{\F}^{(n)}(f') \right| > \epsilon \bigg) \leq \eta.
	\end{equation*}
\end{enumerate}

\proofSubsection{Showing (a) Joint Convergence of Marginals} We can show that (a) above holds for the stochastic process $\big\{ \GG_{\F}^{(n)}(f) : f \in \F \big\}$ by the Theorem \ref{thm:weightedCLT} (Weighted Martingale Triangular Array Central Limit Theorem).

Specifically, by Cramer Wold device, it is sufficient to show that for any $c = [c_1, c_2, \dots, c_K ] \in \real^K$ that
\begin{equation*}
	\sum_{k=1}^K c_k \GG_{\F}^{(n)}(f_k) \Dto \N \left( 0, c^\top\begin{bmatrix}
	\Sigma_{1,1} & \Sigma_{1,2} & \dots & \Sigma_{1,K} \\
	\Sigma_{2,1} & \Sigma_{2,2} & \dots & \Sigma_{2,K} \\
	\vdots & \vdots & \ddots & \vdots \\
	\Sigma_{K,1} & \Sigma_{K,2} & \dots & \Sigma_{K,K} \\
	\end{bmatrix} c \right)
\end{equation*}
where $\Sigma_{k, k'} \triangleq \Estar{2:t} \left[ \GG_{\F}(f_k) \GG_{\F}(f_{k'})\right]$. 

Note that
\begin{equation*}
	\sum_{k=1}^K c_k \GG_{\F}^{(n)}(f_k) 
	= \frac{1}{ \sqrt{n} } \sum_{i=1}^n \bigg( \rhohat{i}{2:t} \sum_{k=1}^K c_k f_k( \history{i}{t} ) 
	- \E\bigg[ \rhohat{i}{2:t} \sum_{k=1}^K c_k f_k( \history{i}{t} ) \bigg] \bigg)
\end{equation*}
\begin{multline*}
    = \frac{1}{ \sqrt{n} } \sum_{i=1}^n \bigg( \WW{i}{2:t}\big( \betastar{}, \betahat{} \big) \rhostar{i}{2:t} \sum_{k=1}^K c_k f_k( \history{i}{t} ) \\
	- \E\bigg[ \WW{i}{2:t}\big( \betastar{}, \betahat{} \big) \rhostar{i}{2:t} \sum_{k=1}^K c_k f_k( \history{i}{t} ) \bigg] \bigg)
	\Dto \N \left( 0, \bar{\Sigma} \right),
\end{multline*}
where
\begin{equation}
    \label{eqn:functionalNormalitySigma}
    \bar{\Sigma} \triangleq \Estar{2:t} \bigg[ \bigg( \rhostar{i}{2:t} \sum_{k=1}^K c_k f_k( \history{i}{t} ) \bigg)^2 \bigg] - \E_{ \pistar{2:t} } \bigg[ \rhostar{i}{2:t} \sum_{k=1}^K c_k f_k( \history{i}{t} ) \bigg]^2.
\end{equation}
The above weak convergence holds by Theorem \ref{thm:weightedCLT} (Weighted Martingale Triangular Array Central Limit Theorem).
When applying Theorem \ref{thm:weightedCLT}, we use the following properties:
\begin{itemize}
    \item Conditions \ref{cond:exploration} (Minimum Exploration) and \ref{cond:lipschitzPolicy} (Lipschitz Policy Functions) hold.
    %%%%%%%%%%%%%%%%%%
    \item $\betahat{t'} - \betastar{t'} = O_P( 1/\sqrt{n} )$ for all $t' \in [1 \colon t-1]$.
    %%%%%%%%%%%%%%%%%%
    \item Recall that for some $\alpha > 0$, $\int_{0}^1 \sqrt{ \log N_{[~]} \big( \epsilon, \F, L_{2+\alpha} ( \Pstar ) \big) } d \epsilon < \infty$. This means that for all $f \in \F$, we can find a bracket $(l_j, u_j)$ such that $l_j(\history{i}{t}) \leq f(\history{i}{t}) \leq u_j(\history{i}{t})$ a.s. Additionally, recall that the brackets are such that $\Estar{2:t} \big[ \big| l \big( \history{i}{t} \big) \big|^{2+\alpha} \big] < \infty$ and $\Estar{2:t} \big[ \big| u \big( \history{i}{t} \big) \big|^{2+\alpha} \big] < \infty$.
    Thus, for all $f \in \F$, 
    \begin{equation*}
        \Estar{2:t} \big[ \big| f \big( \history{i}{t} \big) \big|^{2+\alpha} \big] 
        \leq \Estar{2:t} \big[ \big| l \big( \history{i}{t} \big) \big|^{2+\alpha} \big] + \Estar{2:t} \big[ \big| u \big( \history{i}{t} \big) \big|^{2+\alpha} \big]  < \infty.
    \end{equation*}

    Thus, by repeatedly applying Lemma \ref{lemma:binomialBound}, for some positive constant $c_{2+\alpha} < \infty$, the following result holds:
    \begin{multline*}
        \Estar{2:t} \bigg[ \bigg| \rhostar{i}{2:t} \sum_{k=1}^K c_k f_k( \history{i}{t} ) \bigg|^{2+\alpha} \bigg] 
        \leq c_{2+\alpha}^{K} \sum_{k=1}^K \Estar{2:t} \bigg[ \bigg| \rhostar{i}{2:t} c_k f_k( \history{i}{t} ) \bigg|^{2+\alpha} \bigg] \\
        \leq c_{2+\alpha}^{K} \sum_{k=1}^K \pi_{\min}^{(t-1)(2+\alpha)} |c_k|^{2+\alpha} \Estar{2:t} \left[ \big| f_k( \history{i}{t} ) \big|^{2+\alpha} \right] < \infty.
    \end{multline*}
    The last inequality above holds by Condition \ref{cond:exploration}.
    
    %$\Estar{2:t} \big[ \big| f_{t} \big( \history{i}{t} \big) \big|_1^{2+\alpha} \big] < \infty$ for $\alpha > 0$. This holds because we can find an envelope function $F\big( \history{i}{t} \big)$, i.e., $f_{t} \big( \history{i}{t} \big) \leq F \big( \history{i}{t} \big)$ a.s. such that $\Estar{2:t} \big[ \big| F_{t} \big( \history{i}{t} \big) \big|_1^{2+\alpha} \big] < \infty$; do this by taking the supremum of all finitely many the upper and lower bracketing functions.
\end{itemize}

\medskip
\noindent By the definition of $\bar{\Sigma}$ from display \eqref{eqn:functionalNormalitySigma}, 
\begin{equation*}
    \bar{\Sigma} = \Estar{2:t} \bigg[ \bigg( \rhostar{i}{2:t} \sum_{k=1}^K c_k f_k( \history{i}{t} ) \bigg)^2 \bigg] - \E_{ \pistar{2:t} } \bigg[ \rhostar{i}{2:t} \sum_{k=1}^K c_k f_k( \history{i}{t} ) \bigg]^2
\end{equation*}
\begin{multline*}
    = \sum_{k=1}^K \sum_{k'=1}^K \bigg\{ \Estar{2:t} \bigg[ \rhostar{i}{2:t} c_k f_k( \history{i}{t} )
    \rhostar{i}{2:t} c_{k'} f_{k'}( \history{i}{t} ) \bigg] \\
    - \E_{ \pistar{2:t} } \left[ \rhostar{i}{2:t} c_k f_k( \history{i}{t} ) \right] \E_{ \pistar{2:t} } \left[ \rhostar{i}{2:t} c_{k'} f_{k'}( \history{i}{t} ) \right] \bigg\}
\end{multline*}
\begin{multline*}
    = \sum_{k=1}^K \sum_{k'=1}^K c_k c_{k'
    } \Estar{2:t} \bigg[ \bigg( \rhostar{i}{2:t} f_k( \history{i}{t} )
    - \E_{ \pistar{2:t} } \left[ \rhostar{i}{2:t} f_k( \history{i}{t} ) \right] \bigg) \\
    \bigg( \rhostar{i}{2:t} f_{k'}( \history{i}{t} )
    - \E_{ \pistar{2:t} } \left[ \rhostar{i}{2:t} f_{k'}( \history{i}{t} ) \right] \bigg) \bigg]
\end{multline*}
By the definition of $\E \big[ \GG_{\F}(f_k) \GG_{\F}(f_{k'}) \big]$ from display \eqref{eqn:functionalNormalityVariance},
\begin{equation*}
    = \sum_{k=1}^K \sum_{k'=1}^K c_k c_{k'
    } \E \big[ \GG_{\F}(f_k) \GG_{\F}(f_{k'}) \big]
    = \sum_{k=1}^K \sum_{k'=1}^K c_k c_{k'
    } \Sigma_{k,k'}
    = c^\top \begin{bmatrix}
	\Sigma_{1,1} & \Sigma_{1,2} & \dots & \Sigma_{1,K} \\
	\Sigma_{2,1} & \Sigma_{2,2} & \dots & \Sigma_{2,K} \\
	\vdots & \vdots & \ddots & \vdots \\
	\Sigma_{K,1} & \Sigma_{K,2} & \dots & \Sigma_{K,K} \\
	\end{bmatrix} c.
\end{equation*}
The second equality above holds since recall $\Sigma_{k,k'} \triangleq \E \big[ \GG_{\F}(f_k) \GG_{\F}(f_{k'}) \big]$.

Thus, we have shown that $\bar{\Sigma}$ from display \eqref{eqn:functionalNormalitySigma} is such that
\begin{equation*}
	\bar{\Sigma} = c^\top \begin{bmatrix}
	\Sigma_{1,1} & \Sigma_{1,2} & \dots & \Sigma_{1,K} \\
	\Sigma_{2,1} & \Sigma_{2,2} & \dots & \Sigma_{2,K} \\
	\vdots & \vdots & \ddots & \vdots \\
	\Sigma_{K,1} & \Sigma_{K,2} & \dots & \Sigma_{K,K} \\
	\end{bmatrix} c.
\end{equation*}

\proofSubsection{Showing (b) Asymptotically Tight} The asymptotically tight condition above holds by the same argument used in the proof of Theorem 19.5 from \cite{van2000asymptotic}, but by replacing the use of maximal inequality Lemma 19.34 of \cite{van2000asymptotic} in that proof with our maximal inequality from Lemma \ref{lemma:maximalBracketing} (Maximal Inequality as a Function of the Bracketing Integral). We discuss this argument below.

Let $\epsilon, \delta > 0$. Since $\F$ has finite bracketing integral by assumption, we can find a partition of $\F$ into finitely many sets $\F_1, \F_2, ..., \F_{N_{\delta_0}}$ where $N_{\delta_0} \triangleq N_{[~]} \big( \delta_0, \F, L_{2+\alpha} ( \Pstar ) \big)$ and $\delta_0 \triangleq \delta \sqrt{ \pi_{\min}^{-(t-1)} }$. Note that
\begin{equation*}
	%\limsup_{n \to \infty} 
    \PP^* \bigg( \sup_{j \in [1 \colon N_{\delta_0}]} \sup_{f, f' \in \F_j} \left| \GG_{\F}^{(n)}(f) - \GG_{\F}^{(n)}(f') \right| > \epsilon \bigg)
\end{equation*}
Above $\PP^*$ refers to outer probabilities as defined in Section 18.2 \cite{van2000asymptotic}.
\begin{equation*}
    = \PP^* \bigg( \sup_{j \in [1 \colon N_{\delta_0}]} \sup_{f, f' \in \F_j} \left| \GG_{\F}^{(n)}(f - f') \right| > \epsilon \bigg)
\end{equation*}
By Markov inequality,
\begin{equation*}
	\leq \frac{1}{\epsilon} \E^* \bigg[ \sup_{j \in [1 \colon N_{\delta_0}]} \sup_{f, f' \in \F_j} \left| \GG_{\F}^{(n)}(f - f') \right| \bigg].
\end{equation*}

\noindent Let $\MC{G}_\delta$ be the function class such that $\MC{G}_\delta \triangleq \big\{ f - f' \TN{~~s.t.~~} f, f' \in \F_j, j \in [1 \colon N_{\delta_0}] \big\}$. Note the following observations:
\begin{itemize}
    \item Note that for any $f, f' \in \F_j$, $\Estar{2:t} \left[ \big\{ f(\history{i}{t}) - f'(\history{i}{t}) \big\}^2 \right] \leq \delta_0^2$; thus $\Estar{2:t} \big[ g(\history{i}{t})^2 \big] \leq \delta_0^2$ for any $g \in \MC{G}_\delta$.  Note that by Condition \ref{cond:exploration}, this implies that $\Estar{2:t}\left[ \rhostar{i}{2:t} g(\history{i}{t})^2 \right] 
    \leq \pi_{\min}^{-(t-1)} \Estar{2:t}\left[ g(\history{i}{t})^2 \right]
    \leq \pi_{\min}^{-(t-1)} \delta_0^2 = \delta^2$ for all $g \in \MC{G}_\delta$
    %%%%%%%%%%%
    \item We take as given for now that 
    \begin{equation}
        \label{appFunctional:bracketingIntegral}
        \int_{0}^1 \sqrt{ \log N_{[~]} \big( \epsilon, \MC{G}_\delta, L_2 \big( \Pstar \big) \big) } d \epsilon < \infty.
    \end{equation}
    We show the above holds at the end of this proof.
    %%%%%%%%%%%%%
    \item We also take as given that there exists a non-negative envelope function $G$ where $\big| g(\history{i}{t}) \big| \leq G(\history{i}{t})$ a.s. for all $g \in \MC{G}_\delta$ and $\Estar{2:t} \left[ G(\history{i}{t})^2 \right] < \infty$ (we show this at the end of this proof).
\end{itemize}
Using the above observations and Condition \ref{cond:exploration}, we can apply Lemma \ref{lemma:maximalBracketing} (Maximal Inequality as a Function of the Bracketing Integral) to get that 
\begin{equation}
	\label{eqn:maximalBracketingApply}
	\lesssim \frac{1}{ \epsilon } \bigg\{ \int_{0}^\delta \sqrt{ \log N_{[~]} \big( \eta, \MC{G}_\delta, L_2 \big( \Pstar \big) \big) } d \eta 
	+ \sqrt{n} \Estar{2:t} \left[ \rhostar{i}{2:t} G(\history{i}{t}) \II_{ G(\history{i}{t}) > \sqrt{n} a(\delta) }\right] \bigg\},
\end{equation}
where $a(\delta) \triangleq \delta / \sqrt{ \log N_{[~]} \big( \delta, \G_\delta, L_{2} ( \Pstar ) \big) }$. Above $\lesssim$ means less than or equal to when scaled by universal positive constants.

\begin{itemize}
    \item By display \eqref{appFunctional:bracketingIntegral}, the first term (integral term) in display \eqref{eqn:maximalBracketingApply} converges to zero as $\delta \to 0$.
    %%%%%%%%%%%%%%%%%%%
    \item Regarding the second term in display \eqref{eqn:maximalBracketingApply}, since $\II_{ G(\history{i}{t}) > \sqrt{n} a(\delta) } = 1$ implies that $G(\history{i}{t}) \big\{ \sqrt{n} a(\delta) \big\}^{-1} > 1$, thus, $\big| G(\history{i}{t}) \big\{ \sqrt{n} a(\delta) \big\}^{-1} \big| \geq \II_{ G(\history{i}{t}) > \sqrt{n} a(\delta) }$.
    \begin{multline*}
        \sqrt{n} \Estar{2:t} \left[ \rhostar{i}{2:t} G(\history{i}{t}) \II_{ G(\history{i}{t}) > \sqrt{n} a(\delta) }\right] 
        \leq a(\delta)^{-1} \Estar{2:t} \left[ \rhostar{i}{2:t} G(\history{i}{t})^2 \right] \\
        \leq a(\delta)^{-1} \pi_{\min}^{-(t-1)} \Estar{2:t} \left[ G(\history{i}{t})^2 \right].
    \end{multline*}
    The last inequality above holds by Condition \ref{cond:exploration}.
    The above goes to zero as $n \to \infty$ for every fixed $\delta$.
\end{itemize}
Thus, we have that display \eqref{eqn:maximalBracketingApply} converges to zero; this is sufficient for the Theorem to hold.
All that remains is to show that display \eqref{appFunctional:bracketingIntegral} holds and that we can find an envelope function $G$. We do this below.

%%%%%%%%%%%%%%%%%%%%%%%%%%%%%%%%%%%%%%
\proofSubsubsection{Bracketing Functions for $\MC{G}_\delta$; display \eqref{appFunctional:bracketingIntegral}}
Let $\eta > 0$. Since $\F$ has finite bracketing integral by assumption, we can find  $N_{\eta} \triangleq N_{[~]} \big( \eta, \F, L_{2+\alpha} ( \Pstar ) \big)$ bracketing functions $\big\{ (l_k, u_k) \big\}_{k=1}^{N_{\eta}}$. Note that for any $g \in \MC{G}_\delta$, we can find some $f, f' \in \F$ such that $g = f - f'$. 

We will show that the brackets $\big\{ (l_k - u_{k'}, ~ u_k - l_{k'}) \big\}_{k =1; k'=1}^{k=N_\eta; k' = N_\eta}$ will cover $\MC{G}_\delta$ and be of size $2 \sqrt{c_2} \eta$ in $L_{2+\alpha} ( \Pstar )$ norm for a positive constant $c_2 < \infty$.

\medskip
\noindent \textit{Covering:} We can find brackets $(l_k, u_k)$ and $(l_{k'}, u_{k'})$ such that $l_k(\history{i}{t}) \leq f(\history{i}{t}) \leq u_k(\history{i}{t})$ a.s. and $l_{k'}(\history{i}{t}) \leq f'(\history{i}{t}) \leq u_{k'}(\history{i}{t})$ a.s. Thus, $l_k(\history{i}{t}) - u_{k'}(\history{i}{t}) \leq f(\history{i}{t}) - f'(\history{i}{t}) \leq u_k(\history{i}{t}) - l_{k'}(\history{i}{t})$ a.s. 

\medskip
\noindent \textit{Size:} Note that 
\begin{equation*}
    \sqrt{ \Estar{2:t} \big[ \big| u_k(\history{i}{t}) - l_{k'}(\history{i}{t}) - l_k(\history{i}{t}) + u_{k'}(\history{i}{t}) \big|^2 \big] }
\end{equation*}
By Lemma \ref{lemma:binomialBound} for some positive constant $c_{2} < \infty$,
\begin{equation*}
    \leq \sqrt{ c_2 \Estar{2:t} \big[ \big| u_k(\history{i}{t}) - l_k(\history{i}{t}) \big|^2 \big]
    + c_2 \Estar{2:t} \big[ \big| u_{k'}(\history{i}{t}) - l_{k'}(\history{i}{t}) \big|^2 \big] }
\end{equation*}
Since $\sqrt{a+b} \leq \sqrt{a} + \sqrt{b}$ for $a,b > 0$ (to see this square both sides),
\begin{equation*}
    \leq \sqrt{ c_2 \Estar{2:t} \big[ \big| u_k(\history{i}{t}) - l_k(\history{i}{t}) \big|^2 \big] }
    + \sqrt{ c_2 \Estar{2:t} \big[ \big| u_{k'}(\history{i}{t}) - l_{k'}(\history{i}{t}) \big|^2 \big] }
\end{equation*}
\begin{equation*}
    \leq 2 \sqrt{c_2} \eta.
\end{equation*}

We now discuss why the last inequality above holds. By construction of our bracketing functions, $\Estar{2:t} \big[ \big| u_k(\history{i}{t}) - l_k(\history{i}{t}) \big|^{2+\alpha} \big]^{1/(2+\alpha)} \leq \eta$. Note that $h(x) = x^{1+\alpha/2}$ is convex. By Jensen's inequality,
\begin{multline*}
    \Estar{2:t} \left[ \big| u_k(\history{i}{t}) - l_k(\history{i}{t}) \big|^{2} \right]^{1+\alpha/2} 
    = h \left( \Estar{2:t} \big[ \big| u_k(\history{i}{t}) - l_k(\history{i}{t}) \big|^{2} \big] \right) \\
    \leq \Estar{2:t} \left[ h(\big| u_k(\history{i}{t}) - l_k(\history{i}{t}) \big|^2 \big) \right]
    = \Estar{2:t} \left[ \big| u_k(\history{i}{t}) - l_k(\history{i}{t}) \big|^{2+\alpha} \right].
\end{multline*}
Thus, we have that
\begin{equation*}
    \Estar{2:t} \left[ \big| u_k(\history{i}{t}) - l_k(\history{i}{t}) \big|^{2} \right] 
    \leq \Estar{2:t} \left[ \big| u_k(\history{i}{t}) - l_k(\history{i}{t}) \big|^{2+\alpha} \right]^{1/(1+\alpha/2)}
\end{equation*}
By taking the square root of both sides,
\begin{equation*}
    \Estar{2:t} \left[ \big| u_k(\history{i}{t}) - l_k(\history{i}{t}) \big|^{2} \right]^{1/2}
    \leq \Estar{2:t} \left[ \big| u_k(\history{i}{t}) - l_k(\history{i}{t}) \big|^{2+\alpha} \right]^{1/(2+\alpha)} \leq \eta.
\end{equation*}

\medskip
\noindent \textit{Bracketing Number:} By the above results we have that
\begin{equation*}
    N_{[~]} \left( 2 \sqrt{c_2} \eta, \MC{G}_\delta, L_2 \big( \Pstar \big) \right)
    \leq N_{[~]} \left( \eta, \F, L_{2+\alpha} \big( \Pstar \big) \right)^2.
\end{equation*}
Moreover,
\begin{equation*}
    N_{[~]} \big( \eta, \MC{G}, L_2 \big( \Pstar \big) \big)
    \leq N_{[~]} \big( \eta / (2 \sqrt{c_2}), \F, L_{2+\alpha} \big( \Pstar \big) \big)^2.
\end{equation*}
Thus,
\begin{equation*}
    \int_0^1 \sqrt{ \log N_{[~]} \big( \eta, \MC{G}, L_2 \big( \Pstar \big) \big) } d \eta
    \leq \int_0^1 \sqrt{ \log N_{[~]} \big( \eta / (2 \sqrt{c_2}), \F, L_{2+\alpha} \big( \Pstar \big) \big)^2 } d \eta
\end{equation*}
By exponent property of $\log$,
\begin{equation*}
    = \sqrt{2} \int_0^1 \sqrt{ \log N_{[~]} \big( \eta / (2 \sqrt{c_2}), \F, L_{2+\alpha} \big( \Pstar \big) \big) } d \eta
\end{equation*}
We now use integration by substitution, with $u = \eta / (2 \sqrt{c_2})$; note that $\fracpartial{u}{\eta} = (2 \sqrt{c_2})^{-1}$.
\begin{equation*}
    = \sqrt{2} (2 \sqrt{c_2}) \int_0^1 \sqrt{ \log N_{[~]} \big( \eta / (2 \sqrt{c_2}), \F, L_{2+\alpha} \big( \Pstar \big) \big) } (2 \sqrt{c_2})^{-1} d \eta
\end{equation*}
\begin{equation*}
    = \sqrt{2} (2 \sqrt{c_2}) \int_0^{(2 \sqrt{c_2})^{-1}} \sqrt{ \log N_{[~]} \big( u, \F, L_{2+\alpha} \big( \Pstar \big) \big) } du < \infty.
\end{equation*}
The above is bounded by our assumption that $\int_0^1 \sqrt{ \log N_{[~]} \big( u, \F, L_{2+\alpha} \big( \Pstar \big) \big) } du$. Note that if $(2 \sqrt{c_2})^{-1} \leq 1$, the result above holds directly by this assumption. If $(2 \sqrt{c_2})^{-1} > 1$, the result above holds because $ N_{[~]} \big( \eta, \F, L_{2+\alpha} \big( \Pstar \big) \big) \leq  N_{[~]} \big( 1, \F, L_{2+\alpha} \big( \Pstar \big) \big)$ for all $\eta > 1$.

\bigskip
\noindent \noindent \under{Envelope Function for $\MC{G}_\delta$:} We can construct the envelope function for $\MC{G}_\delta$ using the brackets for $\MC{G}_\delta$ that we constructed above. Specifically, the envelope function $G$ can be taken to be the supremum of the upper and lower bracketing functions for $\MC{G}_\delta$. This envelope function will be such that $\Estar{2:t} \left[ G(\history{i}{t})^2 \right] < \infty$ since the brackets we constructed for $\MC{G}_\delta$ have finite $L_{2+\alpha} \big( \Pstar \big)$ norm (see Section \ref{sec:bracketingNumberDef} for more on the definition of bracketing functions we use). ~~~$\blacksquare$

%%%%%%%%%%%%%%%%%%%%%%%%%%%%%%%%%%%%%%%%%%%%%%%%%%%%%%%%%%%%%%%%%%%%%%%%
%%%%%%%%%%%%%%%%%%%%%%%%%%%%%%%%%%%%%%%%%%%%%%%%%%%%%%%%%%%%%%%%%%%%%%%%

\subsection{Stochastic Equicontinuity (Lemma \ref{lemma:stochasticEquicontinuity})} 
\label{mainapp:stochasticEquicontinuity}

\begin{lemma}[Stochastic Equicontinuity]
    \label{lemma:stochasticEquicontinuity}
    Let $t \in [1 \colon T-1]$ and let $\F$ be a class of real-valued, measurable functions of $\history{i}{t}$ such that $\int_0^1 \sqrt{ \log N_{[~]} \left( \epsilon, \F, L_{2+\alpha}(\Pstar) \right) } d \epsilon < \infty$ for some constant $\alpha > 0$. Let $\hat{f}^{(n)} \in \F$ be a sequence of functions such that $\nu \big( \hat{f}^{(n)}, f_0 \big) \Pto 0$ for some $f_0 \in \F$ where $\nu(f, g) \triangleq \Estar{2:t} \left[ \big| f\big(\history{i}{t} \big) - g\big(\history{i}{t} \big) \big|^2 \right]^{1/2}$.
    
    Under Conditions \ref{cond:exploration} and \ref{cond:lipschitzPolicy}, and the condition that $\betahat{1:t-1} - \betastar{1:t-1} = O_P(1/\sqrt{n})$ we have that
    \begin{equation}
        \label{equicont:mainResult}
        \GG_{\F}^{(n)} \big( \hat{f}^{(n)} \big) - \GG_{\F}^{(n)} \big( f_0 \big) \Pto 0
    \end{equation}
    where
    \begin{equation*}
        \GG_{\F}^{(n)} (f) \triangleq \frac{1}{ \sqrt{n} } \sum_{i=1}^n \left( \rhohat{i}{2:t} f( \history{i}{t} ) - \E\big[ \rhohat{i}{2:t} f( \history{i}{t} ) \big] \right).
    \end{equation*}
\end{lemma}

\startproof{Lemma \ref{lemma:stochasticEquicontinuity}}
We use an argument akin to that of Lemma 19.24 of \cite{van1996weak}, which is for i.i.d. data. We use $\ell^{\infty}(\F)$ to refer to the collection of all bounded functions from $\F$ to $\real$.

By Theorem \ref{thm:functionalAsymptoticNormality} (Functional Asymptotic Normality under Finite Bracketing Integral), the following stochastic process converges weakly to a mean-zero Gaussian Process $\GG_\F \in \ell^{\infty}(\F)$:
\begin{equation}
    \label{mainapp:weakConvergence}
    \GG_{\F}^{(n)} \triangleq \left\{ \GG_{\F}^{(n)}(f) \TN{~~s.t.~~} f \in \F \right\}
    \Dto \GG_\F,
\end{equation}
where the limit $\GG_\F$ has the following covariance function:
\begin{multline}
    \label{equicont:Gcov}
    \E \big[ \GG_{\F}(f) \GG_{\F}(g) \big] \triangleq \Estar{2:t} \bigg[ \left( \rhostar{i}{2:t} f( \history{i}{t} ) - \Estar{2:t} \big[ \rhostar{i}{2:t} f( \history{i}{t} ) ] \right) \\
    \left( \rhostar{i}{2:t} g( \history{i}{t} ) - \Estar{2:t} \big[ \rhostar{i}{2:t} g( \history{i}{t} ) \big] \right) \bigg].
\end{multline}
We are able to apply Theorem \ref{thm:functionalAsymptoticNormality} because of our assumptions that Conditions \ref{cond:exploration} and \ref{cond:lipschitzPolicy} hold, and since we've assumed that $\int_0^1 \sqrt{ \log N_{[~]} \left( \epsilon, \F, L_{2+\alpha}(\Pstar) \right) } d \epsilon < \infty$ and $\betahat{1:t-1} - \betastar{1:t-1} = O_P( 1/\sqrt{n} )$.

By Lemma 18.15 of \cite{van2000asymptotic}, the weak convergence result from display \eqref{mainapp:weakConvergence} implies that the limit $\GG_{\F}$ can be constructed to have almost all sample paths in $\TN{UC}(\F, \rho)$, the collection of all uniformly continuous functions from $\F$ to $\real$; $\rho$ is the standard deviation semi-metric:
\begin{equation*}
    \rho(f, g) \triangleq \sqrt{\Estar{2:t} \left[ \big| \GG_\F(f) - \GG_\F(g) \big|^2 \right]} 
\end{equation*}

For now, we take as given that $\rho \big( \hat{f}^{(n)}, f_0 \big) \Pto 0$ (we show this at the end of this proof). Thus, we have that $\hat{f}^{(n)} \Pto f_0$. By Slutsky's theorem and the convergence result from display \eqref{mainapp:weakConvergence}, we have that $\big( \GG_\F^{(n)}, \hat{f}^{(n)} \big) \Dto \big( \GG_\F, f_0 \big)$. 

Consider the evaluation function $g: \ell^{\infty}(\F) \by \F \mapsto \real$ where $g(\GG, f) \triangleq \GG(f) - \GG(f_0)$. Note that the evaluation mapping $g$ is continuous at $\big( z, f \big) \in \ell^{\infty}(\F) \by \F$ if $z$ is continuous at $f$; this is discussed in the proof of Lemma 18.15 of \cite{van2000asymptotic}. Since, as discussed earlier, $\GG_{\F}$ can be constructed to have almost all sample paths in $\TN{UC}(\F, \rho)$, thus $\GG_\F$ is at $f_0$ for almost all sample paths and the evaluation mapping $g$ is continuous at  $\big( \GG_\F, f_0 \big)$ for almost all sample paths.

Thus, by the continuous mapping theorem, we have that
\begin{equation}
    \label{equicont:CMT}
    \GG_\F^{(n)}\big(  \hat{f}^{(n)} \big) - \GG_0 \big( f_0 \big)
    = g\big( \GG_\F^{(n)},  \hat{f}^{(n)} \big)
    \Dto g \big( \GG_\F, f_0 \big)
    = \GG_\F \big( f_0 \big) - \GG_\F \big( f_0 \big) = 0.
\end{equation}
Since convergence in distribution to a constant implies convergence in probability, display \eqref{equicont:CMT} above implies the main result display \eqref{equicont:mainResult} holds. 

\proofSubsection{We now show that $\rho \big( \hat{f}^{(n)}, f_0 \big) \Pto 0$} 
Note that for any functions $f, g \in \F$,
\begin{equation*}
    \rho(f, g)^2 = \Estar{2:t} \left[ \big| \GG_\F(f) - \GG_\F(g) \big|^2 \right]
\end{equation*}
\begin{equation*}
    = \Estar{2:t} \left[ \GG_\F(f)^2 - 2 \GG_\F(f) \GG_\F(g) + \GG_\F(g)^2 \right]
\end{equation*}
Using the covariance expression from display \eqref{equicont:Gcov}, we have that for \\
$G(\history{i}{t}; f) \triangleq \rhostar{i}{2:t} f( \history{i}{t} ) - \Estar{2:t} \big[ \rhostar{i}{2:t} f( \history{i}{t} ) ]$,
\begin{equation*}
    = \Estar{2:t} \left[ G(\history{i}{t}; f)^2 
    - 2 G(\history{i}{t}; f) G(\history{i}{t}; g) 
    + G(\history{i}{t}; g)^2 \right]
\end{equation*}
\begin{equation*}
    = \Estar{2:t} \left[ \left\{ G(\history{i}{t}; f) 
    - G(\history{i}{t}; g) \right\}^2 \right]
\end{equation*}
By the definition of $G(\history{i}{t}; f)$,
\begin{multline*}
    = \Estar{2:t} \bigg[ \bigg\{ \left( \rhostar{i}{2:t} f( \history{i}{t} ) - \Estar{2:t} \big[ \rhostar{i}{2:t} f( \history{i}{t} ) ] \right) \\
    - \left( \rhostar{i}{2:t} g( \history{i}{t} ) - \Estar{2:t} \big[ \rhostar{i}{2:t} g( \history{i}{t} ) ] \right) \bigg\}^2 \bigg]
\end{multline*}
Let $X = \rhostar{i}{2:t} f( \history{i}{t} ) - \rhostar{i}{2:t} g( \history{i}{t} )$. Since $\Estar{2:t} \big[ ( X - \Estar{2:t}[X] )^2 \big] = \Estar{2:t} \big[ X^2 \big] - \Estar{2:t}\big[X\big]^2$, 
\begin{multline*}
    = \Estar{2:t} \bigg[ \left( \rhostar{i}{2:t} f( \history{i}{t} ) - \rhostar{i}{2:t} g( \history{i}{t} ) \right)^2 \bigg] \\
    - \Estar{2:t} \left[ \rhostar{i}{2:t} g( \history{i}{t} ) - \rhostar{i}{2:t} f( \history{i}{t} ) \right]^2 
\end{multline*}
\begin{equation*}
    \leq \Estar{2:t} \bigg[ \big\{ \pi_{2:t}^{*,(i)} \big\}^{-2} \left\{ f( \history{i}{t} ) -  g( \history{i}{t} ) \right\}^2 \bigg]
\end{equation*}
By Condition \ref{cond:exploration}, $\big\{ \pi_{2:t}^{*,(i)} \big\}^{-2} \leq \pi_{\min}^{2(t-1)}$ a.s., so
\begin{equation*}
    \leq \pi_{\min}^{2(t-1)} \Estar{2:t} \left[ \left\{ f( \history{i}{t} ) -  g( \history{i}{t} ) \right\}^2 \right]
\end{equation*}
The above implies that $\rho \big( \hat{f}^{(n)}, f_0 \big) \Pto 0$ because by assumption of the Lemma, we have that $\nu \big( \hat{f}^{(n)}, f_0 \big) \Pto 0$ where $\nu(f, g) \triangleq \Estar{2:t} \left[ \big|f\big(\history{i}{t} \big) - g\big(\history{i}{t} \big) \big|^2 \right]^{1/2}$. $~~~\blacksquare$

\clearpage
%%%%%%%%%%%%%%%%%%%%%%%%%%%%%%%%%%%%%%%%%%%%
\section{Maximal Inequalities for Adaptively Sampled Data} %%%%%%%%%%%%%%%%%%%%%%%%%%%%%%%%%%%%%%%%%%%%
%%%%%%%%%%%%%%%%%%%%%%%%%%%%%%%%%%%%%%%%%%%%
\label{app:maximalInequalities}

%%%%%%%%%%%%%%%%%%%%%%%%%%%%%%%%%%%%%%%%%%%%
\startOverview{Overview  and Notation for Appendix \ref{app:maximalInequalities} Results}
In this Appendix, we will use the following notation:
\begin{equation*}
    \pi_{2:t}^{*,(i)} \triangleq \prod_{t'=2}^t \pistar{t'}(\action{i}{t'}, \state{i}{t'})
    \TN{~~and~~}
    \hat{\pi}_{2:t}^{(i)} \triangleq \prod_{t'=2}^t \pihat{t'}(\action{i}{t'}, \state{i}{t'}).
\end{equation*}

Consider a function class $\F$ whose complexity is sufficiently controlled. In this section, our goal is to show a maximal inequality to bound the following:
\begin{equation}
    \label{eqnapp:maximalBoundTarget}
    \E^* \bigg[ \sup_{f \in \F} \bigg| \frac{1}{ \sqrt{n} } \sum_{i=1}^n \left( \rhohat{i}{2:t} f( \history{i}{t} ) - \E\left[ \rhohat{i}{2:t} f( \history{i}{t} ) \right] \right) \bigg| \bigg] 
\end{equation}
Above $\E^*$ refers to outer expectations as defined in Section 18.2 \cite{van2000asymptotic}. A bound for the above term in display \eqref{eqnapp:maximalBoundTarget} is used in Theorem \ref{thm:functionalAsymptoticNormality} (Functional Asymptotic Normality under Finite Bracketing Integral for Adaptively Sampled Data). This maximal inequality will be a function of the bracketing integral $\F$ bracketing integral, $\int_{0}^1 \sqrt{ \log N_{[~]} \big( \epsilon, \F, L_p ( \Pstar ) \big) } d \epsilon$, which we assume is finite. 

Note that since $\{\history{i}{T}\}_{i=1}^{n}$ are not independent in our setting, we cannot use classical maximal inequalities for i.i.d. data to bound the term in display \eqref{eqnapp:maximalBoundTarget}. See Section 19.6 of \cite{van2000asymptotic} for information on maximal inequalities for i.i.d. data. Our results build on the ideas used in these results for i.i.d. data.

%empirical process theory for i.i.d. data cannot be used to prove that the stochastic process in display \eqref{equieqn:stochasticProcess} is asymptotically tight
%is continuity of the limiting process converges in distribution to a Gaussian process \sam{ the real property is that you get continuity of the limiting process in terms of  the processes' covariance function but this is getting beyond this work} in $l^\infty( \F_{t,c_t})$ (the collection of all bounded functions from $\F_{t,c_t}$ to $\real$).   We now provide a summary of results in this section and describe the ways in which our results are similar to and differ from classical results in empirical processes.

\medskip
\noindent \bo{Summary of Results in this Section}
\begin{itemize}
    \item \bo{Weighted Martingale Bernstein Inequality (Lemma \ref{lemma:bernstein})} proves a Bernstein inequality for our non-independent, adaptively sampled data type and is the most novel step in this section. The proof leverages the conditional independence of the action selection at each time-step and the fact that the underlying potential outcomes are i.i.d. The proof repeatedly uses a key helper Lemma \ref{lemma:tightnessLemma} (Moving Products out of Expectations using Weights).
    %%%%%%%%%%%%%%%%%%%%%%%%%%%%%%%%%%%%%%%%%%%%%
    \item \bo{Maximal Inequality for Finite Class of Functions (Lemma \ref{lemma:maximalFinite})} proves a maximal inequality to bound the term in display \eqref{eqnapp:maximalBoundTarget} in the case that $| \F | < \infty$. The proof closely follows that of Lemma 19.33 \cite{van2000asymptotic}, but replaces the use of a Bernstein inequality for i.i.d. data with Lemma \ref{lemma:bernstein} (Weighted Martingale Bernstein Inequality).
    %for stochastic processes in the form of display \eqref{equieqn:stochasticProcess} 
    %%%%%%%%%%%%%%%%%%%%%%%%%%%%%%%%%%%%%%%%%%%%%
    \item \bo{Maximal Inequality as a Function of the Bracketing Integral (Lemma \ref{lemma:maximalBracketing})} proves a maximal inequality to bound the term in display \eqref{eqnapp:maximalBoundTarget} as a function of the bracketing integral for $\F$. The proof closely follows that of Lemma 19.34 \cite{van2000asymptotic}, but replaces the use of a maximal inequality for empirical processes for a finite class of functions on i.i.d. data with Lemma \ref{lemma:maximalFinite} (Maximal Inequality for Finite Class of Functions).
\end{itemize}

%%%%%%%%%%%%%%%%%%%%%%%%%%%%%%%%%%%%%%%%%%%%
%%%%%%%%%%%%%%%%%%%%%%%%%%%%%%%%%%%%%%%%%%%%
\subsection{Moving Products out of Expectations using Weights (Lemma \ref{lemma:tightnessLemma})}

\begin{lemma}[Moving Products out of Expectations using Weights]
	\label{lemma:tightnessLemma}
    For any $t \in [2 \colon T]$, let $f$ be any real-valued, measurable function of $\history{i}{t}$ such that $\Estar{2:t} \big[ \big| f(\history{i}{t}) \big| \big] < \infty$. Let $c$ be fixed constants. The following equality holds for any $n \geq 1$: % under Condition \ref{cond:exploration}:
	\begin{equation}
    \label{eqn:weightedConditionalProductKeyResult}
	   \E \bigg[ \prod_{i=1}^n \left( \rhohat{i}{2:t} f(\history{i}{t}) + c \right) \bigg]
	   = \prod_{i=1}^n \Estar{2:t} \left[ \rhostar{i}{2:t} f(\history{i}{t}) + c \right].
	 \end{equation}
\end{lemma}

\begin{remark}[Display \eqref{eqn:weightedConditionalProductKeyResult} Comment]
    Note that regarding the expectation terms on the right hand side above, for any stochastic policies $\pi_{2:t}(\beta_{1:t-1})$,
    \begin{multline*}
        \Estar{2:t} \left[ \rhostar{i}{2:t} f(\history{i}{t}) \right]
        = \Estar{2:t} \bigg[ \bigg( \prod_{t'=2}^t \pistar{t'} ( \action{i}{t'}, \state{i}{t'} ) \bigg)^{-1} f(\history{i}{t}) \bigg] \\
        = \E_{\pi_{2:t}(\beta_{1:t-1})} \bigg[ \bigg( \prod_{t'=2}^t \pi_{t'} ( \action{i}{t'}, \state{i}{t'} ; \beta_{t'-1} ) \bigg)^{-1} f(\history{i}{t}) \bigg].
    \end{multline*}
\end{remark}

\startproof{Lemma \ref{lemma:tightnessLemma} (Conditional Independence using Weights)}
For notational convenience we consider the $t$ set to $T$ case; the argument holds by the same argument for any $t \in [2 \colon T]$.

Let $t \in [2:T]$ and let $g$ be a real-valued, measurable function of $\history{i}{t}, \state{i}{t+1}$. % such that $\Estar{2:t+1} \big[ \big| g(\history{i}{t}, \state{i}{t+1}) \big| \big] < \infty$. 
A key result which we will take as given for now (we prove it at the end of this proof), is that
\begin{multline}
	\label{eqn:tightnessLemmaOnestep}
	\E \bigg[ \prod_{i=1}^n \bigg( \rhohat{i}{2:t} g \big(\history{i}{t}, \state{i}{t+1}\big) + c \bigg) \bigg] \\
	= \E \bigg[ \prod_{i=1}^n \bigg( \rhohat{i}{2:t-1} \E_{\pistar{t}} \bigg[ \rhostar{i}{t} g \big(\history{i}{t}, \state{i}{t+1} \big) \bigg| \history{i}{t-1}, \state{i}{t} \bigg] + c \bigg) \bigg].
	\hspace{-1mm}
\end{multline}

We now show that the desired result holds by repeatedly applying display \eqref{eqn:tightnessLemmaOnestep}.
Applying display \eqref{eqn:tightnessLemmaOnestep} for $t$ set to $T$ and for $g$ set to $f$, we have that 
\begin{equation*}
	\E \bigg[ \prod_{i=1}^n \left( \rhohat{i}{2:T} f(\history{i}{T}) + c \right) \bigg]
\end{equation*}
\begin{equation*}
	= \E \bigg[ \prod_{i=1}^n \left( \rhohat{i}{2:T-1} \E_{\pistar{T}} \left[ \rhostar{i}{T} f(\history{i}{t}) \big| \history{i}{T-1}, \state{i}{T} \right] + c \right) \bigg]
\end{equation*}

Now, note that $\E_{\pistar{T}} \left[ \rhostar{i}{T} f(\history{i}{T}) \big| \history{i}{T-1}, \state{i}{T} \right]$ is a function of $\history{i}{T-1}, \state{i}{T}$; let this be function be $g$ when we apply display \eqref{eqn:tightnessLemmaOnestep} again for $t$ set to $T-1$. %Note that by Jensen's inequality and Condition \ref{cond:exploration},
%$\Estar{2:t+1} \big[ \big| g(\history{i}{t}, \state{i}{t+1}) \big| \big] 
%= \Estar{2:T-1} \left[ \left| \E_{\pistar{T}} \big[ \rhostar{i}{T} f(\history{i}{T}) \big| \history{i}{T-1}, \state{i}{T} \big] \right| \right] \\
%\leq \Estar{2:T} \left[ \big| \rhostar{i}{T} f(\history{i}{T}) \big| \right]
%\leq \pi_{\min}^{-1} \Estar{2:T} \left[ \big| f(\history{i}{T}) \big| \right] < \infty$.
\begin{equation*}
	= \E \bigg[ \prod_{i=1}^n \bigg( \rhohat{i}{2:T-2} \Estar{T-1} \left[ \rhostar{i}{T-1} \E_{\pistar{T}} \left[ \rhostar{i}{T} f(\history{i}{T}) \big| \history{i}{T-1}, \state{i}{T} \right] \bigg| \history{i}{T-2}, \state{i}{T-1} \right] + c \bigg) \bigg]
\end{equation*}
Since $\rhostar{i}{T-1}$ is a constant given $\history{i}{T-1}$,
\begin{equation*}
	= \E \bigg[ \prod_{i=1}^n \bigg( \rhohat{i}{2:T-2} \Estar{T-1} \left[ \E_{\pistar{T}} \left[ \rhostar{i}{T-1:T} f(\history{i}{T}) \big| \history{i}{T-1}, \state{i}{T} \right] \bigg| \history{i}{T-2}, \state{i}{T-1} \right] + c \bigg) \bigg]
\end{equation*}
By law of iterated expectations,
\begin{equation*}
	= \E \bigg[ \prod_{i=1}^n \left( \rhohat{i}{2:T-2} \Estar{T-1:T} \left[ \rhostar{i}{T-1:T} f(\history{i}{T}) \big| \history{i}{T-2}, \state{i}{T-1} \right] + c \right) \bigg].
\end{equation*}

\medskip
\noindent By repeatedly applying display \eqref{eqn:tightnessLemmaOnestep} and the above argument for $t$ set to $T-2, T-3, ..., 2$ we have that
\begin{equation*}
	= \E \bigg[ \prod_{i=1}^n \left( \rhohat{i}{2:T-3} \Estar{T-2:T} \left[ \rhostar{i}{T-2:T} f(\history{i}{T}) \big| \history{i}{3}, \state{i}{T-2} \right] + c \right) \bigg]
\end{equation*}
\begin{equation*}
	= \E \bigg[ \prod_{i=1}^n \left( \rhohat{i}{2:T-4} \Estar{T-3:T} \left[ \rhostar{i}{T-3:T} f(\history{i}{T}) \big| \history{i}{4}, \state{i}{T-3} \right] + c \right) \bigg]
\end{equation*}
\begin{equation*}
	= ...
	= \E \bigg[ \prod_{i=1}^n \left( \Estar{2:T} \left[ \rhostar{i}{2:T} f(\history{i}{T}) \big| \history{i}{1}, \state{i}{1} \right] + c \right) \bigg].
\end{equation*}

Finally, recall that $\big\{ \history{i}{1}, \state{i}{2} \big\}_{i=1}^n = \big\{ \state{i}{1}, \action{i}{1}, \reward{i}{1}, \state{i}{2} \big\}_{i=1}^n$ are independent over $i \in [1 \colon n]$. Thus,
\begin{multline*}
	\E \bigg[ \prod_{i=1}^n \left\{ \Estar{2:T} \left[ \rhostar{i}{2:T} f(\history{i}{T}) \big| \history{i}{1}, \state{i}{1} \right] + c \right\} \bigg] \\
	= \prod_{i=1}^n \E \left[ \Estar{2:T} \left[ \rhostar{i}{2:T} f(\history{i}{T}) \big| \history{i}{1}, \state{i}{1} \right] + c \right]
\end{multline*}
By law of iterated expectations,
\begin{equation*}
	= \prod_{i=1}^n \Estar{2:T} \left[ \rhostar{i}{2:T} f(\history{i}{T}) + c \right].
\end{equation*}
Thus we have shown that the desired result holds and all that is left is to show that display \eqref{eqn:tightnessLemmaOnestep} holds.

\proofSubsection{Proof of display \eqref{eqn:tightnessLemmaOnestep}}
The proof of display \eqref{eqn:tightnessLemmaOnestep} leverages (i) the Radon-Nikodym weights and (ii) conditional independence properties. Pick any $t \in [2:T]$ and let $g$ be a real-valued, measurable function of $\history{i}{t}, \state{i}{t+1}$. By law of iterated expectations,
\begin{equation*}
	\E \bigg[ \prod_{i=1}^n \left( \rhohat{i}{2:t} g \big(\history{i}{t}, \state{i}{t+1} \big) + c \right) \bigg]
\end{equation*}
\begin{equation*}
	= \E \bigg[ \E \bigg[ \prod_{i=1}^n \left( \rhohat{i}{2:t} g \big(\history{i}{t}, \state{i}{t+1} \big) + c \right) \bigg| \history{1:n}{t-1}, \state{1:n}{t} \bigg] \bigg]
\end{equation*}
Note that the conditional expectation $\E \left[ \prod_{i=1}^n \left( \rhohat{i}{2:t} g \big(\history{i}{t}, \state{i}{t+1} \big) + c \right) \big| \history{1:n}{t-1}, \state{1:n}{t} \right]$ is only integrating over $\big\{ \action{i}{t}, \reward{i}{t}, \state{i}{t+1} \big\}_{i=1}^n$ (the policy parameters $\betahat{1:t-1}$ used in $\rhohat{i}{2:t}$ are known given $\history{1:n}{t-1}$). Additionally, note that conditional on $\history{1:n}{t-1}, \state{1:n}{t}$, \\
$\big\{ \action{i}{t}, \reward{i}{t}, \state{i}{t+1} \big\}$ are independent over $i \in [1 \colon n]$. Thus,
\begin{equation*}
	= \E \bigg[ \prod_{i=1}^n \left( \E \left[ \rhohat{i}{2:t} g \big(\history{i}{t}, \state{i}{t+1} \big) \bigg| \history{1:n}{t-1}, \state{1:n}{t} \right] + c \right) \bigg]
\end{equation*}
Since $\rhohat{i}{2:t-1} = \big( \prod_{t'=2}^{t-1} \pihat{t'}(\action{i}{t'}, \state{i}{t'}) \big)^{-1}$ is a constant given $\history{1:n}{t-1}, \state{1:n}{t}$,
\begin{equation*}
	= \E \bigg[ \prod_{i=1}^n \left( \rhohat{i}{2:t-1} \E \left[ \rhohat{i}{t} g \big(\history{i}{t}, \state{i}{t+1} \big) \bigg| \history{1:n}{t-1}, \state{1:n}{t} \right] + c \right) \bigg]
\end{equation*}
Now now show that the above equals the following:
\begin{equation}
    \label{eqn:keyLemmaLastEquality}
	= \E \bigg[ \prod_{i=1}^n \left( \rhohat{i}{2:t-1} \Estar{t} \left[ \rhostar{i}{t} g \big(\history{i}{t}, \state{i}{t+1} \big) \big| \history{i}{T-1}, \state{i}{t} \right] + c \right) \bigg].
\end{equation}
Showing the above will be sufficient for display \eqref{eqn:tightnessLemmaOnestep}. 

Note that since $1 = \pistar{t}(\action{i}{t}, \state{i}{t})^{-1} \pistar{t}(\action{i}{t}, \state{i}{t}) = \rhostar{i}{t} \pistar{t}(\action{i}{t}, \state{i}{t})$,
\begin{multline*}
    \E \left[ \rhohat{i}{t} g \big(\history{i}{T}, \state{i}{t+1} \big) \big| \history{1:n}{t-1}, \state{1:n}{t} \right] \\
    = \E \left[\rhohat{i}{t} \rhostar{i}{t} \pistar{t}(\action{i}{t}, \state{i}{t}) g \big(\history{i}{T}, \state{i}{t+1} \big) \big| \history{1:n}{t-1}, \state{1:n}{t} \right] \\
    = \Estar{t} \left[ \rhostar{i}{t} g \big(\history{i}{t}, \state{i}{t+1} \big) \big| \history{1:n}{t-1}, \state{1:n}{t} \right].
\end{multline*}
The last equality above holds because $\rhohat{i}{t} \pistar{t}(\action{i}{t}, \state{i}{t}) = \WW{i}{t}(\betastar{t-1}, \betahat{t-1})$.

Also, note that the expectation $\E_{\pistar{t}} \left[ \rhostar{i}{t} g \big(\history{i}{t}, \state{i}{t+1} \big) \big| \history{1:n}{t-1}, \state{1:n}{t} \right]$ integrates over $\big\{ \action{i}{t}, \reward{i}{t}, \state{i}{t+1} \big\}$. Since actions are selected using $\pistar{t}$ rather than $\pihat{t}$ in the expectation, the distribution of $\big\{ \action{i}{t}, \reward{i}{t}, \state{i}{t+1} \big\}$ depends only on $\history{1:n}{t-1}, \state{1:n}{t}$ through $\history{i}{t-1}, \state{i}{t}$. This means that 
\begin{equation*}
    \E_{\pistar{t}} \left[ \rhostar{i}{t}  g \big(\history{i}{t}, \state{i}{t+1} \big) \big| \history{1:n}{t-1}, \state{1:n}{t} \right] 
    = \E_{\pistar{t}} \left[ \rhostar{i}{t} g \big(\history{i}{t}, \state{i}{t+1} \big) \big| \history{i}{t-1}, \state{i}{t} \right].
\end{equation*}
Thus we have shown display \eqref{eqn:keyLemmaLastEquality} holds. $\blacksquare$

%%%%%%%%%%%%%%%%%%%%%%%%%%%%%%%%%%%%%%%%%%%%
\subsection{Weighted Martingale Bernstein Inequality (Lemma \ref{lemma:bernstein})} %%%%%%%%%%%%%%%%%%%%%%%%%%%%%%%%%%%%%%%%%%%%
%%%%%%%%%%%%%%%%%%%%%%%%%%%%%%%%%%%%%%%%%%%%

\begin{lemma}[Weighted Martingale Bernstein Inequality]
	\label{lemma:bernstein}
	We assume Condition \ref{cond:exploration} (Minimum Exploration) holds.
	Let $f$ be a real-valued, measurable function of $\history{i}{t}$. % with $0 < \| f \|_\infty < \infty$. 
    Then, for any $x > 0$ and for all $n \geq 1$,
	\begin{multline}
	    \label{berstein:eqn}
		\PP \bigg( \bigg| \frac{1}{ \sqrt{n} } \sum_{i=1}^n \rhohat{i}{2:t} f(\history{i}{t}) - \E \left[ \rhohat{i}{2:t} f(\history{i}{t}) \right] \bigg| \geq x \bigg) \\
		\leq 2 \exp \bigg( -\frac{ \pi_{\min}^{t-1} }{4} \frac{ x^2}{ \Estar{2:t} \left[ \rhostar{i}{2:t} f(\history{i}{t})^2 \right] + x \| f \|_\infty / \sqrt{n} } \bigg).
	\end{multline}
    Above $\| f \|_\infty \triangleq \sup_{h} | f(h) |$.
\end{lemma}

%\noindent \textit{\bo{Remark:}
%Note that regarding the expectation on the right hand side in display \eqref{berstein:eqn}, for any fixed policies $\pi_{2:t}(\beta_{1:t-1})$,}
%\begin{equation*}
%    \Estar{2:t} \left[ \rhostar{i}{2:t} f(\history{i}{t})^2 \right]
%    = \E_{\pi_{2:t}(\beta_{1:T-1})} \bigg[ \bigg( \prod_{t'=2}^t \pi_t(\action{i}{t'}, \state{i}{t'}; \beta_{t'}) \bigg)^{-1} f(\history{i}{t})^2 \bigg].
%\end{equation*}
%we use the distribution under the target policies $\pistar{2:T}$ purely for convenience. This could easily be changed to distributions under another policy that satisfies Condition \ref{cond:exploration}.

%\sam{I thought that the $\pistar{2:T}$ was part of the function $f$ so the term, $\Estar{2:T} \big[ f(\history{i}{T})^2 \big]$ has three copies of 
%$\pistar{2:T}$--see display (8) in body of paper....}

%%%%%%%%%%%%%%%%%%%%%%%%%%%%%%%%%%%%%
\startproof{Lemma \ref{lemma:bernstein} (Weighted Martingale Bernstein Inequality)}
We follow an argument similar to Lemma 19.32 in \cite{van2000asymptotic}. For notational convenience we consider the $t$ set to $T$ case; the argument holds by the same argument for any $t \in [2 \colon T]$.

The leading 2 in display \eqref{berstein:eqn} is due to separate bounds for the upper and lower tail bounds. It is sufficient to show the upper tail bound, because the lower tail bound holds by the upper tail bound applied to $-f$.

Note if $\| f \|_\infty = 0$, then $\PP \left( \frac{1}{ \sqrt{n} } \sum_{i=1}^n \left( \rhohat{i}{2:t} f(\history{i}{t}) - \E \left[ \rhohat{i}{2:t} f(\history{i}{t}) \right] \right) \geq x \right) = 0$, so in this case display \eqref{berstein:eqn} easily holds. Thus, for the remainder of the proof we assume that $\| f \|_\infty > 0$.

Let $x > 0$.
\begin{equation}
\label{bernsteinSAM}
		\PP \bigg( \frac{1}{\ \sqrt{n} } \sum_{i=1}^n \left( \rhohat{i}{2:T} f(\history{i}{T}) - \E \left[ \rhohat{i}{2:T} f(\history{i}{T}) \right] \right) \geq x \bigg)
\end{equation}
Using a Chernoff bound, for any $\lambda > 0$,
\begin{equation*}
		\leq e^{-\lambda x } \E \bigg[ \exp \bigg\{ \frac{\lambda}{ \sqrt{n} } \sum_{i=1}^n \bigg( \rhohat{i}{2:T} f(\history{i}{T}) - \E \left[ \rhohat{i}{2:T} f(\history{i}{T}) \right] \bigg) \bigg\} \bigg]
\end{equation*}
Using properties of exponents, we can change the summation in exponent into a product,
\begin{equation*}
	= e^{- \lambda x } \E \bigg[ \prod_{i=1}^n \exp \left\{ \frac{\lambda}{ \sqrt{n} } \bigg(\rhohat{i}{2:T} f(\history{i}{T}) - \E \left[ \rhohat{i}{2:T} f(\history{i}{T}) \right] \bigg) \right\} \bigg]
\end{equation*}
We now apply Maclaurin series for exponential function, i.e., that $e^z = \sum_{k=0}^\infty \frac{z^k}{k!}$.
\begin{equation*}
	= e^{- \lambda x } \E \bigg[ \prod_{i=1}^n \sum_{k=0}^\infty \frac{1}{k!} \left(\frac{\lambda }{ \sqrt{n} } \right)^k \bigg( \rhohat{i}{2:T} f(\history{i}{T}) - \E \left[ \rhohat{i}{2:T} f(\history{i}{T}) \right] \bigg)^k \bigg]
\end{equation*}
By simplifying the first two terms in the inner summation,
\begin{multline}
    \label{eqn:bernsteinNonnegInequality}
	= e^{- \lambda x } \E \bigg[ \prod_{i=1}^n \bigg\{ 1 + \frac{\lambda}{ \sqrt{n} } \bigg( \rhohat{i}{2:T} f(\history{i}{T}) - \E \left[ \rhohat{i}{2:T} f(\history{i}{T}) \right] \bigg) \\
	+ \sum_{k=2}^\infty \frac{1}{k!} \left(\frac{\lambda}{ \sqrt{n} } \right)^k \bigg( \rhohat{i}{2:T} f(\history{i}{T}) - \E \left[ \rhohat{i}{2:T} f(\history{i}{T}) \right] \bigg)^k \bigg\} \bigg]
\end{multline}

\noindent Now note the following observations:
\begin{itemize}
    \item Let $\epsilon > 0$. By Condition \ref{cond:exploration} (Minimum Exploration),
    %\begin{equation*}
    %    \WW{i}{t'}\big( \betastar{t'}, \betahat{t'} \big) = \frac{\pistar{t'}(\action{i}{t'}, \state{i}{t'})}{\pihat{t'}(\action{i}{t'}, \state{i}{t'})}
    %    \geq \frac{\pistar{t'}(\action{i}{t'}, \state{i}{t'})}{1}
    %\geq \pi_{\min} ~~~ \TN{a.s.}
    %\end{equation*}
    %Thus,
    \begin{equation}
        \label{eqn:rhohatUpperBound}
        \rhohat{i}{2:t} = \bigg[ \prod_{t'=2}^t \pihat{t'}(\action{i}{t'}, \state{i}{t'}) \bigg]^{-1} \leq \pi_{\min}^{-(t-1)} ~~\TN{a.s.},
    \end{equation}
    and
    \begin{multline}
        \label{appBernstein:kminus2Upper}
        \left| \rhohat{i}{2:T} f(\history{i}{T}) - \E \left[ \rhohat{i}{2:T} f(\history{i}{T}) \right] \right| \\
        \leq \left| \rhohat{i}{2:T} f(\history{i}{T}) \right| + \left| \E \left[ \rhohat{i}{2:T} f(\history{i}{T}) \right] \right| 
        \leq 2 \pi_{\min}^{-(T-1)} \| f \|_\infty.
    \end{multline}
    %%%%%%%%%%%%%%%%%%%%%%%%%%%%%%%
    \item We can upper bound the following:
    \begin{equation*}
	    \left( \rhohat{i}{2:T} f(\history{i}{T}) - \E \left[ \rhohat{i}{2:T} f(\history{i}{T}) \right] \right)^2
    \end{equation*}
    \begin{equation*}
	    = \big( \hat{\pi}_{2:T}^{(i)} \big)^{-2} f(\history{i}{T})^2 - 2 \rhohat{i}{2:T} f(\history{i}{T}) \E \left[ \rhohat{i}{2:T} f(\history{i}{T}) \right] + \left( \E \left[ \rhohat{i}{2:T} f(\history{i}{T}) \right] \right)^2
    \end{equation*}
    Note that $\E \left[ \rhohat{i}{2:T} f(\history{i}{T}) \right]
    = \Estar{2:T} \left[ \rhostar{i}{2:T} f(\history{i}{T}) \right]$.
    \begin{multline*}
	    = \rhohat{i}{2:T} \left( \rhohat{i}{2:T} f(\history{i}{T})^2 - 2 f(\history{i}{T}) \Estar{2:T} \left[ \rhostar{i}{2:T} f(\history{i}{T}) \right] \right) \\
        + \left( \Estar{2:T} \left[ \rhostar{i}{2:T} f(\history{i}{T}) \right] \right)^2
    \end{multline*}
    Since $\rhohat{i}{2:T} \leq \pi_{\min}^{-(T-1)}$ a.s. by display \eqref{eqn:rhohatUpperBound},
    \begin{multline*}
	    \leq \rhohat{i}{2:T} \bigg( \underbrace{ \pi_{\min}^{-(T-1)} f(\history{i}{T})^2 - 2 f(\history{i}{T}) \Estar{2:T} \left[ \rhostar{i}{2:T} f(\history{i}{T}) \right] }_{\triangleq g(\history{i}{T})} \bigg) \\
        + \left( \Estar{2:T} \left[ \rhostar{i}{2:T} f(\history{i}{T}) \right] \right)^2
    \end{multline*}
    For $g(\history{i}{T}) \triangleq \pi_{\min}^{-(T-1)} f(\history{i}{T})^2 
    - 2 f(\history{i}{T}) \Estar{2:T} \left[ \rhostar{i}{2:T} f(\history{i}{T}) \right]$,
    \begin{equation*}
	    = \rhohat{i}{2:T} g(\history{i}{T}) + \left( \Estar{2:T} \left[ \rhostar{i}{2:T} f(\history{i}{T}) \right] \right)^2.
    \end{equation*}
    Thus, in summary we have that
    \begin{multline}
        \label{appBernstein:2Upper}
	    \left( \rhohat{i}{2:T} f(\history{i}{T}) - \E \left[ \rhohat{i}{2:T} f(\history{i}{T}) \right] \right)^2 \\
        \leq \rhohat{i}{2:T} g(\history{i}{T}) + \left( \Estar{2:T} \left[ \rhostar{i}{2:T} f(\history{i}{T}) \right] \right)^2.
    \end{multline}
    %%%%%%%%%%%%%%%%%%%%%%%%%%%%%%%
    \item By display \eqref{appBernstein:kminus2Upper} and \eqref{appBernstein:2Upper}, we have that for any $k \geq 2$
    \begin{multline}
        \label{app:innerProductInequalityTerm}
        \left( \rhohat{i}{2:T} f(\history{i}{T}) - \E \left[ \rhohat{i}{2:T} f(\history{i}{T}) \right] \right)^k \\
        \leq \left( \rhohat{i}{2:T} f(\history{i}{T}) - \E \left[ \rhohat{i}{2:T} f(\history{i}{T}) \right] \right)^2 \left| \rhohat{i}{2:T} f(\history{i}{T}) - \E \left[ \rhohat{i}{2:T} f(\history{i}{T}) \right] \right|^{k-2} \\
        \leq \bigg( \rhohat{i}{2:T} g(\history{i}{T}) + \left( \Estar{2:T} \left[ \rhostar{i}{2:T} f(\history{i}{T}) \right] \right)^2 \bigg) \big\{ 2 \pi_{\min}^{-(T-1)} \| f \|_\infty \big\}^{k-2}
    \end{multline}
\end{itemize}

\noindent Note that in display \eqref{eqn:bernsteinNonnegInequality}, each of the terms in the product over $n$ terms is non-negative because the $i^{\TN{th}}$ term in the product equals $\exp \left\{ \frac{\lambda}{ \sqrt{n} } \left(\rhohat{i}{2:T} f(\history{i}{T}) - \E \left[ \rhohat{i}{2:T} f(\history{i}{T}) \right] \right) \right\}$ and $e^x \geq 0$ for all $x$. Thus, by display \eqref{app:innerProductInequalityTerm}, we  can upper bound display \eqref{eqn:bernsteinNonnegInequality} as follows:
\begin{multline}
	\label{eqn:bernsteinPreLemma}
	\leq e^{- \lambda x } \E \bigg[ \prod_{i=1}^n \bigg\{ 1 + \frac{\lambda}{ \sqrt{n} } \bigg( \rhohat{i}{2:T} f(\history{i}{T}) - \E \left[ \rhohat{i}{2:T} f(\history{i}{T}) \right] \bigg) \\
	+ \sum_{k=2}^\infty \frac{1}{k!} \left(\frac{\lambda}{ \sqrt{n} } \right)^k \bigg( \rhohat{i}{2:T} g(\history{i}{T}) + \Estar{2:T} \left[ \rhostar{i}{2:T} f(\history{i}{T}) \right]^2 \bigg) \left( 2 \pi_{\min}^{-(T-1)} \| f \|_\infty \right)^{k-2} \bigg\} \bigg]
\end{multline}

Note that everything in the expectation above in display \eqref{eqn:bernsteinPreLemma} is bounded a.s.; we will show that this is true for the infinite summation over $k$. Let $y = \rhohat{i}{2:T} g(\history{i}{T}) + \Estar{2:T} \left[ \rhostar{i}{2:T} f(\history{i}{T}) \right]^2$ and $z = 2 \pi_{\min}^{-(T-1)} \| f \|_\infty$. Note that both $y$ and $z$ are bounded a.s. Thus, since $\| f \|_\infty > 0$ by assumption (we discussed the $\| f \|_\infty = 0$ case at the beginning of this proof), we have that $z > 0$, so $\sum_{k=2}^\infty \frac{1}{k!} \big(\frac{\lambda}{ \sqrt{n} }\big)^k y z^{k-2}
= y z^{-2} \sum_{k=2}^\infty \frac{1}{k!} \big(\frac{\lambda}{ \sqrt{n} }\big)^k z^k 
= y z^{-2} \left( e^{ z \lambda / \sqrt{n} } - \sum_{k=0}^1 \frac{1}{k!} \big(\frac{\lambda}{ \sqrt{n} }\big)^k z^k \right)$ is also bounded a.s. \\

Moreover, display \eqref{eqn:bernsteinPreLemma} can be written as $e^{- \lambda x }  \E \left[ \prod_{i=1}^n \left( \rhohat{i}{2:T} h(\history{i}{T}) + c \right) \right]$, for some function $h$ and some finite constant $c$. (Note that $\E \left[ \rhohat{i}{2:T} f(\history{i}{T}) \right] = \Estar{2:T} \left[ \rhostar{i}{2:T} f(\history{i}{T}) \right]$)
Thus, we can apply Lemma \ref{lemma:tightnessLemma} (Moving Products out of Expectations using Weights) to get that display \eqref{eqn:bernsteinPreLemma} is equal to
\begin{multline*}
	= e^{- \lambda x } \prod_{i=1}^n \Estar{2:T} \bigg[ 1 + \frac{\lambda}{ \sqrt{n} } \bigg( \rhostar{i}{2:T} f(\history{i}{T}) - \E \left[ \rhohat{i}{2:T} f(\history{i}{T}) \right] \bigg) \\
	+ \sum_{k=2}^\infty \frac{1}{k!} \left(\frac{\lambda}{ \sqrt{n} }\right)^k \bigg( \rhostar{i}{2:T} g(\history{i}{T}) + \Estar{2:T} \left[ \rhostar{i}{2:T} f(\history{i}{T}) \right]^2 \bigg) \left( 2 \pi_{\min}^{-(T-1)} \| f \|_\infty \right)^{k-2} \bigg].
\end{multline*}

Since $\E \left[ \rhohat{i}{2:T} f(\history{i}{T}) \right] = \Estar{2:T} \left[ \rhostar{i}{2:T} f(\history{i}{T}) \right]$, we can cancel terms in the first line above.
\begin{multline*}
	= e^{- \lambda x } \prod_{i=1}^n \bigg\{ 1 
	+ \Estar{2:T} \bigg[ \sum_{k=2}^\infty \frac{1}{k!} \left(\frac{\lambda}{ \sqrt{n} }\right)^k \bigg( \rhostar{i}{2:T} g(\history{i}{T}) + \Estar{2:T} \left[ \rhostar{i}{2:T} f(\history{i}{T}) \right]^2 \bigg) \\
	\left( 2 \pi_{\min}^{-(T-1)} \| f \|_\infty \right)^{k-2} \bigg] \bigg\}
\end{multline*}

Since everything in the expectations above are bounded a.s. (discussed below display \eqref{eqn:bernsteinPreLemma}), we can exchange the expectation with the infinite summation over $k$.
\begin{multline}
	\label{eqn:bernsteinPostLemma}
	= e^{- \lambda x } \prod_{i=1}^n \bigg\{ 1 + \sum_{k=2}^\infty \frac{1}{k!} \left(\frac{\lambda}{ \sqrt{n} }\right)^k \bigg( \Estar{2:T} \left[ \rhostar{i}{2:T} g(\history{i}{T}) \right] + \Estar{2:T} \left[ \rhostar{i}{2:T} f(\history{i}{T}) \right]^2 \bigg) \\
	\left( 2 \pi_{\min}^{-(T-1)} \| f \|_\infty \right)^{k-2} \bigg\}.
\end{multline}
Note the following:
\begin{itemize}
    \item Recall that $g(\history{i}{T}) \triangleq \pi_{\min}^{-(T-1)} f(\history{i}{T})^2 - 2 f(\history{i}{T}) \Estar{2:T} \left[ \rhostar{i}{2:T} f(\history{i}{T}) \right]$. So,
    \begin{equation*}
	   \Estar{2:T} \left[ \rhostar{i}{2:T} g(\history{i}{T}) \right] + \Estar{2:T} \left[ \rhostar{i}{2:T} f(\history{i}{T}) \right]^2
    \end{equation*}
    \begin{equation*}
	   = \pi_{\min}^{-(T-1)} \Estar{2:T} \left[ \rhostar{i}{2:T} f(\history{i}{T})^2 \right] 
	   - 2 \Estar{2:T} \left[ \rhostar{i}{2:T} f(\history{i}{T}) \right]^2 
	   + \Estar{2:T} \left[ \rhostar{i}{2:T} f(\history{i}{T}) \right]^2
    \end{equation*}
    \begin{equation*}
	   = \pi_{\min}^{-(T-1)} \Estar{2:T} \left[ \rhostar{i}{2:T} f(\history{i}{T})^2 \right] - \Estar{2:T} \left[ \rhostar{i}{2:T} f(\history{i}{T}) \right]^2
	  \end{equation*}
    \begin{equation*}  
        \leq \pi_{\min}^{-(T-1)} \Estar{2:T} \left[ \rhostar{i}{2:T} f(\history{i}{T})^2 \right]
    \end{equation*}
\end{itemize}

\noindent Thus display \eqref{eqn:bernsteinPostLemma} can be upper bounded by the following:
\begin{equation*}
	e^{- \lambda x } \prod_{i=1}^n \bigg\{ 1 
	+ \sum_{k=2}^\infty \frac{1}{k!} \left(\frac{\lambda}{ \sqrt{n} }\right)^k \pi_{\min}^{-(T-1)} \Estar{2:T} \left[ \rhostar{i}{2:T} f(\history{i}{T})^2 \right]
	\left( 2 \pi_{\min}^{-(T-1)} \| f \|_\infty \right)^{k-2} \bigg\}
\end{equation*}
By i.i.d. potential outcomes,
\begin{equation*}
	= e^{- \lambda x } \bigg\{ 1 
	+ \sum_{k=2}^\infty \frac{1}{k!} \left(\frac{\lambda}{ \sqrt{n} }\right)^k \pi_{\min}^{-(T-1)} \Estar{2:T} \left[ \rhostar{i}{2:T} f(\history{i}{T})^2 \right]
	\left( 2 \pi_{\min}^{-(T-1)} \| f \|_\infty \right)^{k-2} \bigg\}^n
\end{equation*}
\noindent By rearranging terms,
\begin{equation*}
	= e^{- \lambda x } \bigg\{ 1 
	+ \frac{1}{n} \sum_{k=2}^\infty \frac{1}{k!} \frac{1}{2} \lambda^k \underbrace{ 2 \pi_{\min}^{-(T-1)} \Estar{2:T} \left[ \rhostar{i}{2:T} f(\history{i}{T})^2 \right] }_{ \triangleq \lambda_1^{-1} }
	\bigg( \underbrace{ 2 \pi_{\min}^{-(T-1)} \| f \|_\infty / \sqrt{n} }_{ \triangleq \lambda_2^{-1} } \bigg)^{k-2} \bigg\}^n
\end{equation*}
\begin{equation}
	\label{eqn:bernsteinLambdas}
	= e^{- \lambda x } \bigg\{ 1 + \frac{1}{n} \sum_{k=2}^\infty \frac{1}{k!} \frac{1}{2} \lambda^k 
	\left( \lambda_1^{-1} \lambda_2^{-(k-2)} \right) \bigg\}^n
\end{equation}

\noindent Note that since $\lambda_1^{-1},\lambda_2^{-1} \geq 0$,
\begin{equation*}
	\lambda \triangleq x \left( \lambda_1^{-1} + x \lambda_2^{-1} \right)^{-1}
    \leq \min \left\{ x \left( \lambda_1^{-1} + 0 \right)^{-1}, 
    x \left( 0 + x \lambda_2^{-1} \right)^{-1} \right\}
    = \min \left( x \lambda_1, \lambda_2 \right).
\end{equation*}
Thus we have that $\lambda^k 
	\leq \lambda \min( x \lambda_1, \lambda_2 )^{k-1}
	\leq \lambda x \lambda_1 \lambda_2^{k-2}$. So we can upper bound display \eqref{eqn:bernsteinLambdas} as follows:
\begin{equation*}
	\leq e^{- \lambda x } \bigg\{ 1 + \frac{1}{n} \sum_{k=2}^\infty \frac{1}{k!} \frac{1}{2} \left( \lambda x \lambda_1 \lambda_2^{k-2} \right) \left( \lambda_1^{-1} \lambda_2^{-(k-2)} \right) \bigg\}^n
\end{equation*}
\begin{equation*}
	= e^{- \lambda x } \bigg\{ 1 + \frac{1}{n} \underbrace{ \sum_{k=2}^\infty \frac{1}{k!} }_{\leq 1} \frac{1}{2} \lambda x \bigg\}^n
\end{equation*}
By the Maclaurin series for exponential function, $e^z = \sum_{k=0}^\infty \frac{z^k}{k!}$, we have $\sum_{k=2}^\infty \frac{1}{k!} = e - \frac{1}{0!} - \frac{1}{1!} = e - 2 \leq 1$. 
\begin{equation*}
	\leq e^{- \lambda x } \bigg\{ 1 
	+ \frac{1}{n} \frac{1}{2}  x \lambda \bigg\}^n
\end{equation*}
Again by the Maclaurin series for exponential function, $e^z = \sum_{k=0}^\infty \frac{z^k}{k!}$, so for $z > 0$ we have that $1 + z\leq e^{ z }$. This means that $(1+z)^n \leq e^{z n}$.
\begin{equation*}
	\leq e^{- \lambda x} \exp \bigg( \frac{1}{2} x \lambda \bigg)
	= \exp \bigg( -\frac{1}{2} x \lambda \bigg)
\end{equation*}
Recall that $\lambda \triangleq x \left( \lambda_1^{-1} + x \lambda_2^{-1} \right)^{-1}$, so,
\begin{equation*}
	= \exp \bigg( -\frac{1}{2} x x \left( \lambda_1^{-1} + x \lambda_2^{-1} \right)^{-1} \bigg)
\end{equation*}

\noindent Recall that $\lambda_1^{-1} = 2 \pi_{\min}^{-(T-1)} \Estar{2:T} \left[ \rhostar{i}{2:T} f(\history{i}{T})^2 \right]$ and $\lambda_2^{-1} = 2 \pi_{\min}^{-(T-1)} \| f \|_\infty / \sqrt{n}$. So,
\begin{equation*}
	= \exp \bigg( -\frac{ \pi_{\min}^{T-1} }{4} \frac{ x^2}{ \Estar{2:T} \big[ \rhostar{i}{2:T} f(\history{i}{T})^2 \big] + x \| f \|_\infty / \sqrt{n} } \bigg). ~~ \blacksquare
\end{equation*}

%%%%%%%%%%%%%%%%%%%%%%%%%%%%%%%%%%%%%%%%%%%%
\subsection{Maximal Inequality for Finite Class of Functions (Lemma \ref{lemma:maximalFinite})} %%%%%%%%%%%%%%%%%%%%%%%%%%%%%%%%%%%%%%%%%%%%
%%%%%%%%%%%%%%%%%%%%%%%%%%%%%%%%%%%%%%%%%%%%

\begin{lemma}[Maximal Inequality for Finite Class of Functions]
    \label{lemma:maximalFinite}
    Let $\F$ be a finite class of bounded, real-valued, measurable functions of $\history{i}{t}$ with size $|\F| \geq 2$. For $f \in \F$ we define
	\begin{equation*}
	    \GG_n (f) \triangleq \frac{1}{ \sqrt{n} } \sum_{i=1}^n \left( \rhohat{i}{2:t} f( \history{i}{t} ) - \E\left[ \rhohat{i}{2:t} f( \history{i}{t} ) \right] \right).
	\end{equation*}
	Under Condition \ref{cond:exploration} (Minimum Exploration) for all sufficiently large $n$,
    \begin{multline}
	    \label{eqn:maximalFinite}
	    \E \bigg[ \max_{f \in \F} \big| \GG_n (f) \big| \bigg] 
	    \leq C \bigg\{ \pi_{\min}^{-(T-1)} \max_{f \in \F} \frac{ \| f \|_\infty }{ \sqrt{n} } \log (|\F|) \\
	    + \sqrt{\pi_{\min}^{-(t-1)}} \max_{f \in \F} \sqrt{ \Estar{2:t} \left[ \rhostar{i}{2:t}  f(\history{i}{T})^2 \right] } \sqrt{ \log ( |\F|) } \bigg\},
    \end{multline}
	for some universal positive constant $C$ (specified in the proof).
\end{lemma}

%%%%%%%%%%%%%%%%%%%%%%%%%%%%%%%%%%%%%%%%%%%%
\startproof{Lemma \ref{lemma:maximalFinite} (Maximal Inequality for Finite Class of Functions)} %%%%%%%%%%%%%%%%%%%%%%%%%%%%%%%%%%%%%%%%%%%%
Our proof follows a very similar argument to Lemma 19.33 in \cite{van2000asymptotic}. Specifically, our proof only deviates because we use our Lemma \ref{lemma:bernstein} (Weighted Martingale Bernstein Inequality) to prove displays \eqref{eqn:GunderInequality} and \eqref{eqn:GoverInequality} below. 

For notational convenience we consider the $t$ set to $T$ case; the argument holds by the same argument for any $t \in [2 \colon T]$. 

\proofSubsection{Special cases}
\begin{itemize}
    \item Note that if $f \in \F$ such that $\| f \|_\infty = 0$, then $\GG_n (f) = 0$. These zero functions do not contribute to increasing the upper bound for $\E \big[ \max_{f \in \F} \big| \GG_n (f) \big| \big]$. Thus, we assume that $\| f \|_\infty > 0$ for all $f \in \F$ for the remainder of this proof, as this is the most difficult case.
    \item Note that $f \in \F$ such that $\Estar{2:t} \left[ \rhostar{i}{2:t}  f(\history{i}{T})^2 \right] = 0$, then $f(\history{i}{T})$ is a constant function and $\GG_n (f) = 0$. These constant functions do not contribute to increasing the upper bound for $\E \big[ \max_{f \in \F} \big| \GG_n (f) \big| \big]$. Thus, we assume that $\Estar{2:t} \left[ \rhostar{i}{2:t}  f(\history{i}{T})^2 \right] > 0$ for all $f \in \F$ for the remainder of this proof, as this is the most difficult case.
\end{itemize}

\proofSubsection{Main argument} Let $u, v$ be non-negative, real-valued functions of $f \in \F$ such that
\begin{itemize}
	\item $u(f) = 24 \pi_{\min}^{-(T-1)} \Estar{2:T} \left[ \rhostar{i}{2:T} f(\history{i}{T})^2 \right]$
	\item $v(f) = 24 \pi_{\min}^{-(T-1)} \| f \|_\infty / \sqrt{n}$
\end{itemize}
Note that
\begin{equation*}
	\E \left[ \max_{f \in \F} | \mathbb{G}_n (f) | \right]
	= \E \left[ \max_{f \in \F} \bigg\{ | \mathbb{G}_n (f) | \II_{ | \mathbb{G}_n (f) | > u(f)/v(f)}
	+ | \mathbb{G}_n (f) | \II_{ | \mathbb{G}_n (f) | \leq u(f)/v(f)} \bigg\} \right]
\end{equation*}
\begin{equation*}
	\leq \E \left[ \max_{f \in \F} | \mathbb{G}_n (f) | \II_{ | \mathbb{G}_n (f) | > u(f)/v(f)} \right]
	+ \E \left[ \max_{f \in \F} | \mathbb{G}_n (f) | \II_{ | \mathbb{G}_n (f) | \leq u(f)/v(f)} \right]
\end{equation*}
Let $\under{ \GG_n }(f) \triangleq | \mathbb{G}_n (f) | \II_{ | \mathbb{G}_n (f) | > u(f)/v(f)}$ and $\overline{ \GG_n }(f) \triangleq | \mathbb{G}_n (f) | \II_{ | \mathbb{G}_n (f) | \leq u(f)/v(f)}$.
\begin{equation*}
	= \E \left[ \max_{f \in \F} \under{ \GG_n }(f) \right]
	+ \E \left[ \max_{f \in \F} \overline{ \GG_n }(f) \right]
\end{equation*}
\begin{equation}
	\label{eqn:finiteFinequality1}
	\leq \E \left[ \max_{f \in \F} \under{ \GG_n }(f) / v(f) \right] \left( \max_{f \in \F} v(f) \right)
	+ \E \left[ \max_{f \in \F} \overline{ \GG_n }(f) / \sqrt{u(f)} \right] \left( \max_{f \in \F} \sqrt{u(f)} \right)
\end{equation}
The main results show in this proof are the following:
\begin{equation}
	\label{eqn:GfloorLog}
	\E \left[ \max_{f \in \F} \under{ \GG_n }(f) / v(f) \right] \leq \log \left( 1 + | \F | \right)
\end{equation}
\begin{equation}
	\label{eqn:GceilLog}
	\E \left[ \max_{f \in \F} \overline{ \GG_n }(f) / \sqrt{u(f)} \right] \leq \sqrt{ \log \left( 1 + | \F | \right) }
\end{equation}
For now we take displays \eqref{eqn:GfloorLog} and \eqref{eqn:GceilLog} as given and show why the Lemma holds. Using these two results, we have that display \eqref{eqn:finiteFinequality1} can be upper bounded by the following:
\begin{multline*}
	\leq \log \left( 1 + | \F | \right) \bigg( \max_{f \in \F} \underbrace{ 24 \pi_{\min}^{-(T-1)} \| f \|_\infty / \sqrt{n} }_{v(f)} \bigg) \\
	+ \sqrt{ \log \left( 1 + | \F | \right) } \bigg( \max_{f \in \F} \underbrace{ \sqrt{ 24 \pi_{\min}^{-(T-1)} \Estar{2:T} \left[ \rhostar{i}{2:T} f(\history{i}{T})^2 \right] } }_{\sqrt{u(f)}} \bigg)
\end{multline*}
\vspace{-5mm}
\begin{multline*}
	= 24 \pi_{\min}^{-(T-1)}  \left( \max_{f \in \F} \frac{ \| f \|_\infty }{ \sqrt{n} } \right) \log \left( 1 + | \F | \right) \\
	+ \sqrt{ 24 \pi_{\min}^{-(T-1)}  }\left( \max_{f \in \F} \sqrt{ \Estar{2:T} \left[ \rhostar{i}{2:T} f(\history{i}{T})^2 \right] } \right) \sqrt{ \log \left( 1 + | \F | \right) } 
\end{multline*}
Since $c_{\log} \triangleq \sup_{x \geq 2} \frac{ \log(1+x) }{ \log(x) }$ is bounded, $\frac{\log(1 + |\F|)}{\log(| \F |)} \leq c_{\log}$ so, $1 \leq c_{\log} \frac{\log(| \F |)}{\log(1 + |\F|)}$.
\begin{multline*}
	\leq c_{\log} 24 \pi_{\min}^{-(T-1)} \left( \max_{f \in \F} \frac{ \| f \|_\infty }{ \sqrt{n} } \right) \log \left( | \F | \right) \\
	+ \sqrt{ c_{\log} 24 \pi_{\min}^{-(T-1)}  }\left( \max_{f \in \F} \sqrt{ \Estar{2:T} \left[ \rhostar{i}{2:T} f(\history{i}{T})^2 \right] } \sqrt{ \log \left( | \F | \right) } \right)
\end{multline*}
\begin{multline*}
	\leq 24 \max \big( c_{\log}, c_{\log}^{1/2} \big) \bigg\{ \pi_{\min}^{-(T-1)} \left( \max_{f \in \F} \frac{ \| f \|_\infty }{ \sqrt{n} } \right) \log \left( | \F | \right) \\
	+  \sqrt{ \pi_{\min}^{-(T-1)} } \left( \max_{f \in \F} \sqrt{ \Estar{2:T} \left[ \rhostar{i}{2:T} f(\history{i}{T})^2 \right] } \right) \sqrt{ \log \left( | \F | \right) } \bigg\}.
\end{multline*}
The above implies that the desired result, display \eqref{eqn:maximalFinite}, holds. All that remains is to prove that displays \eqref{eqn:GfloorLog} and \eqref{eqn:GceilLog} hold. 

\proofSubsection{Proving display \eqref{eqn:GfloorLog} holds} %%%%%%%%%%%%%%%%%%%%%%%%%%%%%%
Let $x > 0$. We now state some results and discuss why they hold below. For all $n \geq 1$ we have that for any $f \in \F$, 
\begin{multline}
	\label{eqn:GunderInequality}
	\PP \left( \left| \under{\GG}_n(f) \right| \geq x \right) 
    \underbrace{ \leq }_{(a)} \PP \left( \left| \under{\GG}_n(f) \right| \geq \max \big\{ x, u(f)/v(f) \big\} \right) \\
	\underbrace{ \leq }_{(b)} 2 \exp \bigg( - 6 \frac{\max \big\{ x, u(f)/v(f) \big\}^2}{ u(f) + \max \big\{ x, u(f)/v(f) \big\} v(f) } \bigg) 
    \underbrace{\leq}_{(c)} 2 \exp \left( - 3 \frac{x}{ v(f) } \right).
\end{multline}
\begin{itemize}
    \item Inequality (a) holds because recall that $\under{ \GG_n }(f) \triangleq | \mathbb{G}_n (f) | \II_{ | \mathbb{G}_n (f) | > u(f)/v(f)}$. 
    %%%%%%%%%%%%%%%%%%%%%%
    \item Inequality (b) holds by Lemma \ref{lemma:bernstein} (Weighted Martingale Bernstein Inequality) since Condition \ref{cond:exploration} holds. Recall that $u(f) = 24 \pi_{\min}^{-(T-1)} \Estar{2:T} \left[ \rhostar{i}{2:T} f(\history{i}{T})^2 \right]$ and $v(f) = 24 \pi_{\min}^{-(T-1)} \| f \|_\infty / \sqrt{n}$.
    %%%%%%%%%%%%%%%%%%%%%%
    \item Inequality (c) holds because 
    \begin{multline*}
        6 \frac{\max \big\{ x, u(f)/v(f) \big\}^2}{ u(f) + \max \big\{ x, u(f)/v(f) \big\} v(f) }
        = 6 \frac{\max \big\{ x, u(f)/v(f) \big\}}{ u(f) / \max \big\{ x, u(f)/v(f) \big\} + v(f) } \\
        \geq 6 \frac{x}{ u(f) / \max \big\{ x, u(f)/v(f) \big\} + v(f) }
        \geq 6 \frac{x}{ u(f) / \big\{ u(f)/v(f) \big\} + v(f) }
        = 3 \frac{x}{v(f)}.
    \end{multline*}
\end{itemize}

\noindent We now show that the following is less than or equal to $1$:
\begin{equation*}
	\E \left[ e^{ \left| \under{ \GG_n }(f) \right| / v(f) } \right] - 1
	= \E \bigg[ \int_0^{ \left| \under{ \GG_n }(f) \right| / v(f) } e^x dx \bigg]
	= \E \left[ \int_0^\infty \II_{ x \leq \left| \under{ \GG_n }(f) \right| / v(f) } e^x dx \right]
\end{equation*}
Note the following:
\begin{itemize}
    \item Since $f$ is bounded and since $\rhohat{i}{2:T} \leq \pi_{\min}^{T-1}$ a.s. by Condition \ref{cond:exploration}, thus $\under{ \GG_n }(f)$ is bounded a.s. (remember $n$ is fixed).
    \item Since we are consider the cases in which $\| f \|_\infty > 0$, thus $v(f) > 0$. 
    \item By the above two results, $\under{ \GG_n }(f) / v(f)$ is bounded a.s., so $\E \left[ \int_0^\infty \II_{ x \leq \left| \under{ \GG_n }(f) \right| / v(f) } e^x dx \right]$ is also bounded.
\end{itemize}
Thus, by Fubini's theorem, we can exchange the integrals,
\begin{equation*}
	= \int_0^\infty \E \bigg[ \II_{ x \leq \left| \under{ \GG_n }(f) \right| / v(f) } \bigg] e^x dx 
	= \int_0^\infty \PP \bigg( \left| \under{ \GG_n }(f) \right| \geq x v(f) \bigg) e^x dx 
\end{equation*}
By display \eqref{eqn:GunderInequality},
\begin{equation*}
	\leq 2 \int_0^\infty e^{-3x} e^x dx  
	= 2 \int_0^\infty e^{-2x} dx 
	= 2 \left( \lim_{x \to \infty} -\frac{1}{2} e^{-2x} + \frac{1}{2} e^0 \right)
	= 2 \left( 0 + \frac{1}{2} \right)
	= 1.
\end{equation*}
Thus we have that for $\gamma(x) = e^x - 1$,
\begin{equation}
	\label{eqn:GunderInequality2}
	\E \left[ \gamma \big\{ \left| \under{ \GG_n }(f) \right| / v(f) \big\} \right] 
	= \E \left[ \exp \big\{ \left| \under{ \GG_n }(f) \right| / v(f) \big\} \right] - 1 \leq 1.
\end{equation}
Note that $\gamma(x) = e^x - 1$ is convex,
so by Jensen's inequality,  %$\exp( \E[ X ] ) \leq \E [ \exp(X) ]$, so
\begin{equation*}
	\exp \left( \E \left[ \max_{f \in \F} \left| \under{ \GG_n }(f) \right| / v(f) \right]  \right) - 1
	= \gamma \left( \E \left[ \max_{f \in \F} \left| \under{ \GG_n }(f) \right| / v(f) \right] \right)
\end{equation*}
\begin{equation*}
	\leq \E \left[ \gamma \left( \max_{f \in \F} \left| \under{ \GG_n }(f) \right| / v(f) \right) \right]
	\leq \sum_{f \in \F} \E \left[ \gamma \left\{ \left| \under{ \GG_n }(f) \right| / v(f) \right\} \right]
	\leq | \F |.
\end{equation*}
The last inequality above holds by display \eqref{eqn:GunderInequality2}. 
By adding $1$ and taking the $\log$ on both sides, we have display \eqref{eqn:GfloorLog} holds, i.e., that
\begin{equation*}
	\E \left[ \max_{f \in \F} \left| \under{ \GG_n }(f) \right| / v(f) \right] \leq \log \left( | \F | + 1 \right).
\end{equation*}

\proofSubsection{Proving display \eqref{eqn:GceilLog} holds} %%%%%%%%%%%%%%%%%%%%
Let $x > 0$. We now state some results and discuss why they hold below. For all $n \geq 1$ we have that for any $f \in \F$, 
\begin{multline}
	\label{eqn:GoverInequality}
	\PP \left( \left| \overline{\GG}_n(f) \right| \geq x \right) 
    \underbrace{ = }_{(a)} \begin{cases}
        \PP \left( \left| \overline{\GG}_n(f) \right| \geq x \right) & \TN{~if~} x < u(f) / v(f) \\
        0 & \TN{~if~} x \geq u(f) / v(f)
    \end{cases}
\end{multline}
\begin{equation*}
    \underbrace{ \leq }_{(b)} \begin{cases}
        2 \exp \left(-6 \frac{x^2}{u(f) + x v(f)} \right) & \TN{~if~} x < u(f) / v(f) \\
        0 & \TN{~if~} x \geq u(f) / v(f)
    \end{cases}
\end{equation*}
\begin{equation*}
    \underbrace{ \leq }_{(c)} \begin{cases}
        2 \exp \left(-3 \frac{x^2}{u(f)} \right) & \TN{~if~} x < u(f) / v(f) \\
        0 & \TN{~if~} x \geq u(f) / v(f)
    \end{cases}
    \leq 2 \exp \left(-3 \frac{x^2}{u(f)} \right).
\end{equation*}

\begin{itemize}
    \item Inequality (a) holds because recall that $\overline{ \GG_n }(f) \triangleq | \mathbb{G}_n (f) | \II_{ | \mathbb{G}_n (f) | \leq u(f)/v(f)}$. 
    %%%%%%%%%%%%%%%%%%%%%%
    \item Inequality (b) holds by Lemma \ref{lemma:bernstein} (Weighted Martingale Bernstein Inequality) since Condition \ref{cond:exploration} holds. Recall that $u(f) = 24 \pi_{\min}^{-(T-1)} \Estar{2:T} \left[ \rhostar{i}{2:T} f(\history{i}{T})^2 \right]$ and $v(f) = 24 \pi_{\min}^{-(T-1)} \| f \|_\infty / \sqrt{n}$.
    %%%%%%%%%%%%%%%%%%%%%%
    \item Inequality (c) holds because if $x < u(f) / v(f)$, then
    \begin{equation*}
        6 \frac{x^2}{u(f) + x v(f)}
        \geq 6 \frac{x^2}{u(f) + u(f) / v(f) v(f)}
        = 3 \frac{x^2}{u(f)}.
    \end{equation*}
\end{itemize}
\medskip
\noindent We now show that the following is less than or equal to $1$:
\begin{equation*}
	\E \left[ e^{ \left| \overline{ \GG_n }(f) \right|^2 / u(f) } \right] - 1
	= \E \bigg[ \int_0^{\left| \overline{ \GG_n }(f) \right|^2 / u(f) } e^x dx \bigg]
	= \E \left[ \int_0^\infty \II_{x \leq \left| \overline{ \GG_n }(f) \right|^2 / u(f) } e^x dx \right]
\end{equation*}
\begin{equation*}
	= \E \left[ \int_0^\infty \II_{\sqrt{ x u(f) } \leq \left| \overline{ \GG_n }(f) \right| } e^x dx \right]
\end{equation*}
Note the following:
\begin{itemize}
    \item Since $f$ is bounded and since $\rhohat{i}{2:T} \leq \pi_{\min}^{T-1}$ a.s. by Condition \ref{cond:exploration}, thus $\under{ \GG_n }(f)$ is bounded a.s. (remember $n$ is fixed).
    \item Since are considering the cases in which $\Estar{2:t} \left[ \rhostar{i}{2:t}  f(\history{i}{T})^2 \right] > 0$ (see discussion of special cases at the beginning of this proof), thus $u(f) > 0$. 
    \item By the above two results, $\under{ \GG_n }(f) / \sqrt{u(f)}$ is bounded a.s., so $\E \left[ \int_0^\infty \II_{\sqrt{ x u(f) } \leq \left| \overline{ \GG_n }(f) \right| } e^x dx \right]$ is also bounded.
\end{itemize}
Thus, by Fubini's theorem, we can exchange integrals,
\begin{equation*}
	= \int_0^\infty \E \left[ \II_{ \sqrt{ x u(f) } \leq \left| \overline{ \GG_n }(f) \right| } \right] e^x dx 
	= \int_0^\infty \PP \left( \left| \overline{ \GG_n }(f) \right| \geq \sqrt{x u(f)} \right) e^x dx 
\end{equation*}
By display \eqref{eqn:GoverInequality}, 
\begin{equation*}
	\leq 2 \int_0^\infty e^{-3x + x} dx  
	= 2 \int_0^\infty e^{-2x} dx 
	= 2 \left( \lim_{x \to \infty} -\frac{1}{2} e^{-2x} + \frac{1}{2} e^0 \right)
	= 2 \left( 0 + \frac{1}{2} \right)
	= 1.
\end{equation*}
Thus we have that for $\gamma_2(x) = e^{x^2} - 1$,
\begin{equation}
	\label{eqn:GoverInequality2}
	\gamma_2 \left( \left| \overline{ \GG_n }(f) \right| / \sqrt{u(f)} \right) 
	= \E \left[ \exp \left( \left| \overline{ \GG_n }(f) \right|^2 / u(f) \right) \right] - 1 \leq 1.
\end{equation}
Since $\gamma_2(x) = e^{x^2} - 1$ is convex, by Jensen's inequality,
\begin{equation*}
	\exp \bigg( \E \left[ \max_{f \in \F} \left| \overline{ \GG_n }(f) \right| / \sqrt{ u(f) } \right]^2  \bigg) - 1
	= \gamma_2 \bigg(\E \bigg[ \max_{f \in \F} \left| \overline{ \GG_n }(f) \right| / \sqrt{ u(f) } \bigg] \bigg) 
\end{equation*}
\begin{equation*}
	\leq \E \bigg[ \gamma_2 \left( \max_{f \in \F} \left| \overline{ \GG_n }(f) \right| / \sqrt{ u(f) } \right) \bigg]
	\leq \sum_{f \in \F} \E \bigg[ \gamma_2 \left( \left| \overline{ \GG_n }(f) \right| / \sqrt{ u(f) } \right) \bigg]
	\leq | \F |.
\end{equation*}
The last inequality above holds by display \eqref{eqn:GoverInequality2}. By adding $1$, taking the $\log$ and the square-root on both sides, we have display \eqref{eqn:GceilLog} holds, i.e., that
\begin{equation*}
	\E \bigg[ \max_{f \in \F} \left| \overline{ \GG_n }(f) \right| / \sqrt{u(f)} \bigg]
	\leq \sqrt{ \log \left( | \F | + 1 \right) }. ~~~ \blacksquare
\end{equation*}

%%%%%%%%%%%%%%%%%%%%%%%%%%%%%%%%%%%%%%%%%%%%
\subsection{Maximal Inequality as a Function of the Bracketing Integral (Lemma \ref{lemma:maximalBracketing})} %%%%%%%%%%%%%%%%%%%%%%%%%%%%%%%%%%%%%%%%%%%%
%%%%%%%%%%%%%%%%%%%%%%%%%%%%%%%%%%%%%%%%%%%%

\begin{lemma}[Maximal Inequality as a Function of the Bracketing Integral]
	\label{lemma:maximalBracketing}
	Let $\delta > 0$. Let $\F$ be a class of real-valued measurable functions of $\history{i}{t}$ such that \\
    $\int_{0}^1 \sqrt{ \log N_{[~]} \big( \epsilon, \F, L_{2} ( \Pstar ) \big) } d \epsilon < \infty$ and $\Estar{2:T}\big[ \rhostar{i}{2:T} f(\history{i}{T})^2 \big] \leq \delta^2$ for all $f \in \F$.

	Under Condition \ref{cond:exploration} (Minimum Exploration), for all $n \geq 1$,
	\begin{multline}
		\label{eqn:maximalBracketing}
		\E^* \bigg[ \sup_{f \in \F} \big| \GG_n (f) \big| \bigg] 
		\lesssim \int_{0}^\delta \sqrt{ \log N_{[~]} \big( \delta, \F, L_2 \big( \Pstar \big) \big) } d \epsilon \\
		+ \sqrt{n} \Estar{2:t} \left[ \rhostar{i}{2:t} F(\history{i}{t}) \II_{ F(\history{i}{t}) > \sqrt{n} a(\delta) }\right],
	\end{multline}
	where
	\begin{itemize}
	    \item $a(\delta) = \delta / \sqrt{ \log N_{[~]} \big( \delta, \F, L_2 ( \Pstar ) \big) }$
	    \item $\GG_n (f) \triangleq \frac{1}{ \sqrt{n} } \sum_{i=1}^n \left( \rhohat{i}{2:t} f( \history{i}{t} ) - \E\left[ \rhohat{i}{2:t} f( \history{i}{t} ) \right] \right)$
	    \item $F$ is an envelope where $\sup_{f \in \F} | f(\history{i}{t} ) | < F(\history{i}{t} ) < \infty$ with probability $1$.
	\end{itemize}
	Above $\lesssim$ means less than or equal to when scaled by universal positive constants. Above $\E^*$ refers to outer expectations as defined in Section 18.2 \cite{van2000asymptotic}.
\end{lemma}

%%%%%%%%%%%%%%%%%%%%%%%%%%%%%%%%%%%%%
\startproof{Lemma \ref{lemma:maximalBracketing} (Maximal Inequality as a Function of the Bracketing Integral)}
Our proof is almost identical to that of Lemma 19.34 \cite{van2000asymptotic} except that we use the maximal inequality in Lemma \ref{lemma:maximalFinite} instead of a maximal inequality for i.i.d. data; we include the full proof for clarity and completeness. For notational convenience we consider the $t$ set to $T$ case; the argument holds by the same argument for any $t \in [2 \colon T]$. \\

\noindent Note that by triangle inequality,
\begin{equation}
	\label{eqn:maximalBracketing1}
	\E^* \bigg[ \sup_{f \in \F} \left| \GG_n (f) \right| \bigg] \\
	\leq \E^* \bigg[ \sup_{f \in \F} \left| \GG_n  \left( f \II_{F > \sqrt{n} a(\delta)} \right) \right| \bigg]
	+ \E^* \bigg[ \sup_{f \in \F} \left| \GG_n \left( f \II_{F \leq \sqrt{n} a(\delta)} \right) \right| \bigg].
\end{equation}

\proofSubsection{Bounding First Term in display \eqref{eqn:maximalBracketing1}}
This term is to deal with potentially unbounded functions $f \in \F$.
\begin{equation*}
	\E^* \bigg[ \sup_{f \in \F} \left| \GG_n  \left( f \II_{F > \sqrt{n} a(\delta)} \right) \right| \bigg]
\end{equation*}
By using the definition of $\GG_n$,
\begin{equation*}
	= \E^* \bigg[ \sup_{f \in \F} \bigg| \frac{1}{ \sqrt{n} } \sum_{i=1}^n \bigg( \rhohat{i}{2:T} f(\history{i}{T}) \II_{F(\history{i}{T}) > \sqrt{n} a(\delta)} - \E \left[ \rhohat{i}{2:T} f(\history{i}{T}) \II_{F(\history{i}{T}) > \sqrt{n} a(\delta)} \right] \bigg) \bigg| \bigg]
\end{equation*}
By triangle inequality,
\begin{multline*}
	\leq \frac{1}{ \sqrt{n} } \sum_{i=1}^n \E^* \bigg[ \sup_{f \in \F} \left| \rhohat{i}{2:T} f(\history{i}{T}) \II_{F(\history{i}{T}) > \sqrt{n} a(\delta)} \right| \bigg] \\
	+ \frac{1}{ \sqrt{n} } \sum_{i=1}^n \sup_{f \in \F} \left\{ \left| \E \left[ \rhohat{i}{2:T} f(\history{i}{T}) \II_{F(\history{i}{T}) > \sqrt{n} a(\delta)} \right] \right| \right\}
\end{multline*}
By Jensen's inequality,
\begin{equation*}
	\leq 2 \frac{1}{ \sqrt{n} } \sum_{i=1}^n \E^* \bigg[ \sup_{f \in \F} \left( \rhohat{i}{2:T} \big| f(\history{i}{T}) \big| \II_{F(\history{i}{T}) > \sqrt{n} a(\delta)} \right) \bigg]
\end{equation*}
Recall our envelope function $F$ satisfies $| f(\history{i}{t} ) | < F(\history{i}{t} ) < \infty$ a.s., so 
\begin{equation*}
	\leq 2 \frac{1}{ \sqrt{n} } \sum_{i=1}^n \E \left[ \rhohat{i}{2:T} F(\history{i}{T}) \II_{F(\history{i}{T}) > \sqrt{n} a(\delta)} \right]
\end{equation*}
\begin{equation*}
	= 2 \frac{1}{ \sqrt{n} } \sum_{i=1}^n \E_{ \pistar{2:T} } \left[ \rhostar{i}{2:T} F(\history{i}{T}) \II_{F(\history{i}{T}) > \sqrt{n} a(\delta)} \right]
\end{equation*}
Since the expectation above is indexed by the deterministic policy $\pistar{2:T}$, $\history{i}{T}$ within the expectation are i.i.d.
\begin{equation*}
	= 2 \sqrt{n} \E_{ \pistar{2:T} } \left[ \rhostar{i}{2:T} F(\history{i}{T}) \II_{F(\history{i}{T}) > \sqrt{n} a(\delta)} \right]
\end{equation*}
This us gives us the second part of the bound from display \eqref{eqn:maximalBracketing}.

\proofSubsection{Bounding Second Term in display \eqref{eqn:maximalBracketing1}}
We now focus on bounding the following:
\begin{equation*}
	\E^* \bigg[ \sup_{f \in \F} \left| \GG_n f \II_{ F \leq \sqrt{n} a(\delta)} \right| \bigg]
\end{equation*}

\noindent We now consider the class of functions $\bar{\F} := \left\{ f \II_{F \leq \sqrt{n} a(\delta)} : f \in \F \right\}$.
We first show that \\
$N_{[~]}\big( \epsilon, \bar{\F}, L_2( \Pstar ) \big) \leq N_{[~]}\big( \epsilon, \F, L_2( \Pstar ) \big)$.
\begin{itemize}
	\item By definition of bracketing numbers, we can cover $\F$ with $N_{[~]}\big( \epsilon, \F, L_2( \Pstar ) \big)$ brackets, each with size at most $\epsilon$. Specifically, we can find brackets $[l_j, u_j]$ for \\
    $j \in \left[ 1 \colon N_{[~]}\big( \epsilon, \F, L_2( \Pstar ) \big) \right]$ that cover $\F$ such that $\Estar{2:T} \left[ (u_j - l_j)^2 \right]^{1/2} \leq \epsilon$ for all brackets $[l_j, u_j]$.
	\item Note that brackets $\left[ l_j \II_{F \leq \sqrt{n} a(\delta)}, ~ u_j \II_{F \leq \sqrt{n} a(\delta)} \right]$ for some $j \in \left[ 1 \colon N_{[~]}\big( \epsilon, \F, L_2( \Pstar ) \big) \right]$ cover $\bar{\F}$.
	\item Additionally, note that $\Estar{2:T} \left[ \big( [u_j - l_j] \II_{ F \leq \sqrt{n} a(\delta)} \big)^2 \right]
	\leq \Estar{2:T} \left[ (u_j - l_j)^2 \right]
	\leq \epsilon^2$.
\end{itemize}
Thus, we have that 
\begin{equation}
	\label{eqn:bracketingFbarbound}
	N_{[~]}\big( \epsilon, \bar{\F}, L_2( \Pstar ) \big) \leq N_{[~]}\big( \epsilon, \F, L_2( \Pstar ) \big).
\end{equation}

%%%%%%%%%%%%%%%%%%%%%%%%%%%%%%%%%%%%%%%%%%%
\proofSubsection{Desiderata for Nested Partitions}
We now assume the existence of nested partitions of $\bar{\F}$ that satisfy certain conditions. We will finish the proof assuming these partitions exist and conclude by constructing these partitions.

High level, we assume we have nested partitions of $\bar{\F}$ that are indexed by positive integers $q$. These partitions are designed to become increasingly fine-grained as $q$ increases. Specifically the ``size'' of each piece of the partition will be on the order of $2^{-q}$, i.e., the ``size'' of the partitions will halve as $q$ increases by $1$. The partitions are nested in that each partition piece at level $q+1$ is a subset of some partition piece at level $q$. 

We pick $q_0$ to be a positive integer such that
\begin{equation}
    \label{app:q0deltachoice}
    \delta < 2^{-q_0} \leq 2 \delta.
\end{equation}
For every integer $q \geq q_0$ we have a partition of $\bar{\F}$; we denote these partions as $\{ \bar{\F}_{q,j} \}_{j=1}^{N_q}$. We assume that $N_{q_0} = N_{[~]} \big( 2^{-q_0}, \bar{\F}, L_2( \Pstar ) \big)$. These partitions are nested in that for each $q \geq q_0 +1$ and for every $j \in [1 \colon N_q]$, we have that the partition piece $\bar{\F}_{q,j}$ is a subset of some partition piece $\bar{\F}_{q-1,k}$ for some $k \in [1 \colon N_{q-1}]$. Moreover, we further assume the following:
\begin{itemize}
	\item \bo{Requirement on the ``size'' of partition pieces:}
		For each partition $q$ and partition piece $j \in [1 \colon N_q]$, let $\Delta_{q,j}$ be a measurable function of $\history{i}{T}$ such that $\sup_{f,g \in \bar{\F}_{q,j}} | f- g| \leq \Delta_{q,j}$ a.s. and
		\begin{equation}
			\label{eqn:Deltabound}
			  \Estar{2:T} \left[ \rhostar{i}{2:T} \Delta_{q,j} ( \HH_{T}^{(i)} )^2  \right]
			\leq 2^{-2q}.
		\end{equation}
	\item \bo{Requirement on how the number of partition pieces grows as the ``size'' goes to zero:} 
		\begin{equation}
			\label{eqn:integralBound}
			\sum_{q=q_0}^\infty 2^{-q} \sqrt{ \log N_q } \lesssim 
			\int_0^\delta \sqrt{ \log N_{[~]}\big( \epsilon, \bar{\F}, L_2( \Pstar ) \big)} d \epsilon.
		\end{equation}
\end{itemize}
We construct nested partitions that satisfy the above conditions at the end of this proof. For now, we assume such nested partitions described above exist and we continue with the argument. 

%%%%%%%%%%%%%%%%%%%%%%%%%%%%%%%%%%%%%%%%%%%
\proofSubsection{Main Argument Assuming Desired Nested Partitions Exist}
For every partition piece $\bar{\F}_{q,j}$, we choose a arbitrary point $\bar{f}_{q,j}$ in that partition piece, i.e., for each $q \geq q_0$ and every $j \in [1 \colon N_q]$ we choose a point $\bar{f}_{q,j} \in \F_{q,j}$. We also define functions $\lambda_q : \bar{\F} \mapsto \bar{\F}$ that maps each function $\bar{f} \in \bar{\F}$ to these points $\{ \bar{f}_{q,j} \}_{j=1}^{N_q}$; specifically, for any $\bar{f} \in \bar{\F}$ we can find some partition piece $\bar{\F}_{q,j}$ such that $\bar{f} \in \bar{\F}_{q,j}$ and we map that $\bar{f}$ to the point $\bar{f}_{q,j}$. \\

\noindent Note that for any integer $Q > q_0$, by telescoping series, for any $\bar{f} \in \bar{\F}$,
\begin{equation*}
	\bar{f}(\history{i}{T}) = \lambda_{q_0} \bar{f}(\history{i}{T}) 
	+ \sum_{q=q_0}^{Q} \left\{ \lambda_{q+1} \bar{f}(\history{i}{T}) - \lambda_q \bar{f}(\history{i}{T}) \right\} 
	+ \bar{f}(\history{i}{T}) - \lambda_{Q+1} \bar{f}(\history{i}{T})
\end{equation*}
\begin{multline}
	\label{eqn:chainDecomp}
	= \lambda_{q_0} \bar{f}(\history{i}{T}) 
	+ \sum_{q=q_0}^\infty \II_{q \leq Q} \left\{ \lambda_{q+1} \bar{f}(\history{i}{T}) - \lambda_q \bar{f}(\history{i}{T}) \right\} \\
	+ \sum_{q=q_0}^\infty \II_{q = Q+1} \left\{ \bar{f}(\history{i}{T}) - \lambda_q \bar{f}(\history{i}{T}) \right\}.
\end{multline}

\noindent For any $\bar{f} \in \bar{\F}$, we define $Q_{\bar{f}}(\history{i}{T}) \in [q_0, \infty]$ to be a random variable representing the maximum partition level with no bound violations up to that level. Specifically,
\begin{equation}
    \label{app:nestedPartitionQdef}
	Q_{\bar{f}}(\history{i}{T}) \triangleq \bigg\{ \sup_{q \geq q_0}  \TN{ s.t. } \sum_{j=1}^{N_p} \II_{ \bar{f} \in \bar{\F}_{p,j} } \Delta_{p,j} (\history{i}{T}) \leq \sqrt{n} 2^{-p} / \sqrt{ \log N_{p} } \TN{ for all } p \in [q_0 : q] \bigg\},
\end{equation}
Thus, by replacing $Q$ with $Q_{\bar{f}}$ and by applying $\GG_n$ to both sides of display \eqref{eqn:chainDecomp},
\begin{equation*}
	\GG_n (\bar{f})
	= \GG_n (\lambda_{q_0} \bar{f} ) 
	+ \sum_{q=q_0}^\infty \GG_n \left( \II_{q \leq Q_{\bar{f}}} \left( \lambda_{q+1} \bar{f} - \lambda_q \bar{f} \right) \right)
	+ \sum_{q=q_0}^\infty \GG_n \left( \II_{q = Q_{\bar{f}}+1} \left( \bar{f} - \lambda_q \bar{f} \right) \right)
\end{equation*}

\noindent Thus, we have that by triangle inequality
\begin{multline}
	\label{eqn:equicontinuityBounds}
	\E^* \bigg[ \sup_{f \in \F} \left| \GG_n f \II_{ F \leq \sqrt{n} a(\delta)} \right| \bigg]
    = \E^* \bigg[ \sup_{\bar{f} \in \bar{\F}} \left| \GG_n (\bar{f}) \right| \bigg] \\
	\leq \underbrace{ \E^* \bigg[ \sup_{\bar{f} \in \bar{\F}} \left| \GG_n (\lambda_{q_0} \bar{f}) \right| \bigg] }_{(i)} 
	+ \underbrace{ \E^* \bigg[ \sup_{\bar{f} \in \bar{\F}} \bigg| \sum_{q=q_0}^\infty \GG_n \left( \II_{q \leq Q_{\bar{f}}} \left( \lambda_{q+1} \bar{f} - \lambda_{q} \bar{f} \right) \right) \bigg| \bigg] }_{(ii)} \\
	+ \underbrace{ \E^* \bigg[ \sup_{\bar{f} \in \bar{\F}} \bigg| \sum_{q=q_0}^\infty \GG_n \left( \II_{q = Q_{\bar{f}}+1} \left( \bar{f} - \lambda_q \bar{f} \right) \right) \bigg| \bigg]. }_{(iii)} 
\end{multline}

\noindent Below we will show the following results:
\begin{itemize}
	\item Bounding term (i)
	\begin{equation}
		\label{eqn:equicontinuityBound(i)}
		\E^* \bigg[ \sup_{\bar{f} \in \bar{\F}} \left| \GG_n (\lambda_{q_0} \bar{f}) \right| \bigg]
		\lesssim 2 \pi_{\min}^{-(T-1)} 2^{-q_0} \sqrt{ \log N_{q_0} }
	\end{equation}
	\item Bounding term (ii) 
	\begin{equation}
		\label{eqn:equicontinuityBound(ii)}
		\E^* \bigg[ \sup_{\bar{f} \in \bar{\F}} \bigg| \sum_{q=q_0}^\infty \GG_n \big( \II_{q \leq Q_{\bar{f}}} \left( \lambda_{q+1} \bar{f} - \lambda_{q} \bar{f} \right) \big) \bigg| \bigg]
		\lesssim 2 \pi_{\min}^{-(T-1)} \sum_{q=q_0}^\infty 2^{-q} \sqrt{ \log N_{q} }
	\end{equation}
	\item Bounding term (iii)
	\begin{equation}
		\label{eqn:equicontinuityBound(iii)}
		\E^* \bigg[ \sup_{\bar{f} \in \bar{\F}} \bigg| \sum_{q=q_0}^\infty \GG_n \left( \II_{q = Q_{\bar{f}}+1} \left( \bar{f} - \lambda_q \bar{f} \right) \right) \bigg| \bigg]
		\lesssim 4 \pi_{\min}^{-(T-1)} \sum_{q=q_0}^\infty 2^{-q} \sqrt{ \log N_{q} }.
	\end{equation}
\end{itemize}

\noindent For now we assume the above three displays hold (we show they hold later). Thus, we can upper bound display \eqref{eqn:equicontinuityBounds} as follows:
\begin{equation*}
	\E^* \bigg[ \sup_{\bar{f} \in \bar{\F}} \left| \GG_n (\bar{f}) \right| \bigg]
	\lesssim 2 \pi_{\min}^{-(T-1)} 2^{-q_0} \sqrt{ \log N_{q_0} } 
	+ 6 \pi_{\min}^{-(T-1)} \sum_{q=q_0}^\infty 2^{-q} \sqrt{ \log N_{q} }
\end{equation*}
\begin{equation*}
	\leq 8 \pi_{\min}^{-(T-1)} \sum_{q=q_0}^\infty 2^{-q} \sqrt{ \log N_{q} }
\end{equation*}
By display \eqref{eqn:integralBound},
\begin{equation*}
	\lesssim \pi_{\min}^{-(T-1)} \int_0^{\delta} \sqrt{ \log ( N_{[~]} \big( \epsilon, \bar{\F}, L_2( \Pstar ) \big) } d \epsilon
	\leq \pi_{\min}^{-(T-1)} \int_0^{\delta} \sqrt{ \log ( N_{[~]} \big( \epsilon, \F, L_2( \Pstar ) \big) } d \epsilon.
\end{equation*}
The last inequality above holds by display \eqref{eqn:bracketingFbarbound}.
We now show that displays \eqref{eqn:equicontinuityBound(i)}, \eqref{eqn:equicontinuityBound(ii)}, and \eqref{eqn:equicontinuityBound(iii)} hold.

%%%%%%%%%%%%%%%%%%%%%%%%%%%%%%%%%%%%%%%%%%%
\proofSubsubsection{Display \eqref{eqn:equicontinuityBound(i)}: Bounding term (i) }
\begin{equation*}
	\E^* \bigg[ \sup_{\bar{f} \in \bar{\F}} \left| \GG_n (\lambda_{q_0} \bar{f}) \right| \bigg]
	= \E \bigg[ \max_{ j \in [1 \colon N_{q_0}] } \left| \GG_n (\bar{f}_{q_0, j}) \right| \bigg]
\end{equation*}

\noindent By Lemma \ref{lemma:maximalFinite} (Maximal Inequality for Finite Class of Functions) for any $n \geq 1$,
\begin{multline}
	\label{eqn:asymptoticEquicontinuity1}
	\lesssim \pi_{\min}^{-(T-1)} \max_{ j \in [1 \colon N_{q_0}] } \frac{ \| \bar{f}_{q_0, j} \|_\infty }{ \sqrt{n} } \log N_{q_0} \\
	+ \sqrt{\pi_{\min}^{-(T-1)}} \max_{ j \in [1 \colon N_{q_0}] } \sqrt{ \Estar{2:T} \left[ \rhostar{i}{2:T} \bar{f}_{q_0, j}(\history{i}{T})^2 \right] } \sqrt{ \log N_{q_0} }.
\end{multline}

\begin{itemize}	
	\item Note that since $\bar{f}_{q_0,j}(\history{i}{T}) = f_{q_0,j}(\history{i}{T}) \II_{F(\history{i}{T}) \leq \sqrt{n} a(\delta)} \leq \sqrt{n} a(\delta)$ a.s., we get the first inequality below:
    \begin{equation*}
		\max_{ j \in [1 \colon N_{q_0}] } \big\{ \| \bar{f}_{q_0, j} \|_\infty \big\}
		\leq \sqrt{n} a(\delta) 
        = \sqrt{n} \delta / \sqrt{ \log N_{[~]} \big( \delta, \F, L_2( \Pstar ) \big) }
        \leq \sqrt{n} 2^{-q_0} / \sqrt{ \log N_{q_0} }.
	\end{equation*}
    The last inequality above holds because $\delta < 2^{-q_0} \leq 2 \delta$ from display \eqref{app:q0deltachoice} and
    \begin{equation*}
        N_{[~]} \big( \delta, \F, L_2( \Pstar ) \big)
        \geq N_{[~]} \big( 2^{-q_0}, \F, L_2( \Pstar ) \big)
        \geq N_{[~]} \big( 2^{-q_0}, \bar{\F}, L_2( \Pstar ) \big) 
        = N_{q_0}.
    \end{equation*} 
    The last inequality above holds by display \eqref{eqn:bracketingFbarbound}.
	\item $\max_{j \in [1 \colon N_q]} \sqrt{ \Estar{2:T} \left[ \rhostar{i}{2:T} \bar{f}_{q_0, j}(\history{i}{T})^2 \right] }
	\leq \delta  \leq 2^{-q_0}$; the first inequality holds by assumption of the Lemma and the second inequality holds since we choose $q_0$ such that $\delta < 2^{-q_0} \leq 2 \delta$.
\end{itemize}
Thus, by the above bullets, we can upper bound display \eqref{eqn:asymptoticEquicontinuity1} as follows:
\begin{equation*}
	\leq \pi_{\min}^{-(T-1)} 2^{-q_0} \sqrt{ \log N_{q_0} }
	+ \sqrt{\pi_{\min}^{-(T-1)}} 2^{-q_0} \sqrt{ \log N_{q_0} }
    \leq 2 \pi_{\min}^{-(T-1)} 2^{-q_0} \sqrt{ \log N_{q_0} }.
\end{equation*}

\noindent Thus, we have that display \eqref{eqn:equicontinuityBound(i)} holds.

%%%%%%%%%%%%%%%%%%%%%%%%%%%%%%%%%%%%%%%%%%%
\proofSubsubsection{Bounding term (ii):}
By triangle inequality,
\begin{equation*}
	\E^* \bigg[ \sup_{\bar{f} \in \bar{\F}} \bigg| \sum_{q=q_0}^\infty \GG_n \big( \II_{q \leq Q_{\bar{f}}} \left( \lambda_{q+1} \bar{f} - \lambda_{q} \bar{f} \right) \big) \bigg| \bigg]
	\leq \sum_{q=q_0}^\infty \E^* \bigg[ \sup_{\bar{f} \in \bar{\F}} \bigg| \GG_n \left( \II_{q \leq Q_{\bar{f}}} \left( \lambda_{q+1} \bar{f} - \lambda_{q} \bar{f} \right) \right) \bigg| \bigg]
\end{equation*}
By the definition of $Q_{\bar{f}}$ from display \eqref{app:nestedPartitionQdef}, $\II_{q \leq Q_{\bar{f}}} = \II_{q \leq Q_{\lambda_q \bar{f}}}$,
\begin{equation*}
	= \sum_{q=q_0}^\infty \E \bigg[ \max_{j \in [1 \colon N_q]} \bigg| \GG_n \left( \II_{q \leq Q_{\bar{f}_{q,j}}} \left( \bar{f}_{q+1,j} - \bar{f}_{q,j} \right) \right) \bigg| \bigg]
\end{equation*}
\noindent By Lemma \ref{lemma:maximalFinite} (Maximal Inequality for Finite Class of Functions) for all sufficiently large $n$,
\begin{multline}
	\label{eqn:asymptoticEquicontinuity2}
	\lesssim \sum_{q=q_0}^\infty \pi_{\min}^{-(T-1)} \max_{ j \in [1 \colon N_{q_0}] } \frac{ \left\| \II_{q \leq Q_{\bar{f}_{q,j}}} \left( \bar{f}_{q+1,j} - \bar{f}_{q,j} \right) \right\|_\infty }{ \sqrt{n} } \log N_{q} \\
	+ \sum_{q=q_0}^\infty \sqrt{\pi_{\min}^{-(T-1)}} \max_{ j \in [1 \colon N_{q}] } \sqrt{ \Estar{2:T} \left[ \rhostar{i}{2:T} \II_{q \leq Q_{\bar{f}_{q,j}}} \left( \bar{f}_{q+1,j} - \bar{f}_{q,j} \right)^2 \right] } \sqrt{ \log N_{q} }.
\end{multline}

\begin{itemize}
	\item Note that by the definition of $\II_{q \leq Q_{\bar{f}_{q,j}}}$ ($Q_{\bar{f}}$ was defined in display \eqref{app:nestedPartitionQdef}) and by our nested partitions, we have that 
	\begin{multline*}
	    \left\| \II_{q \leq Q_{\bar{f}_{q,j}}} \left( \bar{f}_{q+1,j} - \bar{f}_{q,j} \right) \right\|_{\infty} 
	    \leq \sup_{\bar{f},\bar{f}' \in \bar{\F}_{q,j}} \left\| \II_{q \leq Q_{\bar{f}_{q,j}}} | f - f' | \right\|_{\infty} \\
	    \leq \left\| \II_{q \leq Q_{\bar{f}_{q,j}}}\Delta_{q,j}( \history{i}{T} ) \right\|_{\infty} 
	    \leq \sqrt{n} 2^{-q} / \sqrt{ \log N_{q} }.
	\end{multline*}
 
	\item By our nested partitions, we have that $\lambda_{q+1} \bar{f}, \lambda_{q} \bar{f}$ are in the same $q^{\mathrm{th}}$-level partition piece, i.e., $\lambda_{q+1} \bar{f}, \lambda_{q} \bar{f} \in \bar{\F}_{q,j}$ for some $\bar{\F}_{q,j}$ with $j \in [1 \colon N_{q}]$. Thus,
	\begin{multline*}
		\Estar{2:T} \left[ \II_{q \leq Q_{\bar{f}_{q,j}}( \history{i}{T} ) } \rhostar{i}{2:T} \left( \bar{f}_{q+1,j}( \history{i}{T} ) - \bar{f}_{q,j}( \history{i}{T} ) \right)^2 \right] \\
		\leq \sup_{\bar{f},\bar{f}' \in \bar{\F}_{q,j}} \Estar{2:T} \left[ \rhostar{i}{2:T} \left( \bar{f}( \history{i}{T} ) - \bar{f}'( \history{i}{T} ) \right)^2 \right] \\
		\leq \Estar{2:T} \left[ \rhostar{i}{2:T} \Delta_{q,j}( \history{i}{T} )^2 \right].
		%\leq 2^{-2q}.
	\end{multline*}
	Moreover, by properties of our partitions,
	\begin{multline*}
	    \max_{j \in [1 \colon N_q]} \sqrt{\Estar{2:T} \left[ \II_{q \leq Q_{\bar{f}_{q,j}}( \history{i}{T} ) } \rhostar{i}{2:T} \left( \bar{f}_{q+1,j}( \history{i}{T} ) - \bar{f}_{q,j}( \history{i}{T} ) \right)^2 \right]} \\
	    \leq \max_{j \in [1 \colon N_q]} \sqrt{ \Estar{2:T} \left[ \rhostar{i}{2:T} \Delta_{q,j}( \history{i}{T} )^2 \right] } \leq 2^{-q}.
	\end{multline*}
	The last inequality above holds by the size property of our partitions from display \eqref{eqn:Deltabound}.
\end{itemize}

%\noindent By Lemma \ref{lemma:maximalFinite}, a maximal inequality for finite classes of functions,
\noindent %Since $a_{q} \triangleq 2^{-q} / \sqrt{ \log N_{q+1} }$ and $N_{q+1} \geq N_{q}$, 
By the above bullets, we have that display \eqref{eqn:asymptoticEquicontinuity2} is upper bounded by the following:
\begin{equation*}
	\leq \sum_{q=q_0}^\infty \left\{ \pi_{\min}^{-(T-1)} 2^{-q} \sqrt{ \log N_{q} } 
	+ \sqrt{\pi_{\min}^{-(T-1)}} 2^{-q} \sqrt{ \log N_q } \right\}
	\leq 2 \pi_{\min}^{-(T-1)} \sum_{q=q_0}^\infty 2^{-q} \sqrt{ \log N_{q} }.
\end{equation*}
Thus, we have that display \eqref{eqn:equicontinuityBound(ii)} holds.

%%%%%%%%%%%%%%%%%%%%%%%%%%%%%%%%%%%%%%%%%%%
\proofSubsubsection{Bounding term (iii):}
By triangle inequality,
\begin{equation}
	\label{eqn:termiii}
	\E^* \bigg[ \sup_{\bar{f} \in \bar{\F}} \bigg| \sum_{q=q_0}^\infty \GG_n \left( \II_{q = Q_{\bar{f}}+1} \left( \bar{f} - \lambda_q \bar{f} \right) \right) \bigg| \bigg]
	\leq \sum_{q=q_0}^\infty \E^* \bigg[ \sup_{\bar{f} \in \bar{\F}} \bigg| \GG_n \left( \II_{q = Q_{\bar{f}}+1} \left( \bar{f} - \lambda_q \bar{f} \right) \right) \bigg| \bigg]
\end{equation}
Note that if some functions $f, g$ are such that $\big| f(\history{i}{T}) \big| \leq g(\history{i}{T})$ a.s., then
\begin{multline}
    \label{app:finalMaxInequalityObs}
	\left| \GG_n (f) \right| \leq \frac{1}{\sqrt{n}} \left| \sum_{i=1}^n \rhohat{i}{2:t} f( \history{i}{T} ) \right| + \sqrt{n} \left| \E \big[ \rhohat{i}{2:t} f( \history{i}{T} ) \big] \right| \\
	\leq \frac{1}{\sqrt{n}} \left| \sum_{i=1}^n \rhohat{i}{2:t} g( \history{i}{T} ) \right| + \sqrt{n} \left| \E \left[ \rhohat{i}{2:t} g( \history{i}{T} ) \right] \right| \\
	\leq \left| \GG_n (g) \right| + 2 \sqrt{n} \left| \E \left[ \rhohat{i}{2:t} g( \history{i}{T} ) \right] \right| \TN{~~a.s.}
\end{multline}

Note that $\left| \II_{q = Q_{\bar{f}}( \history{i}{T} )+1} \big( \bar{f}( \history{i}{T} ) - \lambda_q \bar{f}( \history{i}{T} ) \big) \right|
\leq \II_{q = Q_{\bar{f}}( \history{i}{T} )+1} \sum_{j=1}^{N_q} \II_{\bar{f} \in \bar{\F}_{q,j}} \Delta_{q,j}( \history{i}{T} )$ a.s. by the definition of $\Delta_{q,j}$ from above display \eqref{eqn:Deltabound}. So by the observation from display \eqref{app:finalMaxInequalityObs} we can upper bound display \eqref{eqn:termiii} as follows:
\begin{multline*}
	\leq \sum_{q=q_0}^\infty \E^* \bigg[ \sup_{\bar{f} \in \bar{\F}} \bigg| \GG_n \bigg( \II_{q = Q_{\bar{f}}( \history{i}{T} )+1} \sum_{j=1}^{N_q} \II_{\bar{f} \in \F_{q,j}} \Delta_{q,j}( \history{i}{T} ) \bigg) \bigg| \bigg] \\
	+ 2 \sqrt{n} \sum_{q=q_0}^\infty \sup_{\bar{f} \in \bar{\F}} \bigg| \E \bigg[ \rhohat{i}{2:t} \II_{q = Q_{\bar{f}}( \history{i}{T} )+1} \sum_{j=1}^{N_q} \II_{\bar{f} \in \F_{q,j}} \Delta_{q,j}( \history{i}{T} ) \bigg] \bigg|
\end{multline*}
Since $\II_{q \leq Q_{\bar{f}}+1} = \II_{q \leq Q_{\lambda_q \bar{f}}+1}$ ($Q_{\bar{f}}$ was defined in display \eqref{app:nestedPartitionQdef}),
\begin{multline}
	\label{eqn:termiii2}
	= \sum_{q=q_0}^\infty \E \bigg[ \max_{j \in [1 \colon N_q]} \bigg| \GG_n \left( \II_{q = Q_{\bar{f}_{q,j}}( \history{i}{T} )+1} ~ \Delta_{q,j}( \history{i}{T} ) \right) \bigg| \bigg] \\
	+ 2 \sqrt{n} \sum_{q=q_0}^\infty \max_{j \in [1 \colon N_q]} \E \bigg[ \rhohat{i}{2:t} \II_{q = Q_{\bar{f}_{q,j}}( \history{i}{T} )+1} \Delta_{q,j}( \history{i}{T} ) \bigg]
\end{multline}

\begin{itemize}
    \item Due to the nested property of our partitions (see above display \eqref{eqn:Deltabound} for the definition of $\Delta_{q,j}$),
    \begin{equation*}
        \II_{q = Q_{\bar{f}_{q,j}}( \history{i}{T} )+1} ~ \Delta_{q,j}( \history{i}{T} )
        \leq \II_{q = Q_{\bar{f}_{q,j}}( \history{i}{T} )+1} ~ \Delta_{q-1,j}( \history{i}{T} ) ~~\TN{a.s.}
    \end{equation*}
    Moreover, by the definition of $Q_{\bar{f}_{q,j}}$ from display \eqref{app:nestedPartitionQdef},
    \begin{equation*}
        \II_{q = Q_{\bar{f}_{q,j}}( \history{i}{T} )+1} ~ \Delta_{q-1,j}( \history{i}{T} )
        \leq \sqrt{n} 2^{-(q-1)} / \sqrt{ \log N_{q-1} } ~~\TN{a.s.}
    \end{equation*}
	Thus, by Lemma \ref{lemma:maximalFinite} (Maximal Inequality for Finite Class of Functions) and display \eqref{eqn:Deltabound},
	\begin{multline*}
		\sum_{q=q_0}^\infty \E^* \bigg[ \max_{j \in [1 \colon N_q]} \bigg| \GG_n \big( \II_{q = Q_{\bar{f}_{q,j}}( \history{i}{T} )+1} ~ \Delta_{q,j}( \history{i}{T} ) \big) \bigg| \bigg] \\
		\lesssim \sum_{q=q_0}^\infty \bigg\{ \pi_{\min}^{-(T-1)} 2^{-(q-1)} / \sqrt{ \log N_{q-1} } \log N_q
	    + \sqrt{\pi_{\min}^{-(T-1)}} 2^{-q} \sqrt{ \log N_q } \bigg\}
     \end{multline*}
     Since $N_{q-1} \leq N_q$ (bracketing number for brackets of size $2^{-(q-1)}$ vs $2^{-q}$), thus \\
     $\log N_{q-1} \leq \log N_{q}$ and $1 \leq \sqrt{ \log N_{q} / \log N_{q-1} }$. 
     So,
     \begin{equation*}
        \leq \sum_{q=q_0}^\infty \bigg\{ \pi_{\min}^{-(T-1)} 2^{-(q-1)} \sqrt{ \log N_q }
	    + \sqrt{\pi_{\min}^{-(T-1)}} 2^{-q} \sqrt{ \log N_q } \bigg\}
        \leq \pi_{\min}^{-(T-1)} 2 \sum_{q=q_0}^\infty 2^{-q} \sqrt{ \log N_q }.
	\end{equation*}
 
	\item Note that by our nested partitions and by the definition of $Q_{\bar{f}_{q,j}}$ from display \eqref{app:nestedPartitionQdef}, if $\II_{q = Q_{\bar{f}_{q,j}}( \history{i}{T} )+1} = 1$, then $\Delta_{q,j}( \history{i}{T} ) > \sqrt{n} 2^{-q} / \sqrt{ \log N_{q} }$. Thus, when $\II_{q = Q_{\bar{f}_{q,j}}( \history{i}{T} )+1} = 1$, $\Delta_{q,j}( \history{i}{T} ) \big( \sqrt{n} 2^{-q} / \sqrt{ \log N_{q} } \big)^{-1} \geq 1$. Thus,
    \begin{equation*}
        \II_{q = Q_{\bar{f}_{q,j}}( \history{i}{T} )+1} \Delta_{q,j}( \history{i}{T} )
        \leq \Delta_{q,j}( \history{i}{T} )^2 \big( \sqrt{n} 2^{-q} / \sqrt{ \log N_{q} } \big)^{-1}.
    \end{equation*}
    Thus we have that
	\begin{multline*}
		2 \sqrt{n} \sum_{q=q_0}^\infty \max_{j \in [1 \colon N_q]} \bigg| \E \bigg[ \rhohat{i}{2:t} \II_{q = Q_{\bar{f}_{q,j}}( \history{i}{T} )+1} \Delta_{q,j}( \history{i}{T} ) \bigg] \bigg| \\
		\leq \sum_{q=q_0}^\infty \frac{ 2  \sqrt{ \log N_{q} } }{ 2^{-q} } \max_{j \in [1 \colon N_q]} \E \bigg[ \rhohat{i}{2:t} \Delta_{q,j}( \history{i}{T} )^2 \bigg] \\
		\leq \sum_{q=q_0}^\infty \frac{ 2  \sqrt{ \log N_{q} } }{ 2^{-q} } \cdot 2^{-2q}
        = 2 \sum_{q=q_0}^\infty 2^{-q} \sqrt{ \log N_{q} }.
	\end{multline*}
	The last inequality above holds by the size property of our partitions from display \eqref{eqn:Deltabound}.
\end{itemize}

\noindent By the observations in the above bullets, we can upper bound display \eqref{eqn:termiii2} as follows:
\begin{equation*}
    \lesssim \pi_{\min}^{-(T-1)} 2 \sum_{q=q_0}^\infty 2^{-q} \sqrt{ \log N_q } 
    + 2 \sum_{q=q_0}^\infty 2^{-q} \sqrt{ \log N_{q} }
\end{equation*}
\begin{equation*}
    \leq \pi_{\min}^{-(T-1)} 4 \sum_{q=q_0}^\infty 2^{-q} \sqrt{ \log N_q } 
\end{equation*}
Thus, we have that display \eqref{eqn:equicontinuityBound(iii)} holds.

\proofSubsection{Construct nested partitions}
We now construct nested partitions that satisfy the conditions described previously, particularly displays \eqref{eqn:Deltabound} and \eqref{eqn:integralBound}. \\

\noindent By our bracketing number assumption, for every integer $q \geq q_0$, we can find \\
$N_q^* \triangleq N_{[~]} \big( 2^{-q} \pi_{\min}^{(T-1)/2}, \bar{\F}, L_2( \Pstar ) \big)$ bracketing functions $\big\{ \big[ l_{q,j}^*, u_{q,j}^* \big] \big\}_{j=1}^{N_q^*}$ of size at most $2^{-q} \pi_{\min}^{(T-1)/2}$ that cover $\bar{\F}$, i.e.,
\begin{equation}
    \label{app:finalNonnestedBracketSize}
    \Estar{2:T} \left[ \left( u_{q,j}^*(\history{i}{T}) - l_{q,j}^*(\history{i}{T}) \right)^2 \right]^{1/2} \leq 2^{-q} \pi_{\min}^{(T-1)/2}.
\end{equation}
These brackets form a partition of $\bar{\F}$, which we write as $\big\{ \bar{\F}_{q,j}^* \big\}_{j=1}^{N_q^*}$. Note that these partitions are not necessarily nested. Below we use $\big\{ \bar{\F}_{q,j}^* \big\}_{j=1}^{N_q^*}$ to refer to the potentially \bo{non-nested} partitions (that exist by assumption) and $\big\{ \bar{\F}_{q,j} \big\}_{j=1}^{N_q}$ to refer to the \bo{nested} partitions (that we will construct). \\

\noindent We take intersections of the partitions $\big\{ \bar{\F}_{q,j}^* \big\}_{j=1}^{N_q^*}$ to construct a set of nested partitions $\big\{ \bar{\F}_{q,j} \big\}_{j=1}^{N_q}$ for all integers $q \geq q_0$. 
\begin{itemize}
	\item For partition $\big\{ \bar{\F}_{q_0,j} \big\}_{j=1}^{N_{q_0}}$, we simply set $\bar{\F}_{q_0,j} \triangleq \bar{\F}_{q_0,j}^*$ for all $j \in \big[ 1 \colon N_{q_0}^* \big]$. This means that $N_{q_0} \triangleq N_{q_0}^*$.
	\item For partition $\big\{ \bar{\F}_{q_0+1,j} \big\}_{j=1}^{N_{q_0+1}}$, we set partition pieces $\bar{\F}_{q_0+1,j}$ for all $j \in \big[ 1 \colon N_{q_0}^* \big]$ to be the intersections between all pairs of partition pieces $\bar{\F}_{q_0,k}^*$ and $\bar{\F}_{q_0+1,l}^*$ for $k \in \big[ 1 \colon N_{q_0} \big]$ and $l \in \big[ 1 \colon N_{q_0+1} \big]$. This means that $N_{q_0+1} \triangleq N_{q_0}^* \cdot N_{q_1}^*$.  
	Note that it could be that some partition pieces $\bar{\F}_{q_0+1,j}$ are empty; this is okay, as we simply want to upper bound the bracketing number. %, e.g., if the original partitions $\big\{ \bar{\F}_{q,j}^* \big\}_{j=1}^{N_q^*}$ were already nested for $q= q_0, q_0+1$.
	\item For general $q \geq q_0$, we set partition pieces $\bar{\F}_{q,j}$ for $j \in \big[ 1 \colon N_q \big]$ to be the intersections between all possible combinations in which we take one partition piece from each partition level $\big\{ \bar{\F}_{p,j}^* \big\}_{j=1}^{N_p^*}$ for each $p \in [q_0 \colon q]$. This means that each partition piece $\bar{\F}_{q,j}$ is the intersection between $\bar{\F}_{q_0,k_{q_0}}^*$, $\bar{\F}_{q_0+1,k_{q_0+1}}^*$, ..., $\bar{\F}_{q_0+1,k_{q}}^*$ for $k_{q_0} \in \big[ 1 \colon N_{q_0} \big]$, $k_{q_0+1} \in \big[ 1 \colon N_{q_0}+1 \big]$, ..., $k_{q} \in \big[ 1 \colon N_q \big]$.
	This means that there are $N_{q} \triangleq \prod_{p=q_0}^q N_{p}^*$ total partition pieces (for our constructed nested partitions) at this level.
\end{itemize}

\noindent Recall the potentially non-nested partitions were defined by bracketing functions $\big\{ \big[ l_{q,j}^*, u_{q,j}^* \big] \big\}_{j=1}^{N_q^*}$. Due to how we constructed the nested partitions $\big\{ \big[ l_{q,j}, u_{q,j} \big] \big\}_{j=1}^{N_q}$ (procedure described above), we have the following results:
%Since we constructed\textit{nested} partitions $\big\{ \big[ l_{q,j}, u_{q,j} \big] \big\}_{j=1}^{N_q}$ by taking intersections of the brackets $\big\{ \big[ l_{q,j}^*, u_{q,j}^* \big] \big\}_{j=1}^{N_q^*}$ at different levels, we have the following results:
\begin{itemize}
    \item The nested partitions $\big\{ \big[ l_{q,j}, u_{q,j} \big] \big\}_{j=1}^{N_q}$ must be covering since the non-nested prtitions $\big\{ \big[ l_{q,j}^*, u_{q,j}^* \big] \big\}_{j=1}^{N_q^*}$ are covering and we took all possible intersections of these non-nested partitions of size $p \in [q_0 : q]$ to construct our nested partitions.
    \item The nested partitions $\big\{ \big[ l_{q,j}, u_{q,j} \big] \big\}_{j=1}^{N_q}$ must be at most of the size of the largest bracket out of $\big\{ \big[ l_{q,j}^*, u_{q,j}^* \big] \big\}_{j=1}^{N_q^*}$ again since we took all possible intersections of these non-nested partitions of size $p \in [q_0 : q]$ to construct our nested partitions. Thus, we define $\Delta_{q,j} \triangleq u_{q,j}^* - l_{q,j}^*$; note that this choice of $\Delta_{q,j}$ satisfies the conditions of display \eqref{eqn:Deltabound} since
    \begin{multline*}
        \Estar{2:T} \left[ \rhostar{i}{2:T} \Delta_{q,j} ( \HH_{T}^{(i)} )^2  \right]
        \leq \pi_{\min}^{-(T-1)} \Estar{2:T} \left[ \Delta_{q,j} ( \HH_{T}^{(i)} )^2  \right] \\
        = \pi_{\min}^{-(T-1)} \Estar{2:T} \left[ \left( u_{q,j}^*( \HH_{T}^{(i)} ) - l_{q,j}^*( \HH_{T}^{(i)} ) \right) ^2  \right]
        \leq 2^{-2q}.
    \end{multline*}
    The first inequality above holds by Condition \ref{cond:exploration} and the second inequality holds by how we defined the non-nested partitions in display \eqref{app:finalNonnestedBracketSize}.
\end{itemize}

\noindent We now show that display \eqref{eqn:integralBound} holds, i.e., the number of sets in the partition grows at a bounded rate as the size of the partition pieces goes to zero:
\begin{equation*}
	\sum_{q = q_0}^\infty 2^{-q} \sqrt{ \log N_q }
	= \sum_{q = q_0}^\infty 2^{-q} \sqrt{ \log \bigg( \prod_{p=q_0}^q N_{p}^* \bigg) }
	=  \sum_{q = q_0}^\infty 2^{-q} \sqrt{ \sum_{p=q_0}^q \log N_p^* }
\end{equation*}
Note that $\sqrt{ \sum_{p=q_0}^q \log N_p^* } \leq \sum_{p=q_0}^q \sqrt{ \log N_p^* }$ because $\sqrt{a+b} \leq \sqrt{a} + \sqrt{b}$ for any positive non-negative values $a, b$. 
\begin{equation*}
	\leq \sum_{q = q_0}^\infty 2^{-q} \sum_{p=q_0}^q \sqrt{ \log N_p^* }
	= \sum_{q = q_0}^\infty 2^{-q} \sum_{p=q_0}^\infty \II_{p \leq q} \sqrt{ \log N_p^* }
	= \sum_{p=q_0}^\infty \sqrt{ \log N_p^* } \sum_{q = q_0}^\infty 2^{-q} \II_{p \leq q} 
\end{equation*}
For the last equality above, we can exchange the infinite summations above by Fubini's theorem because the following argument will show that $\sum_{p=q_0}^\infty \sqrt{ \log N_p^* } \sum_{q = q_0}^\infty 2^{-q} \II_{p \leq q}$ is bounded. \\

\noindent Since $\sum_{q = q_0}^\infty 2^{-q} \II_{p \leq q} = \sum_{q = p}^\infty 2^{-q} = 2^{-(p-1)}$,
\begin{equation*}
	= \sum_{p=q_0}^\infty 2^{-(p-1)} \sqrt{ \log N_p^* } 
	= 4 \sum_{p=q_0}^\infty 2^{-(p+1)} \sqrt{ \log N_p^* } 
\end{equation*}
\begin{equation*}
	= 4 \sum_{p=q_0}^\infty 2^{-(p+1)} \sqrt{ \log N_{[~]} \big( 2^{-p} \pi_{\min}^{(T-1)/2}, \bar{\F}, L_2( \Pstar ) \big) } 
\end{equation*}
Since $N_{[~]} \big( 2^{-p} \pi_{\min}^{(T-1)/2}, \bar{\F}, L_2( \Pstar ) \big)$ is monotonically increasing as $p$ increases by lower Darboux sums, we have the following upper bound: % (illustrated below)
\begin{equation*}
	\leq 4 \int_0^{2^{-q_0}} \sqrt{ \log N_{[~]} \big( \epsilon \pi_{\min}^{(T-1)/2}, \bar{\F}, L_2( \Pstar ) \big) } d \epsilon
\end{equation*}

%\begin{figure}[H]
%	\begin{center}
%	\includegraphics[width=8cm]{./figures/integral_inequality}
%	\end{center}
%\end{figure}

\noindent Since we chose $q_0$ such that $\delta < 2^{-q_0} \leq 2 \delta$,
\begin{multline*}
	\leq 4 \int_0^{\delta} \sqrt{ \log N_{[~]} \big( \epsilon \pi_{\min}^{(T-1)/2}, \bar{\F}, L_2( \Pstar ) \big) } d \epsilon
	%\underbrace{\leq}_{(a)} 32 \int_0^\delta \sqrt{ \log N_{[~]} \big( \epsilon, \bar{\F}, L_2( \Pstar ) \big) } d \epsilon \\
    \underbrace{\leq}_{(a)} 4 \int_0^\delta \sqrt{ \log N_{[~]} \big( \epsilon \pi_{\min}^{(T-1)/2}, \F, L_2( \Pstar ) \big) } d \epsilon \\
    = 4 \pi_{\min}^{-(T-1)/2} \int_0^\delta \sqrt{ \log N_{[~]} \big( \epsilon \pi_{\min}^{(T-1)/2}, \F, L_2( \Pstar ) \big) } \pi_{\min}^{(T-1)/2} d \epsilon \\
    \underbrace{=}_{(b)}  4 \pi_{\min}^{-(T-1)/2} \int_0^{\delta \pi_{\min}^{(T-1)/2}} \sqrt{ \log N_{[~]} \big( u, \F, L_2( \Pstar ) \big) } d u
	\underbrace{<}_{(c)} \infty.
\end{multline*}
%Inequality (a) above holds because $N_{[~]} \big( \epsilon, \bar{\F}, L_2( \Pstar )$ is monotonically increasing as $\epsilon$ goes to zero. 

\noindent Inequality (a) above holds by display \eqref{eqn:bracketingFbarbound}.

\noindent Equality (b) above holds by integration by substitution for $u = \pi_{\min}^{(T-1)/2} \epsilon$. 

\noindent Inequality (c) above holds by our finite bracketing integral assumption. $\blacksquare$ 

\end{appendix}

\end{document}